\documentclass[12pt,a4paper,twoside]{diss}

\PassOptionsToPackage{normalem}{ulem}
\usepackage{ulem}
\usepackage{comment}
\usepackage{relsize}
\usepackage{grffile}

\usepackage[multiple]{footmisc}
\usepackage[subfigure,eulermath]{classicthesis_cw}
\usepackage[caption=false]{subfig}
\usepackage[scaled=.9]{beramono}
\usepackage{algorithm}
\usepackage{algorithmic}
\usepackage{boxedminipage}
\usepackage{xcolor}
\usepackage{ucs}
\usepackage[utf8x]{inputenc}
\makeatletter
\@namedef{opt@inputenc.sty}{utf8}
\makeatother
\usepackage{CJKutf8}
\usepackage{supertabular}
\usepackage{amsfonts}
\usepackage{amsmath}
\usepackage{amssymb}
\usepackage{amsthm}
\usepackage{bbding}
\usepackage{placeins}
\usepackage{array}
\usepackage{booktabs}
\usepackage{enumerate}
\usepackage[british]{babel}
\usepackage[T1]{fontenc}
\usepackage{graphicx}
\usepackage{latexsym}
\usepackage{wrapfig}
\usepackage{pifont}
\usepackage{natbib}
\usepackage{wrapfig}
\usepackage{xspace}
\usepackage{tabularx}
\usepackage{multicol}
\usepackage{multirow}
\usepackage{colortbl}
\usepackage{lettrine}
\usepackage{nameref}
\usepackage{remreset}
\usepackage{longtable}
\usepackage{soul}
\usepackage{bm}
\usepackage{bbm}
\usepackage{cleveref}
\usepackage[tight]{minitoc}
\usepackage[nolist,nohyperlinks]{acronym}
\usepackage{rotating}
\usepackage{makecell}
\usepackage[normalem]{ulem}
\usepackage{nicefrac}
\usepackage{appendix}
\usepackage{pdflscape}
\usepackage{afterpage}
\usepackage{bookmark}

\usepackage{adjustbox}
\RequirePackage[hyperpageref]{backref}
   \renewcommand*{\backref}[1]{}
   \renewcommand*{\backrefalt}[4]{
      \ifcase #1
         No cited.
      \or
         Cited on page #2.
      \else
         Cited on pages #2.
      \fi}
\usepackage{pgfplots}
\pgfplotsset{compat=1.14}
\usepackage{sistyle}

\usetikzlibrary{arrows.meta}
\usetikzlibrary{backgrounds}
\usetikzlibrary{positioning, fit, mindmap, trees, calc,tikzmark,shapes}
\usetikzlibrary{shapes.arrows, fadings, automata,tikzmark,decorations.pathreplacing,patterns}
\usepgfplotslibrary{groupplots}
\SIthousandsep{,}

\makeatletter\@removefromreset{footnote}{chapter}\makeatother
\definecolor{tango_green}{RGB}{78, 154, 6}

\definecolor{chamoisee}{rgb}{0.63, 0.47, 0.35}

\newcolumntype{P}[1]{>{\raggedright\arraybackslash}m{#1}}%
\newcolumntype{S}[1]{>{\centering\arraybackslash}m{#1}}%
\newcolumntype{R}[1]{>{\raggedleft\arraybackslash}m{#1}}%

\def\ade20k{\textit{ADE20K}\xspace}
\def\nyud2{\textit{NYUD-v2}\xspace}
\def\sun3d{\textit{SUN3D}\xspace}

\hypersetup{
     colorlinks=false, linktoc=all, pdfstartpage=3, pdfstartview=FitV,%
     breaklinks=true, pdfpagemode=UseNone, pageanchor=true, pdfpagemode=UseOutlines,%
     plainpages=false, bookmarksnumbered, bookmarksopen=true, bookmarksopenlevel=1,%
     hypertexnames=true, pdfhighlight=/O,%
     urlcolor=BrickRed, linkcolor=NavyBlue, citecolor=OliveGreen,%
     linkbordercolor=NavyBlue, citebordercolor=OliveGreen,%
     urlbordercolor=BrickRed,%
}

\definecolor{lightgray}{gray}{0.8}

\newlength{\vfigcaplen}
\newlength{\vtblcaplen}
\newcolumntype{C}{>{$}c<{$}}

 \makeatletter
 \DeclareRobustCommand\onedot{\futurelet\@let@token\@onedot}
 \def\@onedot{\ifx\@let@token.\else.\null\fi\xspace}

 \makeatother

\newcolumntype{I}[1]{>{\centering\arraybackslash}p{#1}} 
\newcolumntype{O}[1]{>{\raggedleft\arraybackslash}p{#1}} 

\usepackage{ifthen}
\makeatletter

\newcounter{rownods}
\providecommand{\rownods}[1][__empty__]{%
\ifthenelse{\equal{#1}{__empty__}}{%
}{%
\setcounter{rownods}{#1}%
\addtocounter{rownods}{-1}%
}%
\renewcommand*\therownods{(\@alph\c@rownods)}
\renewcommand*\theHrownods{(\@alph\c@rownods)}
\refstepcounter{rownods}%
\therownods
}

\newcounter{rownodsb}
\providecommand{\rownodsb}[1][__empty__]{%
\ifthenelse{\equal{#1}{__empty__}}{%
}{%
\setcounter{rownodsb}{#1}%
\addtocounter{rownodsb}{-1}%
}%
\renewcommand*\therownodsb{(\@alph\c@rownodsb)}
\renewcommand*\theHrownodsb{(\@alph\c@rownodsb)}
\refstepcounter{rownodsb}%
\therownodsb
}

\newcounter{rownoteas}
\providecommand{\rownoteas}[1][__empty__]{%
\ifthenelse{\equal{#1}{__empty__}}{%
}{%
\setcounter{rownoteas}{#1}%
\addtocounter{rownoteas}{-1}%
}%
\renewcommand*\therownoteas{(\@alph\c@rownoteas)}
\renewcommand*\theHrownoteas{(\@alph\c@rownoteas)}
\refstepcounter{rownoteas}%
\therownoteas
}

\newcounter{rownoped}
\providecommand{\rownoped}[1][__empty__]{%
\ifthenelse{\equal{#1}{__empty__}}{%
}{%
\setcounter{rownoped}{#1}%
\addtocounter{rownoped}{-1}%
}%
\renewcommand*\therownoped{(\@alph\c@rownoped)}
\renewcommand*\theHrownoped{(\@alph\c@rownoped)}
\refstepcounter{rownoped}%
\therownoped
}

\newcounter{rownoct}
\providecommand{\rownoct}[1][__empty__]{%
\ifthenelse{\equal{#1}{__empty__}}{%
}{%
\setcounter{rownoct}{#1}%
\addtocounter{rownoct}{-1}%
}%
\renewcommand*\therownoct{(\@alph\c@rownoct)}
\renewcommand*\theHrownoct{(\@alph\c@rownoct)}
\refstepcounter{rownoct}%
\therownoct
}

\newcounter{rownofp}
\providecommand{\rownofp}[1][__empty__]{%
\ifthenelse{\equal{#1}{__empty__}}{%
}{%
\setcounter{rownofp}{#1}%
\addtocounter{rownofp}{-1}%
}%
\renewcommand*\therownofp{(\@alph\c@rownofp)}
\renewcommand*\theHrownofp{(\@alph\c@rownofp)}
\refstepcounter{rownofp}%
\therownofp
}
\makeatother

\makeatletter
\long\def\@footnotetext#1{%
      \H@@footnotetext{%
        \ifHy@nesting
         \hyper@@anchor{\@currentHref}{#1}%
       \else
         \Hy@raisedlink{\hyper@@anchor{\@currentHref}{\relax}}#1%
       \fi
     }}

  \def\@footnotemark{%
     \leavevmode
     \ifhmode\edef\@x@sf{\the\spacefactor}\nobreak\fi
     \H@refstepcounter{Hfootnote}%
     \hyper@makecurrent{Hfootnote}%
     \hyper@linkstart{link}{\@currentHref}%
     \@makefnmark
     \hyper@linkend
     \ifhmode\spacefactor\@x@sf\fi
     \relax
   }%

 \ifFN@multiplefootnote%
     \renewcommand*\@footnotemark{%
      \leavevmode
      \ifhmode
        \edef\@x@sf{\the\spacefactor}%
        \FN@mf@check
        \nobreak
      \fi
      \H@refstepcounter{Hfootnote}%
      \hyper@makecurrent{Hfootnote}%
      \hyper@linkstart{link}{\@currentHref}%
      \@makefnmark
      \hyper@linkend
      \ifFN@pp@towrite
        \FN@pp@writetemp
        \FN@pp@towritefalse
      \fi
      \FN@mf@prepare
      \ifhmode\spacefactor\@x@sf\fi
      \relax%
    }%
 \fi
\makeatother

\definecolor{riptide}{RGB}{141,211,199}
\definecolor{pale_prim}{RGB}{255,255,179}
\definecolor{lavender_gray}{RGB}{190,186,218}
\definecolor{salmon}{RGB}{242,131,107}
\definecolor{seagull}{RGB}{128,177,211}
\definecolor{rajah}{RGB}{253,180,98}
\definecolor{yellow_green}{RGB}{198,222,119}
\definecolor{classic_rose}{RGB}{252,205,229}
\definecolor{feijoa}{RGB}{178,223,138}

\definecolor{cruise}{RGB}{179,226,205}
\definecolor{apricot}{RGB}{253,205,172}
\definecolor{periwinkle}{RGB}{203,213,232}
\definecolor{snow_flurry}{RGB}{230,245,201}
\definecolor{buttermilk}{RGB}{255,242,174}

\definecolor{sundown}{RGB}{249, 180, 181}
\definecolor{spindle}{RGB}{179,205,227}
\definecolor{tea_green}{RGB}{204,235,197}
\definecolor{languid_lavender}{RGB}{222,203,228}
\definecolor{champagne}{RGB}{254,217,166}
\definecolor{cream}{RGB}{255,255,204}

\definecolor{monte_carlo}{RGB}{135,204,194}
\definecolor{melon}{RGB}{254,191,181}
\definecolor{granny_smith_apple}{RGB}{150,214,150}
\definecolor{watusi}{RGB}{254,221,207}
\definecolor{see_green}{RGB}{161,228,195}

\definecolor{moss_green}{RGB}{170,216,176}
\definecolor{opal}{RGB}{164,207,190}

\definecolor{pale_turquoise}{RGB}{172,240,242}
\definecolor{Madang}{RGB}{190,235,159}
\definecolor{pixie_green}{RGB}{183,214,170}
\definecolor{coral_andy}{RGB}{243,204,205}
\definecolor{manhattan}{RGB}{226,180,125}
\definecolor{quartz}{RGB}{219,223,238}
\definecolor{spring_sun}{RGB}{242,243,195}
\definecolor{dairy_cream}{RGB}{254,226,189}
\definecolor{surf_crest}{RGB}{205,230,208}
\definecolor{french_pass}{RGB}{195,232,246}
\definecolor{cosmos}{RGB}{248,209,210}
\definecolor{portafino}{RGB}{245,237,160}
\definecolor{sail}{RGB}{163,205,235}
\definecolor{hint_green}{RGB}{226,246,209}

\definecolor{jet_stream}{RGB}{188, 214, 210}

\definecolor{azalea}{RGB}{251, 196, 196}
\definecolor{wewak}{RGB}{244, 143, 150}
\definecolor{bittersweet}{RGB}{255,111,105}
\definecolor{sunset_orange}{RGB}{242,89,75}
\definecolor{light_coral}{RGB}{244, 127, 123}
\definecolor{carnation}{RGB}{245, 80, 86}
\definecolor{flamingo}{RGB}{237, 88, 85}
\definecolor{carmine_pink}{RGB}{231, 76, 60}
\definecolor{deep_carmine_pink}{RGB}{236, 50, 67}
\definecolor{fire_engine_red}{RGB}{210,44,41}
\definecolor{amaranth}{RGB}{234,46,73}
\definecolor{ku_crimson}{RGB}{243, 0, 25}
\definecolor{fire_engine_red}{RGB}{206, 37, 51}
\definecolor{copper_rust}{RGB}{155, 64, 74}

\definecolor{chilean_fire}{RGB}{215, 87, 44}

\definecolor{japanese_laurel}{RGB}{53, 116, 40}

\definecolor{turmeric}{RGB}{211, 178, 76}
\definecolor{saffron}{RGB}{249,193,62}
\definecolor{my_sin}{RGB}{255, 176, 59}
\definecolor{tree_poppy}{RGB}{246, 154, 27}
\definecolor{jaffa}{RGB}{240, 131, 58}
\definecolor{crusta}{RGB}{254, 127, 44}
\definecolor{tahiti_gold}{RGB}{223, 102, 36}
\definecolor{outrageous_orange}{RGB}{255, 100, 45}
\definecolor{safety_orange}{RGB}{254, 106, 0}

\definecolor{turquoise}{RGB}{41,217,194}
\definecolor{puerto_rico}{RGB}{94, 194, 166}
\definecolor{mountain_meadow}{RGB}{0, 163, 136}
\definecolor{free_speech_aquamarine}{RGB}{0, 156, 114}
\definecolor{java}{RGB}{2,190,196}

\definecolor{matisse}{RGB}{25, 104, 167}
\definecolor{shakespeare}{RGB}{85, 154, 193}
\definecolor{mona_lisa}{RGB}{246,152,134}

\definecolor{bgc}{RGB}{245,245,245}
\definecolor{tuatara}{RGB}{67, 67, 67}
\definecolor{aluminum}{RGB}{153,153,153}
\definecolor{silver}{RGB}{191,191,191}
\definecolor{platinum}{RGB}{228,228,228}
\definecolor{mercury}{RGB}{230,230,230}
\definecolor{gallery}{RGB}{240,240,240}
\definecolor{athens_gray}{RGB}{236, 240, 241}
\definecolor{ship_gray}{RGB}{77,77,77}

\definecolor{early_dawn}{RGB}{252,243,218}
\definecolor{egg_shell}{RGB}{238, 234, 215}
\definecolor{midnight}{RGB}{0, 29, 50}
\definecolor{sundown}{RGB}{249, 180, 181}
\definecolor{sun_shade}{RGB}{255, 144, 68}
\definecolor{sushi}{RGB}{117, 168, 47}
\definecolor{tomato}{RGB}{255, 97, 56}
\definecolor{ice_cold}{RGB}{169,232,220}

\definecolor{jelly_bean}{RGB}{45, 126, 150}
\definecolor{celestial_blue}{RGB}{52, 152, 219}
\definecolor{curious_blue}{RGB}{41, 128, 185}
\definecolor{french_blue}{RGB}{0, 112, 182}
\definecolor{matisse}{RGB}{25, 104, 167}

\definecolor{biscay}{RGB}{44, 62, 80}

\definecolor{cosmic_latte}{RGB}{222, 247, 229}
\definecolor{chinook}{RGB}{163, 232, 178}
\definecolor{padua}{RGB}{121, 189, 143}
\definecolor{ocean_green}{RGB}{79, 176, 112}
\definecolor{pastel_green}{RGB}{107, 227, 135}
\definecolor{chateau_green}{RGB}{69, 191, 85}
\definecolor{RoyalBlue}{RGB}{69, 191, 85}
\definecolor{pigment_green}{RGB}{0, 175, 79}
\definecolor{fern}{RGB}{101,197,117}
\definecolor{killarney}{RGB}{56, 113, 66}
\definecolor{viridian}{RGB}{70, 137, 102}
\definecolor{amaranth}{rgb}{0.9, 0.17, 0.31}
\definecolor{kellygreen}{rgb}{77, 186, 23}
\definecolor{azure}{rgb}{0.0, 0.5, 1.0}
\definecolor{gred}{RGB}{219,68,55}
\definecolor{gblue}{RGB}{66,133,244}
\definecolor{gyellow}{RGB}{244,180,0}
\definecolor{ggreen}{RGB}{15,157,88}
\definecolor{ggrey}{RGB}{115,115,115}
\definecolor{blue1}{RGB}{33,113,181}
\definecolor{blue2}{RGB}{66,146,198}
\definecolor{blue3}{RGB}{107,174,214}
\definecolor{blue4}{RGB}{158,202,225}
\definecolor{blue5}{RGB}{198,219,239}
\definecolor{blue6}{RGB}{239,243,255}
\definecolor{blue7}{RGB}{247,251,255}
\newcommand{\error}[1]{\textcolor{gred}{\textbf{#1}}} 
\newcommand{\novelh}[1]{\textcolor{gblue}{\textbf{#1}}} 

\newcolumntype{L}{>{\arraybackslash}m{12cm}}
\crefformat{section}{\S#2#1#3}
\crefformat{subsection}{\S#2#1#3}
\crefformat{subsubsection}{\S#2#1#3}

\DeclareMathOperator*{\argmax}{arg\,max}
\usepackage{xspace} \makeatletter \@ifundefined{onedot}{%
  \DeclareRobustCommand\onedot{\futurelet\@let@token\@onedot}%
  \def\@onedot{\ifx\@let@token.\else.\null\fi\xspace}%
}{} \makeatother

\let\originalparagraph\paragraph
\renewcommand{\paragraph}[2][.]{\originalparagraph{#2#1}}

\usepackage{booktabs} 
\usepackage{array} \usepackage{amssymb} \usepackage{textcomp}

\acrodef{dpm}[\texttt{DPM}]{Deformable Parts Model}
\acrodef{svm}[\texttt{SVM}]{Support Vector Machine}
\acrodef{hog}[\texttt{HOG}]{Histogram of Oriented Gradients}
\acrodef{dish}[\texttt{DiSh}]{Distributed Shape}

\definecolor{lightgray}{gray}{0.8}
\definecolor{verylightgray}{gray}{0.9}

\graphicspath{{./fig/}{./nexus/}{./improve_vae/}{./select/}{./improve_align/}{./segment/}{./diversify/}{./fewshot/}}

\usepackage{silence}
\begin{document}
\pagenumbering{roman}
\hypertarget{title}{}
\pdfbookmark[0]{Title Page}{title}
\begin{titlepage}
\def\docdate{2021}

\newlength{\longskip}
\setlength{\longskip}{0.03\textheight}

\sffamily
\vspace*{\stretch{0.5}}

\begin{center}

\hrulefill\par
\huge{
    \vspace{0.5cm}
    
    \textbf{\scalebox{1}[1.2]{Deep Latent-Variable Models for}}\\ [0.35cm]
    \textbf{\scalebox{1}[1.2]{Text Generation}}\\[0.35cm]

     }
\hrulefill\par
\vspace*{\stretch{1.7}}

\Large
A dissertation submitted towards the degree  \\
Doctor of Engineering\\
(Dr.-Ing.)\\
of the Faculty of Mathematics and Computer Science\\ 
of Saarland University
\vspace*{\stretch{1}}

by \\
\textbf{Xiaoyu Shen} \\[1em]

\end{center}

\end{titlepage}

\cleardoublepage

\hypertarget{abstract}{}
\pdfbookmark[0]{Abstract}{abstract}
\thispagestyle{plain}
\chapter*{\Huge Abstract} 
\emph{Text generation} aims to produce human-like natural language output for down-stream tasks. It covers a wide range of applications like machine translation, document summarization, dialogue generation and so on. Recently deep neural network-based end-to-end architectures have been widely adopted. The end-to-end approach conflates all sub-modules, which used to be designed by complex handcrafted rules, into a holistic encode-decode architecture. Given enough training data, it is able to achieve state-of-the-art performance yet avoiding the need of language/domain-dependent knowledge. Nonetheless, deep learning models are known to be extremely data-hungry, and text generated from them usually suffer from low diversity, interpretability and controllability. As a result, it is difficult to trust the output from them in real-life applications. Deep latent-variable models, by specifying the probabilistic distribution over an intermediate latent process, provide a potential way of addressing these problems while maintaining the expressive power of deep neural networks.

This dissertation presents how deep latent-variable models can improve over the standard encoder-decoder model for text generation. We start from an introduction of encoder-decoder and deep latent-variable models, then go over popular optimization strategies like variational inference, dynamic programming, soft relaxation and reinforcement learning. Finally, we elaborate on:
\begin{enumerate}
    \item How latent variables can \emph{improve the diversity of text generation} by learning holistic, sentence-level latent representations. By doing so, a latent representation can be first sampled, from which diverse text can be generated. We present effective algorithms to simultaneously train the representation learning and text generation via variational inference. Further, to address the uni-modal and inconsistency limitations of variational inference, we propose a wake-sleep variation and mutual information-enhanced training objective. Experiments show they outperform standard variational inference and non-latent-variable models on the dialogue generation tasks.
    \item How latent variables can \emph{improve the controllability and interpretability of text generation} by adding finer-grained, latent specifications on the intermediate generation process. We illustrate using latent variables to stand for word alignment, content selection, text segmentation and field-segment correspondences. We derive efficient training algorithms for them so that the text generation can be explicitly controlled by manipulating the latent variables, which are human interpretable by definition.
    \item How to \emph{overcome the sparsity of training samples} by treating non-parallel text as latent variables. The training can be performed like in the standard EM algorithm which is stable to converge. We show it can be successfully applied in dialogue generation and significantly enrich the topic of generation space by utilizing non-conversational text.
\end{enumerate}
\cleardoublepage

\hypertarget{zusammenfassung}{}
\pdfbookmark[0]{Zusammenfassung}{zusammenfassung}
\hyphenation{Ob-jekt-klas-sen-mo-del-le}
\thispagestyle{plain}
\chapter*{\Huge Zusammenfassung}

\textbf{Textgenerierung} zielt darauf ab, eine menschenähnliche Textausgabe in natürlicher Sprache für Anwendungen zu erzeugen. Es deckt eine breite Palette von Anwendungen ab, wie maschinelle Übersetzung, Zusammenfassung von Dokumenten, Generierung von Dialogen usw. In letzter Zeit werden dafür hauptsächlich End-to-End-Architekturen auf der Basis von tiefen neuronalen Netzwerken verwendet. Der End-to-End-Ansatz fasst alle Submodule, die früher nach komplexen handgefertigten Regeln entworfen wurden, zu einer ganzheitlichen Codierungs-Decodierungs-Architektur zusammen. Bei ausreichenden Trainingsdaten kann eine Leistung auf dem neuesten Stand der Technik erzielt werden, ohne dass sprach- und domänenabhängiges Wissen erforderlich ist. Deep-Learning-Modelle sind jedoch als extrem datenhungrig bekannt und daraus generierter Text leidet normalerweise unter geringer Diversität, Interpretierbarkeit und Kontrollierbarkeit. Infolgedessen ist es schwierig, der Ausgabe von ihnen in realen Anwendungen zu vertrauen. Tiefe Modelle mit latenten Variablen bieten durch Angabe der Wahrscheinlichkeitsverteilung über einen latenten Zwischenprozess eine potenzielle Möglichkeit, diese Probleme zu lösen und gleichzeitig die Ausdruckskraft tiefer neuronaler Netze zu erhalten.

Diese Dissertation zeigt, wie tiefe Modelle mit latenten Variablen Texterzeugung verbessern gegenüber dem üblichen Encoder-Decoder-Modell. Wir beginnen mit einer Einführung in Encoder-Decoder- und Deep Latent Variable-Modelle und gehen dann auf gängige Optimierungsstrategien wie Variationsinferenz, dynamische Programmierung, Soft Relaxation und Reinforcement Learning ein. Danach präsentieren wir Folgendes:
\begin{enumerate}
    \item Wie latente Variablen Vielfalt der Texterzeugung verbessern können, indem ganzheitliche, latente Darstellungen auf Satzebene gelernt werden. Auf diese Weise kann zunächst eine latente Darstellung ausgewählt werden, aus der verschiedene Texte generiert werden können. Wir präsentieren effektive Algorithmen, um gleichzeitig das Lernen der Repräsentation und die Texterzeugung durch Variationsinferenz zu trainieren. Um die Einschränkungen der Variationsinferenz bezüglich Uni-Modalität und Inkonsistenz anzugehen, schlagen wir eine Wake-Sleep-Variation und ein auf Transinformation basierendes Trainingsziel vor. Experimente zeigen, dass sie sowohl die übliche Variationsinferenz als auch nicht-latente Variablenmodelle bei der Dialoggenerierung übertreffen.
    \item Wie latente Variablen die Steuerbarkeit und Interpretierbarkeit der Texterzeugung verbessern können, indem feinkörnigere latente Spezifikationen zum Zwischengenerierungsprozess hinzugefügt werden. Wir veranschaulichen die Verwendung latenter Variablen für Wortausrichtung, Inhaltsauswahl, Textsegmentierung und Feldsegmentkorrespondenz. Wir leiten für sie effiziente Trainingsalgorithmen ab, damit die Texterzeugung explizit gesteuert werden kann, indem die latente Variable, die durch ihre Definition vom Menschen interpretiert werden kann, manipuliert wird.
    \item Überwindung der Seltenheit von Trainingsmustern durch Behandlung von nicht parallelem Text als latente Variablen. Das Training kann wie beim Standard-EM-Algorithmus durchgeführt werden, der stabil konvergiert. Wir zeigen, dass es bei der Dialoggenerierung erfolgreich angewendet werden kann und den Generierungsraum durch die Verwendung von nicht-konversativem Text erheblich bereichert.
\end{enumerate}
\cleardoublepage

\dominitoc
\hypertarget{toc}{}
\pdfbookmark[0]{\contentsname}{toc}
\tableofcontents
\clearoddpage

\pagenumbering{arabic}

\chapter{Introduction}
\label{Introduction}
\lettrine[lines=3]{T}{ext} generation is the subfield of artificial intelligence and computational linguistics that focuses on computer systems that can produce understandable texts in English or other human languages~\citep{reiter2000building}. Specifically, we define text generation as the task of ``\emph{Given some input information, provide coherent human-like text as the output, based on task-specific requirement}". In this 
thesis, we make no assumption about the input. It can be of arbitrary format or even multiple-sourced. Depending on the input context and requirement, it covers a broad range of different tasks such as:
\begin{itemize}
    \item machine translation~\citep{och1999improved}: translate from one language to another
    \item text summarization~\citep{clarke2010discourse}: summarize the gist of the input into concise sentences
    \item data-to-text~\citep{reiter2000building}: describe structured knowledge with human languages
    \item dialogue generation~\citep{vinyals2015neural}: build chatbots that can converse with people like human beings
    \item image captioning~\citep{xu2015show}: describe the contents in an image with human languages
    \item grammar error correction~\citep{dale2012hoo}: correct the grammar errors in the input and output the corrected sentence
    \item paraphrase generation~\citep{bannard2005paraphrasing}: generate paraphrases of the input sentence
    \item style transfer~\citep{shen2017style}: transfer the style of the input (e.g., positive, female, polite) sentence into another (e.g., negative, male, impolite)
\end{itemize}

Early text generation systems are normally rule-based and contain pipe-line structures of content determination, text structuring, sentence aggregation, etc.~\citep{thompson1977strategy,goldberg1994using,reiter1995automatic}, which requires enormous human labor to design hand-crafted rules. The domain knowledge required for one task generally cannot be easily transferred to another one. The recent advances in deep learning have given it a new impetus, where a single encoder-decoder neural architecture is devised to merge the traditional pipe-line architecture~\citep{sutskever2014sequence}. Relying on the powerful generalizing capability of deep neural networks, neural text generation systems can effectively learn the underlying generating distribution of human languages from large-scale dataset without pre-specified domain knowledge~\citep{bahdanau2015neural,xu2015show,rush2015neural,wen2015semantically,shen2017estimation,zhao2017gated,zhao2018comprehensive,hong2019improving,de2018generating,zhuang2017neobility,chang2020unsupervised,wiehr2020safe,chang2020dart,chang2018generative,chang2020dart,su2020moviechats}.

\section{Encoder-Decoder}
Given a specific text generation task, denote by $X=x_1,x_2,\ldots,x_m$ as the input information and $Y=y_1, y_2, \ldots, y_n$ as the output text. $m$ and $n$ are the length of the input and output respectively. All text generation tasks can be converted into an encoder-decoder structure, where an encoder encodes the input information into machine-readable continuous vectors, based on which the decoder estimates the probability of the target output word by word~\footnote{There has been a surge of research interest in unifying all NLP tasks, including understanding and generation, into a single encoder-decoder architecture~\citep{mccann2019the,radford2019language,raffel2019exploring}. Although this dissertation focuses on text generation, relevant techniques can be easily extended to text understanding tasks by replacing the decoder with different classifiers.}. Compared with traditional statistical methods based on n-gram counting and smoothing~\citep{chen1996empirical}, neural network-based encoder-decoder architectures can capture much longer dependency relations and generalize better at unseen scenarios. Normally, the probability is computed autoregressively as:
\begin{equation}
\label{eq: mle}
    p_\theta(Y|X) = \prod_{t=1}^{n} p_\theta(y_t|X,y_{<t})
\end{equation}

\begin{figure}[t]
  \centering
  \includegraphics[]{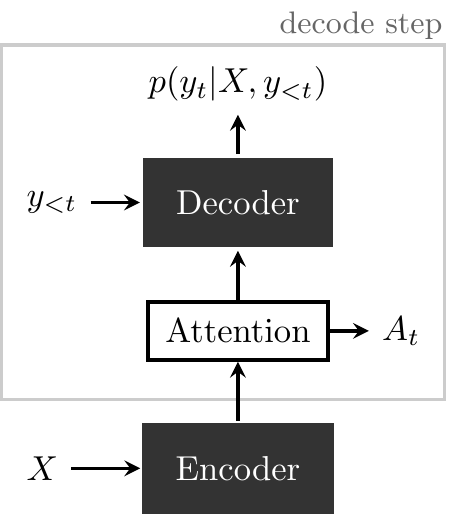}
  \caption{Illustration of a typical encoder-decoder architecture for text generation. Figure taken from \citep{xie2017neural}.}
  \label{fig:encdec}
\end{figure}

The encoder encodes each $x_i$ into a vector $h_i$. 
At each time step $t$, the previously-generated words are compressed into hidden state $d_t$ of the decoder. The input information is usually represented by means of the attention mechanism~\citep{bahdanau2015neural}, which is a weighted average of source vectors. The output probability is defined as:
\begin{equation}
\label{eq: attention}
\begin{split}
p_\theta(y_t|X,&y_{<t}) = p_\theta(y_t|A_t, d_t)\\
p_\theta(y_t|A_t, &d_t) = \text{softmax}(W_1 d_t + W_2 A_t)\\
A_t &= \sum_{i}\alpha_{t,i}h_{i}\\
\alpha_{t,i} &= \frac{e^{f(h_{i}, d_t)}}{\sum_j e^{f(h_{j}, d_t)}}
\end{split}
\end{equation}
$A_{t}$ is the attention vector at time step $t$.
$f$ is a score function to compute the similarity between  $h_i$ and $d_t$~\citep{luong2015effective}. $W_1$ and $W_2$ are learnable parameters. The attention mechanism can be seen as performing an approximation of the latent alignment~\citep{kim2017structured}. If the input information at position $i$ is important for the prediction of word $y_t$, we expect the similarity function $f(h_i, d_t)$ should be large and so as the weight $\alpha_{t,i}$. Figure~\ref{fig:encdec} illustrates the typical architecture of an encoder-decoder model for text generation. 

The choice of the encoder can differ from task to task, e.g. residual networks for images~\citep{he2016deep}, graph convolutional networks~\citep{kipf2016semi} for structured data and transformers~\citep{vaswani2017attention} for sequence input. The decoder is usually implemented as the LSTM~\citep{hochreiter1997long} or transformer neural network for their excellent capability at modelling sequential dependencies.

Details can differ from our definitions above in down-stream variants. For example, non-autoregressive models decode output in parallel instead of recursively to speed up the process~\citep{gu2018nonautoregressive,lee2018deterministic}. The output softmax can be augmented with weight tying~\citep{press2017using,inan2016tying} or pointer generators~\citep{gu2016incorporating,gulcehre2016pointing}. Sparsity constraints can be imposed to improve the interpretability and efficiency of the attention mechanism~\citep{martins2016softmax,child2019generating}. The essential idea is to estimate the probability of $p_\theta(Y|X)$ with an end-to-end encode-decode process. Specifically, in this thesis, we focus on models with \emph{tractable density estimations} of $p_\theta(Y|X)$ and optimized by \emph{maximum likelihood}. Namely, in the training stage, model parameters are updated to maximize the likelihood of parallel input-output pairs, defined by $p_\theta(Y|X)$. Though other variants based on generative adversarial networks~\citep{yu2017seqgan,fedus2018maskgan,Zhou2020Self-Adversarial}, which targets a lower bound of Jensen-Shannon divergence, or energy-based models~\citep{schmidt2019autoregressive,parshakova2019distributional,deng2020residual}, which directly estimate the sequence-level density albeit with intractable partition functions, have shown promising results as for improving the coherence and diversity of generated text, their training process is significantly more complex and unstable. When combined with latent-variable models, the training difficult would be aggregated and therefore very limited attempts have been done.

Though state-of-the-art performances have been achieved, seq2seq models have been constantly criticized for the (1) low diversity of outputs, (2) poor generalization under limited or noisy supervised data and (3) lack of interpretability and controllability~\citep{sriram2018cold,duvsek2020evaluating,holtzman2020the}. Latent variable models, by specifying the probabilistic distribution over a intermediate latent process, provide a tool to address these problems in a principled way. By integrating the latent variable framework into the encoder-decoder architecture (denoted as ``deep latent-variable model"), we are able to inject additional stochasticity, prior knowledge or structured dependencies without sacrificing the model capacity, thereby effectively alleviating the above-mentioned three problems. Recently proposed training algorithms like variational inference~\citep{kingma2014auto,rezende2014stochastic} and automatic differentiation tools like PyTorch~\citep{pytorch} also make it convenient to efficiently train them on large-scale datasets. As a result, improving text generation with deep latent-variable models have been a hot topic in recent years, with applications spreading across different domains~\citep{bowman2016generating,serban2017hierarchical,deng2018latent,shen2019select,he2020a,chang2021does,chang2021jointly,shen2017estimation,su2019improving,shen2020neural,qiu2020easyaug,backes2018simulating}. 
\section{Deep latent-variable model}
The encoder-decoder architecture defined above belongs to the so-called
fully visible belief networks (FVBN)~\citep{frey1996does,frey1998graphical}, which use the chain rule of probability to decompose a probability distribution over an n-dimensional vector into a product of one-dimensional probability distributions~\citep{goodfellow2016nips}. The main feature of FVBN is that all variables (for text generation, the input $X$ and output $Y$) are \emph{explicitly observable}. The only source of stochasticity is the data distribution itself. Given a fixed input $X$, we will always have a deterministic estimation of $p_\theta(Y|X)$.

Latent variable models, on the contrary, hypothesise there exist latent variables which are \emph{not directly observable and are
assumed to affect the response variables}. Applied to text generation, the decoder needs to define two distributions: the prior distribution $ p_\theta(z|X)$ of the latent variable $z$ and the likelihood distribution $p_\theta(Y|X,z)$ of the output text. By making it ``deep", $p_\theta(z|X)$ and $p_\theta(Y|X,z)$ are built upon the deep neural network structure to take advantage its modelling power. $Y$ is now dependent not only on the input $X$ but also the stochastical latent variable $z$. To generate text, latent variables are first sampled from $p_\theta(z|X)$, then text are decoded based on $p_\theta(Y|X,z)$. The process is intuitive since languages are not generated out of blue for humans. Before producing a text, we need to think about the structure, topic, style and so on. $z$ is used to stand for these implicit influencing factors. As these factors are latent and no ground-truth supervision is available, the optimization is usually proceeded by marginalizing over $z$ to find the latent distribution that can best explain the observable $X-Y$ training pairs:
\begin{align}
\label{eq: marginal}
p_\theta(Y|X)=\sum_{z}p_\theta(z|X)p_\theta(Y|X,z)
\end{align}

\begin{figure}[ht]
  \centering
  \includegraphics[width=0.7\textwidth]{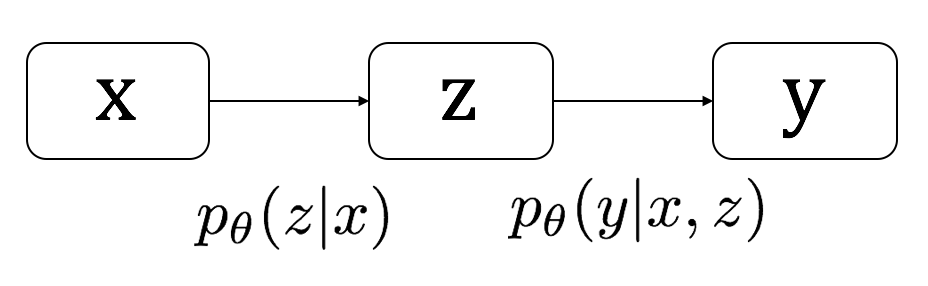}
  \caption{Graphical model of deep latent-variable model. $\theta$ is parameterized by deep neural networks}
  \label{fig:intro_latent}
\end{figure}

The model is then trained still based on the maximum likelihood objective with the density estimated as in Eq.~\ref{eq: marginal}. Figure~\ref{fig:intro_latent} illustrates the graphical model for deep latent-variable models in text generation.

While latent-variable models have a long history in natural language tasks like statistical alignment~\citep{brown1993mathematics} and topic modelling~\citep{blei2003latent}, early research focuses more on the \emph{inference} (i.e., the posterior probability $p_\theta(z|X,Y)$) rather than the \emph{generation} performance (i.e., improving the estimation of $p_\theta(z|X)$ and $p_\theta(Y|X,z)$). To make the inference tractable, independence assumptions (like 0th or 1st order markov) are usually made for the generation process. As a result, they often excel at inferring latent factors but perform rather poorly as generative models~\citep{daume2005induction,angeli2010simple}. By utilizing modern deep neural networks to model the generative distribution, unbounded dependencies across words can be captured so that the  generation performance can be greatly improved. Nonetheless, the power of deep generative models also comes at a cost: latent variables might be ignored given sufficiently powerful generative models, yielding a random inference distribution $p_\theta(z|X,Y)$~\citep{bowman2016generating}. In practice, for deep latent-variable models, there is often a trade-off between the inference and generation, which can be tailored according to specific task requirements~\citep{chen2016variational,pmlr-v80-alemi18a,shen2019select,chang2021time,wiehr2021have}. In this dissertation, as we target at text generation, we will mostly cover techniques improving the generation performance, even if it might sacrifice the inference performance.

Deep neural networks can in theory capture unbounded dependencies to improve the generation, but it comes with the cost: Eq.~\ref{eq: marginal} is usually intractable and no analytical solution exists. Even accurate estimations become extremely expensive since running neural networks sequentially over text is itself very time-consuming. In the next section, we will go over popular strategies to address this optimizing challenge and discuss under which circumstances they should be used.

\section{Contributions and Thesis Outline}
As mentioned above, non-latent encoder-decoder models suffer from the problems of (1) low diversity of generations, (2) poor controllability and interpretability, and (3) requiring large-amount of supervised data. Our main contribution is devising novel latent-variable models within acceptable computational complexity to address the three problems in text generation. Specifically, we propose models to:
\begin{enumerate}
    \item \emph{improve the diversity of text generation} by learning holistic, sentence-level latent representations. By doing so, a latent representation can be first sampled, from which diverse text can be generated. We present effective algorithms to simultaneously train the representation learning and text generation via variational inference. Further, to address the uni-modal and inconsistency limitations of variational inference, we propose a wake-sleep variation (Chapter~\ref{chap: improve-vae}) and mutual information-enhanced (Chapter~\ref{chap: nexus}) training objective. Experiments show they outperform standard variational inference and non-latent-variable models on the dialogue generation tasks.
    \item \emph{improve the controllability and interpretability of text generation} by adding finer-grained, latent specifications on the intermediate generation process. We illustrate using latent variables to stand for word alignment (Chapter~\ref{chap: alignment}), content selection (Chapter~\ref{chap: select}), text segmentation and field-segment correspondences (Chapter~\ref{chap: segment}). We derive efficient training algorithms for them so that the text generation can be explicitly controlled by manipulating the latent variable, which are human interpretable by its definition.
    \item \emph{overcome the sparsity of supervised data} by treating non-parallel text as latent variables. The training can be performed like in the standard EM algorithm which is stable to converge. We apply it on dialogue generation and significantly enrich the generated responses utilizing only non-conversational text (Chapter~\ref{chap: diversify}).
\end{enumerate}

Before delving into detailed works, we first go over popular optimization strategies in Chapter~\ref{chap: optim}, including the cases when exact marginalization is applicable and when variational approximation must be utilized. The pros and cons of different strategies are explained to get a holistic overview of latent-variable models.

Finally, in Chapter~\ref{chap: conclusions}, we draw some conclusions and summarize how latent-variable models can be used to improve text generation.

\section{Publications}
The made contributions mentioned in this dissertation centers around the following publications that includes collaborations of many people. The following list details the authorship, and the papers in which the results were published.

\vspace{1em}\noindent
{[6]} \textit{Improving Variational Encoder-Decoders in Dialogue Generation.}\\
\ul{Xiaoyu Shen}$^*$, Hui Su$^*$, Shuzi Niu, and Vera Demberg.\\
In Proc. of the AAAI Conference on Artificial Intelligence (\textbf{AAAI}), 2018.

The initial idea came after a collaboration work with Hui. After some preliminary discussion, I designed the algorithm for the improved variational encoder-decoder, implemented the system and wrote the final paper. Hui helped evaluate the systems and visualize the T-SNE graph for latent variables. He also constantly provided suggestions through the paper writing. Shuzi and Vera both provided suggestions during our discussions. The details of this work will be discussed at Chapter~\ref{chap: improve-vae}.

\vspace{1em}\noindent
{[5]} \textit{Nexus Network: Connecting the Preceding and the Following.}\\
\ul{Xiaoyu Shen}$^*$, Hui Su$^*$, Wenjie Li, and Dietrich Klakow.\\
In Proc. of the Conference on Empirical Methods in Natural Language Processing (\textbf{EMNLP}), 2018.

Wenjie initializes the idea of utilizing the future utterances in a dialogue. Hui implemented the first version of the system and the evaluation code. Based on it, I designed the algorithm, derived the formulas and drew its connection with the mutual information theory. I also wrote most part of the paper. Dietrich helped check the derivations and proofread the work. The details of this work will be discussed at Chapter~\ref{chap: nexus}.

\vspace{1em}\noindent
{[4]} \textit{Select and Attend: Towards Controllable Content Selection in Text Generation.}\\
\ul{Xiaoyu Shen}, Jun Suzuki, Kentaro Inui, Hui Su, Dietrich Klakow, Satoshi Sekine\\
In Proc. of the Conference on Empirical Methods in Natural Language Processing (\textbf{EMNLP}), 2019.

I designed the algorithms, implemented the whole system and wrote the final paper. Jun suggested using the bpe subword to remove the OOV problem and proofread the paper. Kentaro suggested testing it on the sentence compression task. Satoshi raised the initial idea of the data-to-text task. Hui and Dietrich followed up through discussions. The details of this work will be discussed at Chapter~\ref{chap: select}.

\vspace{1em}\noindent
{[3]} \textit{Improving Latent Alignment in Text Summarization by Generalizing the Pointer Generator.}\\
\ul{Xiaoyu Shen}, Yang Zhao, Hui Su, Dietrich Klakow.\\
In Proc. of the Conference on Empirical Methods in Natural Language Processing (\textbf{EMNLP}), 2019.

I proposed the initial idea, designed the algorithm, implemented the system and wrote the final paper. Yang and Dietrich joined the discussions and provided feedbacks. Hui helped drew the graphs and proofread the paper. The details of this work will be discussed at Chapter~\ref{chap: alignment}.

\vspace{1em}\noindent
{[2]} \textit{Neural Data-to-Text Generation via Jointly Learning the Segmentation and Correspondence.}\\
\ul{Xiaoyu Shen}, Ernie Chang, Hui Su and Dietrich Klakow\\
In Proc. of the Annual Meeting of the Association for Computational Linguistics (\textbf{ACL}), 2020.

I designed the idea, derived the formulas of the marginal likelihood, implemented the system and wrote the final paper. Ernie helped processed the dataset and evaluate the system. Dietrich provided feedbacks, introduced some professionals in the data-to-text area to help answer my questions, and helped proofread the paper. Hui joined the related discussions. The details of this work will be discussed at Chapter~\ref{chap: segment}.

\vspace{1em}\noindent
{[1]} \textit{Diversifying Dialogue Generation with Non-Conversational Text.}\\
Hui Su$^*$, \ul{Xiaoyu Shen}$^*$, Sanqiang Zhao, Xiao Zhou, Pengwei Hu, Rongzhi Zhong, Cheng Niu and Jie Zhou.\\
In Proc. of the Annual Meeting of the Association for Computational Linguistics (\textbf{ACL}), 2020.

I came up with the idea, designed the training algorithms and decide which baseline algorithms we should compare with. I also wrote the final paper. Hui implemented the baseline and our systems, run the whole evaluations. Xiao helped crawl and filter the non-conversational dataset. Sanqiang, Pengwei, Rongzhi, Cheng and Jie joined the discussions and provided feedbacks. The details of this work will be discussed at Chapter~\ref{chap: diversify}.
\cleardoublepage


\chapter{Background on Optimizing Latent-Variable Models}
\label{chap: optim}
\lettrine[lines=3]{T}he main difficulty of optimizing deep latent variable models lies in computing the marginalization in Eq.~\ref{eq: marginal}. When using deep neural networks to parameterize $p_\theta(z|X)$ and $p_\theta(Y|X,z)$, the marginalization is usually intractable. The general idea of optimizing deep latent-variable models can be summarized as follows:
\begin{enumerate}
    \item When computing the marginalization in Eq.~\ref{eq: marginal} is tractable, optimize the exact marginal likelihood by gradient descent. To make it tractable, we can make some necessary simplification and assumptions, which will be discussed in Section~\ref{sec: marg}.
    \item When the marginal likelihood of Eq.~\ref{eq: marginal} is intractable and difficult to simplify, estimate it with variational approximations then optimize over the estimated marginal likelihood. Specifically, when reparameterization tricks~\citep{kingma2014auto} are applicable, we can train the model like variational autoencoders, which will be discussed in Section~\ref{sec: vae}
    \item When reparameterization tricks cannot be easily applied,  the non-differentiability problem exists. We need to resort to soft relaxation, REINFORCE or EM algorithm, which we will cover at Section~\ref{sec: improv}. We will also discuss the potentials and challenges of applying them to text generation. 
\end{enumerate}

\begin{figure}[ht]
  \centering
  \includegraphics[width=\textwidth]{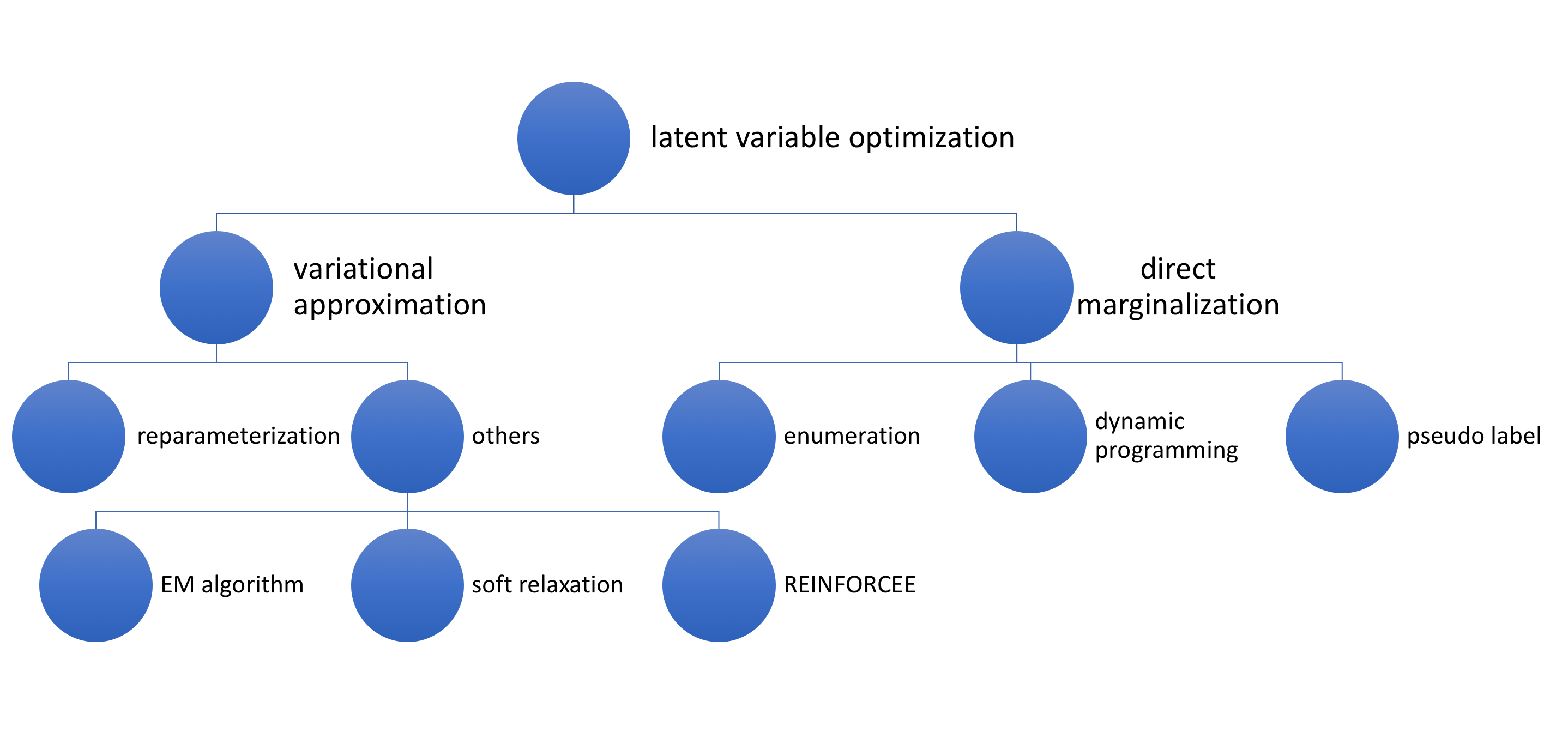}
  \caption{Illustration of optimization technique for latent-variable models}
  \label{fig:optim}
\end{figure}

In practice, one should always try direct marginalization if possible. Though it has been reported that with careful initialization, variational approximations can even outperform exact marginalization by a small margin~\citep{deng2018latent}, variational methods are significantly trickier to train and highly depend on a good initialization and hyperparameter tuning. Figure~\ref{fig:optim} illustrates how we can optimize latent-variable model. In the following section, we will introduce these optimization techniques and explain them with example applications.
\section{Direct Marginalization}
\label{sec: marg}
When the exact marginalization is tractable, the optimization is straightforward. We can perform gradient descent on model parameters $\theta$ to maximize the likelihood of training samples:
\begin{equation}
    \max_{\theta}\mathbb{E}_{X,Y}\log p_\theta(Y|X) = \max_{\theta}\mathbb{E}_{X,Y}\log \sum_{z}p_\theta(Y|X,z)p_\theta(z|X)
\end{equation}
The gradient with respect to $\theta$ is given by:
\begin{equation}
\label{eq: grad_marginal}
\begin{split}
\triangledown_\theta\mathbb{E}_{X,Y}\log p_\theta(Y|X) &= \mathbb{E}_{X,Y}\frac{\triangledown_\theta p_\theta(Y|X)}{p_\theta(Y|X)}\\
&= \mathbb{E}_{X,Y}\sum_{z}\frac{\triangledown_\theta \left(p_\theta(Y|X,z)p_\theta(z|X)\right)}{p_\theta(Y|X)}\\
&=\mathbb{E}_{X,Y}\sum_{z}\frac{p_\theta(Y|X,z)p_\theta(z|X)}{p_\theta(Y|X)}\triangledown_\theta \left(\log (p_\theta(Y|X,z)p_\theta(z|X))\right)\\
&=\mathbb{E}_{X,Y}\mathbb{E}_{z\sim p_\theta(z|X,Y)}\triangledown_\theta (\log( p_\theta(Y|X,z)p_\theta(z|X)))
\end{split}
\end{equation}
The last line of Eq~\ref{eq: grad_marginal} shows its connection with the generalized expectation-maximization (EM) algorithm~\citep{dempster1977maximum,neal1998view}. Maximizing the marginal likelihood is equal to the M-step of the EM algorithm, where $p_\theta(Y|X,z)$ and $p_\theta(z|X)$ are optimized according to the posterior distribution $p_\theta(z|X,Y)$ defined by the current parameter $\theta$~\citep{salakhutdinov2003optimization,sutton2012introduction}. As the maximization step cannot be done analytically for deep neural networks, stochastical gradient descent is performed stepwise as an approximation. By means of the modern automatic differentiation tools for neural networks, we avoid the necessity to calculate the posterior distribution manually. In practice, once the marginal likelihood is computed, we can optimize over it by simply calling the automatic backpropagation function~\citep{eisner2016inside,kim2018tutorial}.
\subsection{Enumeration}
In the simplest case, the exact marginalization is possible by direct enumeration. This usually happens for discrete latent variables following a categorical distribution. A typical example is the pointer generator~\citep{gu2016incorporating,see2017get,merity2016pointer,chang2021selectgen,borgwardt2021logic}.

\begin{figure}[ht]
  \centering
  \includegraphics[width=8cm]{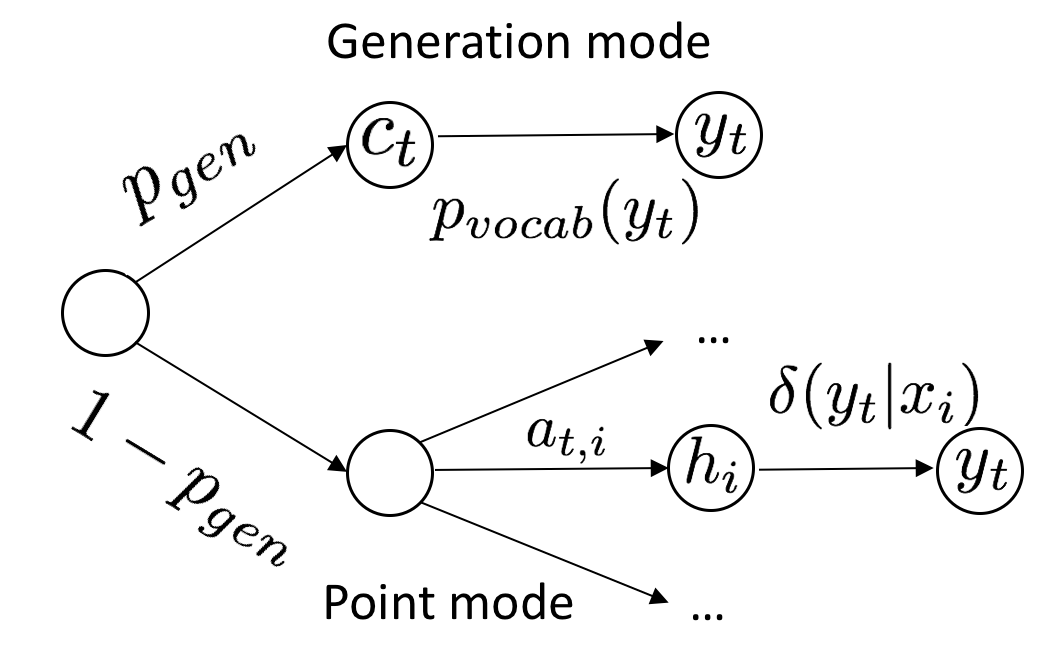}
  \caption{Graphical model of the pointer generator.}
  \label{fig:point_gen}
\end{figure}

Under the pointer generator, the output probability is not simply a softmax distribution over a fixed vocabulary, but further allows the model to directly copy words from the source input. It can be especially beneficial when handling rare or unseen words in the testing stage. At each decoding step $t$, the model first computes a generation probability $p_{gen} \in [0,1]$. $p_{gen}$ is the probability of enabling the generation mode instead of the point mode. In the generation mode, the model computes the probability over the whole vocabulary. In the point mode, the model computes which source word to copy based on the attention distribution $a_t$. The final probability  $p_\theta(y_{t}|X,y_{<t})$ is computed as a combination of the generation probability and the point probability:
\begin{equation*}
\begin{split}
p_{gen} &= \sigma(\text{MLP}_{g}([d_{t}\circ A_{t}]))\\
p_{vocab} &= \text{softmax}(W_1 d_t + W_2 A_t)\\
p_\theta(y_{t}|X,y_{<t}) &= p_{gen}p_{vocab}(y_t)+(1-p_{gen})\sum_{i}a_{t,i}\delta(y_t|x_i)\\
 \delta(y_t|x_i) &= \begin{cases}
    1, & \text{if $y_t=x_i$}.\\
    0, & \text{otherwise}.
  \end{cases}
\end{split}
\end{equation*} 
where $\sigma$ is a sigmoid function and $\text{MLP}_{g}$ is a learnable multi-layer perceptron. $\circ$ denotes vector concatenation. $A_t$ is the attention vector computed as in Eq~\ref{eq: attention}. The graphical model of the pointer generator is depicted in Figure~\ref{fig:point_gen}.

There are two types of latent variables in the pointer generator: a binary variable to decide whether to enable the generation mode or point mode, and a categorical variable $a_{t,i}$ to decide which position to copy suppose the point mode is entered. Since $\delta(y_t|x_i)$ is a simple indicator function, the marginal likelihood can be cheaply computed by enumerating over all paths of latent variables.

Mixture of experts (MoE) model~\citep{quandt1972new,jacobs1991adaptive} is another popular latent-variable model where the exact enumeration is applicable to compute the marginal likelihood. In this case, the latent variable can be considered as a policy network to decide which expert should be assigned to for different inputs. As for text generation, people have tried improving the diversity of generations by sentence-level MoE~\citep{lee2016stochastic,he2018sequence,shen2019mixture} or enhancing the decoder with word-level MoE~\citep{yang2017reference,yang2018breaking,wu2018globaltolocal,wiehr2021have}, both can be easily optimized by enumerating over all possibilities to compute the marginal likelihood.

In some cases, the exact enumeration is expensive, but can be reasonably estimated by enumerating only over the top-$k$ modes. The exact marginalization in Eq.~\ref{eq: marginal} is changed to:
\begin{equation}
\begin{split}
&\max_{\theta}\mathbb{E}_{X,Y}\log \sum_{z}p_\theta(Y|X,z)p_\theta(z|X) \\\approx &\max_{\theta}\mathbb{E}_{X,Y}\log \sum_{z \in \text{TopK}(p_\theta(z|X))}p_\theta(Y|X,z)p_\theta(z|X)    
\end{split}
\end{equation}
This top-$k$ approximation is widely adopted when the latent variable distribution $p_\theta(z|X)$ is  expected to be  sparse and only a  few modes dominate, which is usually the case for discrete latent variables following a categorical distribution, e.g. alignment correspondence in machine translation ~\citep{shankar2018surprisingly,shankar2018posterior}. Since the conditional likelihood $p_\theta(Y|X,z)p_\theta(z|X)$ for $z$ not belonging to the top-k set is expected to be vanishingly small, this usually yields a good approximation of exact marginalization.

Direct enumeration is easy to implement and usually leads to decent performance. However, it is applicable only for very limited scenarios. For continuous latent variables, or discrete variables with complex dependency relations, we have to resort to other methods.
\subsection{Dynamic Programming}
\begin{figure}[ht]
  \centering
  \includegraphics[width=8cm]{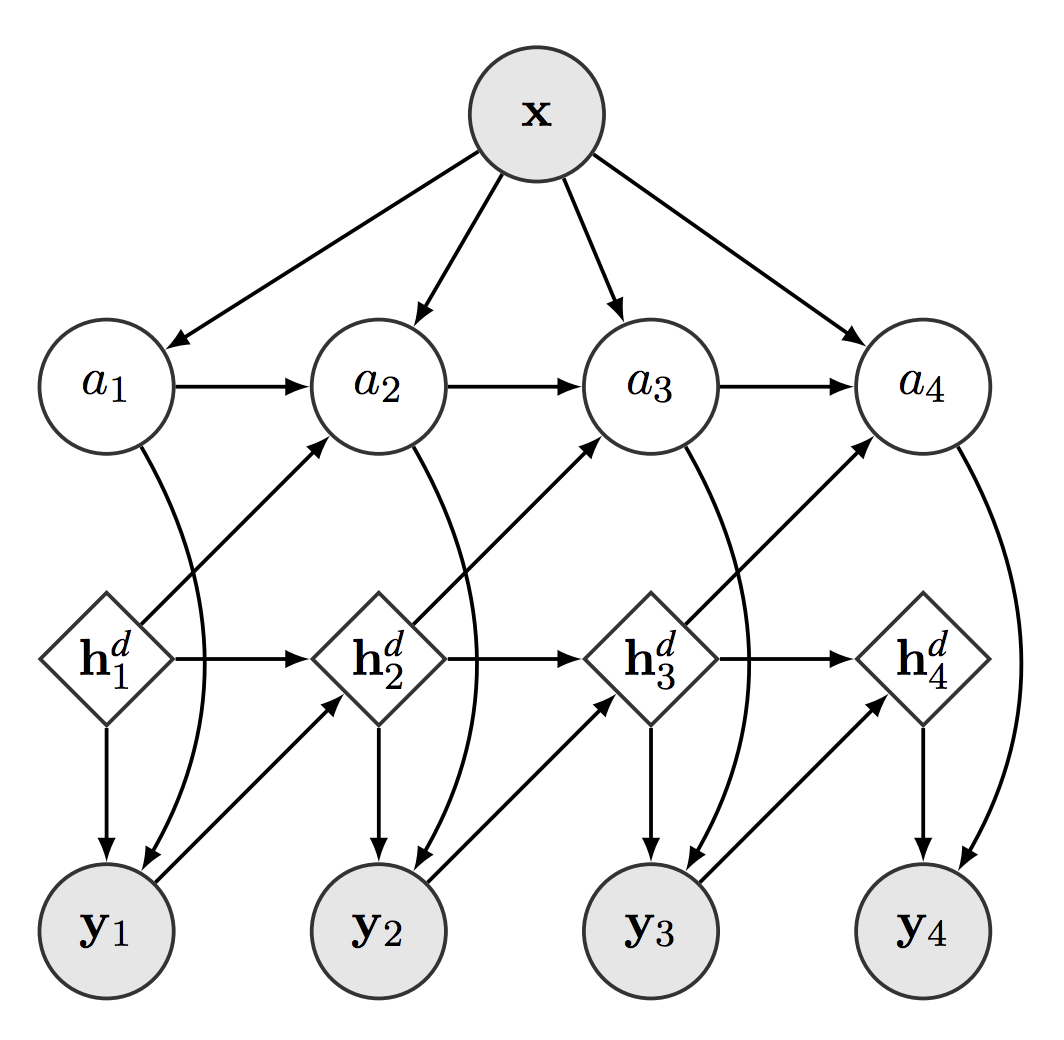}
  \caption{Graphical model of the HMM word alignment model. $h_{t}^{d}$ is the hidden state of the decoder at time $t$ which encapsulates all the history words $y_1,\ldots,y_{t-1}$. Figure taken from \citep{wu2018hard}}
  \label{fig:hmm}
\end{figure}

Sometimes there is structured dependency among latent variables and thereby the computational cost of direct enumeration will be exponentially high. By making appropriate Markov assumptions about the dependency relation, it is possible to efficiently compute the marginal likelihood with dynamic programming under linear complexity. Some classical latent-variable models like hidden markov model, hidden semi-markov model and conditional random field all follow this category. It is straightforward to extend them ``deep" by parameterizing the transition, emission probability and potential functions with deep neural networks. As an example, Figure~\ref{fig:hmm} illustrates the graphical model for the 1st order HMM word alignment model. It has been used in statistical machine translation to induce the word alignment between the source and target language~\citep{och1999improved}.
Nowadays many works incorporate this latent alignment relation into neural encoder-decoder models, yielding comparable or better performance while producing interpretable word alignment. At every time step $t$, the generation process of word $y_t$ is:
\begin{enumerate}
    \item Select a source word $a_t$ to align based on the categorical distribution $p(a_t|h_t,a_{t-1})$, where $a_{t-1}$ is the aligned word at the time stamp of $t-1$ and $h_t$ is the hidden state of the decoder which encapsulates all the history words $y_1,\ldots,y_{t-1}$
    \item Generate word $y_t$ based on the categorical distribution $p(y_t|h_t, a_t)$, where $a_t$ is the aligned source word selected based on  $p(a_t|h_t,a_{t-1})$ and $p(y_t|h_t, a_t)$ is the output distribution of $y_t$ conditioned on the aligned source word and history words $y_1,\ldots,y_{t-1}$.
\end{enumerate}
Here a 1st-order Markov assumption is made for the alignment probability $p(a_t|h_t,a_{t-1})$ where $a_t$ conditions only on the previous aligned position $a_{t-1}$ but irrelevant with $a_1,\ldots, a_{t-2}$. In this case, we can efficiently compute the marginal likelihood with dynamic programming. Specifically, with the forward variable $\beta_{t,i}$ = $p(y_t,a_t=i| h_t)$. We can go through the recursion:
\begin{equation}
\label{eq: forward}
\begin{split}
&\beta_{1,i} = p(y_1|a_1=i, h_1)p(a_1=i)\\
&\beta_{t,i} = p(y_t|a_t=i, h_t)\sum_{j=1}^{m}p(a_t=i|a_{t-1}=j, h_t)\beta_{t-1,j}
\end{split}
\end{equation}
where $h_1$ is the initial hidden state of the decoder, $m$ is the length of the source input and $p(a_1=i)$ is the alignment probability for the first target word.

The recursion goes for $i=1,2,\ldots, n$ where $n$ is the number of the target output. The marginal likelihood is then computed as:
\begin{equation}
    p(y_1,\dots,y_n|x_1,\ldots,x_m) = \sum_{i=1}^{m}\beta_{n,i}
\end{equation}
The recursion in dynamic programming often leads to numerical precision problems, it is therefore necessary to define all the operations in the log space~\citep{kim2017structured}. Note that the Markov assumption is only made for the dependency relation among latent variables $a_1,\ldots,a_n$. The probability $p(a_t|a_{t-1},h_t)$ and $p(y_t|a_t,h_t)$ can condition on all the history words $y_1,\ldots, y_{t-1}$ which are not latent. Therefore, the long-term dependency can usually be implicitly captured through the observable words in the target side and the performance will not significantly drop due to the simplified Markov assumption, as indicated by many works in machine translation~\citep{shankar2018posterior}, character transduction~\citep{wu2018hard} and summarization~\citep{yu2016online}. It is especially useful when inductive bias, e.g., monotonic or proximity bias, needs to be injected to the latent alignment~\citep{yu2016online,shankar2018posterior}. We can easily specify these bias by adding constraints to $p(a_t|a_{t-1}, h_t)$.

Dynamic programming allows us to specify Markov dependency between latent variables yet still permits tractable inference. However, it assumes the generation of each target word conditions only on some specific part of the source input and the current latent variable only conditions on the immediate previous one, which is not the case for many applications.
\subsection{Pseudo Label}
If the dependency among latent variables and the target words are too complex to model and hard to simplify, the marginal likelihood might be intractable even with dynamic programming. For example, we might assume before generating the target output, we should first generate an outline skeleton, syntactic tree or other structured predictions. In these cases, the output needs to wait until the full intermediate predictions are produced, the dynamic programming cannot be easily applied~\footnote{We can still assume the Markov dependency when generating the intermediate prediction. However, this is rarely true as the intermediate prediction is highly structured. Making such simplification will affect the generation performance~\citep{wiseman2018learning}.}. Furthermore, since the intermediate prediction is highly structured, top-k sampling can easily lead to very inaccurate estimations and thus direct enumeration is challenging.

\begin{figure}[ht]
  \centering
  \includegraphics[width=\textwidth]{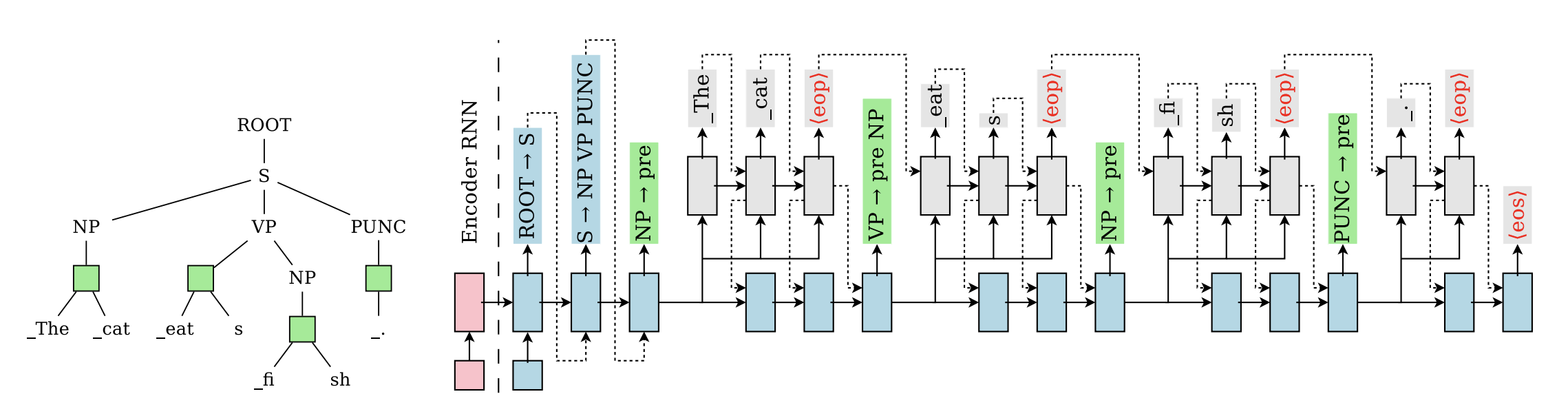}
  \caption{Illustration of the tree decoder for machine translation. Before translating to the target language, a syntactic tree is first generated, based on which the final output is produced from a tree-decoder. Figure taken from \citep{wang2018tree}. }
  \label{fig:tree_decode}
\end{figure}

Suppose we have some oracle, coming either from heuristics, human annotations or an external parser with reasonable accuracy, we can usually treat the oracle as pseudo labels to train the latent variables in a distantly supervised way~\footnote{In the extreme case where the oracle is 100\% accurate, $z$ becomes not ``latent" any more, it basically turns into a fully visible model.}. Formally, the marginal likelihood becomes:
\begin{equation}
    p(y_1,\dots,y_n|x_1,\ldots,x_m) = p(y_1,\dots,y_n|x_1,\ldots,x_m, z = z_{o})p(z=z_o|x_1,\ldots,x_m)
\end{equation}
where $z_o$ is the pseudo label obtained from the oracle. Figure~\ref{fig:tree_decode} provides an example from \citep{wang2018tree}, where $z$ stands for the syntactic tree which is latent. An external parser is utilized to extract the tree structure for each target language sentence. The extracted tree is considered as the pseudo label to train the latent variable. If the external parser can reach a certain degree of accuracy, training the model in this way can usually lead to remarkably good performance. 

Similarly, we can also use z to stand for dependency trees~\citep{elder2019designing}, content planning~\citep{puduppully2019data}, aspects and skeletons~\citep{li2019generating}. In the case where the top-1 pesudo label produced from the oracle might not be accurate enough, it is also possible to use the top-k results produced from the oracle to reduce the variance. Specifically, the marginal likelihood is computed as:
\begin{equation}
\begin{split}
    &p(y_1,\dots,y_n|x_1,\ldots,x_m) \\= \sum_{z_o\in TopK(oracle)}&p(y_1,\dots,y_n|x_1,\ldots,x_m, z = z_{o})p(z=z_o|x_1,\ldots,x_m)
\end{split}
\end{equation}
It is similar to the top-k approximation in the direct enumeration section. However, instead of getting the top-k from the model's own prediction, it derives the top-k from an external oracle.

The method of distant supervision is highly effective when an oracle with reasonable accuracy exists. It essentially brings external knowledge from the oracle into the model so that we do not need to learn the latent variables from scratch. However, assuming the existence of such an oracle basically turns all latent variables into semi-explicit (since now all latent variables are clearly defined through the oracle). We need to know in advance what the latent variables are and how it affects the generation process. The construction of the oracle needs to be carefully handcrafted instead of being learnt automatically. In this sense, it has no big difference from the traditional pipeline model where each intermediate process is manually specified. Due to this limit, a typical approach is to not solely rely on the oracle, but rather only use it as a way to warm up the latent-variable model. After having a decent start, then the model can be further trained with other optimization techniques for a better performance~\cite{shen2019select}.
\section{Variational Approximation}
\label{sec: vae}
When the marginal likelihood is difficult to be computed directly, variational approximations~\citep{kingma2014auto} are a popular strategy to estimate it by projecting the posterior distribution to a tractable parameterized distribution. Recall in Eq.~\ref{eq: grad_marginal} that running gradient descent over the marginal likelihood equals optimizing over the posterior distribution $p_{\theta}(z|X,Y)$. In the case when the marginal likelihood is intractable, $p_{\theta}(z|X,Y)$ will also avoid any analytical expression because the partition function basically requires computing the marginal likelihood a priori. The main idea of variational approximation is that instead of using the real posterior distribution $p_{\theta}(z|X,Y)$, we opt for a surrogate $q_{\phi}(z|X,Y)$ with a \emph{tractable probability density}. The surrogate comes from a distribution family which is much simpler than the real posterior. Therefore, efficiently sampling over the surrogate to estimate the marginal likelihood is possible. In the meantime, to make sure the estimation is within some reasonable error bound, we need to add a regularization which pushes the surrogate to stay close to the real posterior distribution. The commonly chosen regularization imposes a KL-divergence punishment $KL(q_\phi(z|X,Y)||p_\theta(z|X,Y))$. When we use the KL-divergence regularization, it will lead to the well-known evidence lower bound (ELBO)~\citep{jordan1999introduction}: a lower bound of the likelihood in Eq.~\ref{eq: marginal}. 

\begin{figure}[htbp!] 
\centering    
\includegraphics[width=0.5\textwidth]{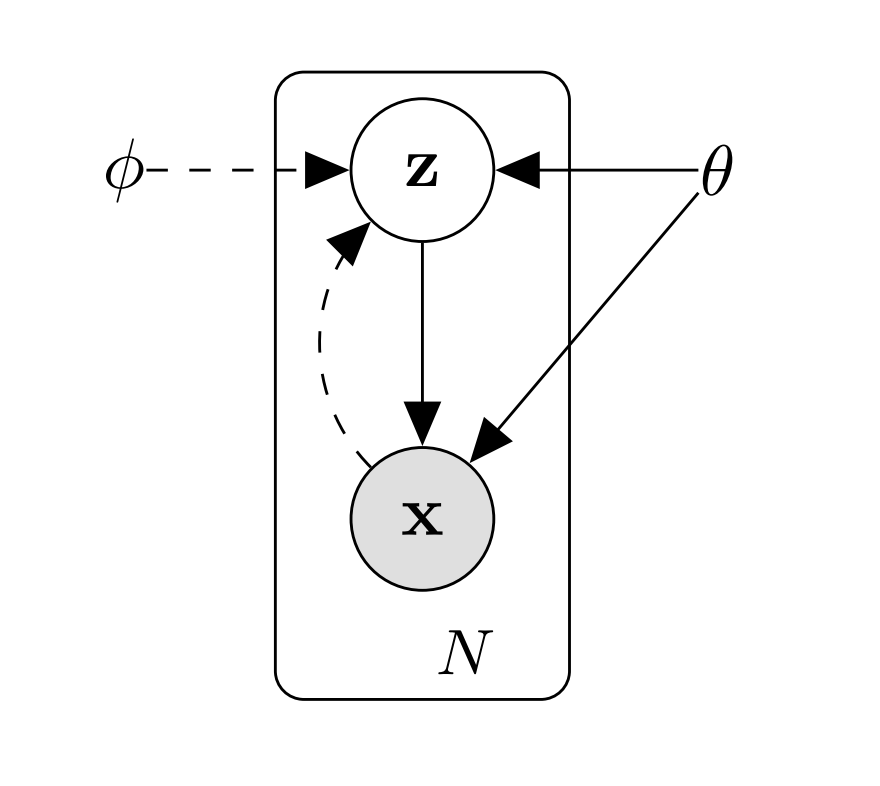}
\caption{Graphical Model of unconditional Latent-variable Model}
\label{fig: vae}
\end{figure}

In this section, for notational simplicity, we assume an unconditional generation process: In the first step, a latent variable $z$ is sampled from a prior distribution $p_\theta(z)$. In the second step, the datapoint $x$ is generated according to the conditional probability $p_\theta(x|z)$. Both distributions are parametrised by $\theta$ and are differentiable with respect to $\theta$ and $z$ almost everywhere. The posterior distribution of $z$ would be $p_\theta(z|x)$. It can be easily extended to the conditional generation process where both $p_\theta(z)$ and $p_\theta(x|z)$ are conditioned on an additional context, like the case in Eq.~\ref{eq: marginal}. The graphical model of an unconditional latent-variable model is illustrated in Figure~\ref{fig: vae}. 
\subsection{Evidence Lower Bound}
To understand how we end up with the ELBO as an approximation of the likelihood, we start from the simplest intuition to estimate the marginal likelihood $\mathbb{E}_{p_\theta(z)}p_\theta(x|z)$ by Monte Carlo sampling:
\begin{align}
\label{eq: direct_monto}
\begin{split}
\log p_\theta(x)&=\log \mathbb{E}_{p_\theta(z)}p_\theta(x|z)\geq \mathbb{E}_{p_\theta(z)} \log p_\theta(x|z) \\&\approx\frac{1}{N}\sum_{i=1}^{N}\log p_\theta(x|z_i;z_i\sim p_\theta(z))
\end{split}
\end{align}

The first line applies the Jensen's equality since the logarithm function is concave. We can then run Monte-Carlo estimation over the logarithm space to avoid underflow. The above equation will lead to a lower bound of the real likelihood too. However, performing this kind of sampling for every training step is too costly. We need a huge amount of samples to get a reasonable estimation because most sampled $z$s from the above equation do not contribute to significant values of $p_\theta(x|z)$.

The evidence lower bound (ELBO) addresses this issue by introducing a proposal $q_\phi(x|z)$ to approximate the true posterior distribution $p_\theta(x|z)$ more accurately:
\begin{align}
\label{eq: elbo}
\begin{split}
\mathbb{E}_{q(x)}[\mathbb{E}_{q_\phi(z|x)}\log \frac{p_\theta(x|z)p_\theta(z)}{q_\phi(z|x)}]&=\mathbb{E}_{q(x)}[\mathbb{E}_{q_\phi(z|x)}\log p_\theta(x|z)-KL(q_\phi(z|x)||p_\theta(z))]\\
&=\mathbb{E}_{q(x)}\log p_\theta(x)-\mathbb{E}_{q(x)}KL(q_\phi(z|x)||p_\theta(z|x))\\
&\leq \mathbb{E}_{q(x)}\log p_\theta(x)
\end{split}
\end{align}
ELBO is a lower bound of the real likelihood $\mathbb{E}_{q(x)}\log p_\theta(x)$. It essentially applies importance sampling over the proposal $q_\phi(x|z)$ to get a tighter lower bound compared with Eq.~\ref{eq: direct_monto}. Notably, as seen in the second line, by maximising the ELBO, we are maximizing the marginal likelihood and minimizing $KL(q_\phi(z|x)||p_\theta(z|x))$ at the same time. This is a nice property. As the training proceeds, the KL-regularization will push the proposal $q_\phi(z|x)$ to be closer and closer to the real posterior distribution $p_\theta(z|x)$. It will lead to a tighter bound of the marginal likelihood and benefit the learning of generative parameters $\theta$ in turn.

Now we get rid of the intractable term $p_\theta(x)$. Though computing the expectation over $q_\phi(z|x)$ is still intractable, we can estimate it by sampling. Unlike the prior $p_\theta(z)$, sampling from the posterior $q_\phi(z|x)$ is more efficient as it is more peaked and contains useful information of the data $x$. 

Eq.~\ref{eq: elbo} can effectively approximate the likelihood. However, the ELBO objective takes the expectation over $q_\phi(z|x)$. $\theta$ is easy to be optimized when $\phi$ is known, but the optimization of $\phi$ is nontrivial because when $\phi$ is parametrised by neural networks, it is impossible to backpropagate the error to $\phi$ through sampled variables.
\citep{hinton1995wake} proposed training the ELBO objective through a separate wake-sleep alternation.
In the wake phase, the objective is:
\begin{align}
max_{\theta}\mathbb{E}_{q(x)}[\mathbb{E}_{q_\phi(z|x)}\log p_\theta(x|z)-KL(q_\phi(z|x)||p_\theta(z))]
\end{align}
which is the same as ELBO except that the inference parameter $\phi$ is fixed. $z$ is sampled according the fixed distribution $q_\phi(z|x)$ and $\theta$ is optimized to reconstruct the original $x$. Since the sampling distribution is fixed, gradient descent can be applied.

In the sleep phase, the objective is:
\begin{align}
max_{\phi}\mathbb{E}_{q(x)}[-KL(p_\theta(z|x)||q_\phi(z|x))]
\end{align}
The generating parameter $\theta$ is fixed. A ``dream" sample is obtained from the generative network by ancestral sampling from $p_\theta(z|x)$ and is used as a target for the maximum likelihood training of the inference network. Again, expectation is taken over a fixed distribution $p_\theta(z|x)$, so gradient descent can be used.

The updating of $\theta$ follows the correct gradient as it uses the ELBO objective. However, $\phi$ is optimized to minimize the reversed KL divergence (As can be seen in the second line of Equation \ref{eq: elbo}, the original EBLO aims at minimizing $KL(q_\phi(z|x)||p_\theta(z|x))$  while the sleep phase minimizes $KL(p_\theta(z|x)||q_\phi(z|x))$. Although both can lead to the same global optimum, in practice this might lead to unexpected model behaviours. Besides, the alternative training mechanism can be easily stuck in a local optimum since errors from last stages will be reinforced in later stages.

Another way to train with ELBO is to directly estimate the gradient with respect to $\phi$ according to:
\begin{align}
\label{eq:vae-direct}
\begin{split}
\triangledown_{\phi}\mathbb{E}_{q_\phi(z|x)}\log p_\theta(z|x)&=\mathbb{E}_{q_\phi(z|x)}\log p_\theta(x|z)\triangledown_{\phi}\log(q_\phi(z|x))\\
&\approx \frac{1}{N}\sum_{i=1}^{N}\log p_\theta(x|z_i)\triangledown_{\phi}\log(q_\phi(z_i|x)); z_i\sim q_\phi(z|x)
\end{split}
\end{align}
\begin{align}
\label{eq: kl}
\begin{split}
&\triangledown_{\phi}KL(q_\phi(z|x)||p_\theta(z))\biggr\rvert_{\phi=\phi_0}\\=&\triangledown_{\phi}\mathbb{E}_{q_\phi(z|x)}\log \frac{q_{\phi_0}(z|x)}{p_\theta(z)}+\triangledown_{\phi}KL(q_\phi(z|x)||q_{\phi_0}(z|x))\biggr\rvert_{\phi=\phi_0}\\
=&\triangledown_{\phi}\mathbb{E}_{q_\phi(z|x)}\log \frac{q_{\phi_0}(z|x)}{p_\theta(z)}\biggr\rvert_{\phi=\phi_0}\\
=&\mathbb{E}_{q_\phi(z|x)}\log \frac{q_{\phi_0}(z|x)}{p_\theta(z)}\triangledown_{\phi}\log(q_\phi(z|x))\biggr\rvert_{\phi=\phi_0}\\
\approx &\frac{1}{N}\sum_{i=1}^{N}\log \frac{q_{\phi_0}(z|x)}{p_\theta(z)}\triangledown_{\phi}\log(q_\phi(z_i|x)); z_i\sim q_\phi(z|x)\biggr\rvert_{\phi=\phi_0}
\end{split}
\end{align}
The second line in Equation \ref{eq: kl} comes from the fact that $\phi_0$ is a local minimum in $KL(q_\phi(z|x)||q_{\phi_0}(z|x))$ thus the gradient with respect to $\phi$ is 0.  Equation \ref{eq: kl} is needed when KL divergence cannot be analytically computed. Estimating the gradient by this kind of sampling suffers from a very high variance and is not practical way of training the ELBO objective~\citep{paisley2012variational}.
\subsection{Reparameterization trick}
The main reason of Eq.~\ref{eq:vae-direct}'s high variance is that it treats $q_\phi(z|x)$ as a black box. Random samples from $q_\phi(z|x)$ are the only way of getting access to the distribution. To utilize the information from not only random samples but also the parameterization format and distribution family of $q_\phi(z|x)$, \citep{kingma2014auto,rezende2014stochastic} proposed a reparameterization trick that allows for efficient estimation of the ELBO and its gradient. When $q_\phi(z|x)$ belongs to the location scale family like a Gaussian distribution, we can reparameterize the random variable $\tilde{z}\sim q_\phi(z|x)$ by a deterministic transformation $g_\phi(x,\epsilon)$ with respect to a noise variable $\epsilon$. For example, if $q_\phi(z|x)$ defines a Gaussian distribution $\mathcal{N}(\mu,\sigma)$, we can  reparameterize $z$ as $\mu+\sigma \epsilon, \epsilon\sim \mathcal{N}(0,1)$. Then the ELBO becomes:
\begin{align}
\begin{split}
&\mathbb{E}_{q(x)}[\mathbb{E}_{q_\phi(z|x)}\log p_\theta(x|z)-KL(q_\phi(z|x)||p_\theta(z))]\\
=&\mathbb{E}_{q(x)}\mathbb{E}_{p(\epsilon)}[\log p_\theta(x,g_\phi(x,\epsilon))-\log q_\phi(g_\phi(x,\epsilon)|x)]\\
\approx&\frac{1}{MN}\sum_{i=1}^{M}\sum_{j=1}^{N}\log p_\theta(x^{i}|z^{(i,j)})p_\theta(z^{(i,j)})-\log q_\phi({z^{(i,j)}|x^{i}})\\
&z^{(i,j)}=g_\phi(\epsilon^{(i,j)},x^{i}), \qquad \epsilon^{(i,j)}\sim p(\epsilon)
\end{split}
\end{align}
When $KL(q_\phi(z|x)||p_\theta(z))$ can be analytically computed, the ELBO can be estimated via a simpler form:
\begin{align}
\label{eq: re-trick}
\begin{split}
ELBO&=\mathbb{E}_{q(x)}[\mathbb{E}_{p(\epsilon)}\log p_\theta(x|g_\phi(x,\epsilon))-KL(q_\phi(z|x)||p_\theta(z)]\\
&\approx \frac{1}{MN}\sum_{i=1}^{M}[\sum_{j=1}^{N}\log p_\theta(x^{i}|z^{(i,j)})-KL(q_\phi(z|x^{(i)})||p_\theta(z))]\\
&z^{(i,j)}=g_\phi(\epsilon^{(i,j)},x^{i}), \qquad \epsilon^{(i,j)}\sim p(\epsilon)
\end{split}
\end{align}
$M$ is the batch size and $N$ is the sample size. \citep{kingma2014auto} shows that when $M$ is large ($M\geq100$), setting $N=1$  is enough. Equation \ref{eq: re-trick} typically has less variance and the derivative with respect to both $\theta$ and $\phi$ can be efficiently estimated. We can use common gradient descent methods for the optimization. In practice, for the sake of training efficiency, $q_\phi(z|x)$ and $p_\theta(z)$ are normally modelled by Gaussian distribution with diagonal covariance matrces, then we can compute the KL divergence via:
\begin{align}
\begin{split}
&KL(q_\phi(z|x)||p_\theta(z))=\frac{1}{2}[\log \frac{|\Sigma_2|}{|\Sigma_1|}-d+tr(\Sigma_2^{-1}\Sigma_1)+(\mu_2-\mu_1)^T\Sigma_2^{-1}(\mu_2-\mu_1)]\\
&q_\phi(z|x)\sim\mathcal{N}(\mu_1,\Sigma_1), p_\theta(z)\sim \mathcal{N}(\mu_2,\Sigma_2)
\end{split}
\end{align}

$d$ is the dimension size of $z$. It can be viewed as a regularised denoising autoencoder. As shown in Equation \ref{eq: re-trick}, the first term is an autoencoder with random noise added on the representation layer, the second term regularises the encoder distribution $q_\phi(z|x)$ to keep it stay close to the prior distribution $p_\theta(z)$.

\subsection{Requirement of Applying Variational Apprroximation}
In summary, variational approximation is a powerful framework to efficiently train both the generative and inference parameters of latent-variable models. To apply this framework, we need to make sure:
\begin{itemize}
\item The latent variable $z$ is continuous. The generative parameter $\theta$ and the inference parameter $\phi$ are differentiable almost everywhere.
\item The inference distribution $q_\phi(z|x)$ belongs to, or can be efficiently approximated by a distribution of, the family that the reparameterization trick can be applied.
\item The probability density of the distribution $q_\phi(z|x)$, $p_\theta(x|z)$, $p_\theta(z)$ and, in the best case, $KL(q_\phi(z|x)||p_\theta(z))$, are all explicitly computable.
\end{itemize}
\section{Improvement of Variational Approximation}
The variational approximation is an effective training framework for latent-variable models and converges to the global optimum when $q_\phi(z|x)$, $p_\theta(x|z)$ and $p_\theta(z)$ have infinite capacity. In practice, as explained above, $q_\phi(z|x)$ and $p_\theta(z)$ are usually modelled with mean-field Gaussian distributions for training efficiency, which inevitably limits its expressiveness to accurately describe the data distribution.
\subsection{Limitation of Variational Approximation}
\label{sec: op-vae}
From the ELBO objective, we have the following equations:
\begin{align}
\label{eq: opvae}
\begin{split}
ELBO&=\mathbb{E}_{q(x)}[\mathbb{E}_{q_\phi(z|x)}\log p_\theta(x|z)-KL(q_\phi(z|x)||p_\theta(z))]\\
&=\mathbb{E}_{q(x)}\log p_\theta(x)-\mathbb{E}_{q(x)}KL(q_\phi(z|x)||p_\theta(z|x))\\
&=-H(x)-\mathbb{E}_{q_\phi(z)}KL(q_\phi(x|z)||p_\theta(x|z))-KL(q_\phi(z)||p_\theta(z))\\
&=-H(x)-\mathbb{E}_{q(x)}KL(q_\phi(z|x)||p_\theta(z|x))-KL(q(x)||p_\theta(x))\\
q_\phi(x|z)&=\frac{q_\phi(z|x)q(x)}{q_\phi(z)}, q_\phi(z)=\mathbb{E}_{q_\phi(z|x)}q(x),p_\theta(z|x)=\frac{p_\theta(x|z)p_\theta(z)}{p_\theta(x)}\\p_\theta(x)&=\mathbb{E}_{p_\theta(x|z)}p_\theta(z)
\end{split}
\end{align}
It is easy to see (from the last line of ELBO) that in the global optimum, we have $q_\phi(z|x)=p_\theta(z|x)$ and $q(x)=p_\theta(x)$ assuming parameters $\theta$ and $\phi$ have infinite capacity. $\theta$ can accurately describe the generating distribution and $\phi$ can accurately infer the real posterior given the generative process defined by $\theta$. 

Now we consider the case when $q_\phi(z|x)$ cannot infer the exact posterior distribution $p_\theta(z|x)$, there is a gap of $\mathbb{E}_{q(x)}KL(q_\phi(z|x)||p_\theta(z|x))$ between the real likelihood and the ELBO objective function we use. $p_\theta(x)$ also has problems with matching the real distribution $q(x)$ when this gap exists. From the third line in Equation \ref{eq: opvae} we can see our objective is to have $p_\theta(z)=q_\phi(z)$ and $p_\theta(x|z)=q_\phi(x|z)$ in the global optimum. Even if we make  $p_\theta(x|z)$ powerful enough such that the second equality is reachable and we have the optimal $p^*_\theta(x|z)=q_\phi(x|z)$, the first equality still cannot be satisfied when $q_\phi(z|x)$ cannot match $p_\theta(z|x)$, because if we have $q_\phi(z)=p_\theta(z)$, then $q_\phi(z|x)\sim q_\phi(x|z)q_\phi(z)=p^*_\theta(x|z)p_\theta(z)$ and we can have $q_\phi(z|x)=p^*_\theta(z|x)$. When $p_\theta(z)\neq q_\phi(z)$, $p^*_\theta(x)=\mathbb{E}_{p^*_\theta(z)}p^*_\theta(x|z)=\mathbb{E}_{p^*_\theta(z)}q_\phi(x|z)\neq \mathbb{E}_{q_\phi(z)}q_\phi(x|z)=q(x)$. Namely, $p_\theta(x)$ does not converge to $q(x)$ even in the global optimum. Therefore, making $q_\phi(z|x)=p_\theta(z|x)$ is important not only for the inference task but also for the generating task if we hope $p_\theta(x)$ converges to the real $q(x)$ in the global optimum. Though in theory we can make $q_\phi(z|x)=p_\theta(z|x)$ by having either a flexible $q_\phi(z|x)$ or $p_\theta(z)$, normally we prefer a more powerful $q_\phi(z|x)$ to approximate the real posterior rather than adjusting the generating distribution to match a limited posterior distribution $q_\phi(z|x)$.

Apart from $q_\phi(z|x)$ and $p_\theta(z)$, a weak $p_\theta(x|z)$ is also problematic for modelling the real distribution of data. For example, in the task of image generation, a common choice is to set $p_\theta(x|z)\sim \mathcal{N}(g_\theta(z),I/2)$, which is a very weak assumption because we are trying to map the posterior distributions of any data point to a fixed variance factored Gaussian family. In this case we have:
\begin{align}
\begin{split}
\label{eq: square-vae}
ELBO=\mathbb{E}_{q(x)}[\mathbb{E}_{q_\phi(z|x)}-||g_\theta(z)-x||^2-KL(q_\phi(z|x)||p_\theta(z))]
\end{split}
\end{align}
which can be seen as a regularised noisy autoencoder with L2 loss. We further have the following proposition~\citep{zhao2017towards}:
\begin{align}
\begin{split}
\text{The optimal solu}&\text{tion to Equation \ref{eq: square-vae} given } \phi \text{ is:}\\
&g_\theta(z)=\mathbb{E}_{q_\phi(z|x)}x
\end{split}
\end{align}

This means unless $q_\phi(z|x)$ is a lossless mapping, $g_\theta(z)$ will always tend to average over all possible data $x$ given a specific latent variable $z$, which is the reason that the VAE often generates realistic but fuzzy natural images. To better model the generating process, a more flexible likelihood distribution $p_\theta(x|z)$ is needed.
Nonetheless, $p_\theta(x|z)$ is normally less a focus than $q_\phi(z|x)$ and $p_\theta(z)$ as its distribution is relatively easy. Besides, $p_\theta(x|z)$ is essentially a fully visible model if we condition on a known $z$ value, then we can use any neural decoders to model it. \textbf{The most difficult thing is to have $q_\phi(z|x)=p_\theta(z|x)$ and most current efforts are concentrated on devising a more flexible, yet still efficient, $q_\phi(z|x)$ to approximate $p_\theta(z|x)$}. In the following section, we would briefly introduce common techniques of improving the flexibility of $q_\phi(z|x)$.

\subsection{Auxiliary Variable}
The auxiliary variable VAE~\citep{salimans2015markov} improves $q_\phi(z|x)$ by imposing a set of auxiliary variables $y=z_0, z_1,..., z_{T-1}$. It introduces a stochastic Markov chain $q_\phi(y,z_T|x)=q_\phi(z_0|x)\prod_{i=1}^{T}q_\phi(z_i|z_{i-1},x)$ and each distribution is modelled by a mean-field Gaussian distribution. Now we have the marginal distribution $q_\phi(z_T|x)=\int_{y}q_\phi(y,z_T|x)$, which is a very rich distribution family and can be used as a close fit to the real posterior distribution $p_\theta(z|x)$. However, the probability density of $q_\phi(z_T|x)$ is intractable. To make the training work we need to define another auxiliary inference distribution $r(y|z_T,x)=\prod_{i=1}^{T}r(z_{i-1}|x,z_i)$. The objective is:
\begin{align}
\begin{split}
\mathcal{L}_{aux}&=\mathbb{E}_{q(x)}\mathbb{E}_{q_\phi(y,z_T|x)}[\log p_\theta(z_T)p_\theta(x|z_T)r(y|z_T,x)-\log q_\phi(y,z_T|x)]\\
&=\mathbb{E}_{q(x)}\log p_\theta(x)-\mathbb{E}_{q(x)}KL(q_\phi(z_T|x)||p_\theta(z_T|x))\\&-\mathbb{E}_{q(x)} \mathbb{E}_{q_\phi(z_T|x)} KL(q_\phi(y|z_T,x)||r(y|z_T,x))\\
&=ELBO-\mathbb{E}_{q(x)} \mathbb{E}_{q_\phi(z_T|x)} KL(q_\phi(y|z_T,x)||r(y|z_T,x))\\
&\leq ELBO\\
&\leq \mathbb{E}_{q(x)}\log p_\theta(x)
\end{split}
\end{align}
Though $q_\phi(z_T|x)$ is theoretically rich, the inference needs a chain of sampling from $q_\phi(z_t|x,z_{t-1})$ which does not allow parallel computation. What's more, because of the additional auxiliary variables, the objective function is a less tight lower bound than ELBO, we are losing more accuracy when maximising the real likelihood.

\subsection{Normalising Flow}
Normalising flow~\citep{dinh2014nice,rezende2015variational} also defines a chain of $z_0, z_1,..., z_{T}$ such that $q_\phi(z_T|x)$ has a much broader distribution family than $q_\phi(z_0|x)$. The difference is that it requires the transformation chain to be \textbf{invertible}. Let $z_i=f_i(z_{i-1},x)$ and $z_0\sim q_\phi(z_0|x)$, then we can derive $q_\phi(z_T|x)$ by using the Jacobian matrices as bellow:
\begin{align}
\begin{split}
\log (q_\phi(z_T|x))=\log (q_\phi(z_0|x))-\sum_{i=1}^T\log det| \frac{\partial f_i}{\partial z_{i-1}}| 
\end{split}
\end{align}
With the known probability density, we can compute the ELBO objective by:
\begin{align}
\begin{split}
\label{eq: invertible}
ELBO&=\mathbb{E}_{q(x)}\mathbb{E}_{q_\phi(z_T|x)}[\log p_\theta(z_T)p_\theta(x|z_T)-\log q_\phi(z_T|x)]\\
&=\mathbb{E}_{q(x)}\mathbb{E}_{q_\phi(z_0|x)}[\log p_\theta(z_0)p_\theta(x|z_T)-\log q_\phi(z_0|x)-2\sum_{i=1}^T\log det| \frac{\partial f_i}{\partial z_{i-1}}| ]\\
&z_T=f_T\circ f_{T-1}\cdots f_1(z_0)
\end{split}
\end{align}
where $\circ$ means the nested function which recursively applies to the variable. $f_i$ is usually defined such that the determinant of the corresponding Jacobian matrix can be efficiently computed. For example, ~\citep{dinh2014nice} proposed a transformation that clipping the whole dimension of $z$ into halves:
\begin{align}
\begin{split}
z_t=(z_t^\alpha,z_t^\beta)=f_t(z_{t-1}^\alpha,z_{t-1}^\beta)=(z_{t-1}^\alpha,z_{t-1}^\beta+m_t(x,z_{t-1}^\alpha))
\end{split}
\end{align}
$m_t$ can be an arbitrarily complex function. Since $f_i$ has a lower triangular Jacobian with all diagonal terms as 1 and the determinant of a lower triangular matrix equals the product of the diagonal terms, we have $det|\frac{\partial f_i}{\partial z_{i-1}}|=1$ for $i=1,2,...,T$.

In ~\citep{rezende2015variational}, the author proposed the planar normalising flow and radial normalising flow in which the Jacobian determinant can also be  efficiently computed and brings more flexibility to the approximated posterior distribution.

In contrast with auxiliary variable VAEs, the normalising flow methods limits the power of transformation by restricting it to be invertible, the resulting distribution is in consequence less flexible, but the computable probability density makes the objective a tighter lower bound, so it is basically deriving a less accurate solution for a more accurate objective.

\subsection{Gaussian Autoregressive Flow}
Under the framework of the normalising flow, the Gaussian Autoregressive Flow~\citep{kingma2016improving} defines a very powerful invertible transformation where each dimension of the new variable is dependent on all the previous dimensions. The transformation function is:
\begin{align}
\begin{split}
\label{eq: autoregressive}
&z_{t,0}=\mu_{t,0}+\sigma_{t,0}z_{t-1,0}\\
z_{t,i}=\mu_{t,i}(z_{t,0:i-1}&)+\sigma_{t,i}(z_{t,0:i-1})z_{t-1,i}, i=1,2,...D
\end{split}
\end{align}
$\mu_{t,i}$ and $\sigma_{t,i}$ can be parametrised by neural networks. Since each dimension is only dependent on the previous ones, we have a lower triangular Jacobian. The determinant of the Jacobian matrix can be easily computed by:
\begin{align}
\begin{split}
\label{eq: jac-autoregressive}
det|\frac{\partial f_i}{\partial z_{i-1}}|=\prod_{i=1}^D \sigma_{t,i}(z_{t,0:i-1})
\end{split}
\end{align}
Then it can be optimized with the objective specified in Equation \ref{eq: invertible}. Though powerful, the transformation needs a sequential transformation process as specified in Equation \ref{eq: autoregressive}, the cost of the transformation process is $\mathcal{O}(DT)$ and grows linearly as the dimension size. As a result, Gaussian Autoregressive Flow models are not practical applications in reality.

\subsection{Inverse Autoregressive Flow}
With respect to the problem of the Gaussian autoregressive flow, \citep{kingma2016improving} proposed the inverse autoregressive flow that can be efficiently trained in parallel while still maintaining the flexibility of $q_\phi(z|x)$. The transformation function is simply the reverse of Equation \ref{eq: autoregressive}:
\begin{align}
\begin{split}
\label{eq: iaf}
&z_{t,0}=\frac{z_{t-1,0}-\mu_{t-1,0}}{\sigma_{t-1,0}}\\
z_{t,i}=&\frac{z_{t-1,i}-\mu_{t-1,i}(z_{t-1,0:i-1})}{\sigma_{t-1,i}(z_{t-1,0:i-1})}, i=1,2,...D
\end{split}
\end{align}
which is still invertible. The advantage is that now $z_t$ is only dependent on $z_{t-1}$ and we do not have to transform dimension by dimension. The transformation can be written in a vectorised version:
\begin{align}
\begin{split}
\label{eq: iaf-vector}
z_{t}=\frac{z_{t-1}-\mu_{t-1}(z_{t-1})}{\sigma_{t-1}(z_{t-1})}
\end{split}
\end{align}
where the division is performed element-wise. Once we have the vector $z_{t-1}$, $\mu_{t-1}(z_{t-1})$ and $\sigma_{t-1}(z_{t-1})$, Every dimension of $z_t$ can be computed in parallel. The determinant of the Jacobian matrix is simply the reciprocal of Equation \ref{eq: jac-autoregressive}.

\begin{figure}[htbp!] 
\centering    
\includegraphics[width=1\textwidth]{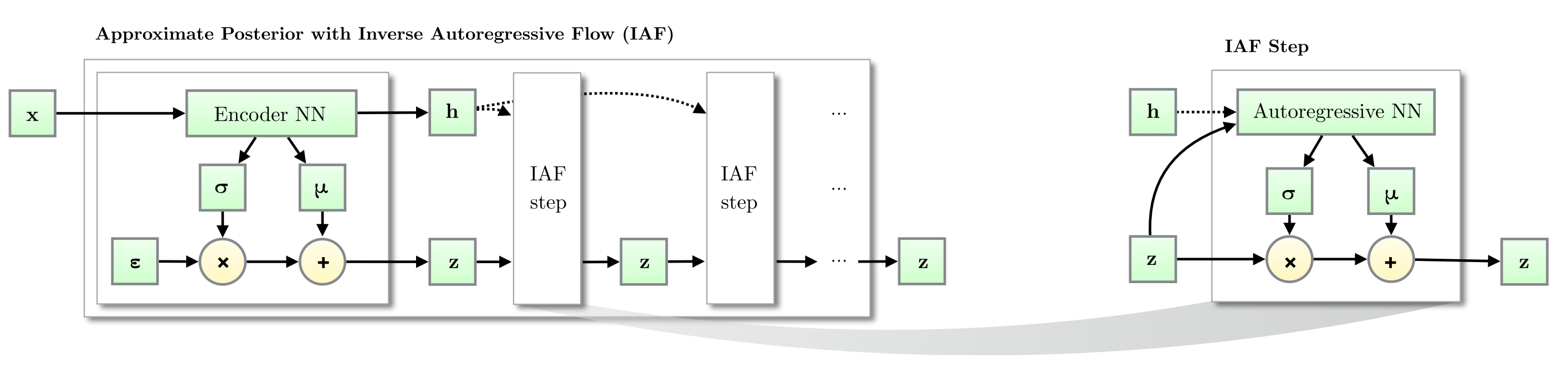}
\caption{Inverse Autoregressive Flow. It consists of an initial sample z drawn from a simple distribution, such as a
Gaussian with diagonal covariance, followed by a chain of nonlinear invertible transformations of z,
each with a simple Jacobian determinants.}
\label{fig: iaf}
\end{figure}

$\mu_{t-1}(z_{t-1})$ and $\sigma_{t-1}(z_{t-1})$ can be computed by the masked autoencoder (MADE)~\citep{germain2015made}, where the output $y_i=f(x_{1:i-1})$ and can be implemented as an MLP with a mask matrix:
\begin{align}
\begin{split}
&y=\sigma((M\odot Wx+b));\\
M=&\begin{vmatrix}
 0&0  &\cdots  &0&0 \\ 
 1&0  &\cdots  &0 &0\\ 
 \cdots& \cdots &\cdots  &\cdots &\cdots\\ 
 1&1  &\cdots  &0 &0\\ 
 1& 1 &  \cdots& 1&0
\end{vmatrix}
\end{split}
\end{align}
In \citep{kingma2016improving}, the author used a transformation that is equivalent to Equation \ref{eq: iaf-vector} up to reparameterization: $z_{t}=\mu_{t-1}(z_{t-1},h)+\sigma_{t-1}(z_{t-1},h)\odot z_{t-1}$. Figure \ref{fig: iaf} shows the detailed structure of the inverse autoregressive flow, where $h$ is an additional output from the initial encoder and are inputted at every transformation step.

In addition to the above mentioned methods, we can also use adversarial learning to get a more flexible distribution, which we will see in the next section.

\section{Generative Adversarial Networks}
Generative adversarial networks (GANs)~\citep{goodfellow2014generative} is another popular technique to approximate an unknown distribution. The main difference of it from previously introduced improving techniques is that it does not directly model the probability density. Instead, they introduce a discriminator to compete with the generator model so that eventually the generator will be driven to recover the real data distribution. In the end, we can get an effective sampler to help us get samples from the unknown distribution.

\subsection{Overview}
The generating process that GANs assume is the same as in variational approximation. First a latent variable is sampled from a prior distribution $z\sim p_\theta(z)$, then a multi-layer-perceptron transformation is applied to directly get a data sample $x=g_\theta(z)$. By this process we implicitly define a data distribution $p_\theta(x)$ whose density distribution is unknown but can be sampled from. The universality of neural networks gives it the potential to fit arbitrary distributions. Our goal is to find the optimal parameter $\theta$ such that this implicitly defined $p_\theta(x)$ is as close as possible to the real distribution $q(x)$. There are usually two perspectives of viewing the generative adversarial network: the game theory perspective and the density ratio perspective. The following section will explain these two perspectives.
\begin{figure}[htbp!] 
\centering    
\includegraphics[width=1\textwidth]{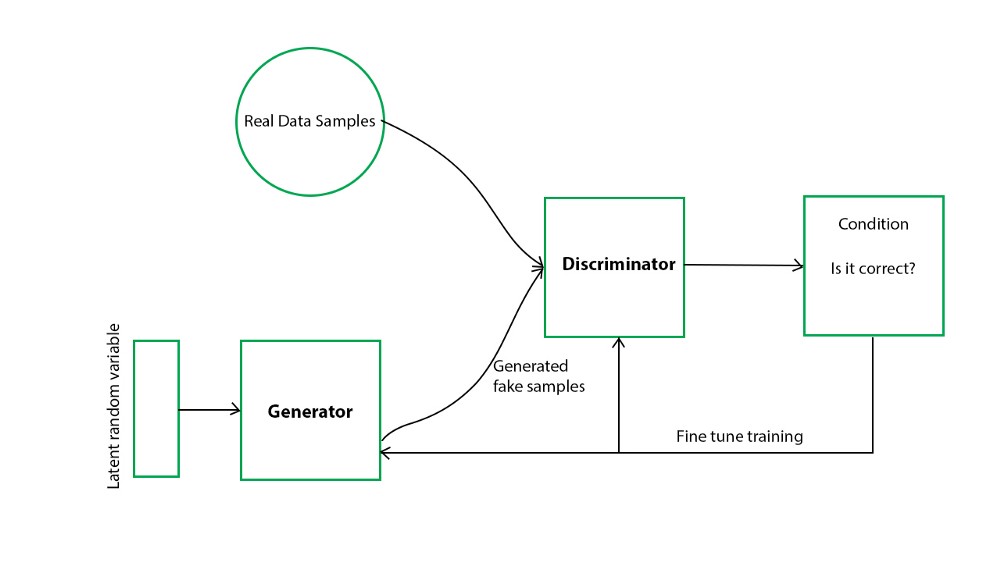}
\caption{Illustration of Generative Adversarial Networks. The discriminator and the generator compete with each other. In the end, the generator will be able to generate real-like samples to fool the discriminator.}
\label{fig: gan_intro}
\end{figure}

\subsection{From the Game Theory Perspective}
The classical definition of generative adversarial networks is from a game theory perspective. According to this definition, the generator, parameterized by $\theta$, generates fake samples with the above-mentioned generating process. We have an extra discriminator D which is trained to distinguish these fake samples from real data samples. The generator and discriminator will compete with each other and form an adversarial game. The generator can be thought of as analogous to a team of counterfeiters, trying to produce fake currency. On the contrary, the discriminative model is analogous to the police, trying to detect the counterfeit currency. Competition in this game drives both teams to improve their skills until the counterfeits are indistiguishable from the genuine articles~\citep{goodfellow2014generative}. The overall objective is:
\begin{align}
\begin{split}
\label{eq: gan}
&\min_{\theta} \max_{D} \mathcal{L}(\theta,D)=\mathbb{E}_{x\sim q(x)}\log D(y=1|x)+\mathbb{E}_{x\sim p_\theta(x)}\log D(y=0|x)\\
=&\min_{\theta} \max_{D} \mathcal{L}(\theta,D)=\mathbb{E}_{x\sim q(x)}\log D(y=1|x)+\mathbb{E}_{z\sim p_\theta(z)}\log D(y=0|g_\theta(z))
\end{split}
\end{align}
$D(y=1|x)$ and $D(y=0|x)$ output the probability that $x$ is a real or fake respectively. $D(y=1|x)+D(y=0|x)=1$. In the inner loop, $D$ is optimized to correctly identify the input sample. In the outer loop, $g_\theta$ is driven to produce real-looking samples to fool the trained discriminator. $D$ and $g_\theta$ are updated alternatively until a Nash equilibrium is achieved.

We can prove for a fixed generator defined by $\theta_0$, the optimal discriminator is:
\begin{align}
\begin{split}
\label{eq: op-dis}
D_{\theta_0}^*(y=1|x)=\frac{q(x)}{q(x)+p_{\theta_0}(x)}
\end{split}
\end{align}
which is intuitive as it outputs a confidence score proportional to the relative probability density. Now suppose the discriminator reaches its optimum for $\theta_0$, let $Q_\theta(x)=\frac{q(x)+p_\theta(x)}{2}$, if we want to update the generator, we have the following results by feeding the optimal discriminator into Equation \ref{eq: gan}:
\begin{align}
\begin{split}
\label{eq: opd-g}
&\triangledown_{\theta} \mathbb{E}_{x\sim q(x)}\log \frac{q(x)}{2Q_{\theta_0}(x)} +\mathbb{E}_{x\sim p_\theta(x)}\log \frac{p_{\theta_0(x)}}{2Q_{\theta_0}(x)}\biggr\rvert_{\theta=\theta_0}\\
=&\triangledown_{\theta}\mathbb{E}_{x\sim p_\theta(x)}\log \frac{p_{\theta_0(x)}}{Q_{\theta_0}(x)}\biggr\rvert_{\theta=\theta_0}\\
=&\triangledown_{\theta}KL(p_\theta||Q_{\theta_0})\biggr\rvert_{\theta=\theta_0}(\text{From Equation \ref{eq: kl}})
\end{split}
\end{align}
Further we can prove:
\begin{align}
\begin{split}
\label{eq: js}
&\triangledown_{\theta} 2JS(p_\theta||q)\biggr\rvert_{\theta=\theta_0}\\
=&\triangledown_{\theta} KL(p_\theta||Q_\theta)+KL(q||Q_\theta)\biggr\rvert_{\theta=\theta_0}\\
=&\triangledown_{\theta} \mathbb{E}_{p_\theta}\log \frac{p_{\theta}}{Q_{\theta_0}}+\mathbb{E}_{p_\theta}\log \frac{Q_{\theta_0}}{Q_\theta}+\mathbb{E}_{q}\log \frac{q}{Q_{\theta_0}}+\mathbb{E}_{q}\log \frac{Q_{\theta_0}}{Q_\theta}\biggr\rvert_{\theta=\theta_0}\\
=&\triangledown_{\theta} \mathbb{E}_{p_\theta}\log \frac{p_{\theta}}{Q_{\theta_0}}-2KL(Q_\theta||Q_{\theta_0})\biggr\rvert_{\theta=\theta_0}\\
=&\triangledown_{\theta}KL(p_\theta||Q_{\theta_0})\biggr\rvert_{\theta=\theta_0}
\end{split}
\end{align}
which is the same as the gradient we just derived.Therefore, if we assume $D(x)$ is optimal for any fixed $\theta_0$, optimizing the generator with respect to Equation \ref{eq: gan} is equivalent to minimizing the Jensen-Shannon divergence between $p_\theta(x)$ and $q(x)$, where the unique global minimum exits if and only if $p_\theta(x)=q(x)$. In \citep{goodfellow2014generative}, it was proved that the Nash equilibrium of this game yields a stationary point where $p_\theta(x)=q(x)$.

However, in practice it is impossible to ensure we can always learn an optimal discriminator. More generally, for any discriminator $D_{\theta_0}$, not necessarily optimal, we have:
\begin{align}
\begin{split}
\label{eq: gd}
\min_{\theta}\mathcal{L}(\theta,D_{\theta_0})=&\min_{\theta}\mathcal{L}(\theta,D_{\theta_0}^*)-\mathbb{E}_{x\sim q(x)}\log \frac{D_{\theta_0}^*(y=1|x)}{D_{\theta_0}(y=1|x)}-\mathbb{E}_{x\sim p_\theta(x)}\log \frac{D_{\theta_0}^*(y=0|x)}{D_{\theta_0}(y=0|x)}\\
=&\min_{\theta}\mathcal{L}(\theta,D_{\theta_0}^*)-2\mathbb{E}_{x\sim Q_\theta(x)}\mathbb{E}_{y\sim D_{\theta_0}^*(y|x)}\log \frac{D_{\theta_0}^*(y|x)}{D_{\theta_0}(y|x)}\\
=&\min_{\theta}JS(p_\theta||q)-2\mathbb{E}_{x\sim Q_\theta(x)}KL(D_{\theta_0}^*(y|x)||D_{\theta_0}(y|x))\\
\leq & \min_{\theta}JS(p_\theta||q)
\end{split}
\end{align}
which means we are basically minimizing a lower bound of the Jensen-Shannon divergence, the bound is tight if and only if the discriminator is optimal. This is not a nice property since normally we would like to minimizing an upper-bound. Minimizing a lower bound instead of an upper bound produces no guarantee about the real divergence. As seen from the last but one line at Equation \ref{eq: gd}, we can optimize by either minimizing the first Jenson-Shannon divergence or maximising the second KL divergence, we cannot guarantee the generator is moving to the correct direction. It could be that the real divergence is still large even if the objective we are minimizing is already zero. Therefore, the discriminator should be trained close to the optimal to get a tighter bound of the Jensen-Shannon divergence.

\subsection{From the Density Ratio Perspective}
The other view is to consider generative adversarial networks as optimizing by means of estimating the density ratio~\citep{mohamed2016learning}. This reverses the game theory perspective. We start directly from our objective: closing the divergence between $p_\theta(x)$ and $q(x)$. However, as explained above, $p_\theta(x)$ has an unknown density and can only be sampled from. We cannot directly apply gradient descent to optimize it. To solve this problem, a discriminator is imposed to estimate the ratio between $p_\theta(x)$ and $q(x)$. With this ratio, it is possible to estimate the gradient of our objective function. From this perspective, the generator and the discriminator are more like collaborating with each other than competing:  the discriminator is trained to freely share the ratio information, with which the generator estimates the gradient value and updates the parameters accordingly.

The density ratio $r_\theta(x)$ can be estimated by:
\begin{align}
\begin{split}
r_\theta(x)=&\frac{p_\theta(x)}{q(x)}=\frac{p(x|y=0)}{p(x|y=1)}=\frac{D_{\theta}^*(y=0|x)p(x)}{p(y=0)}/\frac{D_{\theta}^*(y=1|x)p(x)}{p(y=1)}\\
=&\frac{D_{\theta}^*(y=0|x)}{D_{\theta}^*(y=1|x)}\frac{p(y=1)}{p(y=0)}=\frac{D_{\theta}^*(y=0|x)}{D_{\theta}^*(y=1|x)}\text{ for } p(y=1)=p(y=0)=0.5
\end{split}
\end{align}
which can be easily obtained if we have trained the discriminator $D$ to the optimal. Normally we choose a uniform marginal distribution $p(y=1)=p(y=0)=0.5$. Changing the marginal label distribution is useful if we want more in-depth control: whether we want to precisely fit one of the models (a ``high precision, low recall" task such as generation) or explain all of the modes (a ``high recall, low precision" task such as retrieval)~\citep{creswell2016task}. It has been shown varying the marginal probability value is related to optimizing a generalised Jensen-Shannon divergence~\citep{huszar2015not}. Once we have the ratio, under some circumstances, the gradient information can be estimated. For example, suppose our objective is the KL divergence, we can estimate the gradient with respect to $\theta$ by:
\begin{align}
\begin{split}
\label{eq: density-kl}
&\triangledown_{\theta} KL(q||p_\theta)\biggr\rvert_{\theta=\theta_0}\\
=&\triangledown_{\theta}\mathbb{E}_{x\sim q(x)}\log \frac{q(x)}{p_{\theta}(x)}\biggr\rvert_{\theta=\theta_0}\\
=&\triangledown_{\theta}\mathbb{E}_{x\sim p_{\theta_0}(x)}\frac{q(x)}{p_{\theta_0}(x)}\log \frac{q(x)}{p_{\theta}(x)}\biggr\rvert_{\theta=\theta_0}\\
=&\triangledown_{\theta}\mathbb{E}_{x\sim p_\theta(x)} \frac{q(x)}{p_{\theta_0}(x)}\biggr\rvert_{\theta=\theta_0}\\
=&\mathbb{E}_{\epsilon\sim p(\epsilon)}\triangledown_{\theta}\frac{1}{r_{\theta_0}(g_\theta(\epsilon))}\biggr\rvert_{\theta=\theta_0}
\end{split}
\end{align}
which is the same as maximum likelihood since $KL(q||p_\theta)=-H(q)-\mathbb{E}_qp_\theta$ and the first term is a constant. If our objective is the Jensen-Shannon divergence, according to Equation \ref{eq: js} we have:
\begin{align}
\begin{split}
\label{eq: density-js}
&\triangledown_{\theta} JS(p_\theta||q)\biggr\rvert_{\theta=\theta_0}\\
=&\triangledown_{\theta}\frac{1}{2}KL(p_\theta||Q_{\theta_0})\biggr\rvert_{\theta=\theta_0}\\
=&\frac{1}{2}\mathbb{E}_{p(\epsilon)}\triangledown_{\theta} \log \frac{2r_{\theta_0}(g_\theta(\epsilon))}{1+r_{\theta_0}(g_\theta(\epsilon))}
\end{split}
\end{align}
which is the same as the original GAN objective, we can recover the original form by replacing $r_{\theta_0}$ with the discriminator.

When we train the discriminator and the generator alternatively. Each time the discriminator learn an estimated ratio by distinguishing fake samples from real ones, then the generator uses it to perform gradient descent. In \citep{nowozin2016f}, it was proved the gradient of any distribution that falls into the $f$-divergence family can be estimated by learning such a density ratio. From this perspective, we can more easily understand why an optimal discriminator $D$ is expected: A discriminator close to the optimum can provide a more accurate estimation of the density ratio, the resulting gradient thus has a larger chance of pushing model parameters in the right direction.

\subsection{Problem of GAN}
Though theoretically appealing, GANs have a lot of problems in practice. The training process can be very unstable. Because of the inaccuracy of the discriminator, we have no idea whether the model is becoming better of not. A stopping point is hard to be well defined. A most popular problem is the gradient vanishing. In the earlier training stage, the generator is very naive. The discriminator can be easily trained to reject all fake samples. As a result, the gradient passed to the generator vanishes, the adversarial game ends up with the winning of the discriminator. For example, suppose we define the discriminator as a sigmoid function, which is a common choice for classification:
\begin{align}
\begin{split}
D(y=1|x)=\sigma(x)=\frac{1}{1+e^{-x}}\in(0,1)
\end{split}
\end{align}
The gradient with respect to $\theta$ in the original objective \ref{eq: gan} is:
\begin{align}
\begin{split}
\triangledown_{\theta} \mathcal{L}(\theta, D)=&\triangledown_{\theta}\mathbb{E}_{x\sim p_\theta(x)}\log D(y=0|x)\\
=&\triangledown_{\theta}\mathbb{E}_{\epsilon \sim p(\epsilon)}\log (1-\sigma(g_\theta(\epsilon)))\\
=&\mathbb{E}_{\epsilon \sim p(\epsilon)}\frac{-\sigma(g_\theta(\epsilon))(1-\sigma(g_\theta(\epsilon)))}{1-\sigma(g_\theta(\epsilon))}\triangledown_{\theta}g_\theta(\epsilon)\\
=&\mathbb{E}_{\epsilon \sim p(\epsilon)}(-\sigma(g_\theta(\epsilon))\triangledown_{\theta}g_\theta(\epsilon))\approx 0
\end{split}
\end{align}
The last line comes because if the discriminator can reject almost all fake samples, then $\sigma(g_\theta(\epsilon))=D(y=1|g_\theta(\epsilon))\approx 0$. The gradient above becomes negligible and the generator does not have any motivation to evolve. We can more easily understand this condition from the density ratio perspective. As shown in Equation \ref{eq: density-js}, if the discriminator behaves perfectly, the ratio is a constant and no gradient information will be passed.

As explained in the last section, ideally we would expect an optimal discriminator as it provides a theoretically sound guarantee for the model, but in most cases an optimal discriminator will achieve an accuracy of $100\%$, then the generator will saturate. The model is faced with an awkward dilemma: A good discriminator provides a good density estimator to train the generator in the right direction. However, the better the discriminator is, the smaller the gradient would be. A weak discriminator can avoid the gradient vanishing problem but it will end up with an incorrect objective. In practice we must carefully tune up the degree to which the discriminator should be trained, so the training process is notoriously unstable.

\begin{figure}[htbp!] 
\centering    
\includegraphics[width=1\textwidth]{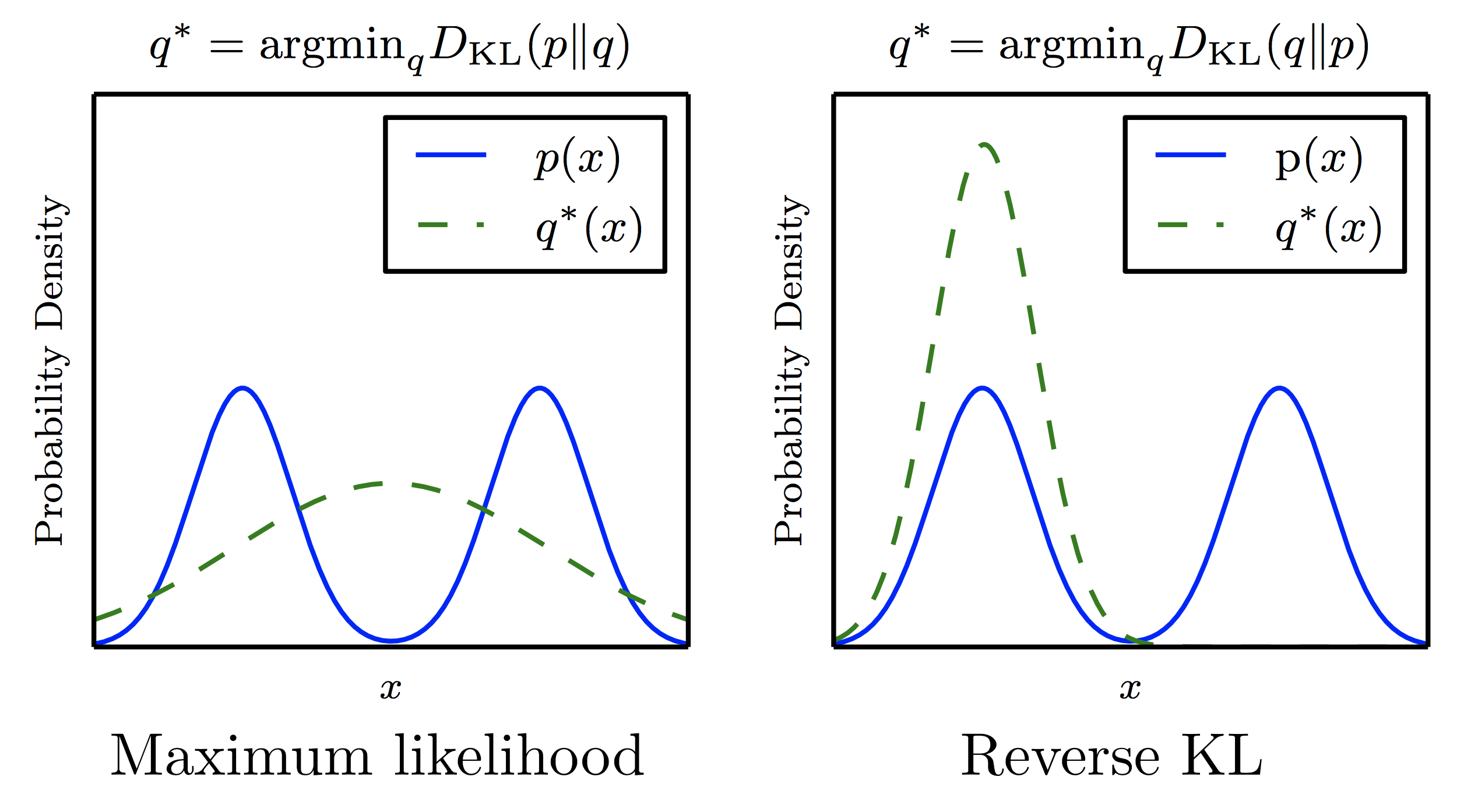}
\caption{Asymmetry of KL divergence}
\label{fig: kl-asy}
\end{figure}

In the original GAN paper, the author proposed flipping the sign on the discriminator's cost to obtain a cost for the generator. The cost for the generator becomes:
\begin{align}
\begin{split}
\min_{\theta}-\mathbb{E}_{p_\theta(x)}\log D(y=1|x)
\end{split}
\end{align}
The only difference is that our objective switches from increasing $D(y=0|x)$ to reducing $D(y=1|x)$. Intuitively it is still doing the same thing, but with this new objective we can get a much stronger gradient. When using a sigmoid function, the gradient is:
\begin{align}
\begin{split}
&\triangledown_{\theta}-\mathbb{E}_{x\sim p_\theta(x)}\log D(y=1|x)\\
=&\triangledown_{\theta}\mathbb{E}_{\epsilon \sim p(\epsilon)}\log \sigma(g_\theta(\epsilon))\\
=&\mathbb{E}_{\epsilon \sim p(\epsilon)}\frac{-\sigma(g_\theta(\epsilon))(1-\sigma(g_\theta(\epsilon)))}{\sigma(g_\theta(\epsilon))}\triangledown_{\theta}g_\theta(\epsilon)\\
=&\mathbb{E}_{\epsilon \sim p(\epsilon)}(-(1-\sigma(g_\theta(\epsilon)))\triangledown_{\theta}g_\theta(\epsilon))\\
\approx&\mathbb{E}_{\epsilon \sim p(\epsilon)}(-\triangledown_{\theta}g_\theta(\epsilon))
\end{split}
\end{align}
Though avoiding the gradient vanishing problem, the training process still has massively unstable updates with this new objective. The reason is that we are not minimizing the Jenson-Shannon divergence any more. In \citep{arjovsky2017towards}, it was proved the new objective is minimizing a reversed KL divergence under an optimal discriminator:
\begin{align}
\begin{split}
&\triangledown_{\theta}(-\mathbb{E}_{x\sim p_\theta(x)}\log D_{\theta_0}^*(y=1|x)\biggr\rvert_{\theta=\theta_0})\\
=&\triangledown_{\theta}( KL(p_\theta||q)-2JS(p_\theta||q)\biggr\rvert_{\theta=\theta_0})
\end{split}
\end{align}
Surprisingly, the second Jensen-Shannon divergence term has an opposite sign from the first reversed KL divergence term, which is pushing the model to move far from the real distribution. What's more, the model easily falls into the mode collapse problem: The generator maps various latent variables to the same data points and has a much less diversity than the real data distribution. Many people believe the problem comes from the asymmetry of the reversed KL divergence. Figure \ref{fig: kl-asy} depicts the asymmetry of KL divergence~\citep{goodfellow2016nips}, where $p(x)$ is the real data distribution and $q^*(x)$ is the optimal model distribution to minimize the KL divergence in two directions. The normal KL divergence objective (same as maximum likelihood) tends to cover the full real distribution (higher recall with lower precision), while the reversed KL divergence objective will focus only on a few modes (higher precision with lower recall). The reason of this asymmetric behaviour lies in the difference of cost:  The normal KL divergence severely punishes low probabilities assigned to real data samples while only moderately punishes high probabilities assigned to fake data samples. In contrast, the reversed KL divergence severely punishes high probabilities assigned to fake data samples while only moderately punishes low probabilities assigned to real data samples. In consequence, the reversed KL divergence objective leads to a generator that often synthesises mode-dropping but real-looking data samples.

However, people have tried training GANs with a normal KL divergence objective by Equation \ref{eq: density-kl}. The converged model still generates real-looking samples but only covers a small amount of modes, some researchers suggest the mode-collapse problem comes more from the adversarial training mechanism than from the choice of divergence. For example, in \citep{metz2016unrolled}, the authors argued an optimal generator for any fixed discriminator is a delta function at the $x$ to which the discriminator assigns highest data probability. Therefore, in standard GAN training, each generator update step is a partial collapse towards a delta function.

\subsection{Requirement for Applying GAN}
In summary, generative adversarial networks are a powerful framework to efficiently train a generative distribution for latent-variable models. There are only two requirement for applying this framewok: 
\begin{itemize}
\item The data distribution $x$ must be continuous. The generative parameter $\theta$ are differentiable almost everywhere.
\item We can sample from, but not necessarily know the exact density function, the likelihood distribution $p_\theta(x|z)$.
\end{itemize}
It is more powerful than variational approximations because it can in theory approximate arbitrary data distribution and does not need to explicitly specify the likelihood density function. In contrast, the function family of the posterior distribution in the variational approximation severely limits its power. We must have a very powerful posterior distribution in order to be able to match the performance of GAN. However, GANs do not allow inference on the known data points as it only models the generative process. Besides, the training of GANs is much more unstable and difficult than using the variational approximation.

Many tricks have been proposed to stabilise the GAN training procedure and avoid the gradient-vanishing or mode-collapse problem. As mentioned above, the GAN adversarial game can be easily stuck into the discriminator-winning equilibrium and the generator loses its gradient, so one obvious solution is to make harder the task for the discriminator. In the folliwng, we breifly introduce some common practices to stabalize the training of GANs.

\section{Improvement of GAN}
\subsection{One-Sided Label Smoothing}
One-sided label smoothing~\citep{salimans2016improved} tries to make the discriminator's task harder by randomly flip the label of real data with probability $\pi<1$, so that each real data sample has some chance of being fake thus crippling the discriminator's decision. The objective for the discriminator becomes:
\begin{align}
\begin{split}
\max_{D} \mathbb{E}_{x\sim q(x)}(\pi\log D(y=1|x)+(1-\pi)\log D(y=0|x))+\mathbb{E}_{x\sim p_\theta(x)}\log D(y=0|x)
\end{split}
\end{align}
The optimal discriminator for this task is:
\begin{align}
\begin{split}
D_{\theta_0}^*(y=1|x)=\frac{(1-\pi) q(x)}{q(x)+p_{\theta_0}(x)}
\end{split}
\end{align}
The good thing is that the crippled optimal discriminator scales down the original optimal discriminator to a constant which does not influence its shape, but now the discriminator can never achieve a $100\%$ accuracy because of the random flipping.

\subsection{Instance Noise}
\citep{sonderby2016amortised} argued one-sided label smoothing did not fundamentally address the gradient-vanishing problem as it punishes all the discriminators equally. There is still a large set of discriminators which achieve near-optimal loss, it is just that the near-optimal loss is now larger. Each of these possibly provides very different gradients to the generator. Thus, training the discriminator $D$ might find a different near-optimal solution each time depending on initialisation.

\begin{figure}[htbp!] 
\centering    
\includegraphics[width=1\textwidth]{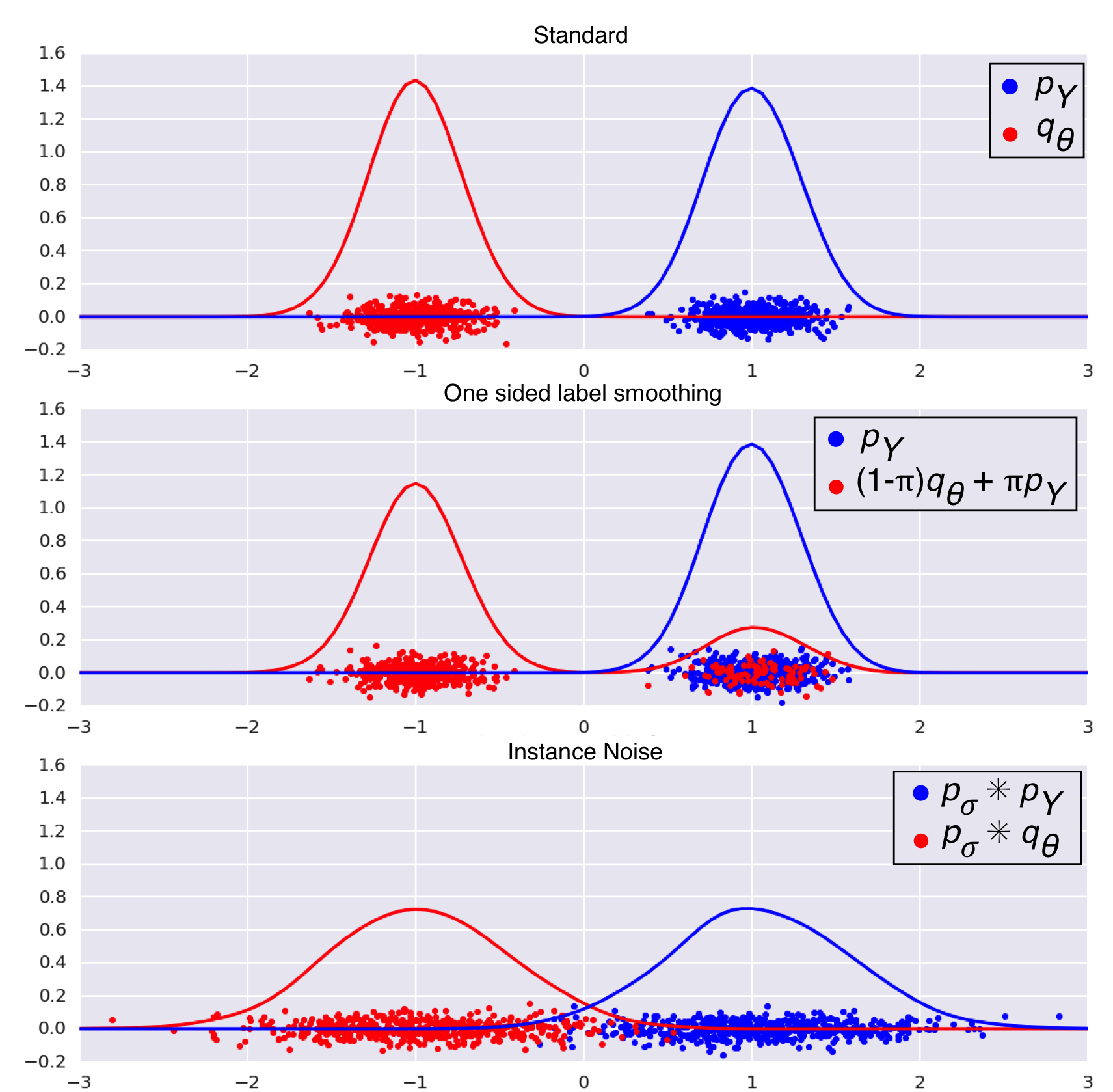}
\caption{Comparison of one-sided label smoothing and instance noise, $p_Y$ denotes real data distribution and $q_\theta$ is the generated distribution}
\label{fig: instance-noise}
\end{figure}

Instead, they proposed adding random noise on the data rather than on the label. Now the objective for the generator becomes:
\begin{align}
\begin{split}
\min_{\theta}d(p_\sigma*p_\theta||p_\sigma*q)
\end{split}
\end{align}
$p_\sigma$ is a noise function. When $d$ is KL divergence and $p_\sigma$ is the Gaussian random noise, minimizing the above objective can be proved to equal minimizing the Bregman divergence between $p_\theta$ and $q$. The advantage of instance noise is that the Bayesian-optimal is unique, the discriminator is less prone to overfitting because it has a wider training distribution, and the log-likelihood-ratio becomes better behaved because there is always an overlap between two distributions and the discriminator can never overwhelmingly win. In \citep{roth2017stabilizing}, it was further shown instance noise is equal to gradient regularisation, where we can achieve the same effect without explicitly adding noise.  Figure \ref{fig: instance-noise} compares one-sided label smoothing and instance noise, where the latter has a broader distribution and fewer sub-optimal discriminators.
\subsection{Minibatch Discrimination}
As mentioned, the model tends to generate samples covering only a few modes of the real data distribution in order to more easily fool the discriminator, which is called mode collapse. Minibatch Discrimination~\citep{salimans2016improved} tackles this problem by handing in multiple data samples to the discriminator at a time. Every time the discriminator sees a minibatch of fake data or real data samples, it should assign a confidence score to the whole set instead of to single samples. If the generator still generates samples only covering a few modes, its entropy within a minibatch is much lower than that of the real data distribution, the discriminator can easily identify it by means of the entropy difference.

We can show~\footnote{http://www.inference.vc/understanding-minibatch-discrimination-in-gans/}:
\begin{align}
\begin{split}
&KL(P^{(N)}||Q^{(N)})=N\cdot KL(P||Q)\\
&JS(P^{(N)}||Q^{(N)})\leq N\cdot JS(P||Q)
\end{split}
\end{align}
where $P^{(N)}$ and $Q^{(N)}$ mean the distribution of minibatches with size $N$. Therefore, when using the KL divergence as the target, the minibatch discriminator does not change our objective, but it does not hold for the Jensen-Shannon divergence.

\subsection{Unrolled GAN}
\begin{figure}[htbp!] 
\centering    
\includegraphics[width=1\textwidth]{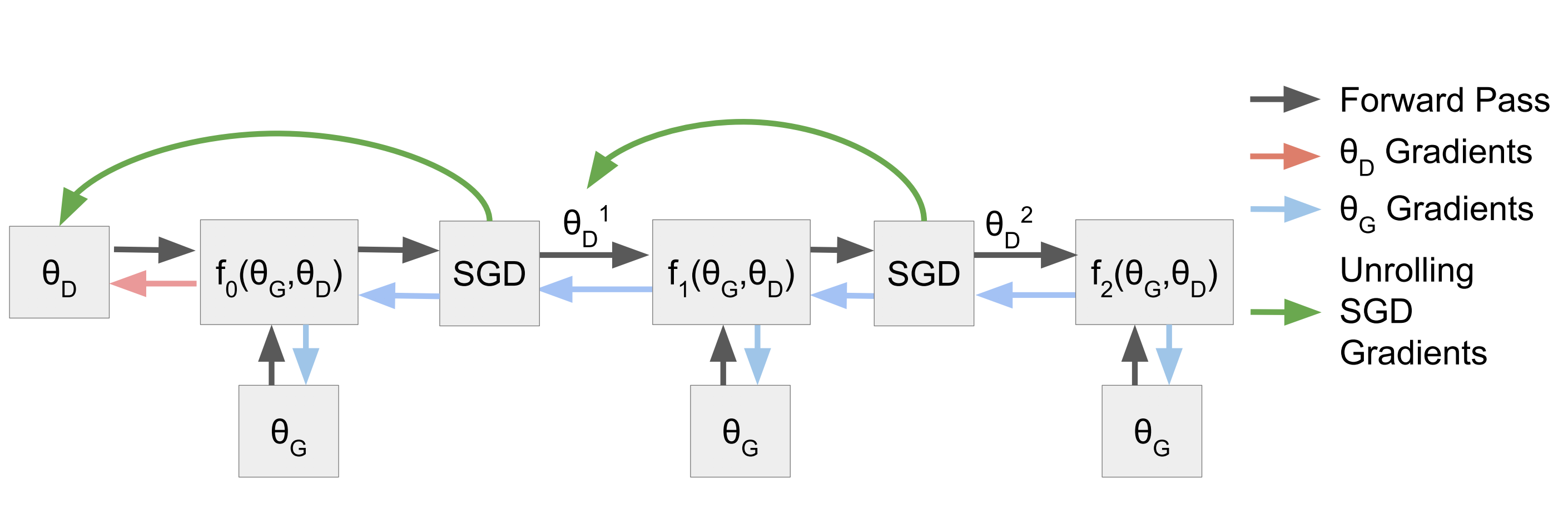}
\caption{Illustration of unrolled gan}
\label{fig: unrolled-gan}
\end{figure}

Unrolled GAN~\citep{metz2016unrolled} addresses the mode-collapse problem by allowing the generator to ``look one step ahead at the future". At each time step, the discriminator is updated only once based on the current generator. The generator, in contrast, receives gradient information from not only the current discriminator, but also from updated discriminators in the next $k$ steps. Intuitively we can think of it as giving some advantages to the generator because it is allowed to ``forsee" what the discriminator will react to it in the next k steps while the discriminator can only make changes according to the current instant generator. Since the target of the generator is to fool not only the current discriminator but also the future discriminators, the mode-collapse problem will be reduced as it will not blindly move towards high-density areas of any single discriminator. The structure is shown in Figure \ref{fig: unrolled-gan}. Specifically, when $k\rightarrow 0$, it becomes the original GAN training method, when $k\rightarrow \infty$, we can assume the generator updates based on a nearly-optimal discriminator and we are minimizing the exact  Jensen-Shannon divergence. Experiments find generally  a larger $k$ value leads to less mode-collapse. This implies an accurate gradient from a close-to-optimal discriminator is clearly expected, if we can find a way to avoid the gradient-vanishing problem.

\subsection{Wasserstein GAN}
\begin{figure}[htbp!] 
\centering    
\includegraphics[width=0.7\textwidth]{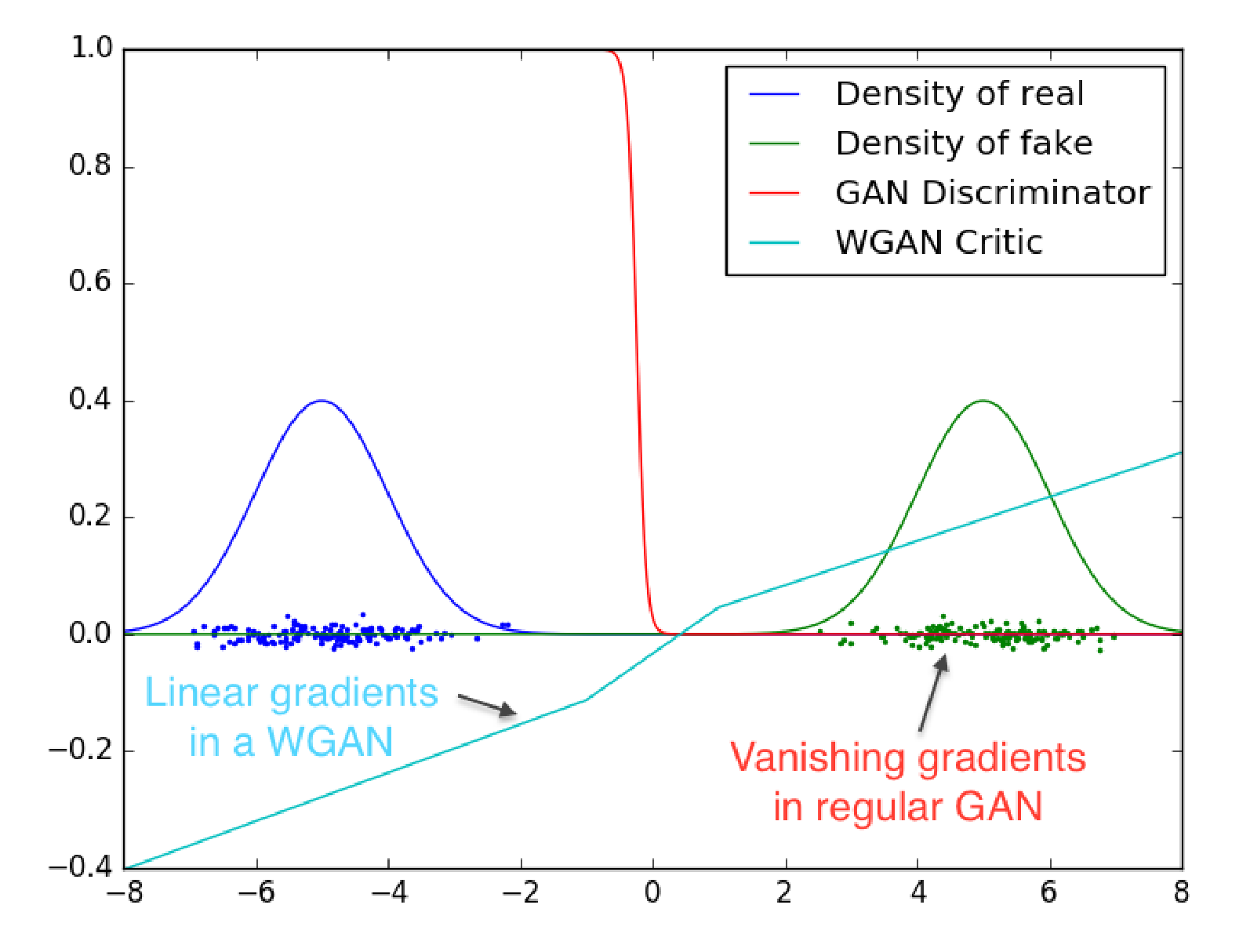}
\caption{Gradient of original GAN and Wasserstein GAN}
\label{fig: wgan}
\end{figure}

~\citep{arjovsky2017wasserstein} fundamentally analysed the training instability of GANs and proposed the Wasserstein GAN framework. It defines the Wasserstein distance, or Earth mover distance as:
 \begin{align}
\begin{split}
W(p_\theta||q)=\inf_{\lambda\sim\Pi(p_\theta,q)}\mathbb{E}_{(x,y)\sim \lambda}||x-y||
\end{split}
\end{align}
where $\Pi(p_\theta,q)$ means the set of all possible joint distributions $p(x,y)$ such that $\sum_{x}p(x,y)=p_\theta(x)$ and $\sum_{y}p(x,y)=q(y)$. Intuitively we can think of each distribution as a set of earths, where more earths indicating higher probability density. The Wasserstein distance basically measures the least amount of earths to be moved to make them equal. The advantage of Wasserstein distance over KL divergence and Jensen-Shannon divergence is that it can still provide meaningful gradient even when two distributions have completely no overlap. Its gradient is smooth and can continuously drive both distributions to move towards each other. As shown in Figure \ref{fig: wgan}, the gradient of the original GAN vanishes when there is no overlap, while Wasserstein GAN has a clean gradients over the whole space.

We can transform Wasserstein distance to a more computable form by applying the Kantorovich-Rubinstein duality~\citep{villani2008optimal}:
\begin{align}
\begin{split}
W(p_\theta||q)=\frac{1}{K}\sup_{||f||_{L}\leq K}\mathbb{E}_{x\sim p_\theta}f(x)-\mathbb{E}_{x\sim q}f(x)
\end{split}
\end{align}
where the supremum is taken over functions $f$ which have a bounded Lipschitz constant $K$. Specifically, $||f||_{L}\leq K$ means for any $x$, we have $||\triangledown_{x}f(x)||\leq K$.

We can approximate the set of $f$ using a feedforward neural network with its weight restricted in a compact space $[-c, c]$. The resulting function space should be a $K$-Lipschitz function with some constant $K$. We do not need to know the exact $K$ value as it only affects the ratio. In practice it can be adjusted by the learning rate. Such a feedforward neural network with weight clipping is referred to as ``critic" in \citep{arjovsky2017wasserstein}, in replacement of the discriminator in the original GAN. The training objective of Wasserstein GAN is then:
\begin{align}
\begin{split}
\min_{\theta}&\max_{f_c}\mathbb{E}_{x\sim q}f_c(x)-\mathbb{E}_{x\sim p_\theta}f_c(x)\\
&st. \text{ weight of }f_c \in [-c, c]
\end{split}
\end{align}
Since the critic $f_c$ can be fully trained towards the optimum without the problem of gradient vanishing, the mode-collapse problem will also be alleviated. As shown in the unrolled GAN, more training steps of the discriminator can reduce the mode-collapse problem. If every time the critic is optimal thus can provide an accurate gradient, the generator will be stably driven to minimize the Wasserstein distance, which in the end leads to $p_\theta(x)=q(x)$.

\citep{gulrajani2017improved} improves Wasserstein GAN by introducing a more advanced way of restricting the function space of feedforward neural networks. Instead of weight clipping, it adds an additional penalty term to directly prevent the function space from crossing the Lipschitz bound, the improved objective is:
\begin{align}
\begin{split}
\min_{\theta}&\max_{f_c}\mathbb{E}_{x\sim q}f_c(x)-\mathbb{E}_{\tilde{x}\sim p_\theta}f_c(\tilde{x})-\lambda\mathbb{E}_{\hat{x}\sim p(\hat{x})}(||\triangledown_{\hat{x}}f_c(\hat{x})||-K)^2\\
&\epsilon \sim U[0,1], x\sim q, \tilde{x}\sim p_\theta, \hat{x}=\epsilon x+(1-\epsilon)\tilde{x}
\end{split}
\end{align}

$\lambda$ is an adjustable weight, where a large $\lambda$ value can ensure $f_c$ is a $K-$Lipschitz function.  Experiments show the new objective can make use of the neural network weights more reasonably and increase the expressive power.

Apart from Wasserstein GAN, there are also efforts which use other divergence metrics. For example, energy-based GAN~\citep{zhao2016energy} minimizes the TV-divergence, least-square GAN~\citep{mao2016least} minimizes the Pearson $\chi^2$ divergence. Boundary-equilibrium GAN~\citep{berthelot2017began} minimizes a lower bound of the Wasserstein distance between auto-encoder loss distributions.

\subsection{VAE+GAN}
\label{sec: vaegan}
There have been lots of attempts of unifying the structure of VAE and GAN to combine the strengths from both sides. Compared with GAN, VAE has an extra encoding process which allows inference over data points. The extra inference capability can be used to improve the generator's performance, for example, by forcing generated data samples to be encoded to the same latent space as the real data to prevent mode collapse~\citep{srivastava2017veegan}. In contrast, GANs have the advantage that it does not need to explicitly define the probability density, which allows more flexibility and can in theory approximate arbitrary distribution.

We can apply GANs to include more flexible distributions within the VAE framework. As defined in Section \ref{sec: vae}, in VAE, there are three distributions we need to model: $p_\theta(z), q_\phi(z|x)$ and $p_\theta(x|z)$. GANs can be applied on any of them:

Firstly, in the latent space, we can implicitly define a posterior distribution $q_\phi$ by transforming a Gaussian random noise through $g_\phi(\epsilon)$, $g_\phi$ is implemented as a multi-layer perceptron. We can optimize with the original VAE objective:
\begin{align}
\begin{split}
\min_{\theta, \phi}\mathbb{E}_{q(x)}[-\mathbb{E}_{q_\phi(z|x)}\log p_\theta(x|z)+KL(q_\phi(z|x)||p_\theta(z))]
\end{split}
\end{align}
The second term about $q_\phi$ is intractable but can be easily estimated by the density ratio method as in Equation \ref{eq: density-kl}. We can train a discriminator to distinguish samples from $q_\phi(z|x)$ and $p_\theta(z)$ to estimate such a ratio. This idea has been extensively explored in \citep{rosca2017variational,mescheder2017adversarial,huszar2017variational}. 

Secondly, in the data space, features learned from the GAN discriminator can be used to define a more reasonable loss function for $p_\theta(x|z)$. For example, in \citep{larsen2015autoencoding}, we assume $p_\theta(D_l(x)|z)=\mathcal{N}(D_l(g_\theta(z)),I)$, $D_l(x)$ denote the hidden representation of the $l$th layer of the discriminator, the objective is defined as:
\begin{align}
\begin{split}
\min_{\theta, \phi}\mathbb{E}_{q(x)}[\frac{1}{2}\mathbb{E}_{q_\phi(z|x)}||D_l(x)-D_l(g_\theta(z))||^2 -KL(q_\phi(z|x)||p_\theta(z))]+\mathcal{L}_{GAN}
\end{split}
\end{align}
The GAN objective is trained simultaneously with the VAE objective, the discriminator is to distinguish real samples from $q(x)$ and fake sample drawn from $p_\theta(x|z)$. In contrast with the Gaussian assumption in Equation \ref{eq: square-vae}, the target here is the hidden representation extracted by the GAN discriminator. We expect the discriminator can learn high-level invariant representative features that can better follow the Gaussian distribution. Similar ideas are also applied in \citep{dosovitskiy2016generating,lamb2016discriminative}.

Lastly, the GAN idea can also be applied in the joint space of latent variables and data samples. For example, in \citep{dumoulin2016adversarially,donahue2016adversarial,pu2017symmetric}, the VAE is optimized through:
\begin{align}
\begin{split}
\min_{\theta, \phi}JS(q(x)q_\phi(z|x)||p_\theta(z)p_\theta(x|z))
\end{split}
\end{align}
It differs from the original ELBO objective but leads to the same global optimum. All the three distributions $p_\theta(z), q_\phi(z|x)$ and $p_\theta(x|z)$ can be implicitly defined without known probability density. The discriminator is applied to the samples from the joint distribution $q_\phi(x,z)$ and $p_\theta(x,z)$. Then the obtained density ratio estimate can be used to approximate the gradient as in Equation \ref{eq: density-js}.

There are many other interesting ways of integrating VAE and GAN. For example, infoGAN~\citep{chen2016infogan} applies the variational inference idea to additionally maximise the lower bound of the mutual information between data $x$ and a subset $c$ of the whole latent space $(c,z)$. The objective is:
\begin{align}
\begin{split}
\min_{\theta, \phi}\mathcal{L}_{GAN}-\lambda \mathbb{E}_{c\sim p(c), x\sim g_\theta(c,z)}\log Q_\phi(c|x)
\end{split}
\end{align}
As proved in Equation \ref{eq: vae-mi}, the second term is a lower bound of the $I(x,c)$. By adding the additional cost, infoGAN effectively prevents the latent variables from being ignored and interpretable features can be encoded into the latent variable subset $c$. Adversarial Generator-Encoder Networks~\citep{ulyanov2017adversarial} define a new adversarial game by using only an encoder module and a generator module. The objective is defined as:
\begin{align}
\begin{split}
\max_{e}\min_{g}d(e(g(Z))||e(X))
\end{split}
\end{align}
It can be proved there is a Nash equilibrium in this adversarial game if and only if $g^*(Z)=p(Z)$. Unlike the previously-mentioned methods, it does not need an additional discriminator to play the adversarial role. The competition is made between the encoder and the decoder. \citep{hu2017unifying} thoroughly analyses the connection between VAEs and GANs and provides a unified view of them. It shows techniques in one model can be easily transferred to the other one and brings performance improvement.

\begin{figure}[htbp!] 
\centering    
\includegraphics[width=1\textwidth]{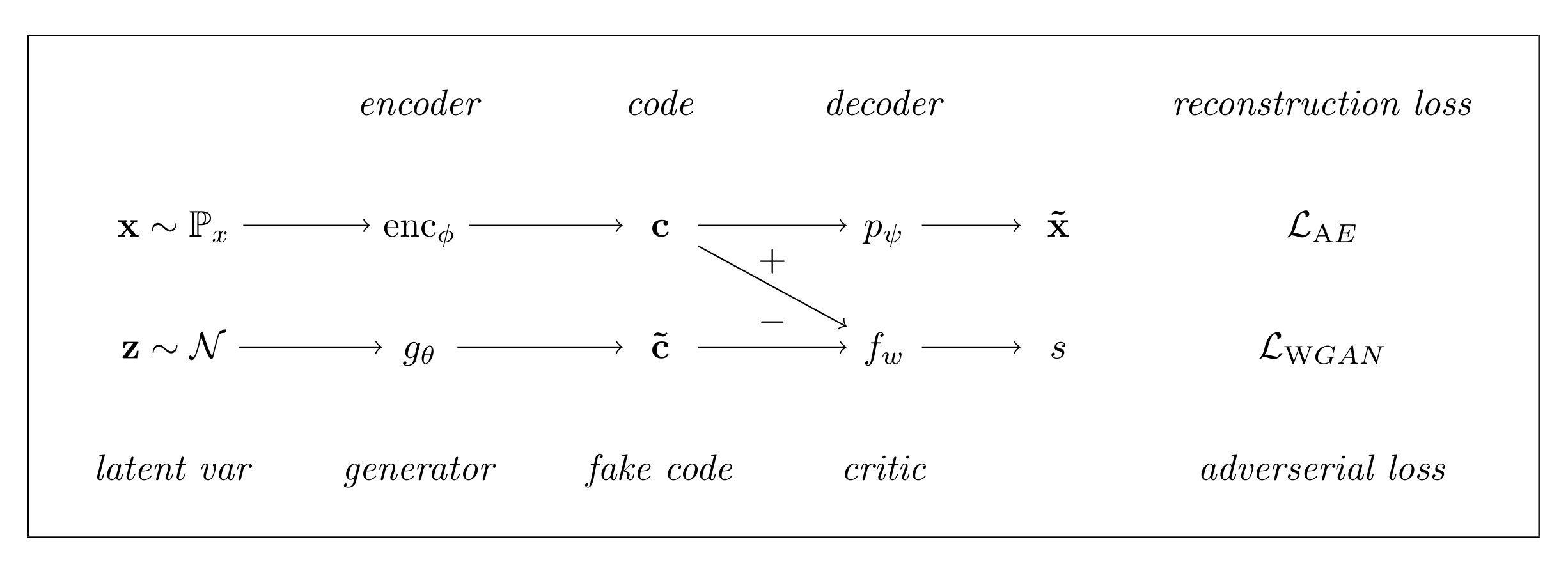}
\caption{Model structure of adversarially regularized autoencoder}
\label{fig: ad-regu}
\end{figure}

\citep{kim2017adversarially} propose the adversarially regularized autoencoder which applied similar ideas on text generation. It builds a normal autoencoder on text plus a GAN component which forces latent codes drawn from a prior distribution to be indistinguishable from those encoded from real text. The final objective contains a reconstruction loss of real text from the autoencoder part and an adversarial loss on the latent space from the GAN part. These two parts are trained simultaneously. The structure is shown in Figure \ref{fig: ad-regu}.

\begin{figure}[htbp!] 
\centering    
\includegraphics[width=1\textwidth]{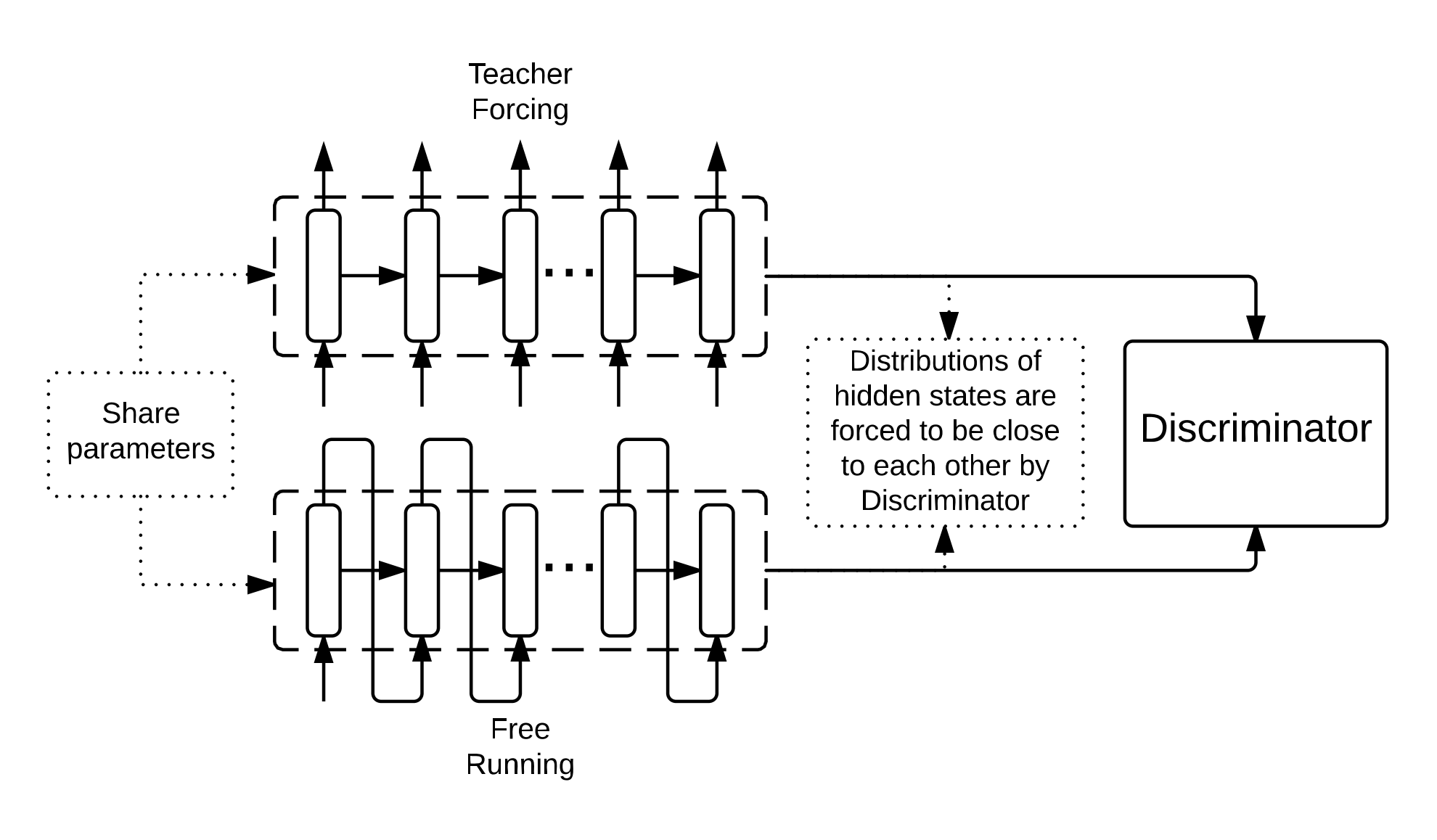}
\caption{Model structure of Professor forcing}
\label{fig: pro-force}
\end{figure}

Similarly, \citep{lamb2016professor} proposed Professor forcing that augmented the maximum-likelihood-based (Teacher Forcing) language models by applying GANs on the hidden state of recurrent neural networks. The discriminator is trained to distinguish RNN hidden states with input from either real text data (teacher forcing) or generated text data (free running). Since the input is continuous hidden states, backpropagation can be applied without approximation. The GAN component is built in order for the RNN hidden state to have similar dynamics facing real data or generated data so that the testing performance can be more close to the training performance. The model structure is shown is Figure \ref{fig: pro-force} 

Unlike the last two sections, applying GANs on latent space functions only as a regularizer and does not directly affect the generating process. The GAN component can bring extra benefits like smoother latent representations or more consistent behaviour between training and testing performance, but it still needs the traditional maximum likelihood training criteria to provide error signals.

\section{When reparameterization not applicable}
\label{sec: improv}
The reparameterization trick offers an effective way of getting a low-varianced, unbiased estimation from the proposal distribution $q_\phi(z|x)$. It can be used in both variational approximation and GANs to backpropagate gradients through sampled latent variables. However, it only applies to some specific distributions. When $q_\phi(z|x)$ comes from the distribution family where the reparameterization trick is not directly applicable, there are generally three ways of circumventing this difficulty: EM algorithm, soft relaxation and reinforcement learning. The following sections will introduce these three methods to train latent-variable models when reparameterization tricks are not applicable.

\subsection{EM algorithm}
EM algorithm~\citep{moon1996expectation} is a classical way of training latent-variable models. It includes two steps, E-step and M-step:
\begin{enumerate}
    \item E (Expectation) Step: Compute the posterior distribution $p_\theta(z|x)$ by fixing the generative model $p_\theta(x|z)$
    \item M (Maximization) Step: Set $\theta=argmax_{\theta}\mathbb{E}_{p_{\theta_0}}(z|x)\log p_\theta(x|z)$. $\theta_0$ is the generative parameter from the last iteration and is fixed in the M-step.
\end{enumerate}
By iteratively performing the E/M step, the log likelihood of the observed data can be gradually improved. In the end, the model is able to find the suitable latent variable $z$ that can explain the generation of $x$.

The M-step of EM algorithm is similar to minimizing the KL-divergence term in ELBO. The only difference is that in ELBO, the expectation is taken over $q_\phi(z|x)$ whose parameters are simultaneously optimized, while the M-step takes expectation over a fixed posterior distribution $p_{\theta_0}(z|x)$ obtained from the E-step.

We can use this idea to train variational autoencoders. Basically, we can optimize $\phi$ and $\theta$ in an iterative way. At each step, the gradient is not backpropagated to the other. By this means, we can sidestep the non-differentiable issue when reparameterization is not applicable. 

Iterative back-translation is a typical application of EM algorithm and has been widely applied in many semi/un-supervised text generation tasks like machine translation and style transfer~\citep{sennrich2016improving,lample2018phrase,hoang2018iterative,subramanian2018multiple}. For many text generation tasks, machine translation for example, it is expensive to obtain parallel text from one language to another. Iterative back translation provides a principled way to effectively utilize unparallel data to facilitate the learning.

\begin{figure}[htbp!] 
\centering    
\includegraphics[width=0.7\textwidth]{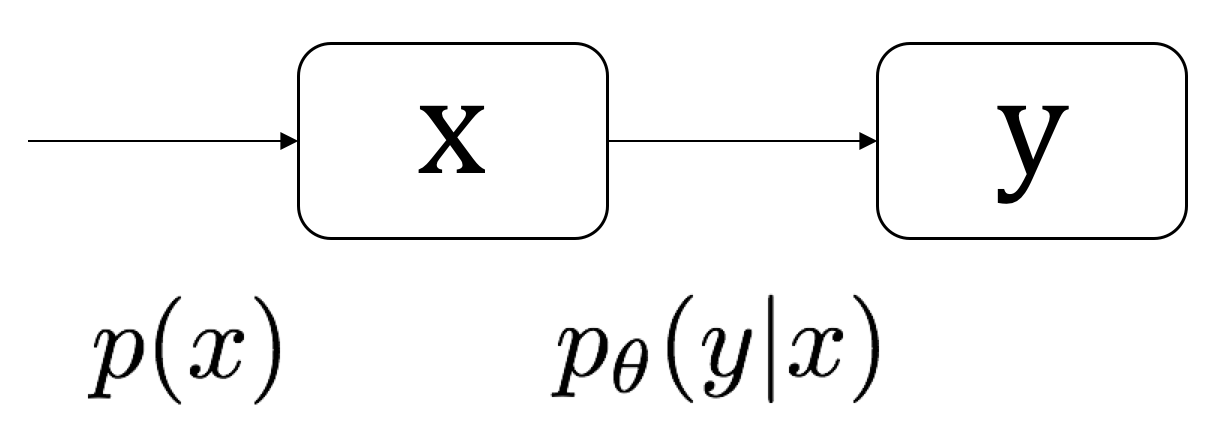}
\caption{Graphical Model of Back Translation}
\label{fig: back_translate}
\end{figure}

Specifically, iterative back-translation assumes the following generation process: Firstly, the condition x is generated from a prior distribution $p(x)$, then the corresponding text $y$ is generated based on $p_\theta(y|x)$. When large amounts of unpaired $y$ is available, we can treat its corresponding $x$ as latent. The generation becomes a latent-variable model. Figure~\ref{fig: back_translate} depicts this generation process. Iterative back-translation essentially applies EM algorithm to train this latent-variable model.

Instead of getting the exact posterior distribution in the E step, iterative back translation resort to a parameterized distribution $q_\phi(x|y)$ to approximate. It utilizes sequence-level knowledge distillation~\citep{kim2016sequence} to minimize $KL(p_\theta(x|y)||q_\phi(x|y))$.

It decompose the training into two steps:
\begin{enumerate}
    \item E-step: generate pseudo pairs $x,y'$ from unpaired $x$ by $p_\theta(y|x)$. Optimize $\phi$ to maximize the likelihood of $q_\phi(x|y)$. We can show this equals minimizing $KL(p_\theta(x|y)||q_\phi(x|y))$. By this means, $q_\phi(x|y)$ will be moved towards the real posterior distribution $p_\theta(x|y)$.
    \item M-step: generate pseudo pairs $x',y$ from unpaired $y$ by $q_\phi(x|y)$. Optimize $\theta$ to maximize the likelihood of $p_\theta(y|x)$. This basically maximizes $\mathbb{E}_{q_\phi(x|y)}\log p_\theta(y|x)$ which treats $q_\phi(y|x)$ as a surrogate of the real posterior distribution $p_\theta(y|x)$.
\end{enumerate}

By repeating the E-step and M-step iteratively, the correspondence between $x$ and $y$ can be established better and better. Many works have shown that even without backpropagating through $\phi$ in the M-step, the performance will not be affected~\citep{lample2018phrase,he2020a}. The training process is essentially the same as the wake-sleep algorithm~\citep{hinton1995wake}. The connection has been anylyzed in \citep{cotterell2018explaining}.

\subsection{Soft Relaxation}
Soft relaxation is another popular solution, which approximates discrete variables with continuous substitutes so that the reparameterization trick can be applied. For example, when generating text, the categorical distribution (defined by the softmax layer) is discrete, we cannot use the reparameterization trick to backpropagate the gradient through it. With the soft relaxation technique, we can generate embeddings instead of hard tokens. People have tried to use the same generating mechanism as images with CNNs~\citep{salimans2016improved,shen2017deconvolutional}. Word tokens are not generated sequentially but independently. The generated sentence is simply a concatenated matrix of every word embedding. By this means, the generated tokens are continuous thus fully differentiable. This method is effective in representation learning, but it cut off the word dependencies thus inappropriate as a generative model. 

\begin{figure}[ht]
  \centering
  \includegraphics[width=\textwidth]{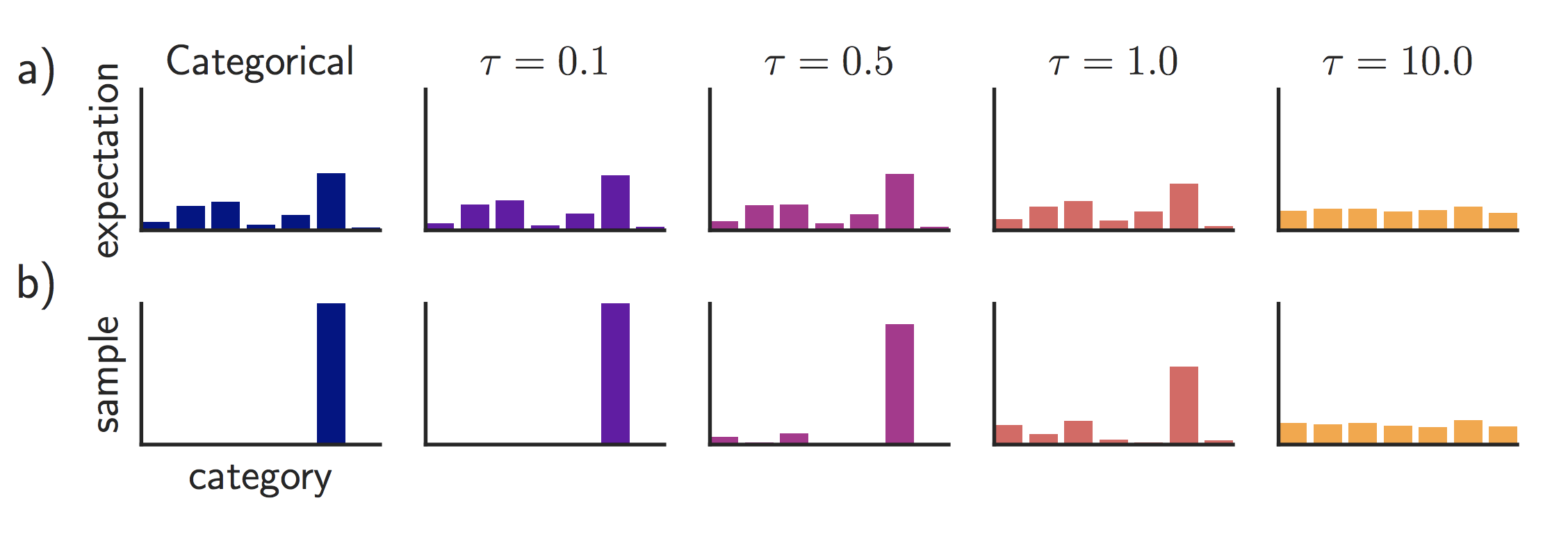}
  \caption{Effects of the temperature in Gumbel-softmax}
  \label{fig:gumbel_softmax}
\end{figure}

\citep{jang2016categorical,maddison2016concrete} proposed Gumbel-softmax to soft-relax samples from the discrete categorical distribution. It has been widely applied in many latent-variable models. Let $u$ be a categorical distribution with probability $\pi_1,\pi_2,...\pi_c$, the samples from this categorical distribution can be approximated by:
\begin{align}
\begin{split}
\label{eq: gumbel}
y_i=\frac{exp((\log (\pi_i)+g_i)/\tau)}{\sum_{j=1}^cexp((\log (\pi_j)+g_j)/\tau)}
\end{split}
\end{align}
$g_i$ follows the distribution Gumbel(0,1) and can be obtained by first sampling $\epsilon_i\sim Uniform(0,1)$ then generating $g_i=-\log(-\log \epsilon)$. $\tau$ is the temperature controlling the sampling accuracy. When $\tau\rightarrow0$, this is an accurate approximation. It becomes more smooth as $\tau$ grows. Figure~\ref{fig:gumbel_softmax} shows the effect of the temperature in Gumbel-softmax. In practice people usually set a high $\tau$ value initially then gradually decrease it as the training goes. This approximation technique has been used in many (semi)supervised NLP tasks when generated sentences need to be inputed to another function~\citep{zhou2017multi,yang2017improved,hu2017controllable}.

With this soft relaxation, we can sequentially generate a sentence word by word. Each time the model outputs a vector with Equation \ref{eq: gumbel}, which in turn serves as input for the next generating steps. The generated sentence is now a smooth transformation and fully differentiable. We can optimize using gradient descent with the GAN objective as in Equation \ref{eq: density-kl} and \ref{eq: density-js}. However, unlike in image generation, real natural sentences are discrete with each position being a unique word. Though the approximation technique helps sidestep the non-differentiable problem, turning generates sentences continuous makes the discriminator's task much easier. It can learn to simply reject all samples without a hard code in each position. Though Wasserstein GAN can still produce meaningful gradients in case of a perfect discriminator, the density ratio is less accurate when two distributions are very different. Moreover, languages are different from images in that there is complex inter-dependencies and linguistic correlations. The transition within words is more delicate and needs the fine-grained concern. Relying only on the binary signal from the discriminator has difficulty learning all the grammar rules and long-range word dependencies. More tricks are needed when training GANs on texts like curriculum learning, pre-training with maximum likelihood~\citep{press2017language,subramanian2017adversarial}, or augmented by feature matching via kernelized discrepancy metric~\citep{zhang2016generating,zhang2017adversarial}. Table \ref{tab: textgan-con} shows generates sentences from \citep{subramanian2017adversarial}, where there are still quite a few inconsistent sentences.
\begin{table*}[htb!]
\begin{center}
\begin{tabular}{|p{2cm}|p{2cm}|p{11cm}|}
\hline
Level & Method & 1-billion-word \\
\hline
\multirow{12}{*}{Word}
      & \multirow{6}{*}{LSTM} &An opposition was growing in China . \\
      & & This is undergoing operation a year . \\
      & & It has his everyone on a blame . \\
      &  &Everyone shares that Miller seems converted President  as Democrat .\\
      &  & Which is actually the best of his children . \\
      & & Who has The eventual policy and weak ? 
\\\cline{2-3}
      & \multirow{4}{*}{CNN} & Companies I upheld , respectively patented saga and Ambac. \\ 
      & & Independence Unit have any will MRI in these Lights \\ 
      & & It is a wrap for the annually of  Morocco \\
      & & The town has Registration matched with unk and the citizens \\
\hline
\multirow{4}{*}{Character} 
      & \multirow{4}{*}{CNN} & To holl is now my Hubby ,\\
      & & The gry timers was faller\\
      & & After they work is jith a\\
      & & But in a linter a revent\\
\hline
\end{tabular}
\end{center}
\caption{Sentences generated by GAN with soft relaxation}
\label{tab: textgan-con}
\end{table*}

To reduce the bias for sequential discrete variables, we can use straight-through Gumbel-softmax. In the forward phase, when generating a sequence, the exact discrete variable is input to the decoder. The soft-relaxation trick of Gumbel-softmax is only used in the backward phase for the gradient computation.

For simpler discrete distributions, like the binary bernoulli distribution, it is even possible to directly use the straight-through estimator. Specifically, let $\alpha$ be the probability of being 1 in the bernoulli distribution. In the forward phase, we set the latent variable as $\mathrm{1}(\alpha>0.5)$ to compute the loss. In the backward phase, we instead use continuous value $\alpha$ to backpropagate the gradient. Likewise, the bias of the gradient computed in this way is related to the peakness of the bernoulli distribution. When $\alpha$ is close to 0 or 1, the bias will be small.

\subsection{Reinforcement Learning}
Reinforcement learning, as have been mentioned in the earlier section, requires only samples from a distribution without getting access to the detailed function. It is able to estimate the gradient without reparameterization, which means we only need an estimated value of the probability density ratio but does not need to take the derivative with it. Specifically, if we want to optimize with respect to $KL(q||p_\theta)$, one method is as in Equation \ref{eq: density-kl}, which requires a continuous output of $g_\theta(\epsilon)$. The other method is through Monte Carlo sampling:
\begin{align}
\begin{split}
\label{eq: kl-rl}
\triangledown_\theta KL(q||p_\theta)\biggr\rvert_{\theta=\theta_0}&=\mathbb{E}_{x\sim q(x)} \triangledown_\theta \log p_\theta(x)\biggr\rvert_{\theta=\theta_0}\\
&=\mathbb{E}_{x\sim q(x)} \triangledown_\theta \log \sum_{z}p_\theta(x|z)p(z)\biggr\rvert_{\theta=\theta_0}\\
\end{split}
\end{align}
Note that now we only need to take the derivative with respect to $p_\theta(x|z)$ and $p_\theta(z)$, which are computable since $z$ is continuous. Estimating through this method has a high variance since we need to marginalize over $z$. \citep{hjelm2017boundary} proposed instead minimizing the KL divergence of the joint distribution:
\begin{align}
\begin{split}
\label{eq: kl-joint}
&\triangledown_\theta KL(q(x)p_{\theta_0}(z|x))||p_\theta(x|z)p(z))\biggr\rvert_{\theta=\theta_0}\\
=&-\mathbb{E}_{x\sim q(x)} \mathbb{E}_{z\sim p_{\theta_0}(z|x)}\triangledown_\theta \log p_\theta(x|z)\biggr\rvert_{\theta=\theta_0}\\
=&- \mathbb{E}_{p\sim p(z)} \mathbb{E}_{x\sim p_{\theta_0}(x|z)}\frac{D^*(y=1|x)}{D^*(y=0|x)}\triangledown_\theta \log p_\theta(x|z)\biggr\rvert_{\theta=\theta_0}\\
\approx&-\mathbb{E}_{p\sim p(z)} \frac{1}{Z_{\rvert_z}}\mathbb{E}_{x\sim p_{\theta_0}(x|z)}\frac{\triangledown_\theta \log p_\theta(x|z)}{r_{\theta_0}(x)}\biggr\rvert_{\theta=\theta_0}
\end{split}
\end{align}
where $Z_{\rvert_z}=\sum_{x}p_\theta(x|z)/r_{\theta_0}(x)$ to ensure $p_\theta(x|z)/r_{\theta_0}(x)$ is a proper density function. In the limit that $r_{\theta_0}(x)=D^*(y=0|x)/D^*(y=1|x)$, $Z_{\rvert_z}=1$ and it becomes the real posterior $p_\theta(z|x)q(x)/p(z)$. Equation \ref{eq: kl-joint} is essentially one kind of reinforce algorithm~\citep{williams1992simple} with the reward as $1/r_{\theta_0}(x)$ and the baseline as $0$. For an optimal discriminator, the reward becomes the density ratio between real and fake samples, the model can get more reward by generating real-like samples.

We can also target the reversed KL divergence by:
\begin{align}
\begin{split}
\label{eq: rkl-joint}
&\triangledown_\theta KL(p_\theta(x|z)p(z)||q(x)p_{\theta_0}(z|x)))\biggr\rvert_{\theta=\theta_0}\\
=&\mathbb{E}_{z\sim p(z)} \triangledown_\theta \mathbb{E}_{x\sim p_\theta(x|z)} \log \frac{p(z)p_{\theta_0}(x|z)}{q(x)p_{\theta_0}(z|x)}\biggr\rvert_{\theta=\theta_0}\\
=& -\mathbb{E}_{p\sim p(z)} \mathbb{E}_{x\sim p_{\theta_0}(x|z)}\log \frac{D^*(y=1|x)}{D^*(y=0|x)}\triangledown_\theta \log p_\theta(x|z)\biggr\rvert_{\theta=\theta_0}\\
\approx& \mathbb{E}_{p\sim p(z)} \mathbb{E}_{x\sim p_{\theta_0}(x|z)}\triangledown_\theta \log p_\theta(x|z)[-\log r_{\theta_0}(x)-\log Z_{\rvert_z}]\biggr\rvert_{\theta=\theta_0}
\end{split}
\end{align}

which is also the same like the policy gradient formula of the reinforce algorithm with reward as $-\log r_{\theta_0}(x)$ and a $z-$dependent baseline as $\log Z_{\rvert_z}$. When applying it on sequential data like text, this can be viewed as a reinforcement learning problem. The model needs to make decisions on which word to generate based on the previously-generated context. The final goal is to fool the discriminator. Since the reward is only known until a full sentence has been generated, we need to estimate the intermediate value function by Monte Carlo search or another parameterized function. There have been many attempts of applying the reinforce algorithm on generating text with GAN and different techniques to reduce the variance have been proposed~\citep{yu2017seqgan,li2016deep,guo2017long,li2017adversarial}.

\begin{figure}[ht]
  \centering
  \includegraphics[width=0.7\textwidth]{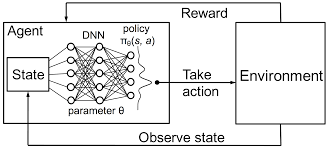}
  \caption{Illustration of Reinforcement Learning}
  \label{fig:reinforcement_learning}
\end{figure}

The above way of estimating the gradient is basically one simplest example of reinforcement learning. Figure~\ref{fig:reinforcement_learning} illustrates the idea of reinforcement learning. The policy distribution can be updated to maximize the reward with only samples from it. As we have mentioned, reinforcement learning treats the distribution as a blackbox and makes no assumption on it, which makes it more universe compared with the reperametrization. However, it also loses useful information to help reduce the variance. Reinforcement learning is itself a hot topic and we can easily borrow techniques like actor-critic or Q-learning to reduce the variance.

\section{VAE in Natural Language Generation}
\label{sec: vae-nlg}
The application of the VAE in natural language generation is quite straightforward. We simply need to replace the likelihood distribution $p_\theta(x|z)$ with a statistical language model that estimates the probability of text sequences. There has been quite a few attempts of applying VAEs in different areas like dialogue generation~\citep{li2017dailydialog,serban2017hierarchical}, machine translation~\citep{zhang2016variational}, sentence generation~\citep{bowman2016generating}, document modelling~\citep{miao2016neural} and topic models~\citep{srivastava2017autoencoding}. As explained in Chapter 1, normally for natural language generation tasks we have an input information to condition on, so most above work used the ``conditional VAE" framework.

\subsection{Conditional VAE}
The idea of conditional VAEs (CVAEs)\citep{yan2016attribute2image,sohn2015learning,shen2017conditional} is to condition every distribution in VAEs on an additional context information $c$. The generating process becomes: $z$ is sampled from a prior distribution $p_\theta(z|c)$, then $x$ is generated from $p_\theta(x|z,c)$. We use $q_\phi(z|x,c)$ to approximate the real posterior distribution $p_\theta(z|x,c)$. The objective is only slightly different:
\begin{align}
\begin{split}
\label{eq: cvae}
ELBO&=\mathbb{E}_{q(x|c)}[\mathbb{E}_{q_\phi(z|x,c)}\log p_\theta(x|z,c)-KL(q_\phi(z|x,c)||p_\theta(z|c))]\\
&=\mathbb{E}_{q(x|c)}[p_\theta(x|c)-KL(q_\phi(z|x,c)||p_\theta(z|x,c))]\\
&\leq \mathbb{E}_{q(x|c)}p_\theta(x|c)\\
&c=f(input)
\end{split}
\end{align}
The conditional VAE can be easily combined with the seq2seq~\citep{sutskever2014sequence} structure when the input and output are both sequences. For example, in dialogue generation, $f$ can be an LSTM encoder that learns a vectorised representation c of the dialogue history, $p_\theta(x|z,c)$ can be an LSTM decoder that estimates the probability of the ground-truth response. Then we have a CVAE-seq2seq framework that combines advantages from both sides --- with the powerful modelling capacity of CVAEs and the end-to-end seq2seq architecture for text modelling.

Though promising results have been achieved, the biggest problem of the VAE application in natural language generation is the ``optimizing challenges".
\subsection{Optimizing Challenges}
The optimizing challenge is that latent variables will be ignored when powerful decoders like LSTMs are used in $p_\theta(x|z)$. In this case, $p_\theta(x|z)$ degenerates to a normal language model with $z$ becoming non-informative to $x$, which is predictable if we look at the ELBO objective again:
\begin{align}
\label{eq: lm-elbo}
ELBO=\mathbb{E}_{q(x)}[\mathbb{E}_{q_\phi(z|x)}\log p_\theta(x|z)-KL(q_\phi(z|x)||p_\theta(z))]
\end{align}
The objective contains reconstruction loss $\mathbb{E}_{q_\phi(z|x)}\log p_\theta(x|z)$ and KL divergence $KL(q_\phi(z|x)||p_\theta(z))$. When using mean-field Gaussian distributions to parameterize $q_\phi(z|x)$ and $p_\theta(z)$, minimizing the KL divergence is rather easy. We can make $KL(q_\phi(z|x)||p_\theta(z))=0$ by turning off the correlation between $x$ and $z$ and set $q_\phi(z|x)=q_\phi(z)=p_\theta(z)$. In contrast, maximising $\mathbb{E}_{q_\phi(z|x)}\log p_\theta(x|z)$ is more difficult because of the complexity of natural language distributions. Since the LSTM is a universal approximator that can in theory describe arbitrary distribution, it has the potential to accurately model $q(x)$ without reliance to $z$. Though taking advantage of $z$ will bring more flexibility in the long term, the inherent short sight of gradient descent will inevitably go after the short rewards by ignoring $z$ to avoid paying the extra KL cost. This problematic tendency in learning is compounded by the LSTM decoder's sensitivity to subtle variation in the hidden states, such as that introduced by the posterior sampling process. In the initial training stage, when $q_\phi(z|x)$ is noisy and does not contain useful information about $x$, $p_\theta(x|z)$ will also tend to recover $x$ by exploiting the internal patterns of $q(x)$ and neglect the noisy $z$. Once this has happened, the decoder ignores the encoder and little to no gradient signal passes between the two, yielding an undesirable stable equilibrium with the KL cost term at zero. In practice, we would expect a small construction error with a non-trivial KL cost such that $z$ helps recover the distribution of $x$. Otherwise there will be no need to impose an additional latent variable.

Similar phenomena is also observed in image generation when expressive $p_\theta(x|z)$ like PixelCNNs~\citep{oord2016pixel} are used. $p_\theta(x|z)$ tends to model $q(x)$ directly without referring to $z$. It is more of a problem in language generation since languages are by nature sequential. Using less expressive languages models will fail to capture inter-language dependencies and severely affect the model performance. Several strategies have been proposed to alleviate this problem.

\subsection{KL cost annealing}
KL cost annealing~\citep{bowman2016generating} is a simple method that adds an additional weight $\epsilon$ to the KL cost term in Equation \ref{eq: lm-elbo}. Initially $\epsilon$ is set to $0$ so that $q_\phi(z|x)$ will try to encode as much information on $x$ as possible to help recover $x$ by $p_\theta(x|z)$. As time goes, $\epsilon$ will gradually increase to 1 to recover the original ELBO objective. Specifically, when $\epsilon=0$, we have:
\begin{align}
\begin{split}
\label{eq: vae-mi}
&\max_{\theta,\phi}\mathbb{E}_{q(x)}\mathbb{E}_{q_\phi(z|x)}\log p_\theta(x|z)-\epsilon KL(q_\phi(z|x)||p_\theta(z))\\
=&\max_{\theta,\phi}\mathbb{E}_{q_\phi(z)}\mathbb{E}_{q_\phi(x|z)}\log q_\phi(x|z)-\log \frac{q_\phi(x|z)}{p_\theta(x|z)}\\
=&\max_{\theta,\phi}\mathbb{E}_{q_\phi(z)}[-H(q_\phi(x|z))]-\mathbb{E}_{q_\phi(z)}KL(q_\phi(x|z)||p_\theta(x|z))\\
=&\max_{\theta,\phi}I_{q_\phi}(x;z)-H(q(x))-\mathbb{E}_{q_\phi(z)}KL(q_\phi(x|z)||p_\theta(x|z))\\
=&\max_{\theta,\phi}I_{q_\phi}(x;z)-\mathbb{E}_{q_\phi(z)}KL(q_\phi(x|z)||p_\theta(x|z))
\end{split}
\end{align}
\begin{figure}[htbp!] 
\centering    
\includegraphics[width=0.7\textwidth]{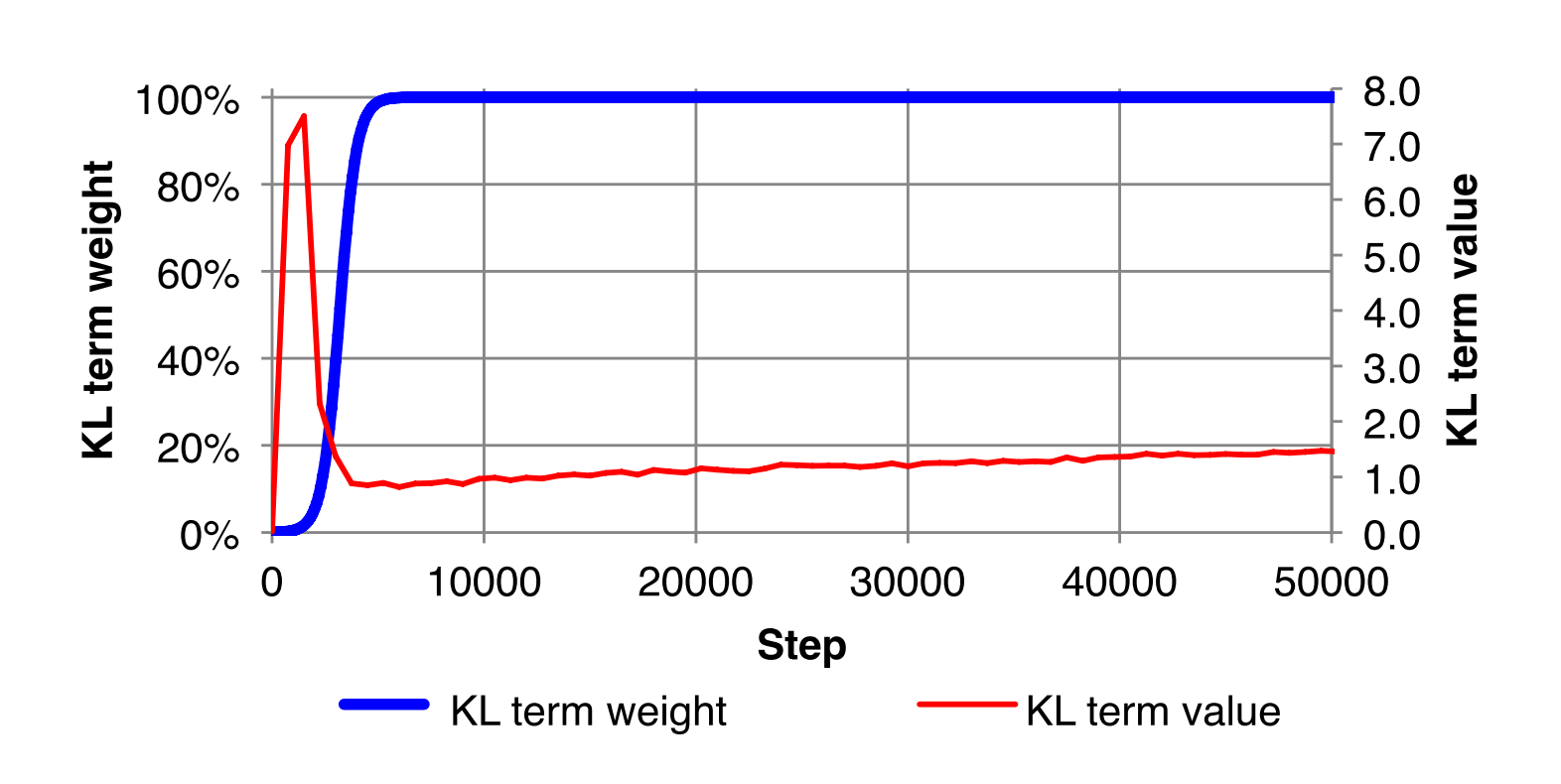}
\caption{Change of KL divergence term}
\label{fig: back-kl}
\end{figure}
At the earlier training stage, when $\epsilon$ is small, we are basically trying to approximate $q_\phi(x|z)$ with $p_\theta(x|z)$ and at the same time maximasing the mutual information between $x$ and $z$ under the distribution defined by $q_\phi(z|x)$, so $q_\phi(z|x)$ will make $z$ informative on $x$. As $\epsilon$ grows to 1, the model will assign more weights to the KL cost term. ~\citep{bowman2016generating} visualises the change of the KL divergence (Figure \ref{fig: back-kl}) when applying it on sentence generation. The KL divergence spikes early in training while the model can encode information in
$z$ cheaply, then drops substantially once it begins paying the full KL divergence penalty, and finally slowly rises again before converging as the model learns to condense more information into $z$. In practice, KL cost annealing normally has to be paired with the word drop-out technique (see next section) or early stop to achieve similar results. Otherwise, the KL divergence still eventually vanishes to zero when the weight grows to 1.

\subsection{Word Drop Out}
\begin{figure}[htbp!] 
\centering    
\includegraphics[width=0.7\textwidth]{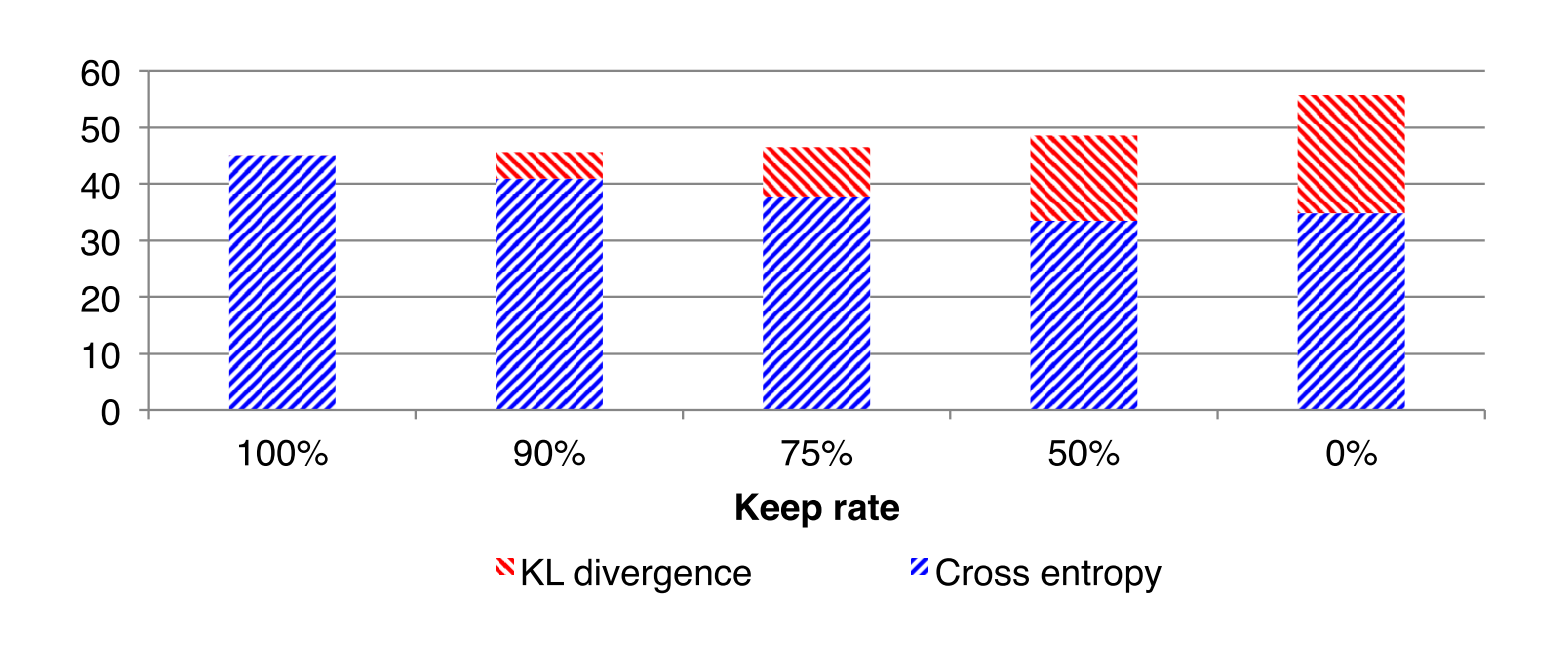}
\caption{Effects of word keep rate. The KL divergence grows as fewer words are kept since the model must reply on more informative $z$ to reconstruct the text.}
\label{fig: back-wdo}
\end{figure}

~\citep{bowman2016generating} also proposed word drop-out to encourage the usage of $z$. In the training phase, some fraction of the history words are randomly replaced with the UNK token indicating unknown words. In this way, we weaken a flexible LSTM decoder by reducing the dependency on history words. The decoder does not have enough history information to recover the exact word thus must turn to the latent variable $z$. Figure \ref{fig: back-wdo} depicts the effect of the word keep rate. When more fractions of history words are dropped, the LSTM decoder becomes weaker and has to seek information from $z$, so the KL divergence grows and $z$ becomes more informative on $x$.

Interestingly, though seemingly weakening the LSTM decoder, word drop-out has been shown to in fact bring improvement to the performance when a proper keep rate is used. In \citep{xie2017data}, word drop-out is explained as a smoothing technique in neural networks. Specifically, randomly replacing some history words with the UNK token can be seen as an interpolation among different n-gram models. To make a prediction, we use the expected probability over different noise of the context:
\begin{align}
\begin{split}
p_\theta(x_t|x_{<t})&=\mathbb{E}_{\tilde{x}_{<t}}p(x_t|\tilde{x}_{<t})=\sum_J \pi(|J|)p(x_t|x_J)\\
\pi(|J|)&=\lambda^{|J|}(1-\lambda)^{t-1-|J|}
\end{split}
\end{align}
$\lambda$ is the work keep rate. $x_{<t}$ is the ground-truth context words, $\tilde{x}_{<t}$ is the noised words after random dropping out. $J\subseteq  \{1,2,...t-1\}$ indicates the indices of un-noised words. $\pi(|J|)$ can be seen as the mixing coefficient of n-gram models $p(x_t|x_J)$. Therefore, word drop-out with a proper rate can not only attenuate the optimizing challenge, but also achieve a lower reconstruction loss by n-gram interpolation.

\subsection{Additional Loss}
We can also prevent the model from ignoring latent variables by imposing additional loss. For example, in \citep{semeniuta2017hybrid,zhao2017learning}, an additional loss is defined to predict each word without any context information, the model is forced to encode information on $z$ to make such a prediction. The new objective is defined as:
\begin{align}
\begin{split}
\label{eq: ad-loss}
&\max_{\theta,\phi}\mathbb{E}_{q(x)}[\mathbb{E}_{q_\phi(z|x)}[\log p_\theta(x|z)+\log p_\lambda(x|z)]-KL(q_\phi(z|x)||p_\theta(z))]\\
=&\max_{\theta,\phi}\mathbb{E}_{q(x)}[\log p_\theta(x) - KL(q_\phi(z|x)||\frac{p_\theta(z|x)p_\lambda(x|z)}{Z_x})+Z_x]\\
&\text{where }p_\theta(x|z)=\prod_{i}p_\theta(x_i|x_{<i},z), p_\lambda(x|z)=\prod_{i}p_\theta(x_i|z),Z_x=\sum_{z}p_\theta(z|x)p_\lambda(x|z)
\end{split}
\end{align}
Since $p_\lambda(x|z)$ is a very weak inputless decoder, the extra reconstruction loss will be high without the help of $z$, pushing the model to devote more efforts on the reconstruction loss than the KL cost.

The new objective does not change the objective for $p_\theta(z)$ and $p_\theta(x|z)$ because the additional loss has nothing to do with both term. We still have $p_\theta(z)=q_\phi(z), p_\theta(x|z)=q_\phi(x|z)$ and $p_\theta(x)=q(x)$ in the nonparametric limit. However, $q_\phi(z|x)$ is no longer $p_\theta(z|x)$ in the global optimum. As can be seen in Equation \ref{eq: ad-loss}, $q_\phi(z|x)$ will be driven to a mixture of $p_\theta(z|x)$ and $p_\lambda(x|z)$, so the additional loss will lead to a biased inference distribution but an unbiased generating distribution.

\begin{figure}[htbp!] 
\centering    
\includegraphics[width=0.7\textwidth]{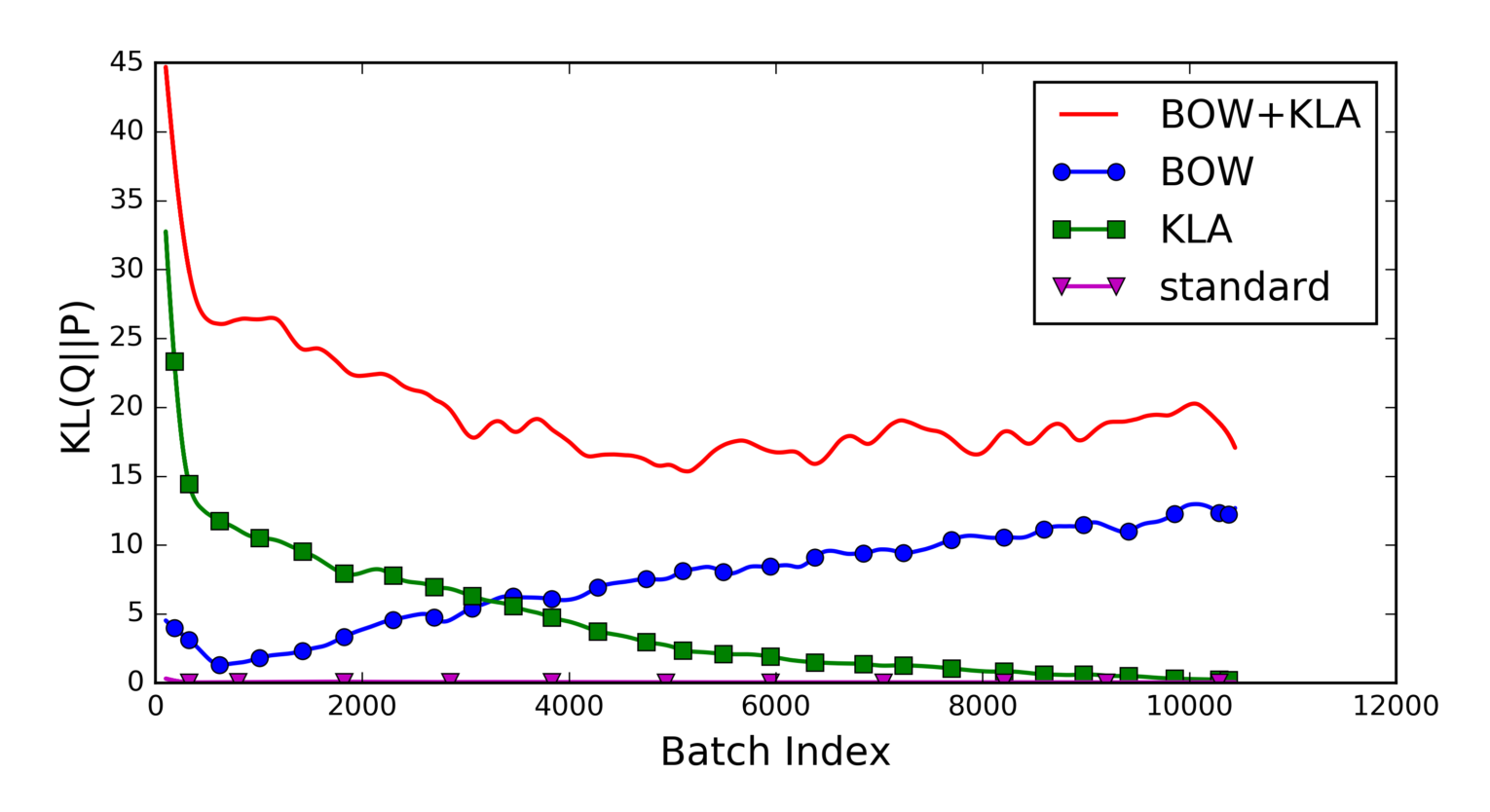}
\caption{Effects of additional loss}
\label{fig: back-bow}
\end{figure}

In \citep{zhao2017learning}, the authors compared the performance using KL cost annealing (KLA) and additional loss, which they call it bag-of-words loss (BOW) in the paper. The results is shown in Figure \ref{fig: back-bow}, in which we can see the additional loss effectively avoids the KL-vanishing problem.

\subsection{Free Bits}
The idea of free bits ``free bits"~\citep{kingma2016improving} is rather simple. We reserve some space for each dimension in the KL cost term, within which the model is free to encode information into latent variables. Only when the KL divergence exceeds the reserved quota will the model pay attention to it. The objective is defined as:
\begin{align}
\begin{split}
\label{eq: fb}
\max_{\theta,\phi}\mathbb{E}_{q(x)}\mathbb{E}_{q_\phi(z|x)}\log p_\theta(x|z)-\sum_{i=1}^D \max (KL(q_\phi(z_i|x)||p_\theta(z_i)),\epsilon)
\end{split}
\end{align}
where $\epsilon$ is the ``free bits" for every dimension.

We can also reserve a quota for the whole KL divergence instead of spreading it over every dimension in average~\citep{yang2017improved}, then the objective is:
\begin{align}
\begin{split}
\max_{\theta,\phi}\mathbb{E}_{q(x)}\mathbb{E}_{q_\phi(z|x)}\log p_\theta(x|z)- \max (KL(q_\phi(z|x)||p_\theta(z)),\epsilon)
\end{split}
\end{align}
where the KL divergence term is restricted within the range of $\epsilon$. Different from Equation \ref{eq: fb}, here we normally have a few dimensions dominating the KL divergence while most dimensions carry little useful information.

There have been also attempts of reducing the KL-vanishing problem by using CNN decoders~\citep{chen2016variational,yang2017improved,shen2017deconvolutional}, where the flexibility of the decoder can be adjusted to trade-off the reconstruction error and KL divergence.

\section{Summary}
In this section, we go over popular optimization techniques for latent-variable models. We also listed the current open challenges, pros and cons of different approaches. In general, the recommended methodology is as stated in the beginning of the section: always start from exact marginalization if possible. When exact marginalization is impossible even with dynamic programming, top-k approximation or pseudo labels, variational approximation can be considered. If variational methods are used, free-bits are currently the most stable regularization to prevent the posterior collapse problem. Advanced techniques, like the flow methods and adversarial training, are not recommended unless the flexibility of the latent variable distribution is confirmed to be a key bottleneck, or the main goal is to explore different alternatives to model the distribution. When the reparameterization trick is applicable, we can simply apply it to estimate the gradient. Otherwise, we can try either EM algorithm, soft relaxation or reinforcement learning. To keep the training more stable, a common practice is to warm up the model with some distant supervision first to initialize it with a decent starting point. In the following contents, we will go through specific applications of latent-variable models and apply techniques from this section to solve them.
\cleardoublepage

\chapter[VAE with Improved decoder]{VAE with Improved decoder}
\label{chap: improve-vae}

\lettrine[lines=3]{T}his chapter presents a novel VAE variant with improved decoding. Variational encoder-decoders (VEDs) have shown promising results in dialogue generation. However, the latent variable distributions are usually approximated by a much simpler model than the powerful RNN structure used for encoding and decoding, yielding the KL-vanishing problem, as mentioned in the last chapter. We propose to separate the training step into two phases: The first phase learns to autoencode discrete texts into continuous embeddings, from which the second phase learns to generalize latent representations by reconstructing the encoded embedding. In this case, latent variables are sampled by transforming Gaussian noise through multi-layer perceptrons and are trained with a separate VED model, which has the potential of realizing a much more flexible distribution. We compare our model with current popular models and the experiment demonstrates substantial improvement in both metric-based and human evaluations~\citep{shen2018improving}.

\section{Introduction}
Recurrent neural networks (RNNs)~\citep{bengio2003neural} are widely used in natural language processing tasks. However, given the history context, RNNs estimate the probability of one word at a time and does not work from a holistic sentence representation~\citep{bowman2016generating}. When applied to dialogue generation, the corresponding result is that it would generate either short, boring responses or long, inconsistent sentences. As the length of generated sentences grows, it would easily deviate from the original intention as such token-level estimation only considers immediate short rewards and neglects global structure consistency. In hence, vanilla RNNs prefer generating generic and safe short responses to avoid the risk of making errors~\citep{vinyals2015neural,serban2015building,shen2017conditional}. One way of improving this deficient generating process is to introduce a sentence-level representation, which can be further conditioned on to ensure the sentence-level consistency.

Deep latent-variable models are a popular way to learn such representations in a generative setting.  Latent representations and generators can be jointly trained in an unsupervised way. By learning the probability of synthesizing real data from intermediate latent variables, they are expected to uncover and disentangle causal factors that are most important to explain the data. The exact log-likelihood normally requires integral in high-dimensional space and cannot be analytically expressed. Current approaches solve this intractability problem by imposing a recognition network to approximate the real posterior probability. Variational autoencoders (VAEs)~\citep{kingma2014auto,rezende2014stochastic} bring scalability and stability to the training procedure, which introduces a reparameterization trick to reduce the variance when estimating the backpropagated gradients.

\citep{serban2016hierarchical} proposed the Latent Variable Hierarchical Recurrent Encoder-Decoder (VHRED) structure which applied the conditional VAE (CVAE)~\citep{sohn2015learning} with RNN encoder-decoders in dialogue generation, in hope of CVAE's advantage of learning global representations being a good complement of RNN's power at modeling local dependencies. However, this simple combination runs into the KL-vanishing problem that the RNN part ends up explaining all the structures without making use of the latent representation. The reason is that RNN is a universal approximator with much more flexibility than the simple gaussian distributed latent variables so that the model lacks enough motivation to utilize them. 

Current approaches normally address this problem by weakening the RNN decoder to match the simpler latent variable distribution, which essentially sacrifices the generating capacity for better representation learning and is inappropriate when our main goal is to learn a generative model. In this paper, on the contrary, we take advantage of the universality of RNNs to help realize a more flexible latent variable distribution. By this means, we can not only add motivation for utilizing latent variables, but also strengthen the expressiveness of the generating model. Specifically, we split the whole structure into a CVAE module and an autoencoder (AE) module. The CVAE module learns to generate latent variables while the AE module builds the connection between them and real dialogue utterances. The outputs of the CVAE serve as input latent variables for the AE module, which is potentially much more flexible than restricting the latent variables to follow a fixed distribution. As the RNN encoder-decoders in the AE module are universal approximators, they are adjusted to extract continuous vectors from the dialogue data that can be more easily modelled by the CVAE module. Combined with a scheduled sampling trick, this structure can significantly improve the generating performance. We show this structure can be compared to an adversarial encoder-decoder which substitutes the GAN step with a VAE alternative. Though theoretically less accurate, our framework is preferred to AED as the training process of VAE is much more reliable than GAN in seq2seq tasks and the universality of RNN ensures this inaccuracy can be controlled within an acceptable range.

\section{VED in Dialogue Generation}
In this section, we review the VAE and VHRED structure, then analyze where the training difficulty comes from when applied in dialogue generation and how current approaches try to solve this problem.
\subsection{VAE and VHRED}
The variational autoencoder (VAE)~\citep{kingma2014auto,rezende2014stochastic} is a popular generative model. Its generating process is as follows: data $x$ is generated by the generative distribution $p_\theta(x|z)$ and $z$ is sampled from the prior distribution $p(z)$. In contrast to calculating the exact log-likelihood, it can be efficiently trained by optimizing a valid lower bound~\citep{jordan1999introduction}. The objective takes the following form:
\begin{align}
\label{eq: vae}
\begin{split}
	&-\log  p_{\theta}(x) \leq -\log p_\theta(x)+\text{KL}(q_\phi(z|x)||p_\theta(z|x)) \\
  &= -\mathbb{E}_{q_{\phi}(z|x)} [\log p_\theta(x|z)]
  	+ \text{KL}(q_{\phi}(z|x || p(z)) 
\end{split}
\end{align}
$p_\theta(z|x)$ is the real posterior distribution of $z$ given the prior distribution $p_\theta(z)$ and the likelihood $p_\theta(x|z)$. The optimizing objective is namely maximizing the likelihood $\log p_\theta(x)$ and at the same time minimizing the mismatch between the approximated posterior $q_\phi(z|x)$, which is parametrized by neural networks, and the real posterior $p_\theta(z|x)$. When the gap $\text{KL}(q_\phi(z|x)||p_\theta(z|x))$ is large, the objective becomes inconsistent and the generating process cannot recover the real data distribution even in the global optimum.

The whole process can be conditioned on an additional context $c$, which leads to the conditional VAE~\citep{sohn2015learning} (CVAE): the output $x$ is generated from the distribution $p_\theta(x|c,z)$, latent variable $z$ is drawn from the prior distribution $p_\theta(z|c)$. The variational lower bound of CVAE is written as follows:
\begin{align}
\label{eq: improv_cvae}
\begin{split}
	-\mathbb{E}_{q_\phi (z|x,c)} [\log p_\theta(x|c,z)]
                       +  \text{KL}(q_\phi (z|x,c) \| p_\theta (z|c))
\end{split}
\end{align}                     
Specially, to some extent, when both the context $c$ and output $x$ are sequential data, CVAE can also be treated as a seq2seq model~\citep{sutskever2014sequence}.

VHRED~\citep{serban2016hierarchical} is a CVAE with hierarchical RNN encoders, where the first-layer RNN encodes token-level variations and the second-layer RNN captures sentence-level topic shifts. In this case, $c$ in Equation. \ref{eq: improv_cvae} stands for dialogue history, $x$ is the response to be decoded and $z$ is the latent variable reflecting the high-level representation of $x$. The distribution $q_\phi(z|x,c)$ and $p_\theta(z|c)$ are usually set as simple Gaussian distributions with diagonal covariance matrix.

\subsection{Optimization Challenges}
\label{sec: why}
In VHRED, straightforwardly optimizing with Equation. \ref{eq: improv_cvae} suffers from the KL-vanishing problem because the RNN decoder $p_\theta(x|c,z)$ is a universal function approximator and tends to represent the distribution without referring to the latent variable. At the beginning of the training process, when the approximate posterior $q_\phi(z|x,c)$ carries little useful information, it is natural for the model to blindly set $q_\phi(z|x,c)$ closer to the Gaussian prior $p_\theta(z|c)$ so that the extra cost from the KL divergence can be avoided~\citep{chen2016variational}.

To better analyze where the optimizing comes from, we can rewrite Equation.~\ref{eq: improv_cvae} as the following:
\begin{align}
\label{eq: why-cvae}
\begin{split}
-\log \int_{z} p_\theta(z|c) p_{\theta}(x|z,c) dz+\text{KL}(q_\phi(z|x,c)||p_\theta(z|x,c))
\end{split}
\end{align}

Let's first take a look at the first item, $\log \int_{z} p_\theta(z|c) p_{\theta}(x|z,c) dz = \log p_\theta(x|c)$. When the family of $p_\theta(x|z,c)$ is complex enough and includes the real distribution of $x$, the optimal value of this item is $p(x|c)$ and the reliance on $z$ is not necessary. However, reliance on $z$ provides the model with a chance of taking advantage of $z's$ distribution and reduces the complexity requirement for the distribution family $p_\theta(x|z,c)$. For example, suppose $p(x|c)=\mathcal{N}(0,1)$ and $p_\theta(z|c)=\mathcal{N}(3,1)$, modeling $p(x|c)$ accurately without reliance on $z$ requires $p_\theta(x|z,c)$ to include the Gaussian distribution, while by means of the linear mapping between $x$ and $z$ $p_\theta(x|z,c)$ can describe the real distribution with only linear complexity. When Gaussian distribution is not covered in the family $p_\theta(x|z,c)$, this model has to exploit the relation between $x$ and $z$ to model the real distribution. Likewise, in dialogue generation, although the RNN decoder $p_\theta(x|c)$ can in theory approximate arbitrary function, perfectly fitting the real dialogue distribution is still difficult due to the optimizing challenge, training corpus size and approximating errors. Therefore, to achieve the global optimum, we believe this first item will always prefer utilizing the latent variables, so long as the decoder $p_\theta(x|z,c)$ is not perfect. The weaker the decoder family is, the more it will be biased to utilizing latent variables. A more flexible prior distribution $p_\theta(z)$ will also increase the chance as it provides more possibilities for utilisation.

The second item is the KL divergence, whose minimum value is 0 if and only if $q_\phi(z|x,c)=p_\theta(z|x,c)$. According to the Bayes theorem, we can express $p_\theta(z|x,c)$ as:
\begin{equation}
p_\theta(z|x,c)=\frac{p_\theta(x|z,c)p_\theta(z|c)}{p_\theta(x|c)}
\end{equation}
By ignoring the latent variable $z$, $p_\theta(x|z,c)$ and $p_\theta(x|c)$ cancel out, setting $q_\phi(z|x,c)=p_\theta(z|c)$ can easily arrive at the global optimum 0. Otherwise, when $p_\theta(z|c)$ is parametrised as a mean-field Gaussian distribution as in VHRED, the real posterior is impossible to fall into the same distribution family. Firstly, the independence relation cannot be satisfied. To make dimensions of $p_\theta(z|x)$ independent of each other, the likelihood $p_\theta(x|z)$ must exactly disentangle the effect of every dimension, which is unrealistic when $p_\theta(x|z)$ is a categorical distribution modelled by the RNN softmax. Secondly, the real posterior distribution can hardly still follow a Gaussian distribution when the likelihood $p_\theta(x|z)$ is based on discrete sequential data. Normally the training process will adjust $p_\theta(x|z)$ to make the real posterior easier to be modelled by $q_\phi(z|x)$~\citep{hinton1995wake}. However, when $x$ represents sentences with variable length, the value of $p_\theta(x|z)$ vanishes greatly when the length grows, which makes the adjusting task much more difficult. This implies the second item will always prefer ignoring the latent variables, so long as the approximated posterior is not powerful enough to perfectly match the real posterior. The weaker the approximating posterior distribution family is, the more it will be biased to ignoring latent variables.

Above all, the objective function of variational encoder-decoders in dialogue generation is essentially the competition of these two items, who is biased to utilizing or ignoring latent variables respectively. The reason of the KL divergence vanishing in the global optimum is that the second term can gain more from ignoring the latent variables than the first term from utilizing them.

\subsection{Current Approaches}
If we use the ELBO objective, as explained, there are two directions to prevent the KL-vanishing problem: improving the advantage of utilising latent variables in $\log \int_{z} p_\theta(z|c) p_{\theta}(x|z,c) dz$ or weakening the advantage of abandoning latent variables in $\text{KL}(q_\phi(z|x,c)||p_\theta(z|x,c))$.

For the former direction, we need to use a smaller distribution family to model the decoder $p_\theta(x|z,c)$. When the decoder is weaker, if ignoring latent variables, it becomes farther from the real distribution at the global optimum thus encouraging latent variables to be exploited. Word drop-out~\citep{bowman2016generating} is a common method to weaken the RNN decoder. At each time step, the input word has a certain chance (drop-out rate) of becoming another word, the RNN decoder therefore cannot store a continuous history context. In \citep{xie2017data}, word drop-out is also explained as a special kind of smoothing. Similarly, for CNN decoders, limiting their power can also encode more information to latent variables~\citep{yang2017improved,chen2016variational}. Bag-of-word loss proposed by \citep{zhao2017learning} can also fall into this category. It imposes an extra loss which forces the latent variable to predict the whole sentence without word inputs, which is essentially increasing the weight of the reconstruction loss with the drop-out rate set to 1.

For the latter direction, we need to use a more flexible prior or posterior distribution for latent variables. Once the approximated posterior distribution is powerful enough, the KL divergence can be close to zero without losing the dependence on latent variables. \citep{serbanpiecewise} applies a piecewise distribution to replace the Gaussian prior distribution. Though can represent multi-modal conditions, it is still limited as a fixed distribution with pre-defined number of modes.  \citep{salimans2015markov} samples latent variables through Markov chains, but it imposed an extra approximation and the objective becomes less accurate. \citep{rezende2015variational,kingma2016improved,chen2016variational} use a normalizing flow. Latent variables are first sampled from a simple distribution then passed through several invertible transformations to get better flexibility. Normalizing flow is computationally more costly and has not been applied in text generation yet.

We can also change the original ELBO objective for easier optimization. KL-annealing~\citep{bowman2016generating} and free bits~\citep{kingma2016improved} are two popular strategies. In KL-annealing, a small weight is added to the KL divergence term in Equation. \ref{eq: improv_cvae}, which starts from zero and gradually increases to 1. This prevents the model from zeroing out the KL divergence at the earlier training stage. Once the KL divergence vanishes, it is difficult to be recovered for the short sight nature of gradient descent. Free bits reserve some space of KL divergence for every dimension of latent variables. KL divergence is only optimized when exceeding the predefined quota. Similar ideas can be found in \citep{yang2017improved}, which reserved space for the total KL divergence instead of for every dimension.

\section{Improving Variational Encoder-Decoders}
\begin{figure*}[!ht]
\centering
\centerline{\includegraphics[height=9cm]{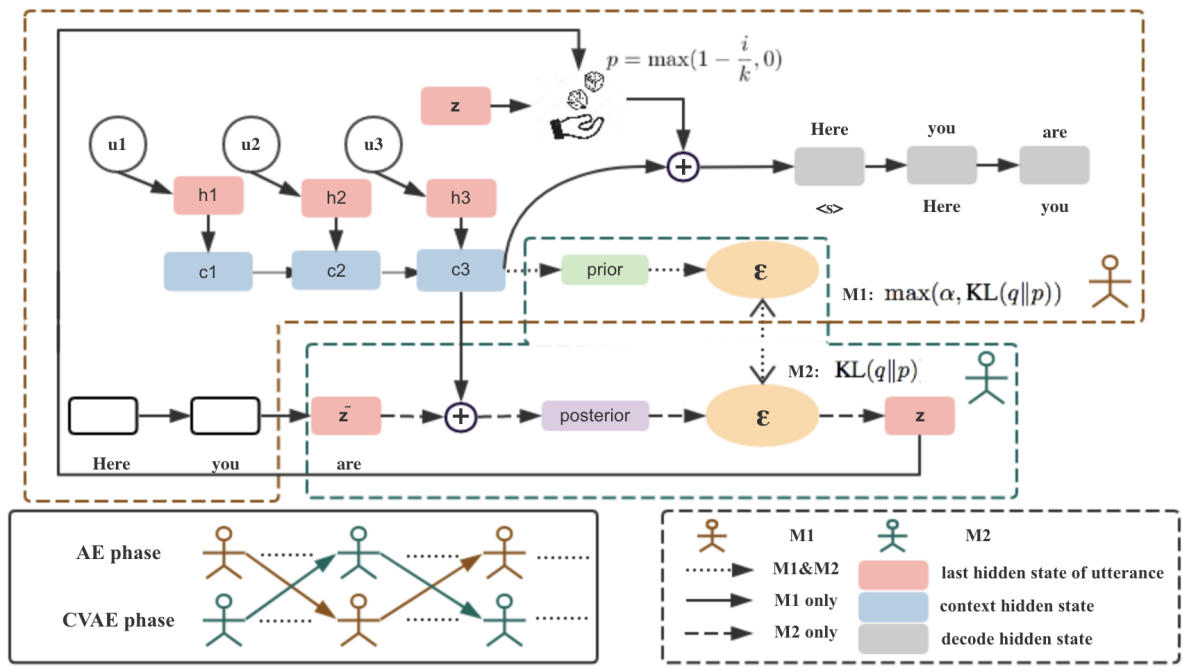}}
\caption{Architecture for collaborative variational encoder-decoder. $\bigoplus$ denotes concatenation of information. $M_1 (AE)$ and $M_2 (CVAE)$ are represented in brown and green respectively.}
\label{fig: improv_model}
\end{figure*}
As discussed above, two ways for alleviating the optimizing challenge includes weakening the RNN decoders and improving the flexibility of latent variable distributions. The latter class is more fundamental since it also brings more expressiveness to the generating model. Weakening the decoders, though attenuating the KL-vanishing problem, will inevitably hurt the overall performance.

\subsection{Adversarial Encoder-Decoder}
An ideal way of representing the latent variable distribution is to use a universal approximator like neural networks. \citep{makhzani2015adversarial} proposed adversarial autoencoder (AAE) which samples posterior latent variables by transforming Gaussian noise through multi-layer-perceptrons. The flexibility of neural networks ensures it can fit arbitrary distribution. However, the probability density is intractable, so adversarial learning~\citep{goodfellow2014generative} must be implemented to replace the original KL divergence term.

\begin{figure}[!ht]
\centering
\centerline{\includegraphics[width=0.6\columnwidth]{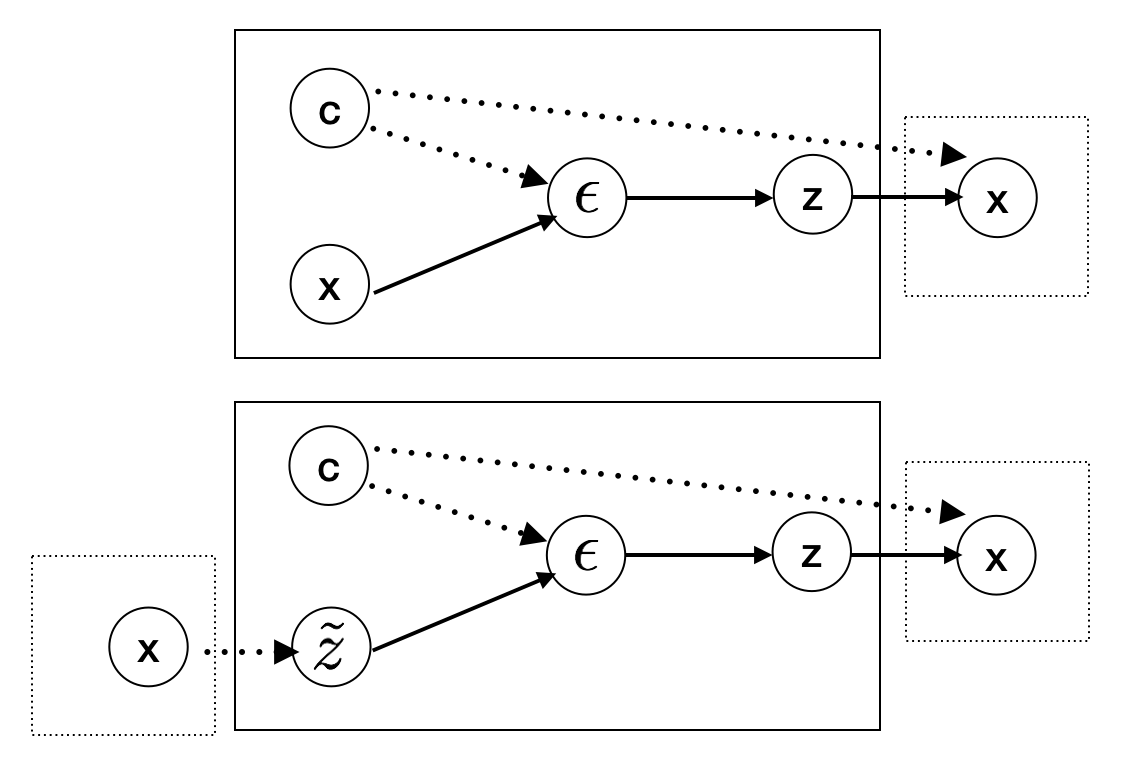}}
\caption{Up: adversarial encoder-decoder, down: adversarial encoder-decoder after replacing GAN with VAE, Full line rectangle: GAN and CVAE phase, Dotted line rectangle: AE phase}
\label{fig:connection}
\end{figure}

We can apply this idea to dialogue generation, where AAE is changed to context-dependent adversarial encoder-decoder (AED). The training objective can be represented as:
\begin{align}
\label{eq: aed}
\begin{split}
-\mathbb{E}_{q_\phi(z|c,x)}p_\theta(x|c,z)+JS(q_\phi(z|c)||p_\theta(z|c))
\end{split}
\end{align}
The training alternates between the autoencoder (AE) phase to optimize $-\mathbb{E}_{q_\phi(z|c,x)}p_\theta(x|c,z)$ and the GAN phase to match the aggregated posterior $q_\phi(z|c)$ and the prior $p_\theta(z|c)$. $q_\phi(z|c,x)$ and $p_\theta(z|c)$ are implicitly defined by passing context-dependent Gaussian random variables $\epsilon$ through multi-layer perceptrons. The graphical model is depicted in Figure. \ref{fig:connection}. It can be shown that this objective differs from the original ELBO by adding an extra punishment to the entropy of $q_\phi(x|z,c)$ and using Jensen-Shannon divergence in lieu of KL divergence. In the non-parametric limit, its generating model can recover the exact data distribution.
\subsection{Replacing GAN with VAE}
The idea of AED sounds appealing, but GAN is notoriously difficult to train, especially when both the prior and posterior need to be updated towards each other, the model becomes extremely sensitive to hyper-parameters and the training is very unstable. In consequence, we try replacing the GAN phase with a CVAE alternative. An RNN encoder is first applied to extract the corresponding latent variable target $\tilde{z}$ for each dialogue turn $x$, based on which a CVAE is trained to reconstruct it through context-dependent Gaussian noise. The connection to AED can be seen in Figure \ref{fig:connection}. Specifically, we just replace the $JS(q_\phi(z|c)||p_\theta(z|c))$ in Equation \ref{eq: aed} with the following CVAE objective:
\begin{align}
\label{eq: aed-cvae}
\begin{split}
-\mathbb{E}_{q_\phi(\epsilon|c,\tilde{z})}p_\theta(z|c,\epsilon)+KL(q_\phi(\epsilon|c,\tilde{z})||p_\theta(\epsilon|c))
\end{split}
\end{align}
$q_\phi(\epsilon|c,\tilde{z})$ is an approximated posterior. It can be easily proved when $q_\phi(\epsilon|c,\tilde{z})$ is powerful enough to cover the real posterior $p_\theta(\epsilon|c,\tilde{z})$, objective \ref{eq: aed-cvae} has the same global optimum as in $JS(q_\phi(z|c)||p_\theta(z|c))$. We can therefore instead alternate between the AE phase and the CVAE phase to achieve the same effect as in AED.
\subsection{Constraining RNN Encoder}
The accuracy of the CVAE objective relies on the matching degree of $q_\phi(\epsilon|c,\tilde{z})$ and $p_\theta(\epsilon|c,\tilde{z})$. Therefore, in the AE phase, apart from encoding representative information to reduce the normal AE reconstruction loss, the RNN encoder should also encode utterances in a manner where the real posterior $p_\theta(\epsilon|c,\tilde{z})$ can be more easily modelled by the distribution defined by $q_\phi(\epsilon|c,\tilde{z})$ in the CVAE phase. To do this, we add a KL divergence constraint to the RNN encoder in the AE phase. The RNN encoder has to keep $KL(q_\phi(\epsilon|c,\tilde{z})||p_\theta(\epsilon|c))$ within a specific range. It is also possible to constrain the value of the whole CVAE objective of Equation. \ref{eq: aed-cvae}, but we find constraining only the KL divergence is enough when the alternating step is not too large. Note that in the encoder phase, the model can only adjust the RNN encoder-decoders to control the KL divergence, the generating parameters for latent variables are fixed.

\subsection{Scheduled Sampling Trick}
In the AE phase, we also find it useful to initially use the ground-truth encoding $\tilde{z}$ then gradually change to noisier CVAE output $z$. We apply the scheduled sampling strategy proposed in \citep{bengio2015scheduled}. Before decoding, a coin is flipped to decide whether to feed the real hidden vector $\tilde{z}$ or the noisy $z$. In the beginning, to make it easy, we mostly pick the real $\tilde{z}$. As the training proceeds, we gradually improve the difficulty by increasing the chance of selecting noisy $z$ until finally all inputs are replaced with the $z$. We decide the chance of selecting the real $\tilde{z}$ with a linear decay function as:
\begin{equation}
\label{eq: sampling}
p=\max (1-\frac{i}{k},0)
\end{equation}
$i$ is the step number and $k$ is a constant controlling the decaying speed. Other decaying functions are also applicable like exponential decay or inverse sigmoid decay.
\subsection{Training Process}
Our model contains a CVAE phase and an AE phase. These two phases are trained iteratively until an equilibrium is achieved. 

In the CVAE phase, A sample $\tilde{z}$ is obtained from the AE by transforming dialogue texts into a continuous embedding and is used as a target for the maximum likelihood training of the CVAE. We assume the generative model $p_\theta(z|\epsilon,c)=\mathcal{N}(\tilde{z},I)$, the loss function is:
\begin{align}
\begin{split}
\min_{\phi} \text{KL}(q_\phi (\epsilon|\tilde{z},c) \| p_\phi &(\epsilon|c))+\frac{1}{2}\mathbb{E}_{q_\phi (\epsilon|\tilde{z},c)}||g_{\phi}(\epsilon)-\tilde{z}||^2_2;\\
&\tilde{z}=f_\theta(x)
\end{split}
\end{align}
$f_\theta$ is the RNN encoder and is fixed as part of the AE module during training. 

In the AE phase, An observation x is sampled from the training data and fed into the transform function to get a continuous vector representation $\tilde{z}=f_\theta(x)$. The corresponding latent variable $z$ is sampled from the posterior distribution $q_\phi(z|\tilde{z},c)$ provided by the CVAE part. The sampled latent variable $z$, together with $x$, forms a target for training the AE. The objective function is:
\begin{align}
\label{eq: m1}
\begin{split}
\min_{\theta}\: &\max(\alpha, \text{KL}(q_\phi (\epsilon|\tilde{z},c) \| p_\phi (\epsilon|c)) )\\&-\mathbb{E}_{q_\phi (z|\tilde{z},c)}[log(p_\theta(x|z,c))];\\
&\tilde{z}=f_\theta(x),z=(1-p)g_\phi(\epsilon)+p\tilde{z}
\end{split}
\end{align}
The first item is used to control KL divergence in a reasonable range such that the transformed $z$ can be more easily modelled by the CVAE phase. $\alpha$ can be used to adjust the leverage between the reconstruction loss and KL divergence, where a lower $\alpha$ value will lead to a lower KL divergence in the end. $p$ is the keeping rate defined in Equation. \ref{eq: sampling}. The detailed architecture is depicted in Figure \ref{fig: improv_model}. We refer to this framework as collaborative VED where the AE and CVAE phase collaborate with each other to achieve a better generating performance.

\subsection{Model Summary}In summary, we replace the GAN phase of AED with a CVAE alternative. The output of the CVAE part are latent variables, which can represent a much broader distribution family than mean-field Gaussian. As CVAE is in theory less accurate than GAN because it needs to approximate the real posterior, we leverage the more powerful RNN encoder-decoders. In the AE phase, they should autoencode utterenaces to make the real posterior easily representable by the CVAE part.
\section{Experiments}
We conduct our experiments on two dialogue datasets: Dailydialog~\citep{li2017dailydialog} and Switchboard~\citep{godfreyswitchboard}. Dailydialog contains 13118 daily conversations under ten different topics. This dataset is crawled from various websites for English learner to practice English in daily life. Statics show that the speaker turns are roughly 8, and the average tokens per utterance is about 15, which are appropriate for training dialog models. Switchboard has 2400 two-sided telephone conversations under 70 specified topics with manually transcribed speech and alignment. Compared with Dailydialog, the turn of every dialogue is much longer and the subject is more disperse. These two datasets are randomly separated into training/validation/test sets with the ratio of 10:1:1.

\subsection{Models and Training Procedures}
For comparison, we also implemented the hred model (seq2seq model with hierarchical RNN encoders), which is the basis of VHRED. Latent variable models are trained by standard KL-annealing with different weights~\citep{bowman2016generating,higgins2017beta}, with additional BOW loss~\citep{zhao2017learning,semeniuta2017hybrid}, word drop-out~\citep{bowman2016generating}, free bits~\citep{kingma2016improved} and our collaborative VED (CO) with the scheduled sampling trick (SS). For our framework, we use the encoder RNN as the transformation function $f_\theta(x)$. We tuned the parameters on the validation set and measure the performance on the test set. In all experiments, the letters are all transformed to the lower-case, the vocabulary size was set as 20,000 and all the OOV words were mapped to a special token $<$unk$>$. We set word embeddings to size of 300 and initialized them with Word2Vec embeddings trained on the Google News Corpus.  The first, second-layer encoder and decoder RNN in the following experiments are single-layer GRU with 512, 1024 and 512 hidden neurons. The dimension of latent variables is set to 512. The batch size is 128 and we fix the learning rate as 0.0002 for all models. Our framework is trained epochwise by alternatively training the CVAE and DAE part. The probability estimators for VAE are 2-layer feedforward neural networks. At test time, we output the most likely responses using beam search with beam size set to 5~\citep{graves2012sequence} and $<$unk$>$ tokens were prevented from being generated. We implemented all the models with the open-sourced Python library Tensorflow~\citep{abadi2016tensorflow} and optimized using the Adam optimizer~\citep{kingma2014adam}. Dialogs are cut into set of slices with each slice containing 80 words then fed into the GPU memory.
\subsection{Metric-based Evaluation}
We compare our model with the basic HRED and several current approaches including KL-annealing (KLA), word drop-out (DO), free-bits (FB) and bag-of-words loss (BOW). The details are summarized in Table \ref{tab: metric} and \ref{tab: embedding}. For KLA, we initialize the weight with 0 and gradually increase to 1 in the first 12000 or 25000 training steps for Dailydialog and Switchboard respectively. The word drop-out rate is fixed to 25\%. Words are dropped out only in the training step. We set the reserved space for every dimension as 0.01 in free bits (FB) and also try reserving 5 bits for the whole dimension space (FB-all). We use an $\alpha$ value 5 for our collaborative model (CO) and set the scheduled sampling (SS) weight $k=2500$ or 5000 for Dailydialog or Switchboard. We also experiment with jointly training the AE and CVAE part in our model and report the results.
\begin{table}[htbp!]
\centering
\caption{Metric Results, left: Dailydialog, right: switchboard}
\label{tab: metric}
\begin{tabular}{llll}
 \textbf{Model}&\textbf{PPL}  &\textbf{KL}  &\textbf{NLL}   \\
 \hline
 HRED&43.4$\vert$48.3  &0.00$\vert$0.00  &229.1$\vert$355.6    \\
 KLA&31.8$\vert$44.5  &4.90$\vert$4.36  &225.0$\vert$331.6  \\
 KLA+DO&29.8$\vert$40.1  &3.80$\vert$4.48  &223.9$\vert$317.0   \\
 KLA+BOW&26.8$\vert$30.9&12.8$\vert$8.92&247.3$\vert$321.1\\
 FB&41.7$\vert$32.1&\textbf{3.34}$\vert$\textbf{3.90}&239.0$\vert$322.7\\
 FB-all&29.4$\vert$21.7&5.01$\vert$4.97&226.1$\vert$308.2\\
 \hline
 CO&26.1$\vert$36.5&4.90$\vert$4.94&223.6$\vert$289.7\\
 CO+DO&25.1$\vert$34.4&5.01$\vert$4.93&218.7$\vert$273.4\\
 \textbf{CO+SS}&\textbf{23.8}$\vert$\textbf{31.8}&4.92$\vert$4.93&\textbf{213.2}$\vert$\textbf{273.4}\\
 CO+SS(joint)&28.5$\vert$39.6&5.16$\vert$5.02&224.3$\vert$301.3\\
\end{tabular}
\end{table}
Table \ref{tab: metric} measured the perplexity (PPL), KL divergence (KL) and negative log-likelihood (NLL). NLL is averaged over all the 80-word slices within every batch. For latent-variable models, NLL is computed as the ELBO, which is the lower bound of the real NLL.

As can be seen, our model CO+SS achieves the lowest NLL over both datasets. The Schedule Sampling (SS) strategy significantly helps brings down the NLL. Word drop-out (DO), though weakening the RNN decoder, improved the performance when combined with both KLA and CO, which verified the assumption that DO can function as a smoothing technique in neural network language models~\citep{xie2017data}. KLA itself needs early stop, otherwise the KL divergence will vanish once the weight increases to 1. BOW avoids the KL-vanishing problem, but the overall performance will significantly decrease because adding an additional loss in theory leads to a biased result for latent variables. BOW information is encoded into the latent variable, but it prevents the decoder from stably learning the word order pattern in the training step thus sacrifices the NLL performance. FB-all performs much better than FB, which suggests most important information is concentrated on a few dimensions. Equally reserving space for every dimension is not suitable. Finally, we also testified the necessity of iteratively training our model. Jointly training the model brings recession on both the perplexity and KL divergence on the two datasets.

\begin{figure}[!ht]
\centering
\includegraphics[width=8cm,height=4cm]{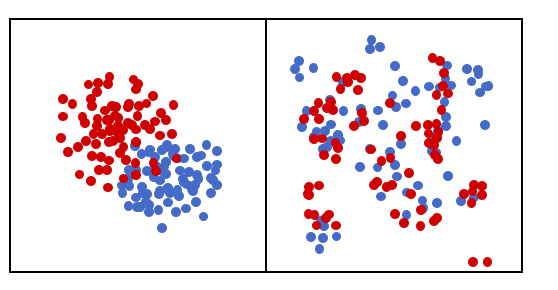}
\caption{T-SNE visualization of sampled latent variables. left: VHRED, right: CO+SS. Red dots correspond to samples from prior distribution, while the blue dots correspond to samples from posterior distribution. Viewable in color mode only.}
\label{fig:scatter}
\end{figure}

Figure. \ref{fig:scatter} visualizes the latent variables drawn from VHRED and our framework. We randomly pick a dialogue context ``I'd like to invite you to dinner tonight , do you have time ?" and apply the information retrieval based method to gather 10 responses with similar context from the corpus. All the 10 responses are verified by humans as appropriate ones, which span over different possibilities like ``Thank you for your invitation. ", ``Don't be silly . Let's go Dutch ." and ``Are you asking me for a date ? ". For each response, 10 samples are drawn from the posterior latent variable distribution, which forms 100 posterior latent variable samples (blue dots) in total. Likewise, 100 samples are drawn from the prior latent variable distribution (dots) given only the dialogue context. The visualization clearly indicates the superiority of our framework in modelling more flexible prior and posterior latent variable distributions. In the VHRED model, both the prior and posterior distributions are limited uni-modal Gaussians with only a little overlap. In our framework, the distributions are more diverse and samples from the prior and posterior distribution share more overlap with each other.

\begin{table}[htbp!]
\centering
\caption{Embedding Results, left: dailydialog, right: switchboard}
\label{tab: embedding}
\begin{tabular}{llll}
 \textbf{Model}&\textbf{Average}  &\textbf{Greedy}  &\textbf{Extrema}   \\
 \hline
 HRED&0.463$\vert$0.334  &0.445$\vert$\textbf{0.399}  &0.356$\vert$0.280    \\
 KLA&0.442$\vert$0.317  &0.436$\vert$0.327  &0.327$\vert$0.267   \\
 KLA+DO&0.458$\vert$0.325  &0.461$\vert$0.341  &0.378$\vert$0.283  \\
 KLA+BOW&0.475$\vert$0.340&0.459$\vert$0.352&0.386$\vert$0.302\\
 FB&0.423$\vert$0.336&0.414$\vert$0.348&0.349$\vert$0.318\\
 FB-all&0.429$\vert$0.341&0.439$\vert$0.352&0.357$\vert$0.325\\
 \hline
 CO&0.465$\vert$0.377&0.465$\vert$0.381&0.394$\vert$0.331\\
 CO+DO&0.489$\vert$0.385&0.471$\vert$0.379&0.397$\vert$0.337\\
 \textbf{CO+SS}&\textbf{0.539}$\vert$\textbf{0.392}&\textbf{0.477}$\vert$0.394&\textbf{0.443}$\vert$\textbf{0.340}\\
  CO+SS(joint)&0.420$\vert$0.347&0.452$\vert$0.360&0.351$\vert$0.308\\
\end{tabular}
\end{table}

Table \ref{tab: embedding} reports the results of the embedding-based topic similarity: Embedding Average (Average), Embedding Extrema (Extrema) and Embedding Greedy
(Greedy)~\citep{liu2016not}. Unlike the NLL, who measures the token-level match, these embedding-based metrics map responses to a vector space and compute the cosine similarly with golden answers, which can to a large extent measure the sentence-level semantic similarity.

We can see our model still achieved the highest topic similarity according to all the three metrics. This suggests our model can bring improvement for both token-level coherence and sentence-level topic match. BOW, though not good at the NLL metric, performed remarkably well on this metric, which implies BOW is beneficial for the decoder to generate the correct high-level meaning but fails to transform the meaning to a fluent sentence. In contrast, FB has a relative lower on-topic similarity score compared with its performance on the token-level likelihood.  
\subsection{Human Evaluation}
\begin{table*}
\caption{{\label{tab: improv_examples}}Examples of context-response pairs for the neural network models. \_\_eou\_\_ denotes end-of-utterance and indicates the start of a new turn.}
\centering
\begin{tabular}{p{60mm}|p{90mm}}
  \textbf{Context} & \textbf{Response}\\
  \hline
  The taxi drivers are on strike again . \_\_eou\_\_ What for ? \_\_eou\_\_ They want the government to reduce the price of the gasoline . \_\_eou\_\_ &  \textbf{KLA:} \ all right . i'll take it . \newline \textbf{KLA+BOW:} \ well , how much is the price of the price price ? \newline \textbf{CO-SS:} \ that's good . i want to buy a new car .  \newline \textbf{Gold:} \ It is really a hot potato .  \\ \hline
   I'd like you to do me a favor . \_\_eou\_\_ What is it ? \_\_eou\_\_ Can you mail this at the post office for me ? I have to go to the doctor for an appointment and I don't have enough time to go to the post office . \_\_eou\_\_ &  \textbf{KLA:} \ i'm sorry , but i'm not sure . i don't want to have a look at the moment . \newline \textbf{KLA+BOW:} \ oh , i'm sorry , but it's a pleasure to be in my room and again-b .  \newline \textbf{CO+SS:} \ ok , i will . do you have any other questions about the company ?   \newline \textbf{Gold:} \ Sure . And I'd like you to get some cold medicine for me when you go to the doctor's .  \\ \hline
   I heard you've found a new job . \_\_eou\_\_ Yes , I'm now working in a company . \_\_eou\_\_ So you're quite happy with your new job ? \_\_eou\_\_ &  \textbf{KLA:} \ to be honest , but i'm not familiar with my friends .  \newline \textbf{KLA+BOW:} \ i'd like to . but i would like my mind .  \newline \textbf{CO+SS:} \ not really , but how about you , sue ?   \newline \textbf{Gold:} \ Right . I enjoy what I'm doing .  \\ \hline
\end{tabular}
\end{table*}

The accurate evaluation of dialogue systems is an open problem. To validate the previous metric-based results, we further conduct a human evaluation on several models. We randomly sampled 100 context from the test corpus and apply 6 different models to generate the best response with beam search. The evaluation is conducted only on the Dailydialog corpus since it is closer to our daily conversation and easier for humans to make the judgement. All the generated responses, together with the dialogue context, are then randomly shuffled and judged on the crowdsourcing website CrowdFlower. People are asked to judge the plausibility of the generated response by giving a binary score in three aspects: grammaticality, coherence with the dialogue context and diversity (ensure the response is not a dull sentence). 54 people are finally involved in evaluating the total 600 responses, each is judged by 3 different people and the score agreed by most people is adopted. We set each person can judge at most 50 responses and filter by manually-set test questions.

The results shows that our model generates highly fluent sentences compared to other approaches. KLA+BOW, as expected, receives the lowest score on fluency. Our model also achieves relative good scores on coherence and diversity, implying novel responses related to the conversation topic can be generated by our model. However, we notice the human evaluation is rather subjective and not reliable enough. If a sentence is influent, humans tend to reject it though the topic might be coherent and the content might be diverse. It is difficult to give an objective score separately for all the three aspects. We can see models with lower scores on fluency normally also receive lower scores on the other two fields like KLA+BOW and FB-all. Therefore, we consider this evaluation only as a complement to the metric-based results, indicating that humans agree with the generations of our models more than with the others.
\begin{table}[htbp!]
\centering
\caption{\label{tab:human scores}Human Judgements for models trained on Dailydialog corpus, F refers to fluent, C refers to coherence and  D refers to diversity.}
\label{tab: human}
\begin{tabular}{llll}
 \textbf{Model}&\textbf{F(\%)}  &\textbf{C(\%)}  &\textbf{D(\%)}   \\
 \hline
 KLA&76  &35  &50   \\
 KLA+DO&80  &41  &\textbf{57}   \\
 KLA+BOW&70&36&48\\
 FB-all&74&29&34\\
 \hline
 CO+DO&82&\textbf{49}&54\\
 \textbf{CO+SS}&\textbf{89}&44&51\\
\end{tabular}

\end{table}

Table \ref{tab: improv_examples} shows exampled generated responses. We can see the our improved collaborative VED model with scheduled sampling can more accurately identify the topic and generate more coherent responses. Standard KL-annealing tends to generate smooth sentences but irrelevant to the context. Imposing an additional BOW loss can increase the probability of correctly capturing the main topic, but the generated responses are sometimes grammatically wrong, as also has been shown from the metric-based results. In the first example, the context is about taxi drivers' request for reducing gasoline price, the response from KLA is a fluent natural sentence but not closely related to the context. Model KLA+BOW starts with a reasonable beginning but ends up with influent continuations. Though influent, KLA+BOW model does capture the main topic about price, indicating it can successfully predict the order-insensitive bag of words but fail to establish a natural sentence. In contrast, our model is not only a fluent sentence, but also close to the topic. More importantly, it brings some new information ``I want to buy a new car" and is helpful to an interactive conversation. Similar conditions can be seen in the other two examples.
\section{Conclusion}

Variational encoder-decoders and recurrent neural networks are powerful in representation learning and natural language processing respectively. Though recently quite a few work has started to apply them on dialogue generation, the training process is still unstable and the performance is hard to be guaranteed. In this work, we thoroughly analyze the reason of the training difficulty and compare different current approaches, then propose a new framework that allows effectively combining these two structures in dialogue generation. We split the whole structure into two parts for more flexible prior and posterior latent variable distributions. The training process is simple, efficient and scales well to large datasets.

We demonstrate the superiority of our model over other popular methods on two dialogue corpus. Experiments show that our model samples latent variables with more flexible distributions without sacrificing recurrent neural network's capability of synthesizing coherent sentences. Without losing generality, our model should be able to apply on any se2seq tasks, which we leave for future work.

The proposed training framework is still limited to the maximum likelihood objective. Later on we will show this objective is not suitable for dialogue generation and explain how we can improve over it.
\cleardoublepage

\chapter[VAE with Mutual Information Maximization]{VAE with Mutual Information Maximization}
\label{chap: nexus}

\newcommand{\indentitem}{\setlength\itemindent{25pt}}

\lettrine[lines=3]{A}s explained in the last chapter, sequence-to-Sequence (seq2seq) models, though being highly efficient in learning the backbone of human-computer communications, they suffer from the problem of strongly favoring short generic responses. In this chapter, we argue that a good response should smoothly connect both the preceding dialogue history and the following conversations. We strengthen this connection through mutual information maximization. To sidestep the non-differentiability of discrete natural language tokens, we introduce an auxiliary continuous code space and map such code space to a learnable prior distribution for generation purpose. Experiments on two dialogue datasets validate the effectiveness of our model, where the generated responses are closely related to the dialogue context and lead to more interactive conversations~\citep{shen2018nexus}.

\section{Introduction}
With the availability of massive online conversational data, there has been a surge of interest in building open-domain chatbots with data-driven approaches. Recently, the neural network based sequence-to-sequence (seq2seq) framework~\citep{sutskever2014sequence,cho2014learning} has been widely adopted. In such a model, the encoder, which is typically a recurrent neural network (RNN), maps the source tokens into a fixed-sized continuous vector, based on which the decoder estimates the probabilities on the target side word by word. The whole model can be efficiently trained by maximum likelihood (MLE) and has demonstrated state-of-the-art performance in various domains.
\begin{figure}
\centering
\fbox{\begin{minipage}{0.47\textwidth}
\textcolor{blue}{\textbf{$A_1$}}: Do you know the movie Star Wars?\\
\textcolor{red}{\textbf{$B_1$}}:  Only a bit. \textcolor{purple}{\underline{You can tell me about it!}}\\
\textcolor{blue}{\textbf{$A_2$}}: Of course! This is about ...
\end{minipage}
}
\caption{A conversation in real life}
\label{fig:example}
\end{figure}
However, this architecture is not suitable for modeling dialogues. Recent research has found that while the seq2seq model generates syntactically well-formed responses, they are prone to being off-context, short, and generic. (e.g., ``I don’t know" or ``I am not sure")~\citep{li2015diversity,serban2015building}. The reason lies in the one-to-many alignments in human conversations, where one dialogue context is open to multiple potential responses. When optimizing with the MLE objective, the model tends to have a strong bias towards safe responses as they can be literally paired with arbitrary dialogue context without semantical or grammatical contradictions. These safe responses break the dialogue flow without bringing any useful information and people will easily lose interest in continuing the conversation.

In this paper, we propose NEXUS Network which aims at producing more on-topic responses to maintain an interactive conversation flow. Our assumption is that a good response should serve as a ``nexus": connecting and being informative to both the preceding dialogue context and the follow-up conversations. For example, in Figure \ref{fig:example}, the response from $B_1$ is a smooth connection, where the first half indicates the preceding context is a ``Do you know" question and the second half informs that the follow-up would be an introduction about \textit{Star Wars}. We establish this connection by maximizing the mutual information (MMI) of the current utterance with both the past and future contexts. In this way, generic responses can be largely discouraged as they contain no valuable information and thus have only weak correlations with the surrounding context. To enable efficient training, two challenges exist.

The first challenge comes from the discrete nature of language tokens, hindering efficient gradient descent. One strategy is to estimate the gradient by methods like Gumbel-Softmax~\citep{maddison2016concrete,jang2016categorical} or REINFORCE algorithm~\citep{williams1992simple}, which has been applied in many NLP tasks~\citep{he2016dual,shetty2017speaking,gu2017neural,paulus2017deep}, but the trade-off between bias and variance of the estimated gradient is hard to reconcile. The resulting model usually strongly relies on sensitive hyper-parameter tuning, careful pre-train and task-specific tricks. \citep{li2015diversity,wang2017steering} avoid this non-differentiability problem by learning a separate backward model to rerank candidate responses in the testing phase while still adhering to the MLE objective for training. However, the candidate set normally suffers from low diversity and a huge sample size is needed for good performance~\citep{li2016simple}.

The second challenge relates to the unknown future context in the testing phase. In our framework, both the history and future context need to be explicitly observed in order to compute the mutual information. When applying it to generating tasks where only the history context is given, there is no way to explicitly take into account the future information. Therefore, reranking-based models do not apply here. \citep{li2016deep} addresses future information by policy learning, but the model suffers from high variance due to the enormous sequential search space. \citep{serban2016hierarchical,zhao2017learning,shen2017conditional} adopt the variational inference strategy to reduce the training variance by optimizing over latent continuous variables. However, they all stick to the original MLE objective and no connection with the surrounding context is considered.

In this work, we address both challenges by introducing an auxiliary continuous code space which is learned from the whole dialogue flow. At each time step, instead of directly optimizing discrete utterances, the current, past and future utterances are all trained to maximize the mutual information with this code space. Furthermore, a learnable prior distribution is simultaneously optimized to predict the corresponding code space, enabling efficient sampling in the testing phase without getting access to the ground-truth future conversation. Extensive experiments have been conducted to validate the superiority of our framework. The generated responses clearly demonstrate better performance with respect to both coherence and diversity.
\section{Model Structure}
\subsection{Motivation}
Let $u_i$ be the $i$th utterance within a dialogue flow. The dialogue history $H_{i-1}$ contains all the preceding context $u_1, u_2,\dots,u_{i-1}$ and $F_{i+1}$ denotes the future conversations $u_{i+1},\dots,u_T$. The objective of our model is to find the decoding probability $p_{\theta}(u_i|H_{i-1},F_{i+1})$ that maximizes the mutual information $I(H_{i-1},u_{i})$ and $I(u_i,F_{i+1})$. Formally, the objective is:

\begin{equation}
\label{eq: mmi}
\begin{gathered}
\max_\theta \lambda_{1}I(H_{i-1},u_{i})+\lambda_{2}I(u_i,F_{i+1})\\
u_i\sim p_{\theta}(u_{i}|H_{i-1},F_{i+1})
\end{gathered}
\end{equation}
$\lambda_{1}$ and $\lambda_{2}$ adjusts the relative weight. Mutual information is defined over $p_{\theta}(u_i|H_{i-1},F_{i+1})$ and the empirical distribution $p(H_{i-1},F_{i+1})$. Now we assume the future context $F_{i+1}$ is known to us when training the decoding probability, we will address the unknown future problem later.

Directly optimizing with this objective is unfortunately infeasible because the exact computation of mutual information is intractable, and backpropagating through sampled discrete sequences is notoriously difficult to train. The discontinuity prevents the direct application of the reparameterization trick~\citep{kingma2014auto}. Low-variance relaxations like Gumbel-Softmax~\citep{jang2016categorical}, semantic hashing~\citep{kaiser2018fast} or vector quantization~\citep{van2017neural} lead to biased gradient estimations, which are accumulated as the sequence becomes longer. The Monte-Carlo-Simulation is unbiased but suffers from high variances. Designing a reasonable control variate for variance reduction is an extremely tricky task~\citep{mnih2014neural, tucker2017rebar}. For this sake, we propose replacing $u_i$ with a continuous code space $c$ learned from the whole dialogue flow.

\begin{figure*}[!ht]
\centering
\centerline{\includegraphics[width=\columnwidth]{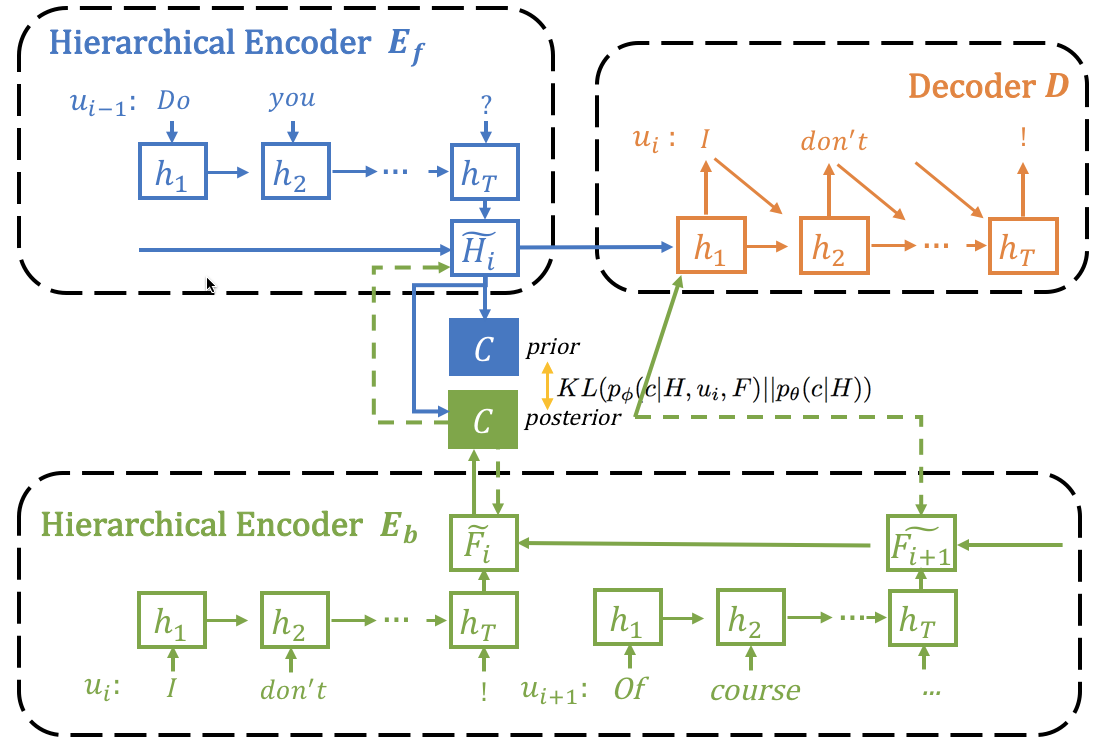}}
\caption{Framework of NEXUS Networks. Full line indicates the generative model to generate the continuous code and corresponding responses. Dashed line indicates the inference model where the posterior code is trained to infer the history, current and future utterances. Both parts are simultaneously trained by gradient descent.}
\label{fig:model}
\end{figure*}
\subsection{Continuous Code Space}
We define the continuous code space $c$ to follow the Gaussian probability distribution with a diagonal covariance matrix conditioning on the whole dialogue:
\begin{equation}
\label{eq: codesample}
c\sim p_{\phi}(c|H_{i-1},F_{i})=\mathcal{N}(\mu_c,\sigma^2_c \mathbb{I}|H_{i-1},F_{i})
\end{equation}
The dialogue history $H_{i-1}$ is encoded into vector $\tilde{H_{i-1}}$ by a forward hierarchical GRU model $E_f$ as in \citep{serban2015building}. The future conversation, including the current utterance, is encoded into $\tilde{F_i}$ by a backward hierarchical GRU $E_b$. $\tilde{H_{i-1}}$ and $\tilde{F_i}$ are concatenated and a multi-layer perceptron is built on top of them to estimate the Gaussian mean and covariance parameters. The code space is trained to infer the encoded history $\tilde{H_{i-1}}$ and future $\tilde{F_{i+1}}$. The full optimizing objective is:
\begin{equation}
\begin{gathered}	
\label{eq: code}
\mathcal{L}(c)=\max_{\phi} \mathbb{E}_{p_{\phi}(H_{i-1},F_{i},c)}[\lambda_{1}\log p_{\phi}(\tilde{H_{i-1}}|c)\\+\lambda_{2}\log p_{\phi}(\tilde{F_{i+1}}|c)]\\
p_{\phi}(H_{i-1},F_{i},c)=p(H_{i-1},F_{i})p_{\phi}(c|H_{i-1},F_{i})\\
p_{\phi}(\tilde{H_{i-1}}|c)=\mathcal{N}(\mu_{H_i}, \sigma^2_{H_i} \mathbb{I}|c)\\
p_{\phi}(\tilde{F_{i+1}}|c)=\mathcal{N}(\mu_{F_{i+1}}, \sigma^2_{F_{i+1}} \mathbb{I}|c)
\end{gathered}
\end{equation}

where $\tilde{H_{i-1}}$ and $\tilde{F_{i+1}}$ are also assumed to be Gaussian distributed given $c$ with mean and covariance estimated from multi-layer perceptrons. We infer the encoded vectors instead of the original sequences for three reasons. Firstly, inferring dense vectors is parallelizable and computationally much cheaper than autoregressive decoding, especially when the context sequences could be unlimitedly long. Secondly, sequence vectors can capture more holistic semantic-level similarity than individual tokens. Lastly, It can also help alleviate the posterior collapsing issue~\citep{bowman2016generating} when training variational inference models on text~\citep{chen2016variational,shen2018improving}, which we will use later. It can be shown that the above objective maximizes a lower bound of $\lambda_{1}I(H_{i-1}, c)+\lambda_{2}I(c,F_{i+1})$, given the conditional probability $p_{\phi}(c|H_{i-1},F_{i})$. The proof is a direct extension of the derivation in \citep{chen2016infogan}, followed by the Data Processing Inequality~\citep{beaudry2012intuitive} that the encoding function can only reduce the mutual information. As the sampling process contains only Gaussian continuous variables, the above objective can be trained through the reparameterization trick~\citep{kingma2014auto}, which is a low-variance, unbiased gradient estimator~\citep{burda2015importance}. After training, samples from $p_{\phi}(c|H_{i-1},F_{i})$ hold high mutual information with both the history and future context. The next step is then transferring the continuous code space to reasonable discrete natural language utterances.
\subsection{Decoding from Continuous Space}
Our decoder transfers the code space $c$ into the ground-truth utterance $u_i$ by defining the probability distribution $p(u_i|H_{i-1},c)$, which is implemented as a GRU decoder going through $u_i$ word by word to estimate the output probability. The encoded history $\tilde{H_{i-1}}$ and code space $c$ are concatenated as an extra input at each time step. The loss function for the decoder is then:
\begin{equation}
\begin{gathered}
\label{eq: decodeobj}
\mathcal{L}(d)=\max_{\phi}\mathbb{E}_{p_{\phi}(H_{i-1},F_{i},c)}\log p_{\phi}(u_i|H_{i-1},c)\\
p_{\phi}(H_{i-1},F_{i},c)=p(H_{i-1},F_{i})p_{\phi}(c|H_{i-1},F_{i})
\end{gathered}
\end{equation}

which can be proved to be the lower bound of the conditional mutual information $I(u_i,c|H_{i-1})$. By maximizing the conditional mutual information, $c_i$ is trained to maintain as much information about the target sequence $u_i$ as possible. 

Combining Eq.~\ref{eq: code} and \ref{eq: decodeobj}, our model until now can be viewed as optimizing a lower bound of the following objective:
\begin{equation}
\label{eq: new-mmi}
\begin{gathered}
\max_{\phi} \lambda_{1}I(H_{i-1}, c)+\lambda_{2}I(c, F_{i+1}) + I(u_i, c|H_{i-1})\\
c\sim p_{\phi}(c|H_{i-1},F_{i})
\end{gathered}
\end{equation}

Compared with the original motivation in Eq.~\ref{eq: mmi}, we sidestep the non-differentiability problem by replacing $u_i$ with a continuous code space $c$, then forcing $u_i$ to contain the same information as maintained in $c$ by additionally maximizing the mutual information between them. 

Nonetheless, Eq.~\ref{eq: new-mmi} and Eq.~\ref{eq: mmi} might lead to different optimums as mutual information does not satisfy the transitive law. In the extreme case, different dimensions of $c$ could individually maintain information about history, current and future conversations and the conversations themselves  do not share any dependency relation. To avoid this issue, we restrict the dimension of $c$ to be smaller than that of the encoded vectors. In this case, optimizing Eq.~\ref{eq: new-mmi} will favor utterances having stronger correlations with the surrounding context to achieve a higher total mutual information.
\subsection{Learnable Prior Distribution for Unknown Future}
The last problem is the sampling mechanism of $c$ in Eq.~\ref{eq: codesample}, which conditions on the ground-truth future conversation. In the testing phase, when we have no access to it, we cannot perform the decoding process as in Eq.~\ref{eq: decodeobj}. To allow for decoding with only the history context, we need to learn an appropriate prior distribution $p_{\theta}(c|H_{i-1})$ for $c$. In the ideal case, we would like
\begin{equation}
p_{\theta}(c|H_{i-1})=\sum_{F_{i}}p_{\phi}(c|H_{i-1},F_{i})=p_{\phi}(c|H_{i-1})
\end{equation}
However, $p_{\phi}(c|H_{i-1})$ is intractable as it integrates over all possible future conversations. We apply variational inference on $c$ to maximize the variational lower bound~\citep{jordan1999introduction}:
\begin{equation}
\begin{gathered}
\label{eq: prior}
\mathcal{L}(p)=\max_{\theta,\phi}\mathbb{E}_{p_{\phi}(c|H_{i-1},F_{i})}\log p_\theta(\tilde{F_{i}}|H_{i-1},c)\\-KL(p_{\phi}(c|H_{i-1},F_{i})||p_\theta(c|H_{i-1}))\\
p_{\theta}(\tilde{F_{i}}|H_{i-1},c)\sim \mathcal{N}(\mu_{F_i}, \sigma^2_{F_i} \mathbb{I}|H_{i-1},c)\\
p_{\theta}(c|H_{i-1})\sim \mathcal{N}(\mu_{prior}, \sigma^2_{prior} \mathbb{I}|H_{i-1}))
\end{gathered}
\end{equation}
It can be reformulated as maximizing:
\begin{equation}
\begin{gathered}
\label{eq: prior-ref}
\mathbb{E}_{p_{\phi}(c|H_{i-1})}KL( p_\phi(\tilde{F_{i}}|H_{i-1},c)||p_\theta(\tilde{F_{i}}|H_{i-1},c))\\-KL(p_{\phi}(c|H_{i-1})||p_\theta(c|H_{i-1}))
\end{gathered}
\end{equation}
We can see it implicitly matches $p_{\phi}(c|H_{i-1})$ to a tractable Gaussian distribution $p_{\theta}(c|H_{i-1})$ by minimizing the KL divergence between them. It also functions as a regularizer to prevent overfitting when learning $p_\phi(c|H_{i-1},F_{i})$. In the testing phase, we can sample $c$ from the learned prior distribution $p_\theta(c|H_{i-1})$, then generate a response based on it.
\subsection{Summary}
To sum up, the total objective function of our model is:
\begin{equation}
\label{eq: obj}
\mathcal{L}=\mathcal{L}(c)+\mathcal{L}(d)+\mathcal{L}(p)
\end{equation}
Weighting can be added to individual loss functions for better performance, but we find it enough to maintain equal weights and avoid extra hyperparameters. All the parameters are simultaneously updated by gradient descent except for the encoders $E_f$ and $E_b$, which only accept gradients from $\mathcal{L}(d)$ since otherwise the model can easily learn to encode no information for a lower reconstruction loss in $\mathcal{L}(c)$ and $\mathcal{L}(p)$. An overview of our training procedure is depicted in Fig.~\ref{fig:model}.

\section{Relationship to Existing Methods}
\paragraph{MMI decoding}
MMI decoder was proposed by \citep{li2015diversity} and further extended in \citep{wang2017steering}. The basic idea is the same as our model by maximizing the mutual information with the dialogue context. However, the MMI principle is applied only at the testing phase rather than the training phase. As a result, it can only be used to evaluate the quality of a generation by estimating its mutual information with the context. To apply it in a generative task, we have to first sample some candidate responses with the seq2seq model, then rerank them by accounting for the MMI score. Our model differs from it in that we directly estimate the decoding probability thus no post-sampling rerank is needed. Moreover, we further include the future context to strengthen the connection role of the current utterances.
\paragraph{Conditional Variational Autoencoder}
The idea of learning an appropriate prior distribution in Eq.~\ref{eq: prior} is essentially a conditional variational autoencoder~\citep{sohn2015learning} where the accumulated posterior distribution is trained to stay close to a prior distribution. It has also been applied in dialogue generation~\citep{serban2016hierarchical,zhao2017learning}. However, all the above methods stick to the MLE objective function and do not optimize with respect to the mutual information. As we will show in the experiment, they fail to learn the correlation between the utterance and its surrounding context. The generation diversity of these models comes more from the sampling randomness of the prior distribution rather than from the correct understanding of context correlation. Moreover, they suffer from the posterior collapsing problem~\citep{bowman2016generating} and require special tricks like KL-annealing, BOW loss or word drop-out~\citep{shen2018improving}. Our model does not have such problems.
\paragraph{Deep Reinforcement Learning Dialogue Generation}
\citep{li2016deep} first considered future success in dialogue generation and applied deep reinforcement learning to encourage more interactive conversations. However, the reward functions are intuitively hand-crafted. The relative weight for each reward needs to be carefully tuned and the training stage is unstable due to the huge search space. In contrast, our model maximizes the mutual information in the continuous space and trains the prior distribution through the reparamaterization trick. As a result, our model can be more easily trained with a lower variance. Throughout our experiment, the training process of NEXUS network is rather stable and much less data-hungry. The MMI objective of our model is theoretically more sound and no manually-defined rules need to be specified.

\section{Experiments}
\FloatBarrier
\begin{table*}[htbp!]
\begin{center}
  \begin{tabular}{lcccccc}
    \toprule
    \multicolumn{1}{l}{\multirow{2}{*}{\textbf{Model}}} & 
    \multicolumn{3}{c}{\textbf{DailyDialog}} & \multicolumn{3}{c}{\textbf{Twitter}} \\ \cmidrule(l){2-7}
    &\textbf{Average} &\textbf{Greedy} &\textbf{Extreme} & \textbf{Average} &\textbf{Greedy} &\textbf{Extreme} \\
    \midrule
    Greedy &0.443 &0.376\phantom{*}&0.328\phantom{*}&0.510\phantom{*} & 0.341&0.356\phantom{*} \\
	Beam&0.437&0.350\phantom{*}&0.369\phantom{*}&0.505\phantom{*} &0.345 &0.352\phantom{*}\\
	MMI&0.457&0.371\phantom{*}&0.371\phantom{*}& 0.518\phantom{*}& 0.353&0.365\phantom{*}\\
    RL& 0.405& 0.329\phantom{*}&0.305\phantom{*}&0.460\phantom{*} &0.349 &0.323\phantom{*}\\
   VHRED&\textbf{0.491}&0.375\phantom{*}&0.313\phantom{*}&0.525\phantom{*}&0.389&0.372\phantom{*}\\
     
     \hline
   \textbf{NEXUS-H}&0.479&0.381*&\textbf{0.385*}&\textbf{0.558*}&0.392&0.373\phantom{*}\\
   \textbf{NEXUS-F}&0.476&0.383*&0.373\phantom{*}&0.549*& 0.393&0.386*\\ 
   \textbf{NEXUS}&0.488&\textbf{0.392*}&0.384*&0.556*&\textbf{0.397*}&\textbf{0.391*}\\
    \bottomrule
  \end{tabular}
\end{center}
\caption{\label{tab: embed}Results of embedding-based metrics. * indicates statistically significant
difference $(p < 0.05)$ from the best baselines. The same mark is used in Table~\ref{tab: bleu}}
\end{table*}
\subsection{Dataset and Training Details}
We run experiments on the DailyDialog~\citep{li2017dailydialog} and Twitter corpus~\citep{ritter2011data}. DailyDialog contains 13118 daily conversations under ten different topics. This dataset is crawled from various websites for English learner to practice English in daily life, which is high-quality, less noisy but relatively smaller. In contrast, the Twitter corpus is significantly larger but contains more noise. We obtain the dataset as used in \citep{serban2016hierarchical} and filter out tweets that have already been deleted, resulting in about 750,000 multi-turn dialogues. The contents have more informal, colloquial expressions which makes the generation task harder. These two datasets are randomly separated into training/validation/test sets with the ratio of 10:1:1. 

In order to keep our model comparable with the state-of-the-art,
we keep most parameter values the same as in \citep{serban2016hierarchical}. We build our vocabulary dictionary based on the most frequent 20,000 words for both corpus and map other words to a UNK token. The dimensionality of the code space $c$ is 100. We use a learning rate of 0.001 for DailyDialog and 0.0002 for Twitter corpus. The batch size is fixed to 128. The word vector dimension is 300 and is initialized with the public Word2Vec~\citep{mikolov2013efficient} embeddings trained on the Google News Corpus. The probability estimators for the Gaussian distributions are implemented as 3-layer perceptrons with the hyperbolic tangent activation function. As mentioned above, when training NEXUS models, we block the gradient from $\mathcal{L}(c)$ and $\mathcal{L}(p)$ with respect to $E_f$ and $E_b$ to encourage more meaningful encodings. The UNK token is prevented from being generated in the test phase. We implemented all the models with the open-sourced Python library Pytorch~\citep{paszke2017automatic} and optimized using the Adam optimizer~\citep{kingma2014adam}.
\subsection{Compared Models}
We conduct extensive experiments to compare our model against several representative baselines. 

\textbf{Seq2Seq}: Following the same implementation as in \citep{vinyals2015neural}, the seq2seq model serves as a baseline. We try both greedy decoding and beam search~\citep{graves2012sequence} with beam size set to 5 when testing.

\textbf{MMI}: We implemented the bidirectional-MMI decoder as in \citep{li2015diversity}, which showed better performance over the anti-LM model. The hyperparameter $\lambda$ is set to 0.5 as suggested. 200 candidates per context are sampled for re-ranking.

\textbf{VHRED}: The VHRED model is essentially a conditional variational autoencoder with hierarchical encoders~\citep{serban2016hierarchical,zhao2017learning}. To alleviate the posterior collapsing problem, we apply the KL-annealing trick and early stop with the step set as 12,000 for the DailyDialog and 75,000 for the Twitter corpus.

\textbf{RL}: Deep reinforcement learning chatbot as in \citep{li2016deep}. We use all the three reward functions mentioned in the paper and keep the relative weights the same as in the original paper. Policy network is initialized with the above-mentioned MMI model.

\textbf{NEXUS-H}: NEXUS network maximizing mutual information only with the history ($\lambda_2=0$).

\textbf{NEXUS-F}: NEXUS network maximizing mutual information only with the future ($\lambda_1=0$).

\textbf{NEXUS}: NEXUS network maximizing mutual information with both the history and future.

NEXUS-H and NEXUS-F are implemented to help us better analyze the effects of different components in our model. The hyperparameters $\lambda_1$ and $\lambda_2$ in NEXUS are set to be 0.5 and 1 respectively as we find history vector is consistently easier to be reconstructed than the future vector (\ref{app: ratio}).

\FloatBarrier
\begin{table*}[htbp!]
\begin{center}
  \begin{tabular}{lcccccc}
    \toprule
    \multicolumn{1}{l}{\multirow{2}{*}{\textbf{Model}}} & 
    \multicolumn{3}{c}{\textbf{DailyDialog}} & \multicolumn{3}{c}{\textbf{Twitter}} \\ \cmidrule(l){2-7}
    &\textbf{BLEU-1} &\textbf{BLEU-2} &\textbf{BLEU-3} & \textbf{BLEU-1} &\textbf{BLEU-2} &\textbf{BLEU-3} \\
    \midrule
    Greedy &0.394\phantom{*} &0.245&0.157\phantom{*}&0.340\phantom{*} & 0.203\phantom{*}&0.116\phantom{*} \\
	Beam&0.386\phantom{*}&0.251&0.163\phantom{*}&0.338\phantom{*} &0.205\phantom{*} &0.112\phantom{*}\\
	MMI&0.407\phantom{*}&0.269&0.172\phantom{*}& 0.347\phantom{*}& 0.208\phantom{*}&0.118\phantom{*}\\
    RL& 0.298\phantom{*}& 0.186&0.075\phantom{*}&0.314\phantom{*} &0.199\phantom{*} &0.103\phantom{*}\\
   VHRED&0.395\phantom{*}&\textbf{0.281}&0.190\phantom{*}&0.355\phantom{*}&0.211\phantom{*}&0.124\phantom{*}\\
     
     \hline
   \textbf{NEXUS-H}&0.418\phantom{*}&0.279&\textbf{0.199*}&\textbf{0.366*}&0.212\phantom{*}&0.126\phantom{*}\\
   \textbf{NEXUS-F}&0.399\phantom{*} &0.260&0.167\phantom{*}&0.359\phantom{*}& 0.213\phantom{*}&0.123\phantom{*}\\
   \textbf{NEXUS}&\textbf{0.424*}&0.276&0.198*&0.363*&\textbf{0.220*}&\textbf{0.131*}\\
    \bottomrule
  \end{tabular}
\end{center}
\caption{\label{tab: bleu}Results of BLEU score. It is computed based on the smooth BLEU algorithm~\citep{lin2004orange}. p-value interval is computed base on the altered bootstrap resampling algorithm~\citep{riezler2005some}}
\end{table*}

\subsection{Metric-based Performance}
\paragraph{Embedding Score} 
We conducted three embedding-based evaluations (average, greedy and extrema)~\citep{liu2016not}, which map responses into vector space and compute the cosine similarity~\citep{rus2012comparison}. The embedding-based metrics can to a large extent capture the semantic-level similarity between generated responses and ground truth. We represent
words using Word2Vec embeddings trained on
the Google News Corpus. We also measure the uncertainty of the score by assuming each data point is independently Gaussian distributed. The standard deviation yields the $95\%$ confidence interval~\citep{barany2007central}. Table~\ref{tab: embed} reports the embedding scores on both datasets. NEXUS network significantly outperforms the best baseline model in most cases. Notably, NEXUS can absorb the advantages from both NEXUS-H and NEXUS-F. The history and future information seem to help the model from different perspectives. Taking into account both of them does not create a conflict and the combination leads to an overall improvement. RL performs rather poorly on this metric, which is understandable as it does not target the ground-truth responses during training~\citep{li2016deep}.
\paragraph{BLEU Score}
BLEU is a popular metric that
measures the geometric mean of the modified n-gram
precision with a length penalty~\citep{papineni2002bleu}. Table~\ref{tab: bleu} reports the BLEU 1-3 scores. Compared with embedding-based metrics, the BLEU score quantifies the word-overlap between generated responses and the ground-truth. One challenge of evaluating dialogue generation by BLEU score is the difficulty of accessing multiple references for the one-to-many alignment relation. Following \citep{sordoni2015neural,zhao2017learning,shen2018improving}, for each context, 10 more candidate references are acquired by using information retrieval methods (see Appendix~\ref{app: ir} for more details). All candidates are then passed to human annotators to filter unsuitable ones, resulting in 6.74 and 5.13 references for DailyDialog and Twitter dataset respectively. The human annotation is costly, so we evaluate it on 1000 sampled test cases for each dataset. As the BLEU score is not the simple mean of individual sentence scores, we compute the $95\%$ significance interval by bootstrap resampling~\citep{koehn2004statistical,riezler2005some}. As can be seen, NEXUS network achieves best or near-best performances with only greedy decoders. NEXUS-H generally outperforms NEXUS-F as the connection with future context is not explicitly addressed by the BLEU score metric. MMI and VHRED bring minor improvements over the seq2seq model. Even when evaluated on multiple references, RL still performs worse than most models.
\FloatBarrier
\begin{table*}[htbp!]
\centering
\resizebox{1\textwidth}{!}{
\begin{tabular}{l|c|c|c|c|c|c|c|c}
\textbf{Model}&AdverSuc  &Neg-PMI&\#Turns&Distinct-1&Distinct-2&Pri&Post&Flu  \\  \hline
Greedy &0.21$\vert$0.13&47.4$\vert$45.8&0.2$\vert$0.6& .019$\vert$.017 &.096$\vert$.072&0.45&0.04&0.92\\
Beam&0.16$\vert$0.12&47.2$\vert$45.3&0.2$\vert$0.7&.026$\vert$.019 & .103$\vert$.086&0.52&0.06&0.90\\
MMI&0.30$\vert$0.19&45.6$\vert$43.2&1.1$\vert$1.6& .042$\vert$.025& .247$\vert$.117&0.56&0.13&0.89\\
RL&0.13$\vert$0.11&45.0$\vert$42.6&2.3$\vert$2.3& .048$\vert$.033& .324$\vert$.287&0.46&0.15&0.69\\
VHRED&0.19$\vert$0.16&46.8$\vert$44.7&1.7$\vert$1.1&.255$\vert$.106&.431$\vert$.311&0.42&0.22&0.92\\
\hline
\textbf{NEXUS-H}&\textbf{0.36}$\vert$\textbf{0.21}&\textbf{44.1}$\vert$41.8&2.0$\vert$1.8&.263$\vert$.108&.454$\vert$.306&0.66&0.20&0.92\\
\textbf{NEXUS-F}&0.22$\vert$0.12&47.1$\vert$45.9&2.6$\vert$2.2&\textbf{.288}$\vert$.117&.466$\vert$.325&0.51&0.31&\textbf{0.94}\\
\textbf{NEXUS}&0.35$\vert$0.18&44.6$\vert$\textbf{41.4}&\textbf{2.8}$\vert$\textbf{2.5}&.282$\vert$\textbf{.119}&\textbf{.470}$\vert$\textbf{.329}&\textbf{0.70}&\textbf{0.33}&0.93\\
\hline
GROUND&0.87$\vert$0.73&40.5$\vert$38.1&4.8$\vert$4.0&.390$\vert$.215&.522$\vert$.495
&0.92&0.67&0.97\\
\end{tabular}
}
\caption{{\label{tab: other-performance}}Coherence, diversity and human evaluations. Left: DailyDialog results, right: Twitter results}
\end{table*}
\paragraph{Connecting the preceding} We define two metrics to evaluate the model's capability of  ``connecting the preceding context": \textbf{AdverSuc} and \textbf{Neg-PMI}. AdverSuc measures the coherence of generated responses with the provided context by learning an adversarial discriminator~\citep{li2017adversarial} on the same corpus to distinguish coherent responses from randomly sampled ones. We encode the context and response separately with two different LSTM neural networks and output a binary signal indicating coherent or not\footnote{We apply the same architecture as in \citep{DBLP:journals/corr/LuKZSB17}. In our experiment, the discriminator performs reasonably well in the 4 scenarios outlined in \citep{li2017adversarial} and thus can be used as a fair evaluation metric.}. The AdverSuc value is reported as the success rate that the model fools the classifier into believing its false generations ($p(generated=coherent)>0.5$). Neg-PMI measures the negative pointwise mutual information value $-\log p(c|r)/p(c)$ between the generated response $r$ and the dialogue context $c$. $p(c|r)$ is estimated by training a separate backward seq2seq model. As $p(c)$ is a constant, we ignore it and only report the value of $-\log p(c|r)$. A good model should achieve a higher AdverSuc and a lower Neg-PMI. The results are listed in Table \ref{tab: other-performance}. We can see there is still a big gap between ground-truth and synthesized responses. As expected, NEXUS-H leads to the most significant improvement. MMI model also performs remarkably well, but it requires post-reranking thus the sampling process is much slower. VHRED and NEXUS-F do not help much here, sometimes even slightly degrade the performance. We also tried removing the history context when computing the posterior distribution in VHRED, the resulting model has similar performance among all metrics, which suggests VHRED itself cannot actually learn the correlation pattern with the preceding context. Surprisingly, though RL explicitly set the coherence score as a reward function, its performance is far from satisfying. We assume RL requires much more data to learn the appropriate policy than other models and the training process suffers from a higher variance. The result is thus hard to be guaranteed.
\paragraph{Connecting the following}
We measure the model's capability of ``connecting the following context" from two perspectives: number of the simulated turns and diversity of generated responses. We apply all models to generate multiple turns until a generic response is reached. The set of generic responses is manually examined to include all utterances providing only passive dull replies\footnote{We use a simple rule matching method (see Appendix ~\ref{app: dull}). We manually inspect it on a validation subset and find the accuracy is more than 90\%. Similar methods are adopted in \citep{li2016deep}.}. The number of generated turns can reflect the time that a model can maintain an interactive conversation. The results are reflected in the \textbf{\#Turns} column in Table \ref{tab: other-performance}. As in \citep{li2015diversity}, we measure the diversity by the percentage of distinct unigrams (\textbf{Distinct-1}) and bigrams (\textbf{Distinct-2}) in all generated responses. Intuitively a higher score on these three metrics implies a more interactive generation system that can better connect the future context. Again, NEXUS network dominates most fields.  NEXUS-F brings more impact than NEXUS-H as it explicitly encourages more interactive turns. Most seq2seq models fail to provide an informative response in the first turn. The MMI-decoder does not change much, possibly because the sampling space is not large enough, a more diverse sampling mechanism~\citep{vijayakumar2016diverse} might help. NEXUS network can effectively continue the conversation for 2.8 turns for DailyDialog and 2.5 turns for Twitter, which is closest to the ground truth (4.8 and 4.0 turns respectively). It also achieves the best diversity score in both datasets. It is worth mentioning that NEXUS-H also improves over baselines, though not as significantly as NEXUS-F, so NEXUS is not a trade-off but more like an enhanced version from NEXUS-H and NEXUS-F.

In summary, NEXUS network clearly generates higher-quality responses in both coherence and diversity, even in a rather small dataset like DailyDialog. NEXUS-H contributes more to the coherence and NEXUS-F more to the diversity.
\FloatBarrier
\begin{table*}
\centering
\begin{tabular}{p{43mm}|p{88mm}}
  \textbf{Context} & \textbf{Response}\\
  \hline
  { Hi , Alice , what are you going to do this sunday ?  \_\_eou\_\_ } & \textbf{MMI:} \ Oh , that' s great !  \newline \textbf{VHRED :} \  Well ,  it seems that i'm going to be late . \newline \textbf{RL :} \  How long he it be ? \newline \textbf{NEXUS:} \  Nothing really , what ' s up ?  \\ \hline
  {Did you watch the Oscars on Sunday night ? \_\_eou\_\_ } & \textbf{MMI:} \ I have no idea . \newline \textbf{VHRED:} \ That's not too bad .\newline \textbf{RL :} \  It's my mum's birthday on Sunday . \newline \textbf{NEXUS:} \  No , nothing much. Could you help me ?  \\ \hline
   {Let's go for a drink ! \_\_eou\_\_ } & \textbf{MMI:} \ Thanks . \newline \textbf{VHRED:} \ all right . let me see you then . \newline \textbf{RL :} \  Thank you inviting me .\newline \textbf{NEXUS:} \  Sounds good . Where are we going ? \\ \hline 
    {You won't want to miss our webinar tomorrow !   \_\_eou\_\_ } & \textbf{MMI:} \  lol ! \newline \textbf{VHRED:} \ That ' s terrific !  \newline \textbf{RL :} \  Sure, where where ?\newline \textbf{NEXUS:} \  Thanks for your invitation! I'm free ! \\ \hline
    {How I read the article ?   \_\_eou\_\_ } & \textbf{MMI:} \  there!!! \newline \textbf{VHRED:} \ What ' s good with it ?  \newline \textbf{RL :} \ : I don’t know what to do with it !\newline \textbf{NEXUS:} \   Maybe the force is yourself ! \\ \hline
\end{tabular}
\caption{{\label{tab: examples}}Examples of context-response pairs. \_\_eou\_\_ denotes end-of-utterance. First three rows are from DailyDialog and the last two rows are from Twitter
  }
\end{table*}
\subsection{Human Evaluation}
We also employed crowdsourced judges to provide evaluations for a random sample of 500 items in the DailyDialog test dataset. Participants are asked to assign a binary score to each context-response pair from three perspectives: whether the response coincides with its preceding context (Pri), whether the response is interesting enough for people to continue (Post) and whether the response itself is a fluent natural sentence (Flu). Each sample gets one point if judged as yes and zero otherwise. Each pair is judged by three participants and the score supported by most people is adopted. We also evaluated the inter-annotator consistency by Fleiss'k score\citep{fleiss1971measuring} and obtained k scores of 0.452 for Pri, 0.459 for Post (moderate agreement) and 0.621 for Flu (substantial agreement), which implies most context-response pairs reach a consensus on the evaluation task. We compute the average human score for each model. Unlike metric-based scores, the human evaluation is conducted only on the DailyDialog corpus as it contains less noise and can be more fairly evaluated by human judges. Table \ref{tab: other-performance} shows the result in the last three columns. As can be seen, the pri and post human scores are highly correlated with the automatic evaluation metric ``coherence" and ``\#turns", verifying the validity of these two metrics. As for fluency, there is no significant difference among most models. As we also manually examined, fluency is not a major problem and all models produce mostly well-formed sentences. Overall, NEXUS network does produce responses that are more acceptable to human judges.

Table \ref{tab: examples} presents some randomly sampled context-response pairs provided by MMI, VHRED, RL and NEXUS model. We see NEXUS network does generate more interactive outputs than the other three. Though reranked by the bidirectional language model, the MMI decoder still produces quite a few generic responses. VHRED's utterances are more diverse, but it only cares about answering to the immediate query and makes no efforts to bring about further topics. Moreover, it also generates more inappropriate responses than the others. RL provides diverse responses but sometimes not fluent or coherent enough. We do observe that NEXUS sometimes generate over-complex questions which are not very natural, as in the second example. But in most cases, it outperforms the others.

\section{Additional Information}
\subsection{Proof of Eq.~\ref{eq: code}}

\begin{equation*}
\begin{split}
&\lambda_1I(H,c)+\lambda_2I(c,F)\\
 \geq&\lambda_1I(\tilde{H},c)+\lambda_2I(c,\tilde{F})\\
 =&\lambda_1\mathbb{E}_{p_\phi(\tilde{H}c)}\log \frac{p_\phi(\tilde{H}|c)}{p(\tilde{H})}+\lambda_1\mathbb{E}_{p_\phi(c\tilde{F})}\log \frac{p_\phi(\tilde{F}|c)}{p(\tilde{F})}\\
 =&\lambda_1\mathbb{E}_{p_\phi(\tilde{H}c)}\log p_\phi(\tilde{H}|c)+\lambda_1\mathbb{H}(\tilde{H})+\lambda_2\mathbb{E}_{p_\phi(c\tilde{F})}\log p_\phi(\tilde{F}|c)+\lambda_2\mathbb{H}(\tilde{F})\\
 \geq&\lambda_1\mathbb{E}_{p_\phi(\tilde{H}c)}\log p_\phi(\tilde{H}|c)+\lambda_2\mathbb{E}_{p_\phi(c\tilde{F})}\log p_\phi(\tilde{F}|c)\\
 =&\lambda_1\mathbb{E}_{p_\phi(\tilde{H}c)}\log p_\gamma(\tilde{H}|c)+\lambda_1KL(p_\phi(\tilde{H}|c)||p_\gamma(\tilde{H}|c))+\lambda_2\mathbb{E}_{p_\phi(c\tilde{F})}\log p_\gamma(\tilde{F}|c)\\+&\lambda_2KL(p_\phi(\tilde{H}|c)||p_\gamma(\tilde{H}|c))\\
 \geq&\lambda_1\mathbb{E}_{p_\phi(\tilde{H}c)}\log p_\gamma(\tilde{H}|c)+\lambda_2\mathbb{E}_{p_\phi(c\tilde{F})}\log p_\gamma(\tilde{F}|c)\\
 =&\mathbb{E}_{p_\phi(\tilde{H}u_{i}\tilde{F,c})}[\lambda_1\log p_\gamma(\tilde{H}|c)+\lambda_2\log p_\gamma(\tilde{F}|c)]
 \end{split}
 \end{equation*}
 \subsection{Proof of Eq.~\ref{eq: decodeobj}}
 \begin{equation*}
 \begin{split}
 I(u_i,c|H)&=\mathbb{E}_{p(H)}\mathbb{E}_{p_\phi(u_{i}c|H)}\log\frac{p_\phi(u_i|Hc)}{p(u_i|H)}\\
 &=\mathbb{E}_{p(H)}\mathbb{E}_{p_\phi(u_{i}c|H)}\log p_\phi(u_i|Hc)+\mathbb{H}(u_i|H)\\
 &\geq \mathbb{E}_{p(Hu_{i}F)}\mathbb{E}_{p_\phi(c|Hu_{i}F)}\log p_\phi(u_i|Hc)\\
 &=\mathbb{E}_{p(Hu_{i}F)}\mathbb{E}_{p_\phi(c|Hu_{i}F)}\log p_\gamma(u_i|Hc)+\mathbb{E}_{p_\phi(HcF)}KL(p_\phi(u_i|Hc)||p_\gamma(u_i|Hc))\\
 &\geq \mathbb{E}_{p(Hu_{i}F)}\mathbb{E}_{p_\phi(c|Hu_{i}F)}\log p_\gamma(u_i|Hc)
 \end{split}
 \end{equation*}
 \subsection{Derivation of Eq.~\ref{eq: prior-ref}}
 \begin{equation*}
 \begin{split}
 &\mathbb{E}_{p(\tilde{u_iF}|\tilde{H})}[\mathbb{E}_{p_{\phi}(c|\tilde{H}\tilde{u_{i}F})}\log p_\phi(\tilde{u_{i}F}|c)-KL(p_{\phi}(c|\tilde{H}\tilde{u_{i}F})||p_\theta(c|\tilde{H}))]\\
 =&\mathbb{E}_{p(\tilde{u_iF}|\tilde{H})}[\mathbb{E}_{p_{\phi}(c|\tilde{H}\tilde{u_{i}F})}\log \frac{p_\phi(\tilde{u_iF}|c)p_\theta(c|\tilde{H})}{p_\phi(c|\tilde{H}\tilde{u_iF})}]\\
 =&\mathbb{E}_{p(\tilde{u_iF}|\tilde{H})}[[\mathbb{E}_{p_{\phi}(c|\tilde{H}\tilde{u_{i}F})}\log \frac{p_\phi(\tilde{u_iF}|c)p_\theta(c|\tilde{H})p(\tilde{u_iF}|\tilde{H})}{p_\phi(\tilde{u_iF}|\tilde{H}c)p_\phi(c|\tilde{H})}]\\
 =&\mathbb{E}_{q_\phi(c|\tilde{H})}KL( p_\phi(\tilde{u_{i}F}|\tilde{H}c)||p_\phi(\tilde{u_{i}F}|\tilde{H}c))-KL(p_{\phi}(c|\tilde{H})||p_\theta(c|\tilde{H}))-\mathbb{H}(\tilde{u_iF}|\tilde{H})
 \end{split}
 \end{equation*}

\FloatBarrier
\subsection{Information Retrieval Technique for Multiple References}
\label{app: ir}
We collected multiple reference responses for each dialogue context in the test set by information retrieval techniques. References are retrieved based on their similarity with the provided context. Responses to the retrieved utterances are used as references. The process of retrieving similar context is as follows: First, we select 1000 candidate utterances using the tf-idf score. These candidates are then mapped to a vector space by summing their contained word vectors. After that, they are reranked based on the average of cosine similarity, Jaccard distance and Euclidean distance with the ground-truth context. The top 10 retrieved responses are passed to human annotators to judge the appropriateness.

\subsection{Phrases that count as forming dull responses}
\label{app: dull}
\begin{enumerate}[1)]
\indentitem
\item i know
\item no \_\_eou\_\_(yes \_\_eou\_\_)
\item no problem
\item lol
\item thanks \_\_eou\_\_
\item don't know 
\item don't think 
\item what ?
\item of course
\item wtf
\end{enumerate}
Utterances matching one of these phrases are treated as dull responses.
\subsection{Effect of hyperparameter $\lambda_1 / \lambda_2$}
\label{app: ratio}
\begin{figure*}[!ht]
\centering
\centerline{\includegraphics[width=16cm]{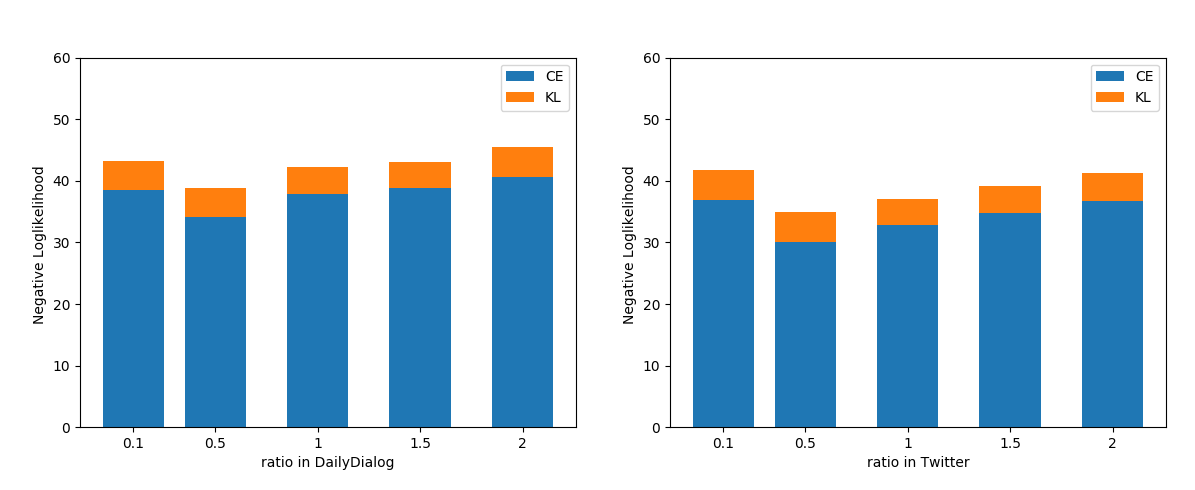}}
\caption{Effect of hyperparameter ratio $\lambda_1 / \lambda_2$ on two datasets. }
\label{fig:parameter}
\end{figure*}
Figure~\ref{fig:parameter} visualizes the effects of hyperparameters $\lambda_1$ and $\lambda_2$. The negative-log-likelihood is decomposed into two parts: decoding cross entropy (CE) as in Eq.~\ref{eq: decodeobj} and KL divergence as in Eq.~\ref{eq: prior}. The sum is a lower bound of the true log-likelihood. The optimal ratio is around 0.5 for both datasets, which means only half weights should be given to the history compared with the future context. Two reasons can explain this phenomena. Firstly, future vector is harder to infer than history as it is not explicitly exposed as an input in Eq.~\ref{eq: code}. Secondly, minimizing the KL divergence in Eq.~\ref{eq: prior} pushes the code space to discard information from the future context so that it could vanish to zero. Therefore, more weights should be given to the future context to maintain a balance.

\section{Conclusion}
In this chapter, we propose ``NEXUS Network" to enable more interactive human-computer conversations. The main goal of our model is to strengthen the ``nexus" role of the current utterance, connecting both the preceding and the following dialogue context. We compare our model with MMI, reinforcement learning and CVAE-based models. Experiments show that NEXUS network consistently produces higher-quality responses. The model is easier to train, requires no special tricks and demonstrates remarkable generalization capability even in a very small dataset.

Our model can be considered as combining the objective of MMI and CVAE and is compatible with current improving techniques. For example, mutual information can be maximized under a tighter bound using Donsker-Varadhan or f-divergence representation~\citep{donsker1983asymptotic,nowozin2016f,belghazi2018mutual}. Extending the   code space distribution to more than Gaussian by importance weighted autoencoder~\citep{burda2015importance}, inverse autoregressive flow~\citep{kingma2016improved} or VamPrior~\citep{tomczak2018vae} should also help with the performance.

Until now, we explain improved techniques for the vanilla VAE with continuous latent variables on the dialogue generation task. The learned continuous latent variables, however, are non-interpretable. This might not be good for some generation tasks where we hope to have finer-grained control on. In the next chapters we will show how we can use latent variables to stand for discrete, interpretable factors.
\cleardoublepage

\chapter[Latent Variable as Content Selection]{Latent Variable as Content Selection}
\label{chap: select}

\lettrine[lines=3]{M}any text generation tasks naturally contain two steps: content selection and surface realization. Current neural encoder-decoder models conflate both steps into a black-box architecture. As a result, the content to be described in the text cannot be explicitly controlled. This paper tackles this problem by decoupling content selection from the decoder. The decoupled content selection is human interpretable, whose value can be manually manipulated to control the content of generated text. The model can be trained end-to-end without human annotations by treating content selection as the latent variable. Different from previous chapters. Latent variables in this chapter has an explicit, interpretable meaning, i.e., content selection of the input. We further propose an effective way to trade-off between performance and controllability with a single adjustable hyperparameter. In both data-to-text and headline generation tasks, our model achieves promising results, paving the way for controllable content selection in text generation~\citep{shen2019select}.\footnote{The source code is available on \url{https://github.com/chin-gyou/controllable-selection} }

\section{Introduction}
\label{sec: intro}
\begin{table}
\setlength{\columnwidth}{4pt}
\center
\footnotesize
\begin{tabular}{@{}l}
{ \parbox{7.5cm}{{\bf Source Sentence: }The \textcolor{ggreen}{sri lankan} government on \textcolor{gblue}{Wednesday} \textcolor{orange}{announced} the \textcolor{ggreen}{closure} of government \textcolor{ggreen}{schools} with immediate effect as a \textcolor{red}{military campaign} against tamil separatists escalated in the {north} of the country.}}\\\midrule

{ \parbox{7.5cm}{{\bf Selected : } \textcolor{ggreen}{sri lankan}, \textcolor{ggreen}{closure}, \textcolor{ggreen}{schools}}}\\
{ \parbox{7.5cm}{{\bf Text: } \textcolor{ggreen}{sri lanka closes schools} .}}\\\midrule
{ \parbox{7.5cm}{{\bf Selected : } \textcolor{ggreen}{sri lankan}, \textcolor{gblue}{Wednesday}, \textcolor{ggreen}{closure}, \textcolor{ggreen}{schools}}}\\
{ \parbox{7.5cm}{{\bf Text: } \textcolor{ggreen}{sri lanka closes schools} on \textcolor{gblue}{Wednesday}.}}\\\midrule
{ \parbox{7.5cm}{{\bf Selected : } \textcolor{ggreen}{sri lankan}, \textcolor{ggreen}{closure}, \textcolor{ggreen}{schools}, \textcolor{red}{military campaign}}}\\
{ \parbox{7.5cm}{{\bf Text: } \textcolor{ggreen}{sri lanka shuts down schools} amid \textcolor{red}{war} fears.}}\\\midrule
{ \parbox{7.5cm}{{\bf Selected : } \textcolor{ggreen}{sri lankan}, \textcolor{orange}{announced}, \textcolor{ggreen}{closure}, \textcolor{ggreen}{schools}}}\\
{ \parbox{7.5cm}{{\bf Text: } \textcolor{ggreen}{sri lanka} \textcolor{orange}{declares} \textcolor{ggreen}{closure} of \textcolor{ggreen}{ schools}.}}\\\midrule

\end{tabular}
\caption{\label{tab: intro}Headline generation examples from our model. We can generate text describing various contents by sampling different content selections. The selected source word and its corresponding realizations in the text are highlighted with the same color.} 
\end{table}
Many text generation tasks, e.g., data-to-text, summarization and image captioning, can be naturally divided into two steps: content selection and surface realization.  The generations are supposed to have two levels of diversity: (1) content-level diversity reflecting multiple possibilities of content selection (what to say) and (2) surface-level diversity reflecting the linguistic variations of verbalizing the selected contents (how to say)~\citep{reiter2000building,nema2017diversity}. Recent neural network models handle these tasks with the encoder-decoder (Enc-Dec) framework~\citep{sutskever2014sequence,bahdanau2015neural}, which simultaneously performs selecting and verbalizing in a black-box way. Therefore, both levels of diversity are entangled within the generation. This entanglement, however, sacrifices the controllability and interpretability, making it diffifcult to specify the content to be conveyed in the generated text ~\citep{qin2018learning,wiseman2018learning}.

With this in mind, this paper proposes decoupling content selection from the Enc-Dec framework to allow finer-grained control over the generation. Table~\ref{tab: intro} shows an example. We can easily modify the content selection to generate text with various focuses, or sample multiple paraphrases by fixing the content selection.

Though there has been much work dealing with content selection for the Enc-Dec, none of them is able to address the above concerns properly. Current methods can be categorized into the following three classes and have different limits:
\begin{enumerate}
    \item \textbf{Bottom-up}: Train a separate content selector to constrain the attention to source tokens~\citep{gehrmann2018bottom}, but the separate training of selector/generator might lead to discrepancy when integrating them together.
    \item \textbf{Soft-select}: Learn a soft mask to filter useless information~\citep{mei2016talk,zhou2017selective}. However, the mask is \emph{deterministic} without any probabilistic variations, making it hard to model the content-level diversity.
    \item \textbf{Reinforce-select}: Train the selector with reinforcement learning~\citep{chen2018fast}, which has high training variance and low diversity on content selection.
\end{enumerate}
In this chapter, we treat the content selection as latent variables and train with amortized variational inference~\citep{kingma2014auto, mnih2014neural}. This provides a lower training variance than Reinforce-select. The selector and generator are co-trained within the same objective, the generations are thus more faithful to the selected contents than Bottom-up methods.
The selector works by simply masking the attention score in the decoding process without extra burden. 
Our model is task-agnostic, end-to-end trainable and can be seamlessly inserted into any encoder-decoder architecture. On both the data-to-text and headline generation task, we show our model outperforms others regarding content-level diversity and controllability while maintaining comparable performance. The performance/controllability trade-off can be effectively adjusted by adjusting a single hyperparameter in the training stage, which constrains an upper bound of the conditional mutual information (CMI) between the selector and generated text~\citep{pmlr-v80-alemi18a,zhao2018information}. A higher CMI leads to stronger controllability with a bit more risk of text disfluency. 

In summary, our contributions are \textbf{(1)} systematically studying the problem of controllable content selection for Enc-Dec text generation, \textbf{(2)} proposing a task-agnostic training framework achieving promising results and \textbf{(3)} introducing an effective way to achieve the trade-off between performance and controllability.

\section{Background and Notation}
Let $X,Y$ denote a source-target pair. $X$ is a sequence of $x_1,x_2,\ldots,x_n$ and can be either some structured data or unstructured text/image depending on the task. $Y$ corresponds to $y_1,y_2,\ldots,y_m$ which is a text description of $X$. The goal of text generation is to learn a distribution $p(Y|X)$ to automatically generate proper text.

The Enc-Dec architecture handles this task with an encode-attend-decode process~\citep{bahdanau2015neural,xu2015show}. The encoder first encodes each $x_i$ into a vector $h_i$. At each time step, the decoder pays attentions to some source embeddings and outputs the probability of the next token by $p(y_t|y_{1:t-1},C_t)$. $C_t$ is a weighted average of source embeddings:
\begin{equation}
\label{eq: select_attention}
\begin{split}
C_t &= \sum_{i}\alpha_{t,i}h_{i}\\
\alpha_{t,i} &= \frac{e^{f(h_{i}, d_t)}}{\sum_j e^{f(h_{j}, d_t)}}
\end{split}
\end{equation}
$d_{t}$ is the hidden state of the decoder at time step $t$. $f$ is a score function to compute the similarity between  $h_i$ and $d_t$~\citep{luong2015effective}.

\section{Content Selection}
\begin{figure}[ht]
\centering
\centerline{\includegraphics[width=\columnwidth]{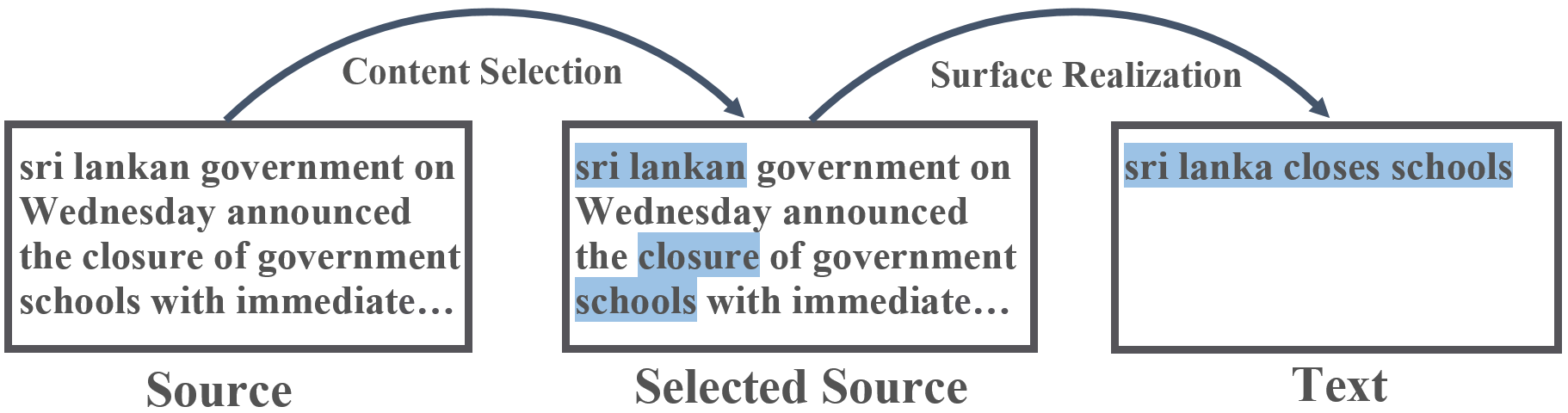}}
\caption{Model will select contents based on $\mathbf{B}(\gamma)$, then decode with $p_\theta(Y|X,\beta)$. Source-text pairs are available for training, but the ground-truth content selection for each pair is unknown.}
\label{fig: structure}
\end{figure}

Our goal is to decouple the content selection from the decoder by introducing an extra content selector. We hope the content-level diversity can be fully captured by the content selector for a more interpretable and controllable generation process. Following \citet{gehrmann2018bottom,yu2018operation}, we define content selection as a sequence labeling task. Let $\beta_1,\beta_2,\ldots,\beta_n$ denote a sequence of binary selection masks. $\beta_i=1$ if $h_i$ is selected and 0 otherwise. $\beta_i$ is assumed to be independent from each other and is sampled from a bernoulli distribution $\mathbf{B}(\gamma_i)$\footnote{\citet{devlin2018bert} have shown that excellent performance can be obtained by assuming such conditionally independence given a sufficiently expressive representation of $x$, though modelling a richer inter-label dependency is for sure beneficial~\citep{lei2016rationalizing, nallapati2017summarunner}.}. $\gamma_i$ is the bernoulli parameter, which we estimate using a two-layer feedforward network on top of the source encoder. Text are generated by first sampling $\beta$ from $\mathbf{B}(\gamma)$ to decide which content to cover, then decode with the conditional distribution $p_\theta(Y|X,\beta)$. The text is expected to faithfully convey all selected contents and drop unselected ones. Fig.~\ref{fig: structure} depicts this generation process. Note that the selection is based on the token-level \emph{context-aware} embeddings $h$ and will maintain information from the surrounding contexts. It encourages the decoder to stay faithful to the original information instead of simply fabricating random sentences by connecting the selected tokens.
For each source-target pair, the ground-truth selection mask is unknown, so training is challenging. In the following session, we discuss several training possibilities and introduce the proposed model in detail.

\subsection{Bottom-up}
The most intuitive way is training the content selector to target some heuristically extracted contents. For example, we can train the selector to select overlapped words between the source and target~\citep{gehrmann2018bottom}, sentences with higher tf-idf scores~\citep{li2018improving} or identified image objects that appear in the caption~\citep{wang2017diverse}. A standard encoder-decoder model is independently trained. In the testing stage, the prediction of the content selector is used to hard-mask the attention vector to guide the text generation in a bottom-up way. Though easy to train, Bottom-up generation has the following two problems: \textbf{(1)} The heuristically extracted contents might be coarse and cannot reflect the variety of human languages and \textbf{(2)} The selector and decoder are independently trained towards different objectives thus might not adapt to each other well.

\textbf{$\bm{\beta}$ as Latent Variable:} Another way is to treat $\beta$ as a latent variable and co-train selector and generator by maximizing the marginal data likelihood. By doing so, the selector has the potential to automatically explore optimal selecting strategies best fit for the corresponding generator component.

With this in mind. We design $p_\theta(Y|X,\beta)$ by changing the original decoder in the following way: (1) We initialize hidden states of the decoder from a mean pooling over selected contents to inform the decoder which contents to cover and (2) Unselected contents will be prohibited from being attended to:
\begin{align}
\label{eq: mask-attention}
\begin{split}
    &d_0 = \text{MLP}\left(\frac{1}{n}\left(\sum_{i}^{n}\beta_i h_i\right)\right)\\
&\alpha_{t,i} = \frac{e^{f\left(h_{i}, d_t\right)}\beta_i}{\sum_j e^{f\left(h_{j}, d_t\right)}\beta_j}
\end{split}
\end{align}
$d_0$ is the initial decoder hidden state and MLP denotes multi-layer-perceptron. 

Since computing the exact marginal likelihood $\log \mathbb{E}_{\beta \sim \mathbf{B}(\gamma)} p_\theta(Y|X,\beta)$ requires enumerating over all possible combinations of $\beta$ (complexity $\mathbf{O}(2^n)$), we need some way to efficiently estimate the likelihood.

\subsection{Soft-Select}
\label{sec: soft-select}
Soft-select falls back on a \emph{deterministic} network to output the likelihood function's first-order Taylor series approximation expanded at $\mathbb{E}_{\beta \sim \mathbf{B}(\gamma)} \beta$:
\begin{equation*}
\begin{split}
&\log \mathbb{E}_{\beta \sim \mathbf{B}(\gamma)} p_\theta(Y|X,\beta)\\\approx & \log [p_\theta(Y|X, \gamma)+\mathbb{E}_{\beta\sim \mathbf{B}(\gamma)}(\beta-\gamma)p'_\theta(Y|X, \gamma)]\\
=&\log p_\theta(Y|X, \gamma)
\end{split}
\end{equation*}
By moving the expectation into the decoding function, we can deterministically compute the likelihood by setting $\beta_i=\gamma_i$, reducing complexity to $\mathbf{O}(1)$. Each attention weight will first be ``soft-masked" by $\gamma$ before being passed to the decoder. soft-select is fully differentiable and can be easily trained by gradient descent. However, this soft-approximation is normally inaccurate, especially when $\mathbf{B}(\gamma)$ has a high entropy, which is common in one-to-many text generation tasks. The gap between $\log \mathbb{E}_{\beta \sim \mathbf{B}(\gamma)} p_\theta(Y|X,\beta)$ and $\log p_\theta(Y|X, \mathbb{E}_{\beta\sim \mathbf{B}(\gamma)})$ will be large~\citep{ma2017dropout,deng2018latent}. In practice, this would lead to unrealistic generations when sampling $\beta$ from the deterministically trained distribution.
\subsection{Reinforce-Select}
\label{sec: reinforced-select}
 Reinforce-select (RS)~\citep{ling2017coarse,chen2018fast} utilizes reinforcement learning to approximate the marginal likelihood. Specifically, it is trained to maximize a lower bound of the likelihood by applying the Jensen inequalily:
\begin{equation*}
\label{eq: hard-mask}
\log \mathbb{E}_{\beta \sim \mathbf{B}(\gamma)} p_\theta(Y|X,\beta)
\geq\mathbb{E}_{\beta \sim \mathbf{B}(\gamma)}\log p_\theta(Y|X,\beta)
\end{equation*}

The gradient to $\gamma$ is approximated with Monte-Carlo sampling by applying the REINFORCE algorithm~\citep{williams1992simple,glynn1990likelilood}. To speed up convergence, we pre-train the selector by some distant supervision, which is a common practice in reinforcement learning. REINFORCE is unbiased but has a high variance. Many research have proposed sophisticated techniques for variance reduction~\citep{mnih2014neural,tucker2017rebar,grathwohl2018backpropagation}. In text generation, the high-variance problem is aggravated because there exists multiple valid selections. Accurately estimating the likelihood becomes difficult. Another issue is its tendency to avoid stochasticity~\citep{raiko2014techniques}, which we will show in Sec~\ref{sec: results} that it results in low content-level diversity. 
\subsection{Variational Reinforce-Select}
 We propose Variational Reinforce-Select (VRS) which applies variational inference~\citep{kingma2014auto} for variance reduction. Instead of directly integrating over $\mathbf{B}(\gamma)$, it imposes a proposal distribution $q_\phi$ for importance sampling. The marginal likelihood is lower bounded by:
\begin{equation}
\label{eq: vhard-mask}
\begin{split}
&\log \mathbb{E}_{\beta \sim \mathbf{B}(\gamma)} p_\theta(Y|X,\beta)\\
&=\log \mathbb{E}_{\beta \sim q_\phi} \frac{p_\theta(Y,\beta|X)}{q_\phi(\beta)}\\
&\geq\mathbb{E}_{\beta \sim q_\phi}\log \frac{p_\theta(Y,\beta|X)}{q_\phi(\beta)}\\
&=\mathbb{E}_{\beta \sim q_\phi}\log p_\theta(Y|X,\beta) - KL(q_\phi||\mathbf{B}(\gamma))
\end{split}
\end{equation}
By choosing a proper $q_\phi$, the bound will be improved and the variance can be largely reduced compared with REINFORCE. If $q_\phi$ equals the posterior distribution $p_\theta(\beta|X,Y)$, the bound is tight and the variance would be zero~\citep{mnih2016variational}. We define $q_\phi(\beta|X,Y)$ as a mean-field distribution parameterized by a set of global parameters $\phi$ to approach the true posterior distribution. $\phi$, $\theta$ and $\gamma$ are simultaneously trained by maximizing the last line of Eq.~\ref{eq: vhard-mask}. $q_\phi(\beta|X,Y)$ also allows us to further perform posterior inference: Given an arbitrary text $Y$ for a source $X$, we can infer which source contents are included in $Y$. 

In Eq.\ref{eq: vhard-mask},  the KL divergence term can be computed analytically. As for the independence assumption, it can be summed over each individual $\beta_i$. 
The likelihood term is differentiable to $\theta$ but not to $\phi$, we estimate the gradient to $\phi$ in Eq~\ref{eq: vhard-mask} by applying the REINFORCE estimator:
\begin{equation*}
\label{eq: vhard-train}
\begin{gathered}
\nabla_\phi\mathbb{E}_{\beta \sim q_\phi}\log p_\theta(Y|X,\beta)=\\
\mathbb{E}_{\beta \sim q_\phi} \nabla_\phi \log q_\phi(\beta|X,Y)(\log p_\theta(Y|X,\beta) - B)
\end{gathered}
\end{equation*}
$B$ is the control variate~\citep{williams1992simple}. The optimal $B$ would be~\citep{weaver2001optimal}:
\begin{equation*}
\label{eq: op-baseline}
B^{\ast} = \mathbb{E}_{\beta \sim q_\phi}\log p_\theta(Y|X,\beta)
\end{equation*}
which we set as a soft-select approximation:
\begin{equation*}
\label{eq: baseline}
B=\log p_\theta(Y|X,\mathbb{E}_{\beta \sim q_\phi} \beta)
\end{equation*}
 We estimate Eq.~\ref{eq: vhard-train} with a single sample from $q_\phi$ for efficiency. Though multiple-sample could potentially further tighten the bound and reduce the variance~\citep{burda2015importance,lawson2018learning,tucker2019doubly}, it brings significant computational overhead, especially in text generation tasks where the whole sentence needs to be decoded.
\subsection{Degree of Controllability}
\label{sec: trade-off}
In practice, when treating content selection as latent variables, the model tends to end up with a trivial solution of always selecting all source tokens~\citep{shen2018reinforced,pmlr-v80-ke18a}.
This behavior is understandable since Eq.~\ref{eq: mask-attention} strictly masks unselected tokens. Wrongly unselecting one token will largely deteriorate the likelihood. Under the maximum likelihood (MLE) objective, this high risk pushes the selector to take a conservative strategy of always keeping all tokens, then the whole model degenerates to the standard Enc-Dec and the selection mask loses effects on the generation. Usually people apply a penalty term to the selecting ratio when optimizing the likelihood:
\begin{equation}
    \label{eq: hh-alpha}
    \mathcal{L} + \lambda \abs{(\Bar{\gamma}-\alpha)}
\end{equation}
 $\mathcal{L}$ is the MLE loss function, $\Bar{\gamma}$ is the mean of $\gamma$ and $\alpha$ is the target selecting ratio. This forces the selector to select the most important $\alpha$ tokens for each source input instead of keeping all of them. 
 
 In our VRS model, we can easily adjust the degree of controllability by limiting an upper bound of the conditional mutual information (CMI) $I(\beta,Y|X)$~\citep{zhao2018information}. Specifically, we can change our objective into:
\begin{equation}
\label{eq: vhard-obj}
\begin{split}
    \max_{\phi,\theta,\gamma}\mathbb{E}_{\beta \sim q_\phi}\log p_\theta(Y|X,\beta) \\- \lambda\abs{KL(q_\phi||\mathbf{B}(\gamma))-\epsilon)}
\end{split}
\end{equation}
$\lambda$ is a fixed lagrangian multiplier. Eq.~\ref{eq: vhard-obj} can be proved equal to maximum likelihood with the constraint $I(\beta,Y|X)=\epsilon$ given proper $\lambda$~\citep{pmlr-v80-alemi18a}. A higher $\epsilon$ indicates $\beta$ has more influences to $Y$ (higher controllability) while always safely selecting all tokens will lead $I(\beta,Y|X)=0$.\footnote{We also tried adding a coverage constraint to ensure the decoder covers all the selected tokens~\citep{wen2015semantically,wang2019toward}, but we find it brings no tangible help since a higher CMI can already discourage including redundant tokens into the selection.} It is preferred over Eq.~\ref{eq: hh-alpha} because (a) CMI directly considers the dependency between the selection and \emph{multiple}-possible text while limiting the ratio aims at finding the \emph{single} most salient parts for each source. (b) Unlike CMI, limiting the ratio is coarse. It considers only the total selected size and ignores its internal distribution.

\begin{algorithm}[tb]
   \caption{Variational Reinforce-Select (VRS)}
   \label{alg}
\begin{algorithmic}
   \STATE {\bfseries Parameters:} $\theta,\phi,\gamma$
   \STATE $pretrain \leftarrow$ TRUE
   \REPEAT
   \STATE Sample X,Y from the corpus;
   \STATE Encode X into $(h_1,h_2,\ldots,h_n)$;
   
   \IF{$pretrain$}
   \STATE Update $\phi$ with distant supervision;\\
   Update $\theta,\gamma$ by  $\nabla_{\theta,\gamma}\mathbb{E}_{\beta \sim q_\phi}\log p_\theta(Y|X,\beta) - KL(q_\phi||\mathbf{B}(\gamma)$ (Eq.~\ref{eq: vhard-mask});
   \ELSE
   \STATE Update $\theta,\gamma,\phi$ by  $\nabla_{\theta,\gamma,\phi} \mathbb{E}_{\beta \sim q_\phi}\log p_\theta(Y|X,\beta) - \lambda\abs{KL(q_\phi||\mathbf{B}(\gamma))-\epsilon)}$ (Eq.~\ref{eq: vhard-obj});
   \ENDIF
   \STATE $pretrain \leftarrow$ FALSE if Eq.~\ref{eq: vhard-mask} degrades
   \UNTIL{convergence and $pretrain$ is False}
\end{algorithmic}
\end{algorithm}

 In practice, we can set $\epsilon$ to adjust the degree of controllability we want. Later we will show it leads to a trade-off with performance. The final algorithm is detailed in Algorithm~\ref{alg}. To keep fairness, we train RS and VRS with the same control variate and pre-training strategy.\footnote{The only extra parameter of VRS is $\phi$ which is a simple MLP structure. The actual training complexity is similar to RS because they both use the REINFORCE algorithm for gradient estimation.}

\section{Related Work}
Most content selection models train the selector with heuristic rules~\citep{hsu2018unified,li2018improving,yu2018operation,gehrmann2018bottom,yao2018plan,moryossef2019step}, which fail to fully capture the relation between selection and generation. \citet{mei2016talk,zhou2017selective,lin2018global,li2018improving} ``soft-select" word or sentence embeddings based on a gating function. The output score from the gate is a deterministic vector without any probabilistic variations, so controlling the selection to generate diverse text is impossible. Very
few works explicitly define a bernoulli distribution for the selector, then train with the REINFORCE algorithm~\citep{ling2017coarse,chen2018fast}, but the selection targets at a high recall regardless of the low precision, so the controllability over generated text is weak. \citet{fan2017controllable} control the generation by manually concatenating entity embeddings, while our model is much more flexible by explicitly defining the selection probability over all source tokens. 

Our work is closely related with learning discrete representations with variational inference~\citep{wen2017latent,van2017neural,kaiser2018fast,lawson2018learning}, where we treat content selection as the latent representation. 
Limiting the KL-term is a common technique to deal with the ``posterior collapse" problem ~\citep{kingma2016improved,yang2017improved,shen2018improving}. We adopt a similar approach and use it to further control the selecting strategy.
\section{Experiments}
For the experiments, we focus on comparing
(1) Bottom-up generation (Bo.Up.), (2) soft-select (SS), (3) Reinforce-select (RS) and (4) Variational-Reinforce-select (VRS) regarding their performance on content selection. SS and RS are trained with the selecting ratio constraint in Eq.~\ref{eq: hh-alpha}. For the SS model, we further add a regularization term to encourage the maximum value of $\gamma$ to be close to 1 as in \citet{mei2016talk}. We first briefly introduce the tasks and important setup, then present the evaluation results. 

\subsection{Tasks and Setup}
We test content-selection models on the headline and data-to-text generation task. Both tasks share the same framework with the only difference of source-side encoders. 

\textbf{Headline Generation}:
We use English Gigaword preprocessed by \citet{rush2015neural}, which pairs first sentences of news articles with their headlines. We keep most settings same as in \citet{zhou2017selective}, but use a vocabulary built by byte-pair-encoding~\citep{sennrich2016neural}. We find it speeds up training with superior performance.

\textbf{Data-to-Text Generation}:
We use the Wikibio dataset~\citep{lebret2016neural}.
The source is a Wikipedia infobox and the target is a one-sentence biography description. Most settings are the same as in \citet{liu2018table}, but we use a bi-LSTM encoder for better performance. 
 
\textbf{Heuristically extracted content}: This is used to train the selector for bottom up models and pre-train the RS and VRS model. For wikibio, we simply extract overlapped words between the source and target. In Gigaword, as the headline is more abstractive, we select the closest source word for each target word in the embedding space. Stop words and punctuations are prohibited from being selected.
\begin{table*}[t!]
\centering
\resizebox{\columnwidth}{!}{
\begin{tabular}{lrrr|rrrr}
\multirow{2}{*}{{Gigaword}} & \multicolumn{3}{c}{\small Oracle upper bound of} &{ \% unique }&{ \% unique}& {\% \small{Effect of}}&{\small Entropy of}\\
    & \small{ROUGE-1} & \small{ROUGE-2} &\small{ROUGE-L}{}& {Generation}& {Mask}&{Selector} & {Selector}\\
  \hline
  \small{Bo.Up.} &42.61 & 22.32 &38.37& 84.28& 95.87 & 87.91 & 0.360\\
  \small{SS} &33.15 & 14.63 &30.68& 82.06&\textbf{96.23} & 85.27 & \textbf{0.392}\\
  {\small RS} & 36.62 & 18.34 &34.60& 3.01& 6.23 & 48.31 & 0.018 \\
  \small{VRS} &\textbf{54.73} & \textbf{33.28} &\textbf{51.62}&\textbf{89.23}&92.51 & \textbf{96.45} & 0.288\\
  \hline
  {Wikibio} & \small{ROUGE-4} & \small{BLEU-4} &\small{NIST}& {Generation}& {Mask} & {Selector} & Selector \\
  \small{Bo.Up.} &47.28 & 49.95 &11.06& 31.57& 77.42 & 40.78 & 0.177 \\
  \small{SS} &41.73 & 43.94 &9.82& \textbf{63.09}& \textbf{89.42} & 70.55 & \textbf{0.355}\\
  {\small RS} & 44.07 & 46.89 &10.31& 4.55& 43.83 & 10.38&0.105 \\
  \small{VRS} &\textbf{52.41} & \textbf{55.03} &\textbf{11.89}& 57.62& 77.83 & \textbf{74.03} & 0.181\\
\end{tabular}
}
\caption{\label{tab: diverse}Diversity of content selection. The \% effect of selector is defined as the ratio of unique generation and mask, which reflects the rate that changing the selector will lead to corresponding changes of the generated text.
  }
\end{table*}

\textbf{Choice of $\alpha/\epsilon$}: As seen in Sec~\ref{sec: trade-off}, we need to set the hyperparameter $\alpha$ for RS/SS and $\epsilon$ for VRS. $\alpha$ corresponds to the selecting ratio. We set them as $\alpha=0.35$ for Wikibio and $0.25$ for Gigaword. The value is decided by running a human evaluation to get the empirical estimation. To keep comparison fairness, we tune $\epsilon$ to make VRS select similar amount of tokens with RS. The values we get are $\epsilon=0.15n$ for Wikibio and $\epsilon=0.25n$ for Gigaword. $n$ is the number of source tokens.~\footnote{$\epsilon$ corresponds to the KL divergence of the selection mask, which scales linearly with the number of source tokens, so we set it proportinally w.r.t. $n$.} 

\subsection{Results and Analysis}
Ideally we would expect the learned content selector to (1) have reasonable diversity so that text with various contents can be easily sampled, (2) properly control the contents described in the generated text and (3) not hurt performance. The following section will evaluate these three points in order.
\label{sec: results}

\textbf{Diversity}: We first look into the diversity of content selection learned by different models. For each test data, 50 selection masks are randomly sampled from the model's learned distribution. Greedy decoding is run to generate the text for each mask. We measure the entropy of the selector, proportion of unique selection masks and generated text in the 50 samples. We further define the ``effect" of the selector as the ratio of sampled unique text and mask. This indicates how often changing the selection mask will also lead to a change in the generated text. The results are averaged over all test data. Following \citet{rush2015neural} and \citet{lebret2016neural}, we measure the quality of generated text with ROUGE-1, 2, L F-score for Gigaword and ROUGE-4, BLEU-4, NIST for Wikibio. As there is only one reference text for each source, we report
an oracle upper bound of these scores by assuming an ``oracle" that can choose the best text among all the candidates~\citep{mao2014deep,wang2017diverse}. Namely, out of each 50 sampled text, we pick the one with the maximum metric score. The final metric score is evaluated on these ``oracle" picked samples. The intuition is that if the content selector is properly trained, at least one out of the 50 samples should describe similar contents with the reference text, the metric score between it and the reference text should be high. Table~\ref{tab: diverse} lists the results. We can have the following observations:
\begin{itemize}
    \item RS model completely fails to capture the content-level diversity. Its selector is largely deterministic, with a lowest entropy value among all models.
    In contrast, the selector from SS, VRS and Bo.Up. have reasonable diversity, with over $90\%$ and $75\%$ unique selection masks for Gigaword and Wikibio respectively.
    \item The selector from VRS has the strongest effect to the generator, especially on the Gigaword data where modifying the content selection changes the corresponding text in more than 95\% of the cases. RS has the lowest effect value, which indicates that even with the selecting ratio constraint, its generator still ignores the selection mask to a large extent.
    \item The oracle metric score of VRS is much higher than the other two. This is beneficial when people want to apply the model to generate a few candidate text then hand-pick the suitable one. VRS has more potential than the other three to contain the expected text. SS performs worst. The gap between the soft approximation and the real distribution, as mentioned before, indeed results in a large drop of performance.
\end{itemize}
In short, compared with others, the content selector of VRS is \emph{(1) diverse, (2) has stronger effect on the text generation and (3) with a larger potential of producing an expected text}.

\textbf{Controllability}: We have shown the content selector of VRS is diverse and has strong effect on the text generation. This section aims at examining whether such effect is desirable, i.e., whether the selector is able to properly control the contents described in the text. We measure it based on the self-bleu metric and a human evaluation.

 The self-bleu metric measures the controllability by evaluating the ``intra-selection" similarity of generated text. Intuitively, by fixing the selection mask, multiple text sampled from the decoder are expected to describe the same contents and thereby should be highly similar to each other. The decoder should only model surface-level diversity without further modifying the selected contents. With this in mind, for each test data, we randomly sample a selection mask from the selector's distribution, then fix the mask and run the decoder to sample 10 different text. The self-BLEU-1 score~\citep{zhu2018texygen} on the sampled text is reported, which is the average BLEU score between each text pair. A higher self-BLEU score indicates the sampled text are more similar with each other. The results are shown in Table~\ref{tab: self-bleu}. We can see generations from VRS have a clearly higher intra-selection similarity. SS performs even worse than RS, despite having a high effect score in Table~\ref{tab: diverse}. The selector from SS affects the generation in an undesirable way, which also explain why SS has a lowest oracle metric score though with a high score on content diversity and effect.
 
\begin{table}[!ht]
\centering
\begin{tabular}{@{}llllll@{}}
\toprule
Method & Bo.Up. &SS & RS & VRS \\ \midrule
Gigaword & 46.58 & 37.20 & 48.13 & \textbf{61.14}\\
Wikibio & 38.30 & 13.92 & 25.99 & \textbf{43.81} \\
\bottomrule
\end{tabular}
\caption{Self-Bleu score by fixing selection mask. Higher means better controllability of content selection}
\label{tab: self-bleu}
\end{table}

\begin{table}[t]
\centering
\begin{tabular}{@{}cccc@{}}
\toprule
Method & \small{Fluency} & \small{intra-consistency} & \small{inter-diversity}\\ \midrule
\small{Reference}& 0.96 &-  &- \\
\small{Enc-Dec} & 0.83 &  -&-  \\\hline
\small{Bo.Up.} & 0.46 & 0.48 & 0.61 \\
\small{SS} & 0.27 & 0.41 & 0.54\\
\small{RS} & \textbf{0.78} & 0.39 & 0.47\\
\small{VRS} &0.74 & \textbf{0.72} & \textbf{0.87}\\
\bottomrule
\end{tabular}
\caption{Human evaluation on fluency, intra-consistency and inter-diversity of content selection on DUC 2004.}
\label{tab: human-eval}
\end{table}

 We further run a human evaluation to measure the text-content consistency among different models. 100 source text are randomly sampled from the human-written DUC 2004 data for task 1\&2~\citep{over2007duc}. Bo.Up, SS, RS and VRS are applied to generate the target text by first sampling a selection mask, then run beam search decoding with beam size 10. We are interested in seeing (1) if multiple generations from the same selection mask are paraphrases to each other (intra-consistent) and (2) if generations from different selection masks do differ in the content they described (inter-diverse). The results in Table~\ref{tab: human-eval} show that VRS significantly outperforms the other two in both intra-consistency and inter-diversity. RS has the lowest score on both because the selector has very weak effects on the generation as measured in the last section. Bo.Up and SS lay between them. Overall VRS is able to \emph{maintain the highest content-text consistency} among them.
 
 \begin{table}[ht]
\centering
\begin{tabular}{@{}lllll@{}}
\toprule
\small{Method} & \small{R-1} & \small{R-2} & \small{R-L} &\small{\%Word} \\ \midrule
\small{\citet{zhou2017selective}} & 36.15 & 17.54 & 33.63 & 100
\\
\small{Enc-Dec} & 35.92 & 17.43 & 33.42 & 100\\\hline
\small{SS} & 20.35 & 4.78 & 16.53 &24.82\\
\small{Bo.Up} & 28.17 & 10.32 & 26.68 &24.54\\
\small{RS} & 35.45 & 16.38 & 32.71 &25.12\\
\small{VRS($\epsilon=0$)-pri} &\textbf{36.42} & \textbf{17.81} & \textbf{33.86} &78.63\\
\small{VRS($\epsilon=0.25$)-pri} & 34.26 & 15.11 & 31.69 &24.36\\\hline\hline
\small{VRS($\epsilon=0$)-post} &37.14 & 18.03 & 34.26 &78.66\\
\small{VRS($\epsilon=0.25$)-post} &\textbf{56.72} & \textbf{33.24} & \textbf{51.88} &24.53\\
\bottomrule
\end{tabular}
\caption{Gigaword best-select results. Larger $\epsilon$ leads to more controllable selector with a bit degrade of performance. (-post) means selecting from the posterior $q_\phi(\beta|X,Y)$, (-pri) is from the prior $\mathbf{B}(\gamma_i)$.}
\label{tab: giga-accuracy}
\end{table}

\begin{table}[t]
\centering
\begin{tabular}{@{}lllll@{}}
\toprule
\small{Method} & \small{R-4} & \small{B-4} & \small{NIST} &\small{\%Word} \\ \midrule
\small{\citet{liu2018table}} & 41.65 & 44.71 &  & 100\\
\small{Enc-Dec} & 42.07 & 44.80 & 9.82 & 100\\\hline
\small{SS} & 5.10 & 5.73 & 0.24 &35.12\\
\small{Bo.Up} & 8.07 & 9.52 & 0.42 &38.79\\
\small{RS} & 42.64 & 45.08 & 10.01 &34.53\\
\small{VRS($\epsilon=0$)-pri} &\textbf{43.01} & \textbf{46.01} & \textbf{10.24} &84.56\\
\small{VRS($\epsilon=0.15$)-pri} & 42.13 & 44.51 & 9.84 &34.04\\\hline\hline
\small{VRS($\epsilon=0$)-post} &43.84 & 46.60 & 10.27 &85.34\\
\small{VRS($\epsilon=0.15$)-post} &\textbf{49.68} & \textbf{52.26} & \textbf{11.48} &34.57\\
\bottomrule
\end{tabular}
\caption{Wikibio best-select results.}
\label{tab: wiki-accuracy}
\end{table}

\textbf{Performance $\&$ Trade-off}: To see if the selector affects performance, we also ask human annotators to judge the text fluency. The fluency score is computed as the average number of text being judged as fluent. We include generations from the standard Enc-Dec model. Table~\ref{tab: human-eval} shows the best fluency is achieved for Enc-Dec. Imposing a content selector always affects the fluency a bit. The main reason is that when the controllability is strong, the change of selection will directly affect the text realization so that a tiny error of content selection might lead to unrealistic text. If the selector is not perfectly trained, the fluency will inevitably be influenced. When the controllability is weaker, like in RS, the fluency is more stable because it will not be affected much by the selection mask. For SS and Bo.Up, the drop of fluency is significant because of the gap of soft approximation and the independent training procedure. In general, VRS does properly decouple content selection from the enc-dec architecture, with only tiny degrade on the fluency.

Table~\ref{tab: giga-accuracy}/\ref{tab: wiki-accuracy} further measure the metric scores on Gigaword/Wikibio by decoding text from the best selection mask based on the selector's distribution (set $\beta_i=1$ if $\mathbf{B}(\gamma_i)>0.5$ and $0$ otherwise). We include results from VRS model with $\epsilon=0$, which puts no constraint on the mutual information. We further report the score by generating the best selection mask from the learned posterior distribution $q_\phi(\beta|X,Y)$ for VRS model. Two current SOTA results from \citet{zhou2017selective} and \citet{liu2018table} and the proportion of selected source words for each model are also included. We have the following observations:
\begin{itemize}
    \item As the value of $\epsilon$ decreases, the performance of VRS improves, but the selector loses more controllability because the model tends to over-select contents (over $75\%$ source words selected). The text-content consistency will become low.
    \item Increasing $\epsilon$ sacrifices a bit performance, but still comparable with SOTA. Especially on Wikibio where the performance drop is minor. The reason should be that Wikibio is relatively easier to predict the selection but Gigaword has more uncertainty.
    \item Increasing $\epsilon$ improves the accuracy of the posterior selection. This would be useful when we want to perform posterior inference for some source-target pair.
    \item Setting $\epsilon=0$ can actually outperform SOTA seq2seq which keeps all tokens, suggesting it is still beneficial to use the VRS model even if we do not care about the controllability.
\end{itemize}

\begin{figure}[!ht]
\centering
\centerline{\includegraphics[width=\columnwidth]{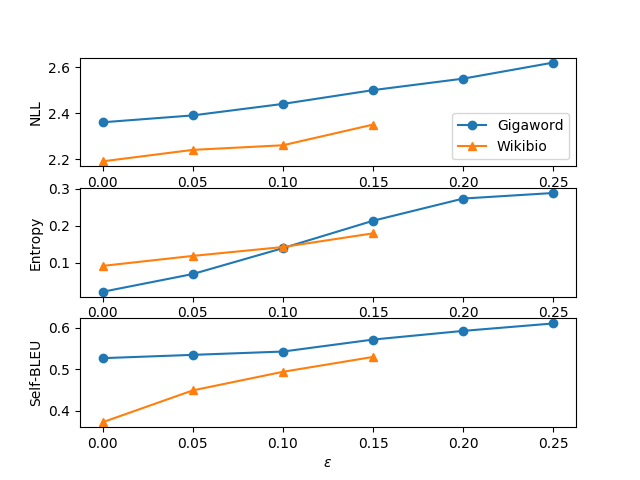}}
\caption{Negative log likelihood (NLL), selection entropy and self-BLEU as $\epsilon$ changes. NLL and self-bleu on Wikibio are added by 1 for better visualization. Lower NLL suggests higher performance. Higher entropy/self-BLEU means higher diversity/controllability.}
\label{fig: epsilon}
\end{figure}

Figure~\ref{fig: epsilon} visualizes how changing the value of $\epsilon$ affects the negative log likelihood (NLL), entropy of the selector and self-bleu score, which roughly correlates with performance, diversity and controllability. NLL is evaluated based on the lower bound in Eq~\ref{eq: vhard-mask}~\citep{sohn2015learning}.
We can see as $\epsilon$ increases, the performance decreases gradually but the content selection gains more diversity and controllability. In practice we can tune the $\epsilon$ value to achieve a trade-off. 

\begin{figure}[!ht]
\begin{boxedminipage}{\columnwidth}
\small
\textbf{Source:} \textcolor{gred}{indian prime minister} p.v. narasimha rao 's \textcolor{gred}{promise} of more autonomy for troubled \textcolor{gred}{kashmir} and his plea for early state elections has \textcolor{gred}{sparked} a \textcolor{gred}{violent reaction} from provincial moslem and opposition parties .
\newline 
\textbf{Samples from SS:}\newline
\textbf{t1:}  indian indian calls for end to violence in kashmir .\newline
\textbf{t2:} indian pm calls for end to violence in afghanistan .\newline
\textbf{t3:} indian pm calls for boycott of pakistan 's ruling party .\newline
\textbf{Samples from Bo.Up:}\newline
\textbf{t1:}  india promises more autonomous more autonomy .\newline
\textbf{t2:} indian pm promises autonomy for kashmir autonomy .\newline
\textbf{t3:} indian pm 's promise sparks violent reaction .\newline
\textbf{Samples from RS:}\newline
\textbf{t1:}  indian pm 's kashmir promises sparks violent reaction.\newline
\textbf{t2:} indian pm 's promise sparks violent reaction .\newline
\textbf{t3:} kashmir parties blast pm 's promise .\newline
\textbf{Samples from VRS:}
\newline\textbf{t1:}  indian pm 's promise on kashmir sparks uproar .\newline
\textbf{t2:} indian pm 's promise on kashmir sparks protests .\newline
\textbf{t3:} indian pm 's promise for kashmir sparks controversy .
\newline
\rule{\columnwidth}{0.4pt}
\textbf{Source:}  \textcolor{gred}{factory orders} for manufactured goods \textcolor{gred}{rose \#.\# percent in september} , the commerce department said here thursday . \newline
\textbf{Samples from SS:}\newline
\textbf{t1:}  u.s. consumer confidence down in january in january.\newline
\textbf{t2:} u.s. wholesale prices up \#.\# percent in october .\newline
\textbf{t3:} u.s. jobless rate rises to \#.\# percent in march .
\newline 
\textbf{Samples from Bo.Up.:}\newline
\textbf{t1:}  september u.s. factory orders up \#.\# percent .\newline
\textbf{t2:} september u.s. factory orders increase .\newline
\textbf{t3:} factory orders up in september .
\newline
\textbf{Samples from RS:}\newline
\textbf{t1:}  u.s. factory orders up \#.\# percent in september .\newline
\textbf{t2:} factory orders for manufactured goods rise .\newline
\textbf{t3:} factory orders up in september from the year .
\newline
\textbf{Samples from VRS:}\newline
\textbf{t1:}  september factory orders up \#.\# percent .\newline
\textbf{t2:} september factory orders rise \#.\# percent .\newline
\textbf{t3:} september factory orders increase \#.\# pct .
\end{boxedminipage}
\caption{Text generation examples from Gigaword. \textcolor{gred}{Highlighted} words are selected. t1-3 are sampled from the decoder based on the selected content. Generations from VRS are more faithful to selected contents.}
\label{fig: gigaword-example}
\end{figure}
\textbf{Generation Example}: Figure~\ref{fig: gigaword-example} shows some examples from Gigaword. As can be seen, decodings from the VRS model are largely consistent with each other, in most cases only replacing one or two words with corresponding synonyms. Samples are able to faithfully convey all selected contents. In contrast, generations from SS. Bo.Up. and RS are unpreditable, differing in both selected contents and also the way of saying. SS and Bo.Up also suffer more from the text disfluency. The generations from them are largely uncertain.

\section{Conclusion}
In this chapter, we tackle the unaddressed problem of controllable content selection in text generation. We propose a general framework based on variational inference that can be potentiall applied to arbitrary tasks. On both the headline generation and data-to-text tasks, our model outperforms state-of-the-art models regarding the diversity and controllability of content selection. We further introduce an effective way to achieve a performance/controllability trade-off, which can be easily tuned to meet specific requirement. 
The proposed model can be further improved by, for example, using a more flexible content selector or multi-sample estimation,  which we leave for future work.

\cleardoublepage

\chapter[Latent Variable as Alignment]{Latent Variable as Alignment}
\label{chap: alignment}

\lettrine[lines=3]{T}he last chapter uses latent variables to model the content selection. This chapter goes one step further by modeling the word-level alignment between the input and the output, so that humans can easily interpret how each word is generated. We base our system with the pointer Generator, which has been the de facto standard for modern summarization systems. However, this architecture faces two major
drawbacks: Firstly, the pointer is limited to copying the exact words while ignoring possible inflections or abstractions, which restricts its power of capturing richer latent alignment. Secondly, the copy mechanism results in a strong bias towards extractive generations, where most sentences are produced by simply copying from the source text. In this chapter, we address these problems by allowing the model to ``edit" pointed tokens instead of always hard copying them. The editing is performed by transforming the pointed word vector into a target space with a learned relation embedding. On three large-scale summarization dataset, we show the model is able to (1) capture more latent alignment relations than exact word matches, (2) improve word alignment accuracy, allowing for better model interpretation and controlling, (3) generate higher-quality summaries validated by both qualitative and quantitative evaluations and (4) bring more abstraction to the generated summaries~\citep{shen2019improving}.\footnote{The source code is available at \url{https://github.com/chin-gyou/generalized-PG}.}

\section{Introduction}
\begin{figure}
\centering
\centerline{\includegraphics[width=\columnwidth]{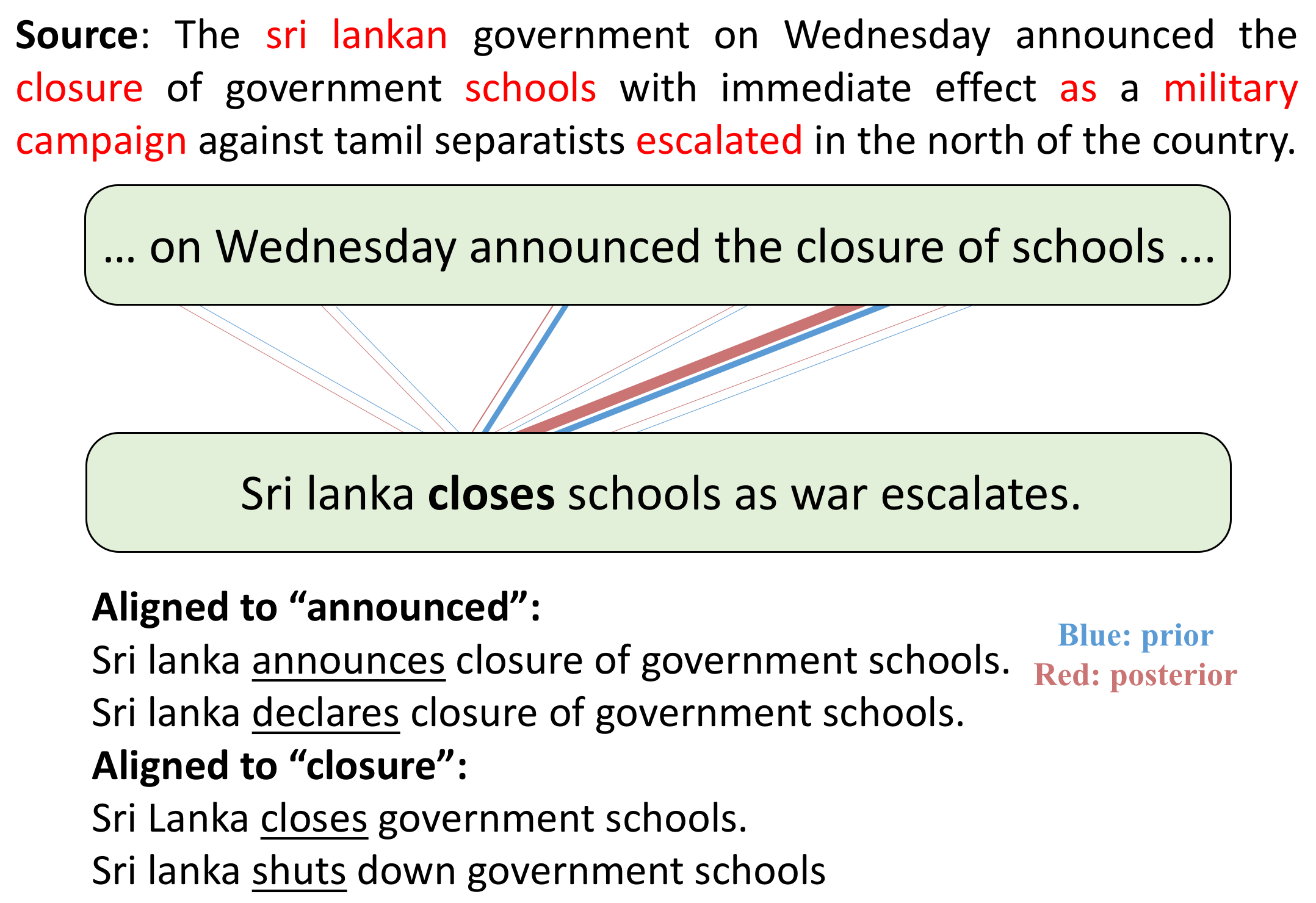}}
\caption{Alignment visualization of our model when decoding ``closes". Posterior alignment is more accurate for model interpretation. In contrast, the prior alignment probability is spared to ``announced" and ``closure", which can be manually controlled to generate desired summaries. Decoded samples are shown when aligned to ``announced" and ``closure" respectively. \textcolor{red}{Highlighted source words} are those that can be directly aligned to a target token in the gold summary.}
\label{fig: intro}
\end{figure}
Modern state-of-the-art (SOTA) summarization models are built
upon the pointer generator architecture~\citep{see2017get}. At each decoding step, the model generates a sentinel to decide whether to sample words based on the neural attention (generation mode), or directly copy from an aligned source context (point mode)~\citep{gu2016incorporating, merity2016pointer, yang2017reference,ge2021learning,wang2020cross}. Though outperforming the vanilla attention models, the pointer generator only captures exact word matches. As shown in Fig.~\ref{fig: intro}, for abstractive summarization, there exists a large number of syntactic inflections (escalated $\rightarrow$ escalates) or semantic transformations (military campaign $\rightarrow$ war), where the target word also has an explicit grounding in the source context but
changes its surface. In standard pointer generators, these words are not covered by the point mode. This largely restricts the application of the pointer generator, especially on highly abstractive dataset where only a few words are exactly copied. Moreover, the hard copy operation biases the model towards extractive summarization, which is undesirable for generating more human-like summaries~\citep{kryscinski2018improving,shen2018towards}. 

To solve this problem, we propose \textbf{G}eneralized \textbf{P}ointer \textbf{G}enerator (\textbf{GPG}) which replaces the hard copy component with a more general soft ``editing" function. We do this by learning a relation embedding to transform the pointed word into a target embedding. For example, when decoding ``closes" in Figure~\ref{fig: intro}, the model should first point to ``closure" in the source, predict a relation to be applied (noun $\rightarrow$ third person singular verb), then transform ``closure" into ``closes" by applying the relation transformation. The generalized point mode is encouraged to capture such latent alignment which cannot be identified by the standard pointer generator. 

This improved alignment modelling is intriguing in that (a) people can better control the generation by manipulating the alignment trajectory, (b) posterior alignment can be inferred by Bayes' theorem~\citep{deng2018latent,shankar2018posterior} to provide a better tool for interpretation\footnote{The induced alignment offers useful annotations for people to identify the source correspondence for each target word. News editors can post-edit machine-generated summaries more efficiently with such annotation. For summary readers, it also helps them track back the source context when they are interested in some specific details.} and finally (c) explicitly capturing the alignment relation should improve generation performance. (Figure~\ref{fig: intro} shows an example of how latent alignment can improve the controllability and interpretation. Pointer generators fail to model such alignment relations that are not exact copies.)
To eliminate the OOV problem, we utilize the byte-pair-encoding (BPE) segmentation~\citep{sennrich2016neural} to split rare words into sub-units, which has very few applications in summarization so far~\citep{fan2018controllable,kiyono2018unsupervised,zhang2021knowledge}, though being a common technique in machine translation~\citep{wu2016google,vaswani2017attention, gehring2017convolutional}. 

Our experiments are conducted on three summarization datasets: CNN/dailymail~\citep{hermann2015teaching}, English Gigaword~\citep{rush2015neural,huang2021dependency} and XSum~\citep{narayan2018don} (a newly collected corpus for extreme summarization). We further perform human evaluation and examine the word alignment accuracy on the manually annotated DUC 2004 dataset. Overall we find our model provides the following benefits:
\begin{enumerate}
    \item It can capture richer latent alignment and improve the word alignment accuracy, enabling better controllability and interpretation.
    \item The generated summaries are more faithful to the source context because of the explicit alignment grounding.
    \item It improves the abstraction of generations because our model allows editing the pointed token instead of always copying exactly.
\end{enumerate}
In the next section, we will first go over the background, then introduce our model and finally present the experiment results and conclusion.
\section{Background}
Let $X,Y$ denote a source-target pair where $X$ corresponds to a sequence of words $x_1,x_2,\ldots, x_n$ and $Y$ is its corresponding summary $y_1,y_2,\ldots,y_m$. In this section, we introduce two baseline models for automatically generating $Y$ from $X$:
\subsection{Seq2seq with Attention}
In a seq2seq model, each source token $x_i$ is encoded into a vector $h_i$. At each decoding step $t$, the decoder computes an attention distribution $a_t$ over the encoded vectors based on the current hidden state $d_t$~\citep{bahdanau2015neural}:
\begin{equation}
\label{eq: align_attention}
a_{t} = \text{softmax}(f(h_i, d_t))
\end{equation}
$f$ is a score function to measure the similarity between $h_i$ and $d_t$. 
The context vector $c_t$ and the probability of next token are computed as below.
\begin{gather}
\begin{split}
\label{eq: p_vocab}
c_{t} &= \sum_{i}h_{i}a_{t,i}\\
y_t^{*} &= [d_{t}\circ c_{t}]L\\
p_{vocab} &= \text{softmax}(y_t^{*}W^{T})
\end{split}
\end{gather}
$\circ$ means concatenation and $L, W$ are trainable parameters. We tie the parameters of $W$ and the word embedding matrix as in \citet{press2017using,inan2016tying,tang2021ast}. Namely, a target vector $y_t^{*}$ is predicted, words having a higher inner product with $y_t^{*}$ will have a higher probability.
\subsection{Pointer Generator}
The pointer generator extends the seq2seq model to support copying source words~\citep{vinyals2015pointer}. At each time step $t$, the model first computes a generation probability $p_{gen} \in [0,1]$ by:
\begin{equation}
\label{eq: gen_prob}
p_{gen} = \sigma(\text{MLP}_{g}([d_{t}\circ c_{t}]))
\end{equation}
$\sigma$ is a sigmoid function and $\text{MLP}_{g}$ is a learnable multi-layer perceptron. 
$p_{gen}$ is the probability of enabling the generation mode instead of the point mode. In the generation mode, the model computes the probability over the whole vocabulary as in Eq.~\ref{eq: p_vocab}. In the point mode, the model computes which source word to copy based on the attention distribution $a_t$ from Eq.\ref{eq: align_attention}. The final probability is marginalized over $a_{t,i}$:
\begin{gather}
p(y_{t}) = p_{gen}p_{vocab}(y_t)+(1-p_{gen})\sum_{i}a_{t,i}\delta(y_t|x_i)\nonumber\\
\label{eq: npg}
 \delta(y_t|x_i) = \begin{cases}
    1, & \text{if $y_t=x_i$}.\\
    0, & \text{otherwise}.
  \end{cases}
\end{gather}
If we know exactly from which mode each word comes from,  {\em e.g.}, by assuming all co-occurred words are copied, then the marginalization can be omitted~\citep{gulcehre2016pointing,wiseman2017challenges,adelani2021preventing}, but normally $p_{gen}$ is treated as a latent variable~\citep{gu2016incorporating,see2017get}.
\section{Generalized Pointer Generator (GPG)}
\label{sec: gpg}
\begin{figure*}[!ht]
\centering
\centerline{\includegraphics[width=\columnwidth]{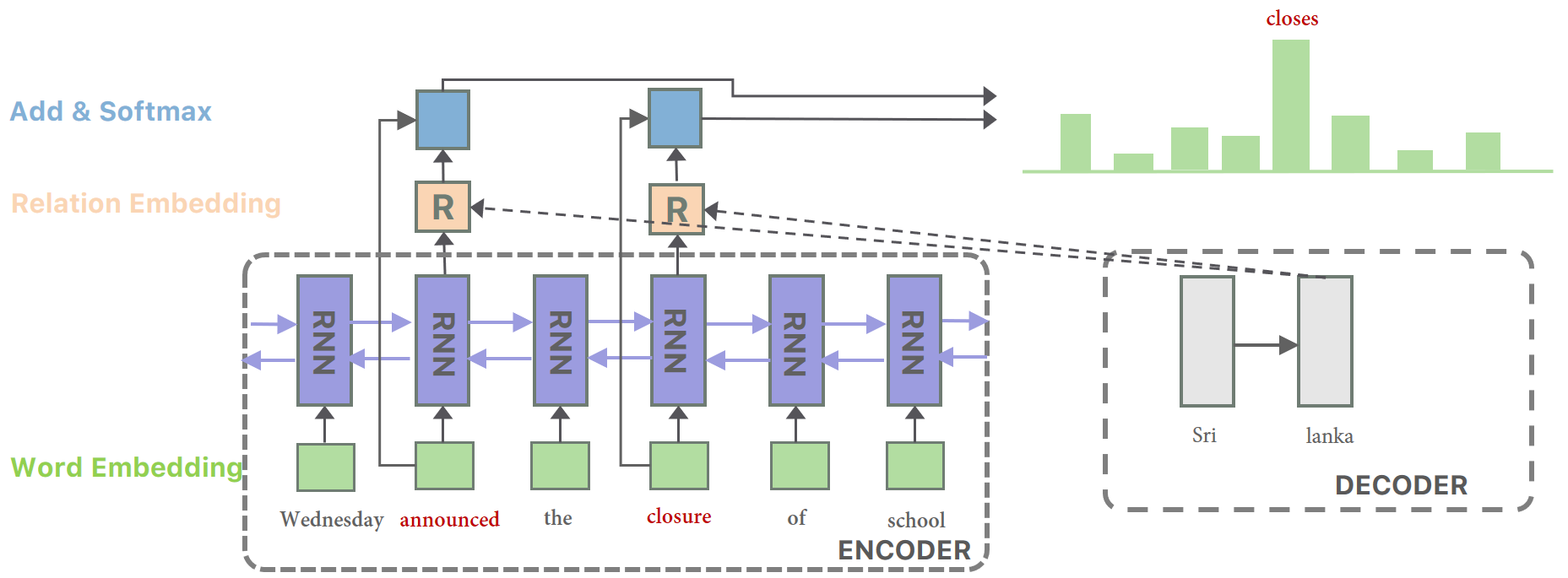}}
\caption{Architecture of the generalized pointer. The same encoder is applied to encode the source and target. When decoding ``closes", we first find top-k source positions with the most similar encoded state. For each position, the decoding probability is computed by adding its word embedding and a predicted relation embedding.}
\label{fig: model}
\end{figure*}
As seen in Eq~.\ref{eq: npg}, $\delta(y_t|x_i)$ is a 0-1 event that is only turned on when $y_t$ is exactly the same word as $x_i$. This restricts the expressiveness of the point mode, preventing it from paying attention to inflections, POS transitions or paraphrases. This section explains how we generalize pointer networks to cover these conditions.

\textbf{Redefine $\boldsymbol{\delta(y_t | x_i)}$}: We extend $\delta(y_t | x_i)$ by defining it as a smooth probability distribution over the whole vocabulary. It allows the pointer to edit $x_i$ to a different word $y_t$ instead of simply copying it. 
Following Eq.~\ref{eq: p_vocab}, we derive $\delta(y_t | x_i)$ by first predicting a target embedding $y_{t,i}^{*}$, then applying the softmax. The difference is that we derive $y_{t,i}^{*}$ as the summation of the pointed word embedding $\overrightarrow{x_i}$ and a relation embedding $r(d_t, h_t)$:
\begin{gather}
\begin{split}
    y_{t,i}^{*} &= r(d_t, h_i) + \overrightarrow{x_i}\\
\delta(y_t | x_i) &= \text{softmax}(y_{t,i}^{*}W^{T})
\end{split}
\label{eq: point}
\end{gather}
$\overrightarrow{x_i}$ denotes the embedding of the word $x_i$. $r(d_t, h_t)$ can be any function conditioning on $d_t$ and $h_t$, which we parameterize with a multi-layer-perceptron in our experiments. The computation of $y_{t,i}^{*}$ is similar to the classical TransE model~\citep{bordes2013translating} where an entity vector is added by a relation embedding to translate into the target entity. The intuition is straightforward: After pointing to $x_t$, humans usually first decide which relation should be applied (inflection, hypernym, synonym, etc) based on the context $[d_t, h_i]$, then transform $x_i$ to the proper target word $y_t$. Using addition transformation is backed by the observation that vector differences often reflect meaningful word analogies~\citep{mikolov2013distributed,pennington2014glove} $(``man"-``king"\approx``woman"-``queen")$ and they are effective at encoding a great amount of word relations like hypernym, meronym and morphological changes~\citep{vylomova2016take,hakami2018does,allen2019analogies,shen2021deep}. These word relations reflect most alignment conditions in text summarization. For example, humans often change the source word to its hypernym (boy $\rightarrow$ child), to make it more specific (person $\rightarrow$ man) or apply morphological transformations (liked $\rightarrow$ like). Therefore, we assume $\delta(y_t | x_i)$ can be well modelled by first predicting a relation embedding to be applied, then added to $\overrightarrow{x_i}$. If $x_i$ should be exactly copied like in standard pointer generators, the relation embedding is a zero vector meaning an identity transition. We also tried applying more complex transitions to $\overrightarrow{x_i}$ like diagonal mapping~\citep{trouillon2016complex}, but did not observe improvements. Another option is to estimate $\delta(y_t | x_i)$ directly from $(d_t, h_i)$ by an MLP regardless of $\overrightarrow{x_i}$. However, this leads to poor alignment and performance drop because $y_t$ is not explicitly grounded on $x_i$\footnote{$h_i$ can contain context information from surrounding words and thus not necessarily relates to word $x_i$. It is part of the reason that neural attention has a poor alignment~\citep{koehn2017six}. Making it grounded on $x_i$ improves alignment and performance is also observed in machine translation~\citep{nguyen2018improving,kuang2018attention}}.

\textbf{Estimate Marginal Likelihood}: Putting Eq.~\ref{eq: point} back to Eq.~\ref{eq: npg}, the exact marginal likelihood is too expensive for training. The complexity grows linearly with the source text length $n$ and each computation of $\delta(y_t | x_i)$ requires a separate softmax operation. One option is to approximate it by sampling like in hard attention models~\citep{xu2015show,deng2017image}, but the training becomes challenging due to the non-differentiable sampling process. In our work, we take an alternative strategy of marginalizing only over $k$ most likely aligned source words. This top-$k$ approximation is widely adopted when the target distribution is expected to be sparse and only a few modes dominate~\citep{britz2017efficient,ke2018sparse,shankar2018surprisingly}. We believe this is a valid assumption in text summarization since most source tokens have a vanishingly small probability to be transferred into a target word. 

For each target word, how to determine the $k$ most likely aligned source words is crucial. An ideal system should always include the gold aligned source word in the top-$k$ selections. We tried several methods and find the best performance is achieved when encoding each source/target token into a vector, then choosing the $k$ source words that are closest to the target word in the encoded vector space. The closeness is measured by the vector inner product\footnote{We compared several strategies for choosing the top-k words. Note that the top-k approximation is only used for training, so we can spot the whole target text to decide top-k candidates.}. The encoded vector space serves like a contextualized word embedding~\citep{mccann2017learned,peters2018deep,shen2017wake}. Intuitively if a target word can be aligned to a source word, they should have similar semantic meaning and surrounding context thus should have similar contextualized word embeddings. The new objective is then defined as in Eq.~\ref{eq: knn}:
\begin{align}
&p(y_{t}) = p_{gen}p_{vocab}(y_t)+(1-p_{gen})p_{point}(y_t)\nonumber\\
\begin{split}
\label{eq: knn}
    &p_{point}(y_t)=\sum_{i}a_{t,i}\delta(y_t|x_i)\\&\phantom{p_{point}(y_t)}\approx \sum_{i; h_i^{T}e(y_t) \in \text{TopK}}a_{t,i}\delta(y_t|x_i)
\end{split}
\end{align}
  $e(y_t)$ is the encoded vector for $y_t$. The marginalization is performed only over the $k$ chosen source words. Eq.~\ref{eq: knn} is a lower bound of the data likelihood because it only marginalizes over a subset of $X$. In general a larger $k$ can tighten the bound to get a more accurate estimation and we analyze the effect of $k$ in Section~\ref{sec: align_results}. Note that the only extra parameters introduced by our model are the multi-layer-perceptron to compute the relation embedding $r(d_t, h_i)$. The marginalization in Eq.~\ref{eq: knn} can also be efficiently parallelized. An illustration of the generalized pointer is in Figure~\ref{fig: model}.
  
\section{Related Work}
Neural attention models~\citep{bahdanau2015neural} with the seq2seq architecture~\citep{sutskever2014sequence} have achieved impressive results in text summarization tasks. However, the attention vector comes from a weighted sum of source information and does not model the source-target alignment in a probabilistic sense. This makes it difficult to interpret or control model generations through the attention mechanism. In practice, people do find the attention vector is often blurred and suffers from poor alignment~\citep{koehn2017six, kiyono2018unsupervised,jain2019attention}. Hard alignment models, on the other hand, explicitly models the alignment relation between each source-target pair. Though theoretically sound, hard alignment models are hard to train. Exact marginalization is only feasible for data with limited length~\citep{yu2016online,aharoni2017morphological,deng2018latent,backes2018simulating}, or by assuming a simple copy generation process~\citep{vinyals2015pointer,gu2016incorporating,see2017get}. Our model can be viewed as a combination of soft attention and hard alignment, where a simple top-k approximation is used to train the alignment part~\citep{shankar2018surprisingly,shankar2018posterior}. The hard alignment generation probability is designed as a relation summation operation to better fit the summarization task. In this way, the generalized copy mode acts as a hard alignment component to capture the direct word-to-word transitions. On the contrary, the generation mode is a standard soft-attention structure to only model words that are purely functional, or need fusion, high-level inference and can be hardly aligned to any specific source context~\citep{daume2005induction}.
\section{Experiments and Results}
In the experiment, we compare seq2seq with attention, standard pointer generators and the proposed generalized pointer generator (GPG). To further analyze the effect of the generalized pointer, we implement a GPG model with only the point mode (GPG-ptr) for comparison. We first introduce the general setup, then report the evaluation results and analysis.
\subsection{General Setup}
\textbf{Dataset}: We perform experiments on the CNN/dailymail~\citep{hermann2015teaching}, English Gigaword~\citep{rush2015neural} and XSum~\citep{narayan2018don} dataset. CNN/DM contains online news with multi-sentence summaries (We use the non-anonymized version from \citet{see2017get}). English Gigaword paired the first sentence of news articles with its headline. XSum corpus provides a single-sentence summary for each BBC long story. We pick these three dataset as they have different properties for us to compare models. CNN/DM strongly favors extractive summarization~\citep{kryscinski2018improving}. Gigaword has more one-to-one word direct mapping (with simple paraphrasing)~\citep{napoles2012annotated} while XSum needs to perform more information fusion and inference since the source is much longer than the target~\citep{narayan2018don}.

\textbf{Model}: We use single-layer bi-LSTM encoders for all models. For comparison, hidden layer dimensions are set the same as in \citet{zhou2017selective} for Gigaword and \citet{see2017get} for CNN/DM and XSum. We train with batch size 256 for gigaword and 32 for the other two. The vocabulary size is set to 30k for all dataset. Word representations are shared between the encoder and decoder. We tokenize words with WordPiece segmentation~\citep{wu2016google} to eliminate the OOV problem.

\textbf{Inference}: We decode text using beam search~\citep{graves2012sequence} with beam size 10. We apply length normalization to rescale the score. Unlike \citet{see2017get,gehrmann2018bottom}, we do not explicitly impose coverage penalty since it brings extra hyper-parameters. Instead, for CNN/Dailymail, we use a simple tri-gram penalty~\citep{paulus2017deep} to prevent repeated generations. GPG models use an exact marginalization for testing and decoding, while for training and validation we use the top-$k$ approximation mentioned above. The decoder will first decode sub-word ids then map them back to the normal sentence. All scores are reported on the word level and thus comparable with previous results. When computing scores for multi-sentence summaries. The generations are split into sentences with the NLTK sentence tokenizer.

\subsection{Results and Analysis}
\label{sec: align_results}
The results are presented in the following order: We first study the effect of the hyperparameter $k$, then evaluate model generations by automatic metrics and look into the generation's level of abstraction. Finally, we report the human evaluation and word alignment accuracy.

\begin{figure}
\centering
\centerline{\includegraphics[width=0.6\columnwidth]{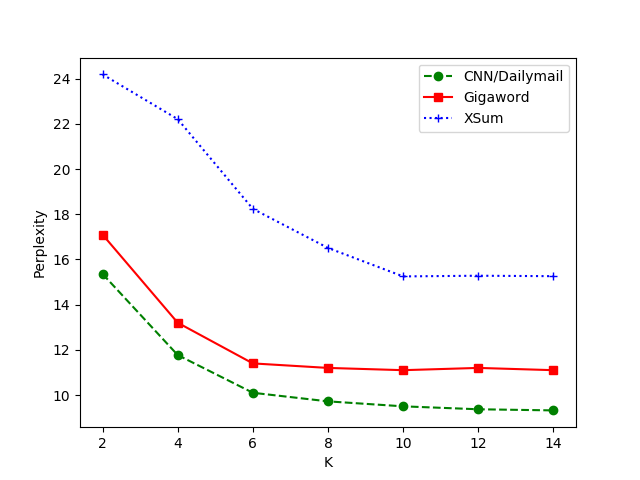}}
\caption{Test perplexity when increasing $k$}
\label{fig: k}
\end{figure}

\textbf{Effect of K}: $k$ is the only hyperparameter introduced by our model. Figure~\ref{fig: k} visualizes the effect of $k$ on the test perplexity. As mentioned in Sec~\ref{sec: gpg}, a larger $k$ is expected to tighten the estimation bound and improve the performance. The figure shows the perplexity generally decreases as increasing $k$. The effect on Gigaword and XSum saturates at $k=6$ and $10$ respectively, so we fix such $k$ value for later experiments. For the longer dataset CNN/Dailymail, the perplexity might still decrease a bit afterwards, but the improvement is marginal, so we set $k=14$ for the memory limit.

\begin{table}[t]
\centering
\begin{tabular}{@{}lcccc@{}}
\toprule
Method & R-1 & R-2 & R-L &PPL \\ \midrule
\small{Point.Gen.+Cov*} & 39.53 & 17.28 & 36.38 &\\
\small{Bottom Up$^\dagger$} & \textbf{41.22} & \textbf{18.68} & \textbf{38.34} &\\\hline
\small{seq2seq} & 39.79 & 17.37 & 36.34 &17.49\\
\small{Point.Gen.} & 40.03 & 17.52 & 36.77 &12.36\\
\small{GPG-ptr} & \underline{40.54} & \textbf{\underline{18.05}} & 37.19 &\underline{10.23}\\
\small{GPG} & \textbf{\underline{40.95}}& \underline{18.01} & \textbf{\underline{37.46}} &\underline{\textbf{9.37}}\\
\bottomrule
\end{tabular}
\caption{ROUGE score on CNN/Dailymail. * marks results from \citet{see2017get}, and $^\dagger$ from \citet{gehrmann2018bottom}. Underlined values are significantly better than Point.Gen. with $p=0.05$.}
\label{tab: cnn-accuracy}
\end{table}
\begin{table}[t]
\centering
\begin{tabular}{@{}lcccc@{}}
\toprule
Method & R-1 & R-2 & R-L &PPL \\ \midrule
\small{seq2seq*} & 34.04 & 15.95 & 31.68 &\\
\small{DRGD$^\dagger$} & \textbf{36.27} & \textbf{17.57} & \textbf{33.62} &\\\hline
\small{seq2seq} & 36.01 & 17.52 & 33.60 &18.92\\
\small{Point.Gen.} & 36.14 & 17.68 & 33.56 &14.90\\
\small{GPG-ptr} & 37.14 & \textbf{19.05} & \textbf{34.67} &12.32\\
\small{GPG} & \textbf{37.23} & 19.02 & 34.66 &\textbf{11.41}\\
\bottomrule
\end{tabular}
\caption{ROUGE score on Gigaword. * marks results from the word-based seq2seq implementation of \citet{zhou2017selective}, and $^\dagger$ from \citet{li2017deep}.}
\label{tab: align_gigaaccuracy}
\end{table}
\begin{table}[t]
\centering
\begin{tabular}{@{}lcccc@{}}
\toprule
Method & R-1 & R-2 & R-L &PPL \\ \midrule
\small{Point.Gen*} & 29.70 & 9.21 & 23.24 &\\
\small{T-CONVS2S*} & \textbf{31.89} & \textbf{11.54} & \textbf{25.75} &\\\hline
\small{seq2seq} & 31.90 & 11.15 & 25.48 &22.87\\
\small{Point.Gen.} & 31.87 & 11.20 & 25.42 &17.83\\
\small{GPG-ptr} & 31.49 & 11.02 & 25.37 &18.62\\
\small{GPG} & \textbf{\underline{33.11}} & \textbf{\underline{12.55}} & \textbf{\underline{26.57}} &\underline{\textbf{15.28}}\\
\bottomrule
\end{tabular}
\caption{ROUGE score on XSum. * marks results from \citet{narayan2018don}. Underlined values are significantly better than Point.Gen. with $p=0.05$.}
\label{tab: xsum-accuracy}
\end{table}
\begin{table*}[t!]
\centering
\resizebox{1\textwidth}{!}{
\begin{tabular}{lrrrr|rrr|rrr}
\multirow{2}{*}{\textbf{Models}} & \multicolumn{4}{c}{\small \% of NNs in CNN/Dailymail} &\multicolumn{3}{c}{\small \% of NNs in Gigaword}&\multicolumn{3}{c}{\small \% of NNs in XSum} \\
    & \textbf{NN-1\phantom{*}} & \textbf{NN-2\phantom{*}} &\textbf{NN-3\phantom{*}}&\textbf{NN-S\phantom{*}}& \textbf{NN-1} &\textbf{NN-2} &\textbf{NN-3}& \textbf{NN-1} & \textbf{NN-2} &\textbf{NN-3}\\
  \hline
  {\small Seq2seq} & 0.38\phantom{*} & 3.56\phantom{*} &7.98\phantom{*}&54.97\phantom{*}& 16.15 & 52.84 &73.76& 27.05 & 76.54 &92.07\\
  \small{Point.Gen.} & 0.04\phantom{*} & 1.51\phantom{*} &4.29\phantom{*}&35.82\phantom{*}& 13.99 & 47.79 &68.53& 19.45 & 66.68 &84.59\\
  \small{GPG-ptr} & 0.17\phantom{*} & 2.05\phantom{*} &5.08\phantom{*}&41.64\phantom{*}& 14.05 & 48.09 &70.70& 20.03 & 69.54 &87.14 \\
  \small{GPG} & 0.35\phantom{*} & 2.91\phantom{*} &5.66\phantom{*}&49.24\phantom{*}& 15.14 & 52.07 &72.73& 24.16 & 71.93 &87.94\\\hline
  \small{Reference} & 9.08\phantom{*} & 46.71\phantom{*} &67.99\phantom{*}&97.78\phantom{*}& 48.26 & 84.53 &94.43& 32.24 & 84.12 &95.92\\
\end{tabular}
}
\caption{\label{tab: novel}Proportion of novel n-grams (NN-1,2,3) and sentences (NN-S) on generated summaries. GPG generate more novel words compared with standard pointer generators, though still slightly lower than seq2seq.
}
\end{table*}
\textbf{Automatic Evaluation}: The accuracy is evaluated based on the standard metric ROUGE ~\citep{lin2004rouge} and the word perplexity on the test data. We report the ROUGE-1, ROUGE-2 and ROUGE-L F-score measured by the official script.
Table~\ref{tab: cnn-accuracy}, \ref{tab: align_gigaaccuracy} and \ref{tab: xsum-accuracy} lists the results for CNN/Dailymail, Gigaword and XSum respectively. Statistically significant results are underlined\footnote{Results on the Gigaword test set is not significant due to the smalle test size (1951 article-summary pairs).}. On the top two rows of each table, we include two results taken from current state-of-the-art word-based models. They are incomparable with our model because of the different vocabulary, training and decoding process, but we report them for completeness. Lower rows are results from our implemented models. Pointer generators bring only slight improvements over the seq2seq baseline. This suggests that after eliminating the OOV problem, the naive seq2seq with attention model can already implicitly learn most copy operations by itself. GPG models outperform seq2seq and pointer generators on all dataset. The improvement is more significant for more abstractive corpus Gigaword and XSum, indicating our model is effective at identifying more latent alignment relations. 

Notably, even the pure pointer model (GPG-ptr) without the generation mode outperforms standard pointer generators in CNN/DM and Gigaword, implying most target tokens can be generated by aligning to a specific source word. The finding is consistent with previous research claiming CNN/DM summaries are largely extractive~\citep{zhang2018abstractiveness,kryscinski2018improving}. Though the Gigaword headline dataset is more abstractive, most words are simple paraphrases of some specific source word, so pure pointer GPG-ptr can work well. This is different from the XSum story summarization dataset where many target words require high-level abstraction or inference and cannot be aligned to a single source word, so combining the point and generation mode is necessary for a good performance. 

The word perplexity results are consistent over all dataset (GPG $<$ GPG-ptr $<$ Point.Gen. $<$ seq2seq). The reduction of perplexity does not necessarily indicate an increase for the ROUGE score, especially for pointer generators. This might attribute to the different probability computation of pointer generators, where the probability of copied words are only normalized over the source words. This brings it an inherent advantage over other models where the normalization is over the whole 30k vocabularies.

\textbf{Level of Abstraction}: In Tab.~\ref{tab: novel}, we look into how abstractive the generated summaries are by calculating the proportion of novel unigram, bigram and trigrams that do not exist in the corresponding source text. On CNN/DM, as the generations contain multiple sentences, we further report the proportion of novel sentences (obtained with NLTK sent\_tokenize).
\begin{figure}
\centering
\centerline{\includegraphics[width=0.6\columnwidth]{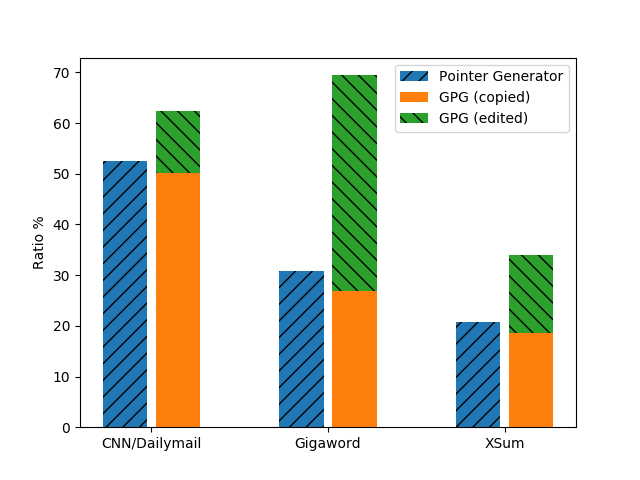}}
\caption{Pointing Ratio of the standard pointer generator and GPG (evaluated on the test data). GPG enables the point mode more often, but quite a few pointed tokens are edited rather than simply copied. }
\label{fig: point-ratio}
\end{figure}

Tab.~\ref{tab: novel} reflects the clear difference in the level of abstraction (seq2seq $>$ GPG $>$ GPG-ptr $>$ Point.Gen.). Though the seq2seq baseline generated most novel words, many of them are hallucinated facts (see Fig~\ref{fig: gen-example}), as has also been noted in \citet{see2017get}. The abstraction of GPG model is close to seq2seq and much higher than copy-based pointer generators. We believe it comes from their ability of ``editing" pointed tokens rather than simple copying them.
\begin{figure}[ht]
\begin{boxedminipage}{\columnwidth}
\small
\textbf{Article:} (...) and bild \novelh{reported} that the (...) \newline
\textbf{Summary:} (...) and bild \novelh{report} that the (...) $[0.964]$
\newline \rule{\columnwidth}{0.4pt}
\textbf{Article:} (...) thousands more \novelh{death} row prisoner (...) \newline
\textbf{Summary:} (...) thousands more \novelh{deaths} row (...) $[0.981]$
\newline \rule{\columnwidth}{0.4pt}
\textbf{Article:} (...) surviving \novelh{relatives} of a woman who (...) 
\newline
\textbf{Summary:} (...) \novelh{family} of woman who (...) $[0.984]$
\newline \rule{\columnwidth}{0.4pt}
\textbf{Article:} \novelh{jordan} 's crown prince (...) \newline
\textbf{Summary:} \novelh{jordanian} crown prince (...) $[0.993]$
\newline \rule{\columnwidth}{0.4pt}
\textbf{Article:} a  middle-aged man and a young \novelh{girl} (...) $[0.814]$ \newline
\textbf{Summary:} a man and a \novelh{child} died (...)
\newline \rule{\columnwidth}{0.4pt}
\textbf{Article:} (...) , was \novelh{abducted} in 2012 (...) \newline
\textbf{Summary:} (...) was \novelh{kidnapped} in (...) $[0.924]$
\end{boxedminipage}
\caption{Examples of summaries produced by GPG. Each two samples from CNN/DM, Gigaword and XSum (up to down). \novelh{bold} denotes novel words and their pointed source tokens. Bracketed numbers are the pointing probability ($1-p_{gen}$) during decoding.}
\label{fig: novel-example}
\end{figure}

To examine the pointing behavior of the GPG model, we visualize the average pointing ratio on three dataset in Fig.~\ref{fig: point-ratio}. The pointing ratio can be considered as the chance that a word is generated from the point mode instead of the generation mode. We compute it as $(1-p_{gen})\sum_{i}a_{t,i}\delta(y_t|x_i)/p(y_t)$, averaged over all target tokens in the test data. For the GPG model, we further split it into the copy ratio (words that are exactly copied) and the editing ratio (words that are edited). We find the GPG model enables the point mode more frequently than standord pointer generators, especially on the Gigaword dataset ($40\%$ more). This also explains why a pure pointer model is more effective for Gigaword and CNN/DM. More than $60\%$ target tokens can be generated from the point mode, while for XSum the ratio is less than $40\%$. Quite a few pointing operation includes text rewriting (\textcolor{ggreen}{green bar} in Fig.~\ref{fig: point-ratio}). This could explain why our model is able to generate more novel words.

A few examples are displayed in Fig~\ref{fig: novel-example}. We find our model frequently changes the tense (reported $\rightarrow$ report), singular/plural (death $\rightarrow$ deaths) or POS tag (jordan $\rightarrow$ jordanian) of words. Sometimes it also paraphrases (relatives $\rightarrow$ family) or abstracts a noun to its hypernym (girl $\rightarrow$ child). The word editing might be wrong though. For example, ``death row prisoner" is wrongly changed to ``deaths row prisoner" in the second example, possibly because this phrase is rarely seen so that the model made an error by mistaking ``death" as the main subject after ``thousands more"\footnote{It also reveals a limit of GPG model in that it only models token-level alignment. For phrases like death row prisoner, it cannot align it based on its compositional meaning.}.

\textbf{Human evaluation}: We further perform a human evaluation to assess the model generations. We focus on evaluating the fluency and faithfulness since the ROUGE score often fails to quantify them~\citep{schluter2017limits,cao2018faithful}. 100 random source-target pairs are sampled from the human-written DUC 2004 data for task 1\&2~\citep{over2007duc}. Models trained on Gigaword are applied to generate corresponding summaries. The gold targets, together with the model generations are randomly shuffled then assigned to 10 human annotators. Each pair is evaluated by three different people and the most agreed score is adopted. Each pair is assigned a 0-1 score to indicate (1) whether the target is fluent in grammar, (2) whether the target faithfully conveys the source information without hallucination and (3) whether the target is considered human-generated or machine-generated (like a 0/1 Turing test). The averaged score is reported in Table~\ref{tab: align_humaneval}. All models generally achieve high scores in fluency, but generations from GPG models are more faithful to the source information thereby have a larger chance of fooling people into believe they're human-generated (over 0.1 higher score on the 0/1 Turing test). This can be explained by GPG's capability at capturing more latent alignments. As shown in Figure ~\ref{fig: point-ratio}, GPG generates over half of the target words by its point mode. Words are generated by explicitly grounding on some source context instead of fabricating freely.
\begin{figure*}[ht]
\begin{boxedminipage}{\columnwidth}
\small
\textbf{Article:} (...) marseille prosecutor brice robin told cnn that `` so far no videos were used in the crash investigation . '' he added , `` a person who has such a video needs to immediately give it to the investigators . '' robin 's comments follow claims by two magazines , german daily bild and french paris match (...)
\newline 
\textbf{Seq2seq:} marseille prosecutor brice robin tells cnn that `` so far no videos were used in the crash investigation ''
\newline 
\textbf{Point.Gen:} \colorbox{blue1}{robin 's comments follow claims by two magazines , german daily bild and} \colorbox{blue1}{french} (..)
\newline 
\textbf{GPG:} \colorbox{blue1}{`` so far no videos were used in the crash investigation , '' prosecutor brice robin} \colorbox{blue3}{says} (..)
\newline \rule{\columnwidth}{0.4pt}
\textbf{Article:} surviving relatives of a woman who claimed she was raped \#\# years ago by the british queen 's representative in australia are seeking to withdraw a lawsuit against him , after the case drew widespread publicity in australia . \newline
\textbf{Seq2seq:} family of british queen 's representative in australia seeking to withdraw lawsuit against him .\newline
\textbf{Point.Gen:} \colorbox{blue1}{surviving relatives} \colorbox{blue6}{of} \colorbox{blue4}{british} \colorbox{blue1}{queen} \colorbox{blue6}{'s} \colorbox{blue1}{representative seeking} \colorbox{blue4}{to} \colorbox{blue2}{withdraw} \colorbox{blue1}{lawsuit} \colorbox{blue4}{against} \colorbox{blue2}{him} \colorbox{blue6}{.}\newline
\textbf{GPG:} \colorbox{blue1}{family} \colorbox{blue3}{of} \colorbox{blue1}{woman who claimed} \colorbox{blue2}{she} \colorbox{blue1}{was} \colorbox{blue2}{victim} \colorbox{blue5}{of} \colorbox{blue1}{british queen} \colorbox{blue6}{'s} \colorbox{blue1}{representative} \colorbox{blue2}{seeks} \colorbox{blue4}{to} \colorbox{blue1}{withdraw lawsuit} \colorbox{blue6}{.}
\newline \rule{\columnwidth}{0.4pt}
\textbf{Article:} police were called to love ranch brothel in crystal , nevada , after he was found unresponsive on tuesday . the american had to be driven to hospital (...) mr odom , 35 , has played       basketball for (...) lakers and clippers . he (...)  was suspended from the nba for violating its anti-drug policy (...) was named nba sixth man of the year (...) 
\newline 
\textbf{Seq2seq:} basketball legend odom odom has died at the age of 83 , police have confirmed .
\newline 
\textbf{Point.Gen:} \colorbox{blue6}{a former} \colorbox{blue1}{nba sixth man} \colorbox{blue6}{has been arrested on suspicion of} \colorbox{blue1}{anti-drug} \colorbox{blue6}{offences} \colorbox{blue5}{in} \colorbox{blue6}{the us state of california .}
\newline 
\textbf{GPG:} \colorbox{blue6}{the american} \colorbox{blue3}{basketball} \colorbox{blue6}{association} \colorbox{blue6}{(}\colorbox{blue1}{lakers}\colorbox{blue6}{)} \colorbox{blue6}{star lamar} \colorbox{blue1}{odom} \colorbox{blue6}{has} \colorbox{blue3}{been found} \colorbox{blue4}{unconscious} \colorbox{blue6}{in the us state of} \colorbox{blue4}{nevada}.
\end{boxedminipage}
\caption{Examples of generated summaries. Examples are taken from CNN/DM, Gigaword and XSum (from up to down). Darker means higher pointing probability.}
\label{fig: gen-example}
\end{figure*}

\begin{table}[t]
\centering
\begin{tabular}{c|c|c|c}
\toprule
 & Fluency & Faithfulness & 0/1  \\\hline
seq2seq & \textbf{0.83} & 0.61 & 0.53\\
Point.Gen. & 0.78 & 0.65 & 0.55\\
GPG-ptr & 0.79 & \textbf{0.78} & 0.67\\
GPG & 0.82 & 0.74 & \textbf{0.69}\\\hline
Gold & 0.96 & 0.92 & 0.96\\
\bottomrule
\end{tabular}
\caption{Human evaluation results on DUC 2004. 0/1 is the score for the 0/1 Turing test.}
\label{tab: align_humaneval}
\end{table}
Fig.~\ref{fig: gen-example} compares some generation snippets. As can be observed, seq2seq models tend to freely synthesize wrong facts not grounded on the source text, especially on the more difficult XSum dataset. In the last example, seq2seq only capture the subject ``odom" and some keywords ``police", ``basketball" then start to freely fabricate random facts. Pointer generators are slightly better as it is trained to directly copy keyword from the source. However, once it starts to enter the generation mode (``of {british}" in example 2 and ``has been arrested" in example 3), the generation also loses control. GPG largely alleviates the problems because it can point to an aligned source word, then transform it by a learned relation embedding. The explicit alignment modelling encourages the model to stay close to the source information.

\begin{table}[t]
\centering
\begin{tabular}{@{}cccc@{}}
\toprule
 & seq2seq & Point.Gen. & GPG  \\ \midrule
Prec & 0.361 & 0.435 (0.512) & 0.533 (0.628)\\
\bottomrule
\end{tabular}
\caption{Word Alignment Precision on DUC 2004. Number in bracket is the posterior alignment precision.}
\label{tab: alignment}
\end{table}
\textbf{Alignment Accuracy}: We also manually annotate the word alignment on the same 100 DUC 2004 pairs. Following \citet{daume2005induction}, words are allowed to be aligned with a specific source word, phrase or a ``null" anchor meaning that it cannot be aligned with any source word. The accuracy is only evaluated on the target words with a non-null alignment. For each target token, the most attended source word is considered as alignment~\citep{ghader2017does}. For the pointer generator and GPG, we also induce the posterior alignment by applying the Bayes' theorem. We report the alignment precision~\citep{och2000comparison} in Table~\ref{tab: alignment}, i.e., an alignment is considered as valid if it matches one of the human annotated ground truth.

The results show that GPG improves the alignment precision by 0.1 compared with the standard pointer generator. The posterior alignment is more accurate than the prior one (also reflected in Figure~\ref{fig: intro}), enabling better human interpretation.
\section{Conclusion}
In this chapter, we propose generalizing the pointer generator to go beyond exact copy operation. At each decoding step, the decoder can either generate from the vocabulary, copy or edit some source words by estimating a relation embedding. Experiments on abstractive summarization show the generalized model generates more abstract summaries yet faithful to the source information. The generalized pointer is able to capture richer latent alignment relationship beyond exact copies. This helps improve the alignment accuracy, allowing better model controllability and interpretation.

We believe the generalized pointer mechanism could have potential applications in many fields where tokens are not exactly copied. By integrating off-the-shelf knowledge bases to clearly model the transition relation embedding, it should further improve the interpretability and might be especially helpful under low-resource settings, which we leave for future work.

A major drawback of the proposed approach is that it can only model word-level alignment. In many cases, a more natural alignment relationship is at the segment-level (phrases or word spans). In the next chapter, we will show it is possible to use latent variables to jointly model segment-level and word-level alignment with a tractable marginal likelihood.
\cleardoublepage

\chapter[Latent Variable as Segmentation and Correspondence]{Latent Variable as Segmentation and Correspondence}
\label{chap: segment}

\lettrine[lines=3]{T}he neural attention model has achieved great success in data-to-text generation tasks. 
Though usually excelling at producing fluent text, it suffers from the problem of information missing, repetition and ``hallucination''. 
Due to the black-box nature of the neural attention architecture, avoiding these problems in a systematic way is non-trivial. 
To address this concern, we propose to explicitly segment target text into fragment units and align them with their data correspondences. 
The segmentation and correspondence are jointly learned as latent variables without any human annotations. The major different with the last chapter is that we can model both segment and word level alignment. This makes it appealing for specific tasks like data-to-text where the input is naturally segmented into several records. 
We further impose a soft statistical constraint to regularize the segmental granularity. 
The resulting architecture maintains the same expressive power as neural attention models, while being able to generate fully interpretable outputs with several times less computational cost. 
On both E2E and WebNLG benchmarks, we show the proposed model consistently outperforms its neural attention counterparts~\citep{shen2020neural}.

\section{Introduction}
Data-to-text generation aims at automatically producing natural language descriptions of structured database~\citep{reiter1997building}. 
Traditional statistical methods usually tackle this problem by breaking the generation process into a set of local decisions that are learned separately~\citep{belz2008automatic,angeli2010simple,kim2010generative,oya2014template}. 
Recently, neural attention models conflate all steps into a single end-to-end system and largely simplify the training process~\citep{mei2016talk,lebret2016neural,shen2017estimation,shen2018nexus,su2019improving,chang2020unsupervised}. 
However, the black-box conflation also renders the generation uninterpretable and hard to control~\citep{wiseman2018learning,shen2019select}. 
Verifying the generation correctness in a principled way is non-trivial. 
In practice, it often suffers from the problem of information missing, repetition and ``hallucination''~\citep{duvsek2018findings,duvsek2020evaluating}.
\begin{figure}[ht]
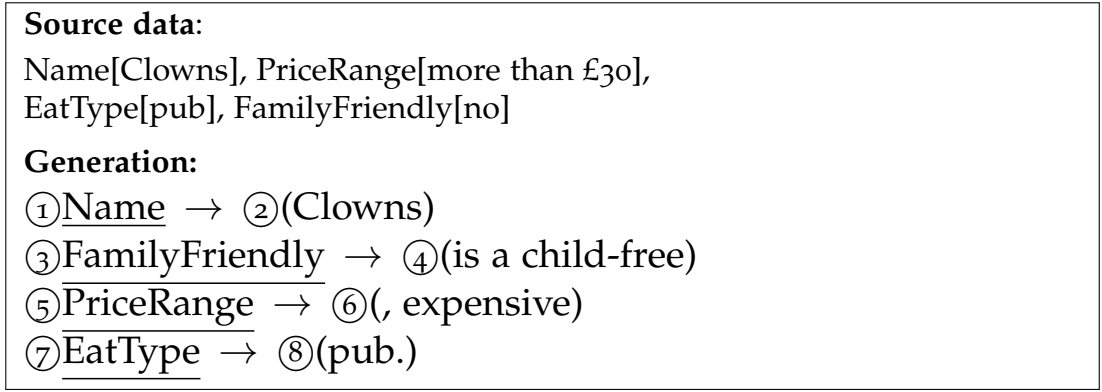

\centering
\fbox{
\parbox{0.9\linewidth}{
\textbf{Source data}:
\vspace{0.1cm}

Name[Clowns], PriceRange[more than \pounds 30],\\
EatType[pub], FamilyFriendly[no]

\vspace{0.2cm}

\textbf{Generation:}

\large \textcircled{\small 1}\underline{Name} $\,\to\,$ \large \textcircled{\small 2}(Clowns)

\large \textcircled{\small 3}\underline{FamilyFriendly} $\,\to\,$ \large \textcircled{\small 4}(is a child-free)

\large \textcircled{\small 5}\underline{PriceRange} $\,\to\,$ \large \textcircled{\small 6}(, expensive)

\large \textcircled{\small 7}\underline{EatType} $\,\to\,$ \large \textcircled{\small 8}(pub.)
}}
    \caption{Generation from our model on the E2E dataset. Decoding is performed segment-by-segment. Each segment realizes one data record. {\large{\textcircled{\small 1}}\raisebox{-0.9ex}{\~{}}\large{\textcircled{\small 8}}} mark the decision order in the generation process.
    }
    \label{fig: segment_intro}
\end{figure}

In this work, we propose to explicitly exploit the \emph{segmental structure} of text. 
Specifically, we assume the target text is formed from a sequence of segments. 
Every segment is the result of a two-stage decision: 
(1) Select a proper data record to be described and 
(2) Generate corresponding text by \emph{paying attention only to the selected data record}. 
This decision is repeated until all desired records have been realized. 
Figure~\ref{fig: segment_intro} illustrates this process.

Compared with neural attention, the proposed model has the following advantages: 
(1) We can monitor the corresponding data record for every segment to be generated. 
This allows us to easily control the output structure and verify its correctness\footnote{For example, we can perform a similar constrained decoding as in \citet{balakrishnan2019constrained} to rule out outputs with undesired patterns.}. 
(2) Explicitly building the correspondence between segments and data records can potentially reduce the hallucination, as noted in \citep{wu2018hard,deng2018latent} that hard alignment usually outperforms soft attention. 
(3) When decoding each segment, the model pays attention only to the selected data record instead of averaging over the entire input data. 
This largely reduces the memory and computational costs~\footnote{Coarse-to-fine attention~\citep{ling2017coarse,deng2017image} was proposed for the same motivation, but they resort to reinforcement learning which is hard to train, and the performance is sacrificed for efficiency.}.

To train the model, we \emph{do not} rely on any human annotations for the segmentation and correspondence, but rather marginalize over all possibilities to maximize the likelihood of target text, which can be efficiently done within polynomial time by dynamic programming. 
This is essentially similar to traditional methods of inducing segmentation and alignment with semi-markov models~\citep{daume2005induction,liang2009learning}. 
However, they make strong independence assumptions thus perform poorly as a generative model~\citep{angeli2010simple}. In contrast, the transition and generation in our model condition on \emph{all previously generated text}. 
By integrating an autoregressive neural network structure, our model is able to capture unbounded dependencies while still permitting tractable inference. 
The training process is stable as it does not require any sampling-based approximations. 
We further add a soft statistical constraint to control the segmentation granularity via posterior regularization~\citep{ganchev2010posterior}. On both the E2E and WebNLG benchmarks, our model is able to produce significantly higher-quality outputs while being several times computationally cheaper. 
Due to its fully interpretable segmental structure, it can be easily reconciled with heuristic rules or hand-engineered constraints to control the outputs.

\section{Related Work}
\label{sec: related}
Data-to-text generation is traditionally dealt with using a pipeline structure containing content planning, sentence planning and linguistic realization~\citep{reiter1997building}. 
Each target text is split into meaningful fragments and aligned with corresponding data records, either by hand-engineered rules~\citep{kukich1983design,mckeown1992text} or statistical induction~\citep{liang2009learning,koncel2014multi,qin2018learning}. 
The segmentation and alignment are used as supervision signals to train the content and sentence planner~\citep{barzilay2005collective,angeli2010simple}. 
The linguistic realization is usually implemented by template mining from the training corpus~\citep{kondadadi2013statistical,oya2014template}. 
Our model adopts a similar pipeline generative process, but integrates all the sub-steps into a single end-to-end trainable neural architecture. 
It can be considered as a neural extension of the PCFG system in \citet{konstas2013global}, with a more powerful transition probability considering inter-segment dependence and a state-of-the-art attention-based language model as the linguistic realizer. 
\citet{wiseman2018learning} tried a similar neural generative model to induce templates. 
However, their model only captures loose data-text correspondence and adopts a weak markov assumption for the segment transition probability. 
Therefore, it underperforms the neural attention baseline as for generation. Our model is also in spirit related to recent attempts at separating content planning and surface realization in neural data-to-text models~\citep{zhao2018comprehensive,puduppully2019data,moryossef2019step, ferreira2019neural}. 
Nonetheless, all of them resort to \emph{manual annotations or hand-engineered rules applicable only for a narrow domain}. 
Our model, instead, automatically learn the optimal content planning via exploring over exponentially many segmentation/correspondence possibilities.

There have been quite a few neural alignment models applied to tasks like machine translation~\citep{wang2018neural,deng2018latent}, character transduction~\citep{wu2018hard,shankar2018posterior} and summarization~\citep{yu2016online,shen2019improving}. 
Unlike word-to-word alignment, we focus on learning the alignment between data records and text segments.
Some works also integrate neural language models to jointly learn the segmentation and correspondence, e.g., phrase-based machine translation~\citep{huang2017towards}, speech recognition~\citep{wang2017sequence} and vision-grounded word segmentation~\citep{kawakami2018unsupervised}. 
Data-to-text naturally fits into this scenario since each data record is normally verbalized in one continuous text segment.

\section{Background: Data-to-Text}
Let $X,Y$ denote a source-target pair. $X$ is structured data containing a set of records and $Y$ corresponds to $y_1,y_2,\ldots,y_m$ which is a text description of $X$. The goal of data-to-text generation is to learn a distribution $p(Y|X)$ to automatically generate proper text describing the content of the data.

The neural attention architecture handles this task with an encode-attend-decode process~\citep{bahdanau2015neural}. 
The input $X$ is processed into a sequence of $x_1,x_2,\ldots,x_n$, normally by flattening the data records~\citep{wiseman2017challenges}. 
The encoder encodes each $x_i$ into a vector $h_i$. 
At each time step, the decoder attends to encoded vectors and outputs the probability of the next token by $p(y_t|y_{1:t-1},A_t)$. $A_t$ is a weighted average of source vectors:
\begin{equation}
\label{eq: segment_attention}
\begin{split}
A_t &= \sum_{i}\alpha_{t,i}h_{i}\\
\alpha_{t,i} &= \frac{e^{f(h_{i}, d_t)}}{\sum_j e^{f(h_{j}, d_t)}}
\end{split}
\end{equation}
$d_{t}$ is the hidden state of the decoder at time step $t$. 
$f$ is a score function to compute the similarity between  $h_i$ and $d_t$~\citep{luong2015effective}.

\section{Approach}

\begin{figure}[t!]
  \centering
\includegraphics[width=0.9\columnwidth]{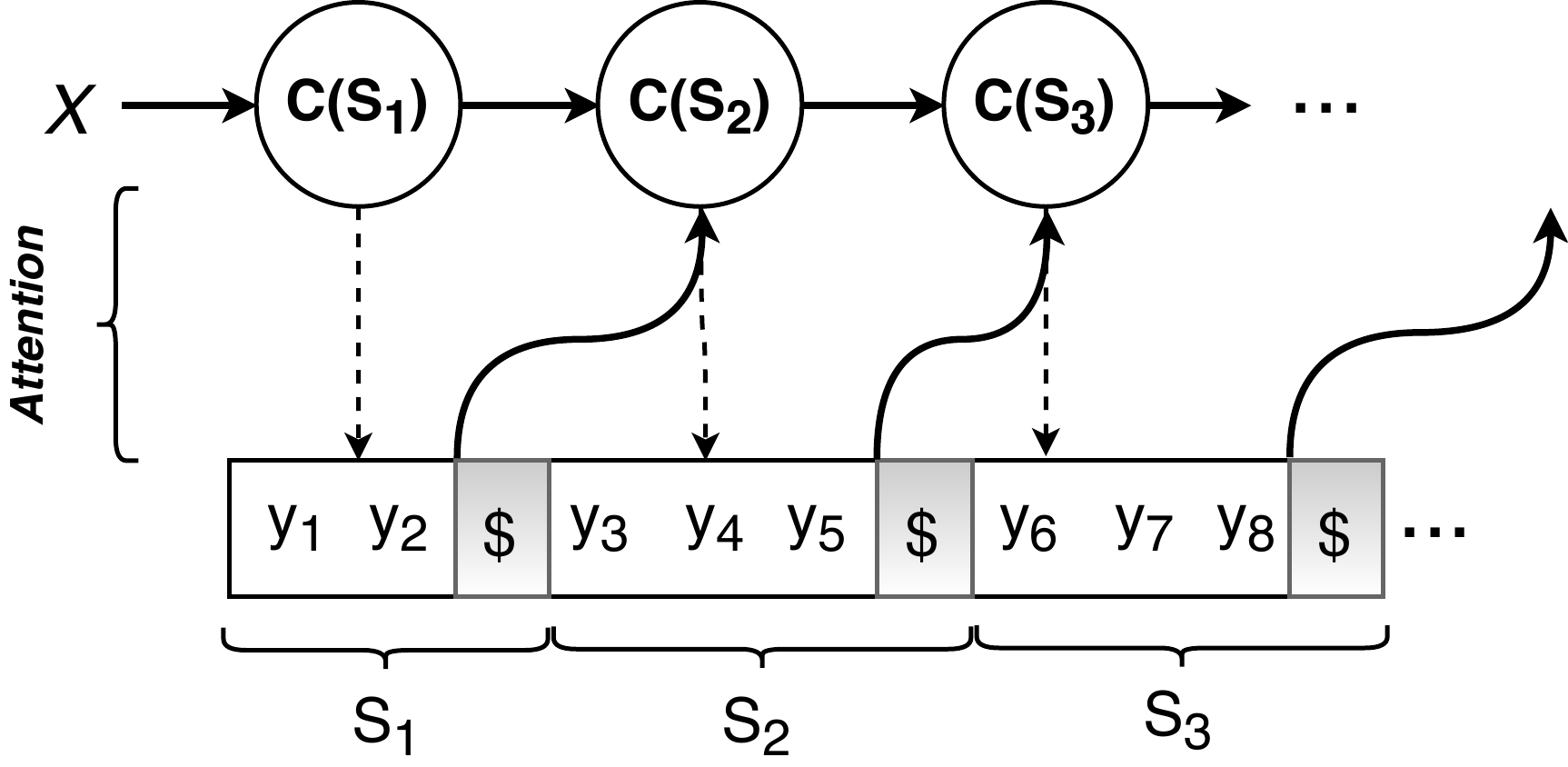}
\caption{\small{Generation process of our approach. 
Segment end symbol $\$$ is ignored when updating the state of the decoder. 
\textit{Solid arrows} indicate the transition model and \textit{dashed arrows} indicate the generation model. 
Every segment $s_o$ is generated by attending only to the corresponding data record $c(s_o)$.}}
\label{fig: architecture}
\end{figure}

Suppose the input data $X$ contains a set of records $r_1, r_2,..., r_K$. 
Our assumption is that the target text $y_{1:m}$ can be segmented into a sequence of fragments. 
Each fragment corresponds to one data record. As the ground-truth segmentation and correspondence are not available, we need to enumerate over all possibilities to compute the likelihood of $y_{1:m}$. 
Denote by $\mathcal{S}_y$ the set containing all valid segmentation of $y_{1:m}$. 
For any valid segmentation $s_{1:\tau_{s}}\in \mathcal{S}_y$, $\pi(s_{1:\tau_{s}})=y_{1:m}$, where $\pi$ means concatenation and $\tau_{s}$ is the number of segments. 
For example, let $m=5$ and $\tau_{s}=3$. 
One possible segmentation would be $s_{1:\tau_{s}}=\{\{ y_1, y_2, \$ \}, \{ y_3, \$ \}, \{ y_4, y_5, \$ \}\}$. $\$$ is the end-of-segment symbol and is removed when applying the $\pi$ operator. 
We further define $c(*)$ to be the corresponding data record(s) of $*$. 
The likelihood of each text is then computed by enumerating over all possibilities of $s_{1:\tau_{s}}$ and $c(s_{1:\tau_{s}})$:
\begin{equation}
\label{eq: segment_marginal}
\begin{split}
&p(y_{1:m}|X) = \sum_{s_{1:\tau_{s}}\in \mathcal{S}_y}p(s_{1:\tau_{s}}|X)\\
 &=\sum_{s_{1:\tau_{s}}\in \mathcal{S}_y}\prod_{o=1}^{\tau_{s}}\sum_{c(s_{o})=r_1}^{r_K}p(s_{o}|\pi(s_{<o}),c(s_{o}))\\&\times p(c(s_o)|\pi(s_{<o}),c(s_{<o}))
\end{split}
\end{equation}
Every segment is generated by first selecting the data record based on the \emph{transition probability} $p(c(s_o)|\pi(s_{<o}),c(s_{<o}))$, then generating tokens based on the word \emph{generation probability} $p(s_{o}|\pi(s_{<o}),c(s_{o}))$. Figure~\ref{fig: architecture} illustrates the generation process of our model.

\paragraph{Generation Probability} We base the generation probability on the same decoder as in neural attention models. The only difference is that \emph{the model can only pay attention to its corresponding data record}. The attention scores of other records are masked out when decoding $s_o$:
\begin{equation*}
    \alpha_{t,i} = \frac{e^{f(h_{i}, d_t)}\mathbbm{1}(x_i\in c(s_o))}{\sum_j e^{f(h_{j}, d_t)}\mathbbm{1}(x_j\in c(s_o))}
\end{equation*}
where $\mathbbm{1}$ is the indicator function. This forces the model to learn proper correspondences and enhances the connection between each segment and the data record it describes.

Following the common practice, we define the output probability with the pointer generator~\citep{see2017get,wiseman2017challenges}:
\begin{equation*}
\begin{split}
p_{gen} &= \sigma(\text{MLP}_{g}([d_{t}\circ A_{t}]))\\
p_{vocab} &= \text{softmax}(W_1 d_t + W_2 A_t)\\
p_\theta(y_{t}|y_{<t}) &= p_{gen}p_{vocab}(y_t)\\&+(1-p_{gen})\sum_{i:y_t=x_i}a_{t,i}
\end{split}
\end{equation*} 
$d_t$ is the decoder's hidden state at time step $t$. $\circ$ denotes vector concatenation. $A_t$ is the context vector. MLP indicates multi-layer perceptron and $\sigma$ normalizes the score between $(0,1)$. $W_1$ and $W_2$ are trainable matrices. $p_{gen}$ is the probability that the word is generated from a fixed vocabulary distribution $p_{vocab}$ instead of being copied. The final decoding probability $p_\theta(y_t)$ is marginalized over $p_{vocab}$ and the copy distribution. The generation probability of $s_o$ factorizes over the words within it and the end-of-segment token:
\begin{equation*}
    p(s_{o}|\pi(s_{<o}),c(s_{o})) =  p_\theta(\$|y_{1:t}) \prod_{y_t \in s_o} p_\theta(y_t|y_{<t})
\end{equation*}

\paragraph{Transition Probability} We make a mild assumption that $c(s_o)$ is dependent only on $c(s_{o-1})$ and $\pi(s_{1:o-1})$ but irrelevant of $c(s_{<o-1})$, which is a common practice when modelling alignment~\citep{och1999improved,yu2016online,shankar2018posterior}. The transition probability is defined as:
\begin{equation}
\label{eq: transit}
\begin{split}
&p(c(s_o)=r_i|c(s_{<o}), \pi(s_{<o}))\\ \approx & p(c(s_o)=r_i|c(s_{o-1}), \pi(s_{<o}))\\
 \propto&f(r_i)^T [M^T A_{s_{o-1}} + N^T d_{s_{o-1}}]
\end{split}
\end{equation}
A softmax layer is finally applied to the above equation to normalize it as a proper probability distribution. $f(r_i)$ is a representation of $r_i$, which is defined as a max pooling over all the word embeddings contained in $r_i$. $A_{s_{o-1}}$ is the attention context vector when decoding the last token in $s_{o-1}$, defined as in Equation~\ref{eq: segment_attention}. It carries important information from $c(s_{o-1})$ to help predict $c(s_o)$. $d_{s_{o-1}}$ is the hidden state of the neural decoder which goes through all history tokens $\pi(s_{1:o-1})$. $M,N$ are trainable matrices to project $A_{s_{o-1}}$ and $d_{s_{o-1}}$ into the same dimension as $f(r_i)$.

We further add one constraint to prohibit \emph{self-transition}, which can be easily done by zeroing out the transition probability in Equation~\ref{eq: transit} when $c(s_{o})=c(s_{o-1})$. This forces the model to group together text describing the same data record.

Since Equation~\ref{eq: transit} conditions on all previously generated text, it is able to capture more complex dependencies as in semi-markov models~\citep{liang2009learning,wiseman2018learning}.

\paragraph{Null Record} In our task, we find some frequent phrases, e.g., ``it is", ``and", tend to be wrongly aligned with some random records, similar to the garbage collection issue in statistical alignment~\citep{brown1993mathematics}. This hurt the model interpretability. Therefore, we introduce an additional null record $r_0$ to attract these non-content phrases. The context vector when aligned to $r_0$ is a zero vector so that the decoder will decode words based solely on the language model without relying on the input data.
\paragraph{Training} Equation~\ref{eq: segment_marginal} contains exponentially many combinations to enumerate over. Here we show how to efficiently compute the likelihood with the forward algorithm in dynamic programming~\citep{rabiner1989tutorial}. We define the forward variable $\alpha(i,j)=p(y_{1:i},c(y_i)=j|X)$. With the base $\alpha(1,j)=p(y_1|c(y_{1})=j)$. The recursion goes as follows for $i=1,2,\ldots,m-1$:
\begin{equation}
\label{eq: segment_forward}
\begin{split}
\alpha&(i+1,j)=\sum_{p=1}^{i}\sum_{q=r_0}^{r_K}\alpha(p,q)\\&\times p(c(y_{p+1})=j|c(y_p)=q, y_{1:p})\\
&\times p(y_{p+1:i+1}|c(y_{p+1:i+1})=q, y_{1:p})\\
&\times p(\$|c(y_{p+1:i+1})=q, y_{1:i+1})
\end{split}
\end{equation}
The final likelihood of the target text can be computed as $p(y_{1:m}|X)=\sum_{j=r_0}^{r_K}\alpha(m,j)$. As the forward algorithm is fully differentiable, we maximize the log-likelihood of the target text by backpropagating through the dynamic programming. The process is essentially equivalent to the generalized EM algorithm~\citep{eisner2016inside}. By means of the modern automatic differentiation tools, we avoid the necessity to calculate the posterior distribution manually~\citep{kim2018tutorial}.

To speed up training, we set a threshold $L$ to the maximum length of a segment as in \citet{liang2009learning,wiseman2018learning}. This changes the complexity in Equation~\ref{eq: segment_forward} to a constant $O(LK)$ instead of scaling linearly with the length of the target text. Moreover, as pointed out in \citet{wang2017sequence}, the computation for the longest segment can be reused for shorter segments. We therefore first compute the generation and transition probability for the whole sequence in one pass. The intermediate results are then cached to efficiently proceed the forward algorithm without any re-computation.

One last issue is the numerical precision, it is important to use the log-space binary operations to avoid underflow~\citep{kim2017structured}.

\begin{table}[ht]
\small
\centering
\begin{tabular}{@{}l@{}}
Near[riverside], Food[French], EatType[pub], Name[Cotto]\\
\hline
\begin{tabular}[c]{@{}l@{}}
1. [Near]$_{\text{Near}}$[the]$_{\text{Null}}$[riverside]$_{\text{Near}}$[is a]$_{\text{Null}}$[French]$_{\text{Food}}$\\
\; [pub]$_{\text{EatType}}$[called]$_{\text{Null}}$[Cotto]$_{\text{Name}}$[.]$_{\text{Null}}$ \\
2. [Near the riverside]$_{\text{Near}}$[is]$_{\text{Null}}$[a French]$_{\text{Food}}$[pub]$_{\text{EatType}}$\\
\; [called Cotto]$_{\text{Name}}$[.]$_{\text{Null}}$ \\
3. [Near the riverside]$_{\text{Near}}$[is a French]$_{\text{Food}}$[pub]$_{\text{EatType}}$\\
\; [called Cotto .]$_{\text{Name}}$ \\
4. [Near the riverside]$_{\text{Near}}$[is a French pub]$_{\text{Food}}$\\
\; [called Cotto .]$_{\text{Name}}$
\end{tabular} \\
\hline
\end{tabular}
\caption{\small Segmentation with various granularities. 1 is too fine-grained while 4 is too coarse. We expect a segmentation like 2 or 3 to better control the generation.}
\label{tab: granularity}
\end{table}

\paragraph{Segmentation Granularity}
\label{para: granu}
There are several valid segmentations for a given text. As shown in Table~\ref{tab: granularity}, when the segmentation (example 1) is too fine-grained, controlling the output information becomes difficult because the content of one data record is realized in separate pieces~\footnote{The finer-grained segmentation might be useful if the focus is on modeling the detailed discourse structure instead of the information accuracy~\citep{reed2018can,balakrishnan2019constrained}, which we leave for future work.}. When it is too coarse, the alignment might become less accurate (as in Example 4, ``pub" is wrongly merged with previous words and aligned together to the ``Food" record). In practice, we expect the segmentation to stay with accurate alignment yet avoid being too brokenly separated. To control the granularity as we want, we utilize posterior regularization~\citep{ganchev2010posterior} to constrain the expected number of segments for each text~\footnote{We can also utilize some heuristic rules to help segmentation. For example, we can prevent breaking syntactic elements obtained from an external parser~\citep{yang-etal-2019-low} or match entity names with handcrafted rules~\citep{chen2018sheffield}. The interpretability of the segmental structure allows easy combination with these rules. We focus on a general \emph{domain-agnostic} method in this paper, though heuristic rules might bring further improvement under certain cases.}, which can be calculated by going through a similar forward pass as in Equation~\ref{eq: segment_forward}~\citep{eisner2002parameter}. Most computation is shared without significant extra burden. The final loss function is:
\begin{equation}
\label{eq: loss}
-\log \mathbbm{E}_{\mathcal{S}_y}p(s_{1:\tau_s}|X) + \max(\abs{\mathbbm{E}_{\mathcal{S}_y}\tau_s-\eta}, \gamma)
\end{equation}
$\log \mathbbm{E}_{\mathcal{S}_y}p(s_{1:\tau_s}|X)$ is the log-likelihood of target text after marginalizing over all valid segmentations. $\mathbbm{E}_{\mathcal{S}_y}\tau_s$ is the expected number of segments and $\eta, \gamma$ are hyperparameters. We use the max-margin loss to encourage $\mathbbm{E}_{\mathcal{S}_y}\tau_s$ to stay close to $\eta$ under a tolerance range of $\gamma$. 


\paragraph{Decoding}
\label{para: decoding}
The segment-by-segment generation process allows us to easily constrain the output structure. Undesirable patterns can be rejected before the whole text is generated. We adopt three simple constraints for the decoder:
\begin{enumerate}
    \item Segments must not be empty.
    \item The same data record cannot be realized more than once (except for the null record).
    \item The generation will not finish until all data records have been realized.
\end{enumerate}
Constraint 2 and 3 directly address the information repetition and missing problem. When segments are incrementally generated, the constraints will be checked against for validity. Note that adding the constraints hardly incur any cost, the decoding process is still finished \emph{in one pass}. No post-processing or reranking is needed.

\paragraph{Computational Complexity} Suppose the input data has $M$ records and each record contains $N$ tokens. The computational complexity for neural attention models is $O(MN)$ at each decoding step where the whole input is retrieved. Our model, similar to chunkwise attention~\citep{chiu2017monotonic} or coarse-to-fine attention~\citep{ling2017coarse}, reduces the cost to $O(M+N)$, where we select the record in $O(M)$ at the beginning of each segment and attend only to the selected record in $O(N)$ when decoding every word. For larger input data, our model can be significantly cheaper than neural attention models.
\section{Experiment Setup}
\paragraph{Dataset}
We conduct experiments on the E2E~\citep{novikova2017e2e} and WebNLG~\citep{gardent2017webnlg} datasets. E2E is a crowd-sourced dataset containing 50k instances in the restaurant domain. The inputs are dialogue acts consisting of three to eight slot-value pairs. WebNLG contains 25k instances describing entities belonging to fifteen distinct DBpedia categories. The inputs are up to seven RDF triples of the form \emph{(subject, relation, object)}. 
\paragraph{Implementation Details}
We use a bi-directional LSTM encoder and uni-directional LSTM decoder for all experiments. Input data records are concatenated into a sequence and fed into the encoder. We choose the hidden size of encoder/decoder as 512 for E2E and 256 for WebNLG. The word embedding is with size 100 for both datasets and initialized with the pre-trained Glove embedding~\footnote{\url{nlp.stanford.edu/data/glove.6B.zip}}~\citep{pennington2014glove}. We use a drop out rate of $0.3$ for both the encoder and decoder. Models are trained using the Adam optimizer~\citep{kingma2014adam} with batch size 64. The learning rate is initialized to 0.01 and decays an order of magnitude once the validation loss increases. All hyperparameters are chosen with grid search according to the validation loss. Models are implemented based on the open-source library PyTorch~\citep{pytorch}.
We set 
the hyperparameters in Eq.~\ref{eq: loss} as $\eta=K, \gamma=1$ (recall that $K$ is the number of records in the input data). The intuition is that every text is expected to realize the content of all $K$ input records. It is natural to assume every text can be roughly segmented into $K$ fragments, each corresponding to one data record. A deviation of $K\pm 1$ is allowed for noisy data or text with complex structures.
\paragraph{Metrics}
We measure the quality of system outputs from three perspectives: (1) \emph{word-level overlap} with human references, which is a commonly used metric for text generation. We report the scores of BLEU-4~\citep{papineni2002bleu}, ROUGE-L~\citep{lin2004rouge}, Meteor~\citep{banerjee2005meteor} and CIDEr~\citep{vedantam2015cider}
. (2) \emph{human evaluation}. Word-level overlapping scores usually correlate rather poorly with human judgements on fluency and information accuracy~\citep{reiter2009investigation,novikova2017we}. Therefore, we passed the input data and generated text to human annotators to judge if the text is fluent by grammar (scale 1-5 as in \citet{belz2006comparing}), contains wrong fact inconsistent with input data, repeats or misses information. We report the \emph{averaged score} for fluency and \emph{definite numbers} for others. The human is conducted on a sampled subset from the test data. To ensure the subset covers inputs with all possible number of records ($K\in[3,8]$ for E2E and $K\in[1,7]$ for WebNLG), we sample 20 instances for every possible $K$. Finally,we obtain 120 test cases for E2E and 140 for WebNLG~\footnote{The original human evaluation subset of WebNLG is randomly sampled, most of the inputs contain less than 3 records, so we opt for a new sample for a thorough evaluation.}. (3) \emph{Diversity of outputs}. Diversity is an important concern for many real-life applications. We measure it by the number of unique unigrams and trigrams over system outputs, as done in \citet{duvsek2020evaluating}.
\section{Results}
\begin{figure}[t]
  \centering
\includegraphics[width=\columnwidth]{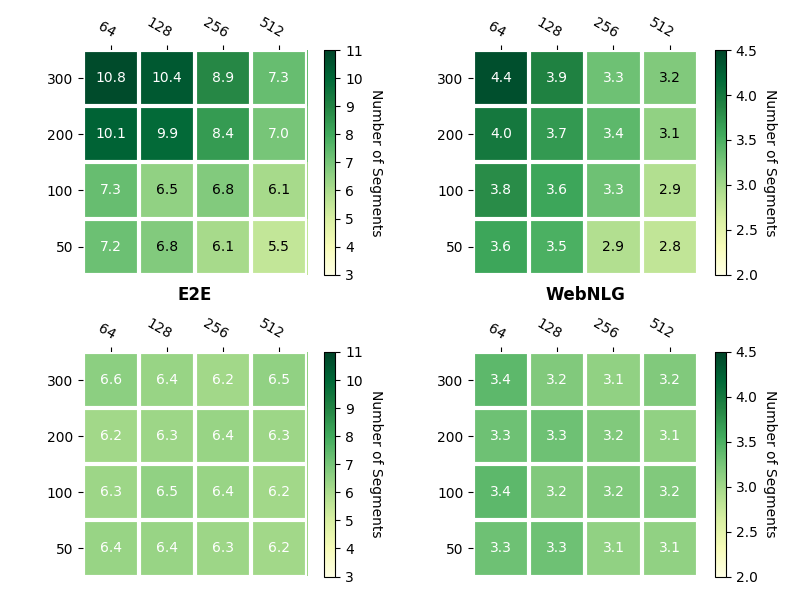}
\caption{\small Average expected number of segments with varying hyperparameters. x-axis is the encoder/decoder hidden size and y-axis is the word embedding size. Upper two figures are without the granularity regularization and the bottom two are with regularization.}
\label{fig: n_segment}
\end{figure}
\begin{table*}[t!]
  \small
  \begin{center}
    \resizebox{1\textwidth}{!}{
    \begin{tabular}{l|cccc|cccc|cc}
    \hline 
     \textbf{Metrics} &
     \multicolumn{4}{c|}{\textbf{Word Overlap}} & \multicolumn{4}{c|}{\textbf{Human Evaluation}} & \multicolumn{2}{c}{\textbf{Diversity}} \\ 
     \textbf{Models} & \textbf{BLEU} & \textbf{R-L} & \textbf{Meteor} & \textbf{CIDEr} & \textbf{Fluent} & \textbf{Wrong} & \textbf{Repeat} & \textbf{Miss}  & \textbf{Dist-1} & \textbf{Dist-3} \\
     \hline
     \textbf{\textsc{Slug}}  & \textbf{0.662} & \textbf{0.677} & 0.445 & \textbf{2.262} & 4.94 & 5 & \textbf{0} & 17 & 74 & 507 \\
     \textbf{\textsc{DANGNT}} & 0.599 & 0.663  & 0.435 & 2.078 & 4.97 & \textbf{0} & \textbf{0} & 21 & 61 & 301 \\
     \textbf{\textsc{TUDA}}  & 0.566 & 0.661  & \textbf{0.453} & 1.821 & \textbf{4.98} & \textbf{0} & \textbf{0} & \textbf{10} & 57  & 143 \\
      \textbf{\textsc{N\_Temp}}  & 0.598 & 0.650  & 0.388 & 1.950 & 4.84 & 19 & 3 & 35 & \textbf{119} & \textbf{795}\\
     \hline
     \textbf{\textsc{PG}} & 0.638 & 0.677 & 0.449 & 2.123 & 4.91 & 15 &  1 & 29 & 133 & 822 \\
    \textbf{\textsc{Ours}}  & 0.647 & \textbf{0.683}  & 0.453 & 2.222 & \textbf{4.96} & \textbf{0} & 1 & 15 & 127 & 870\\
     \textbf{\textsc{Ours (+R)}}  & 0.645 & 0.681  & 0.452 & 2.218 & 4.95 & \textbf{0} & \textbf{0} & 13 & 133  & 881\\
     \textbf{\textsc{Ours (+RM)}}  & \textbf{0.651} & 0.682 & \textbf{0.455} & \textbf{2.241} & 4.95 & \textbf{0} & \textbf{0} & \textbf{3} & \textbf{135} & \textbf{911}\\
     \hline
    \end{tabular}
    }
  \end{center}
   \caption{\small Automatic and human evaluation results on E2E dataset. \textbf{SLUG}, \textbf{DANGNT}, \textbf{TUDA} and \textbf{N\_TEMP} are from previous works and the other models are our own implementations.}
     \label{tab: e2eresults}%
     \vspace{-0.3cm}
\end{table*}%
\begin{table*}[t!]
  \small
  \begin{center}
    \resizebox{1\textwidth}{!}{
    \begin{tabular}{l|cccc|cccc|cc}
    \hline 
     \textbf{Metrics} &
     \multicolumn{4}{c|}{\textbf{Word Overlap}} & \multicolumn{4}{c|}{\textbf{Human Evaluation}} & \multicolumn{2}{c}{\textbf{Diversity}} \\ 
     \textbf{Models} & \textbf{BLEU} & \textbf{R-L} & \textbf{Meteor} & \textbf{CIDEr} & \textbf{Fluent} & \textbf{Wrong} & \textbf{Repeat} & \textbf{Miss}  & \textbf{Dist-1} & \textbf{Dist-3} \\
     \hline
     \textbf{\textsc{Melbourne}}  & 0.450 & \textbf{0.635} & 0.376 & \textbf{2.814} & 4.16 & 42 & 22 & 37 & 3167 & \textbf{13,744} \\
     \textbf{\textsc{UPF-FORGe}} & 0.385 & 0.609  & 0.390 & 2.500 & 4.08 & \textbf{29} & \textbf{6} & \textbf{28} & \textbf{3191} & 12,509 \\\hline
     \textbf{\textsc{PG}} & 0.452 & 0.652 & 0.384 & 2.623 & 4.13 & 43 &  26 & 42 & 3,218 & 13,403 \\
    \textbf{\textsc{Ours}}  & 0.453 & 0.656  & 0.388 & 2.610 & 4.23 & 26 & 19 & 31 & 3,377  & 14,516\\
     \textbf{\textsc{Ours (+R)}}  & 0.456 & \textbf{0.657}  & 0.390 & \textbf{2.678} & \textbf{4.28} & \textbf{18} & \textbf{2} & 24 & 3,405 & 14,351\\
     \textbf{\textsc{Ours (+RM)}}  & \textbf{0.461} & 0.654  & \textbf{0.398} & 2.639 & 4.26 & 23 & 4 & \textbf{5} & \textbf{3,457}  & \textbf{14,981}\\
     \hline
    \end{tabular}
    }
  \end{center}
   \caption{\small Automatic and human evaluation results on WebNLG dataset. \textbf{MELBOURNE} and \textbf{UPFUPF-FORGE} are from previous works and the other models are our own implementations.}
     \label{tab: webnlgresults}%
     \vspace{-0.3cm}
\end{table*}%
\begin{figure*}[ht]
\footnotesize{
\begin{center}
\resizebox{\textwidth}{!}{
\begin{tabular}{ll}
\cellcolor{red!30}{Egg Harbor Township, New Jersey \underline{isPartOf} New Jersey}&\cellcolor{orange!30}{Atlantic City International Airport \underline{Location Identifier} ``KACY" ICAO}\\\cellcolor{yellow!50}{Atlantic City International Airport \underline{location} Egg Harbor Township, New Jersey}&\cellcolor{green!40}{Egg Harbor Township, New Jersey \underline{country} United States}
 \\\cellcolor{blue!20}{Egg Harbor Township, New Jersey \underline{isPartOf} Atlantic County, New Jersey}&
 \\
\\
\end{tabular}
}
{\def\arraystretch{2}\tabcolsep=3pt
\begin{tabular}{l L}
PG         & Atlantic City International Airport is located in Egg Harbor Township , New Jersey , United States . \textbf{It is located in Egg Harbor Township , New Jersey} . \\ \hline
Ours & \colorbox{orange!30}{KACY is the ICAO location identifier of Atlantic City International Airport ,} \colorbox{yellow!50}{which is located at Egg Harbor Township , New jersey}\colorbox{green!40}{, in the United States]} \colorbox{orange!30}{\textbf{. The ICAO location identifier of Atlantic City International Airport is KACY .}} \\
\hline 
Ours (+R)     & \colorbox{orange!30}{KACY is the ICAO location identifier of Atlantic City International Airport ,} \colorbox{yellow!50}{which is located at Egg Harbor Township , New jersey}\colorbox{green!40}{, in the United States]} . \\
\hline
Ours (+RM) &  \colorbox{orange!30}{KACY is the ICAO location identifier of Atlantic City International Airport ,}\colorbox{yellow!50}{which is located at} \colorbox{yellow!50}{Egg Harbor Township , New jersey}\colorbox{green!40}{, in the United States} .\colorbox{blue!30}{The Egg Harbor Township is a part} \colorbox{blue!20}{of Atlantic County , New Jersey} . \colorbox{red!30}{Egg Harbor Township is a part of New Jersey .}
\end{tabular}}
\end{center}
\caption{\small Example generations from WebNLG. Relation types are \underline{underlined} and repeated generations are \textbf{bolded}. Segments and corresponding records in our model are marked in the same color. By adding explicit constraints to the decoding process, repetition and missing issues can be largely reduced. (better viewed in color)}
\label{fig: webnlgexample}
}
\end{figure*}
\begin{table*}[!htbp]
\fontsize{200pt}{300pt}\selectfont
\resizebox{\textwidth}{!}{
\begin{tabular}{|ll|}
\hline
Input:   & \colorbox{blue!5}{\colorbox{blue!15}{\strut [name the mill]}\colorbox{blue!18}{\strut [eattype} \colorbox{blue!30}{\strut restaurant]}\colorbox{blue!17}{\strut [food} \colorbox{blue!40}{\strut english]}\colorbox{blue!10}{\strut [pricerange} \colorbox{blue!7}{\strut moderate]}\colorbox{blue!8}{\strut [customerrating} \colorbox{blue!60}{\strut 1} \colorbox{blue!16}{\strut out of 5]}[area \colorbox{blue!17}{\strut riverside]}} ...  \\
PG: & the mill is a \textbf{\colorbox{blue!50}{\strut low} - priced} restaurant \textbf{in the city centre that delivers take - away} . it is located near caf\'e rouge.\\
Input:   & \colorbox{red!0}{\colorbox{blue!0}{[name the mill]}\colorbox{blue!0}{[eattype} \colorbox{blue!0}{restaurant]}\colorbox{blue!0}{[food} \colorbox{blue!0}{english]}\colorbox{red!10}{\strut [pricerange} \colorbox{red!50}{\strut moderate]}\colorbox{blue!0}{[customerrating} \colorbox{blue!0}{1} {out of 5]}\colorbox{blue!0}{[area}\colorbox{blue!0}{riverside]} ...}  \\ 
Ours:     &      [the mill][restaurant][near caf\'e rouge][in riverside][serves english food][at \colorbox{red!50}{\strut moderate} prices][. it is kid friendly and]...\\\hline      
\end{tabular}
}
\caption{\label{tab: attention}\small{(E2E) Attention map when decoding the word ``low" in the PG model and ``moderate" in our model. Hallucinated contents are \textbf{bolded}. The PG model wrongly attended to other slots thereby ``hallucinated" the content of ``low-priced". Our model always attends to one single slot instead of averaging over the whole inputs, the chance of hallucination is largely reduced.}}
\end{table*}
In this section, we first show the effects of the granularity regularization we proposed, then compare model performance on two datasets and analyze the performance difference. Our model is compared against the neural attention-based pointer generator (\textbf{PG}) which does not explicit learn the segmentation and correspondence. To show the effects of the constrained decoding, we run our model with only the first constraint to prevent empty segments (denoted by \textbf{ours} in experiments), with the first two constraints to prevent repetition (denoted by \textbf{ours (+R)}), and with all constraints to further reduce information missing (denoted by \textbf{ours (+RM)}).
\paragraph{Segmentation Granularity}
We show the effects of the granularity regularization (\cref{para: granu}, \nameref{para: granu}) in Fig~\ref{fig: n_segment}. When varying the model size, the segmentation granularity changes much if no regularization is imposed. Intuitively if the generation module is strong enough (larger hidden size), it can accurately estimate the sentence likelihood itself without paying extra cost of switching between segments, then it tends to reduce the number of transitions. Vice versa, the number of transitions will grow if the transition module is stronger (larger embedding size). With the regularization we proposed, the granularity remains what we want regardless of the hyperparameters. We can thereby freely decide the model capacity without worrying about the difference of segmentation behavior.
\paragraph{Results on E2E}
 On the E2E dataset, apart from our implementations, we also compare agianst outputs from the \textbf{SLUG}~\citep{juraska2018deep}, the overall winner of the E2E challenge (seq2seq-based), \textbf{DANGNT}~\citep{nguyen2018structurebased}, the best grammar rule based model, \textbf{TUDA}~\citep{puzikov2018e2e}, the best template based model, and the autoregressive neural template model (\textbf{N\_TEMP}) from \citet{wiseman2018learning}. SLUG uses a heuristic slot aligner based on a set of handcrafted rules and combine a complex pipeline of data augmentation, selection, model ensemble and reranker, while our model has a simple end-to-end learning paradigm with no special delexicalizing, training or decoding tricks. 
 Table~\ref{tab: e2eresults} reports the evaluated results. Seq2seq-based models are more diverse than rule-based models at the cost of higher chances of making errors. As rule-based systems are by design always faithful to the input information, they made zero wrong facts in their outputs. Most models do not have the fact repetition issue because of the relatively simple patterns in the E2E dataset. therefore, adding the (+R) constraint only improves the performance minorly. The (+RM) constraint reduces the number of information missing to 3 without hurting the fluency. All the 3 missing cases are because of the wrong alignment between the period and one data record, which can be easily fixed by defining a simple rule. We put the error analysis in \cref{sec: error}.  N\_Temp performs worst among all seq2seq-based systems because of the restrictions we mentioned in \cref{sec: related}. As also noted by the author, it trades-off the generation quality for interpretability and controllability. In contrast, our model, despite relying on no heuristics or complex pipelines, \emph{made zero wrong facts with the lowest information missing rate, even surpassing rule-based models}. It also maintains interpretable and controllable without sacrificing the generation quality.
\paragraph{Results on WebNLG}
Table~\ref{tab: webnlgresults} reports the results evaluated on the WebNLG dataset. We also include results from \textbf{MELBOURNE}, a seq2seq-based system achieving highest scores on automatic metrics in the WebNLG challenge and \textbf{UPF-FORGE}, a classic grammar-based system that wins in the human evaluation
WebNLG contains significantly more distinct types of attributes than E2E, so the chance of making errors or repetitions increases greatly. Nevertheless, our model still \emph{performs on-par on automatic metrics with superior information adequacy and output diversity}. The (+R) decoding constraint becomes important since the outputs in WebNLG are much longer than those in E2E, neural network models have problems tracking the history generation beyond certain range. Models might repeat facts that have been already generated long back before. The (+R) constraint effectively reduces the repetition cases from 19 to 2. These 2 cases are intra-segment repetitions and failed to be detected since our model can only track inter-segment constraints (examples are in \cref{sec: error}). The (+RM) constraint brings down the information missing cases to 5 with slightly more wrong and repeated facts compared with (+R). Forcing models to keep generating until coveraging all records will inevitably increase the risk of making errors.
\paragraph{Discussions}
In summary, our models generates \emph{most diverse outputs, achieves similar or better performances in word-overlap automatic metrics while significantly reduces the information hallucination, repetition and missing problems}. An example of hallucination is shown in Table~\ref{tab: attention}. The standard PG model ``hallucinated" the contents of ``low-priced", ``in the city center" and ``delivers take-away". The visualized attention maps reveal that it failed to attend properly when decoding the word ``low". The decoding is driven mostly by language models instead of the contents of input data. In contrast, as we explicitly align each segment to one slot, the attention distribution of our model is \emph{concentrated on one single slot rather than averaged over the whole input}, the chance of hallucinating is therefore largely reduced.

Figure~\ref{fig: webnlgexample} shows some example generations from WebNLG. Without adding the decoding constraints, PG and our model both suffer from the problem of information repetition and missing. However, the interpretability of our model enables us to easily avoid these issues by constraining the segment transition behavior. For the attention-based PG model, there exists no simple way of applying these constraints. We can also explicitly control the output structure similar to \citet{wiseman2018learning}, examples are shown in \cref{sec: cont}.
\section{Conclusion}
In this work, we exploit the segmental structure in data-to-text generation. The proposed model significantly alleviates the information hallucination, repetition and missing problems without sacrificing the fluency and diversity. It is end-to-end trainable, domain-independent and allows explicit control over the structure of generated text. As our model is interpretable in the correspondence between segments and input records, it can be easily combined with hand-engineered heuristics or user-specific requirements to further improve the performance.

\clearpage

\begin{subappendices}

\onecolumn

\section{Error Analysis}
\label{sec: error}
We analyze common errors below. 

\textbf{Missing:} Even with the coverage decoding constraint, the model can still occasionally miss information. 
We show one example in Table \ref{tab: miss}. The segments cover all input records, but the segment aligned to ``familyfriendly" only generates a period symbol. This happens 3 times on E2E and twice on WebNLG. On the other 3 cases of missing on WebNLG, some segments only generate one end-of-sentence symbol. Both conditions can be easily fixed by some simple filtering rules.

\textbf{Repeating:} There are still some repeating cases on the WebNLG dataset. Table~\ref{tab: repeat} shows one example. ``amsterdam-centrum is part of amsterdam" is repeated twice within a segment. As our constraint decoding can only prevent inter-segment repetition, it cannot fully avoid the repetition problem resulting from the intra-segment errors of RNNs.
\begin{table*}[!htbp]
\fontsize{200pt}{300pt}\selectfont
\resizebox{\textwidth}{!}{
\begin{tabular}{ll}
\hline
Input:   & \colorbox{red!20}{\strut name the phoenix} \colorbox{orange!30}{\strut eattype pub} \colorbox{yellow!60}{\strut food french} \colorbox{green!50}{\strut pricerange \pounds 20 - 25} \colorbox{gray!50}{\strut customerrating high} \colorbox{blue!20}{\strut area riverside}\\& \colorbox{red!60}{\strut familyfriendly yes} \colorbox{blue!40}{\strut near crowne plaza hotel}
  \\\hline
Output: & \colorbox{red!20}{\strut the phoenix} \colorbox{orange!30}{ \strut pub} \colorbox{blue!20}{\strut is located in riverside} \colorbox{blue!40}{\strut near crowne plaza hotel .} \colorbox{yellow!60}{\strut it serves french food} \colorbox{green!50}{\strut in the \pounds 20 -}\\& \colorbox{green!50}{\strut 25 price range} \colorbox{gray!50}{\strut . it has a high customer rating} \colorbox{red!60}{\strut .}\\
\hline      
\end{tabular}
}
\caption{\label{tab: miss}\small{Example of missing in E2E. The ``familyfriendly" is wligned to the period symbol.}}
\end{table*}

\begin{table*}[!htbp]
\fontsize{200pt}{300pt}\selectfont
\resizebox{\textwidth}{!}{
\begin{tabular}{|ll|}
\hline
Input:   & \colorbox{red!50}{\strut Amsterdam ground AFC Ajax (amateurs)} \colorbox{blue!40}{\strut Eberhard van der Laan leader Amsterdam} \colorbox{orange!30}{\strut Amsterdam-Centrum part Amsterdam}
  \\
Output: &  \colorbox{orange!30}{\strut amsterdam-centrum is part of amsterdam and amsterdam-centrum is part of amsterdam , the country where} \colorbox{blue!40}{\strut eberhard van}\\& \colorbox{blue!40}{\strut der laan is the leader and } \colorbox{red!50}{\strut the ground of afc ajax ( amateurs ) is located.} \\
\hline      
\end{tabular}
}
\caption{\label{tab: repeat}\small{(E2E) Example of repeatition in WebNLG. The phrase ``amsterdam-centrum is part of amsterdam" is repeated twice.}}
\end{table*}
    
\section{Controlling output structure}
\label{sec: cont}
As our model learns interpretable correspondence of each segment, it can control the output structures same as in \citet{wiseman2018learning}. Table~\ref{tab: control} shows example generations by sampling diverse segment structures.
\begin{table*}[!htbp]
\fontsize{200pt}{300pt}\selectfont
\resizebox{\textwidth}{!}{
\begin{tabular}{ll}
\hline
Input:   & \colorbox{red!20}{\strut name the phoenix} \colorbox{orange!30}{\strut eattype pub} \colorbox{yellow!60}{\strut food french} \colorbox{green!50}{\strut pricerange \pounds 20 - 25} \colorbox{gray!50}{\strut customerrating high} \colorbox{blue!20}{\strut area riverside}\\& \colorbox{red!60}{\strut familyfriendly yes} \colorbox{blue!40}{\strut near crowne plaza hotel}
  \\\hline
Output1: & \colorbox{red!20}{\strut the phoenix} \colorbox{orange!30}{ \strut pub} \colorbox{blue!20}{\strut is located in riverside} \colorbox{blue!40}{\strut near crowne plaza hotel .} \colorbox{yellow!60}{\strut it serves french food} \colorbox{green!50}{\strut in the \pounds 20 -}\\& \colorbox{green!50}{\strut 25 price range} \colorbox{gray!50}{\strut . it has a high customer rating} \colorbox{red!60}{\strut .}\\
Output2: & \colorbox{red!20}{\strut the phoenix} \colorbox{blue!20}{\strut is located in riverside} \colorbox{blue!40}{\strut near crowne plaza hotel .} \colorbox{red!60}{\strut it is a family - friendly} \colorbox{yellow!60}{\strut french} \colorbox{orange!30}{ \strut pub} \colorbox{green!50}{\strut with}\\& \colorbox{green!50}{\strut the price range of \pounds 20 - 25} \colorbox{gray!50}{\strut . it has a high customer rating .}\\
Output3: &  \colorbox{blue!20}{\strut located in riverside} \colorbox{blue!40}{\strut near crowne plaza hotel ,} \colorbox{red!20}{\strut the phoenix} \colorbox{yellow!60}{\strut is a french} \colorbox{orange!30}{ \strut pub} \colorbox{gray!50}{\strut with a high customer rating}\\& \colorbox{green!50}{\strut and a price range of \pounds 20 - 25.} \colorbox{red!60}{\strut It is family - friendly .}\\
\hline      
\end{tabular}
}
\caption{\label{tab: control}\small{Example of generations with diverse structures.}}
\end{table*}
\end{subappendices}
\cleardoublepage

\chapter[Diversify Dialogue with Non-Paired Text]{Diversify Dialogue with Non-Paired Text}
\label{chap: diversify}

\lettrine[lines=3]{T}he previous chapters all inject latent variables between the input and output to improve the diversity of interpretability. This chapter shows that latent-variable models are also a powerful tool when we have no parallel input-output pairs. We focus again on the task of open-domain dialogue generation where neural network-based sequence-to-sequence (seq2seq) models strongly suffer from the low-diversity problem. As bland and generic utterances usually dominate the frequency distribution in our daily chitchat, avoiding them to generate more interesting responses requires complex data filtering, sampling techniques or modifying the training objective. In this chapter, we propose a new perspective to diversify dialogue generation by leveraging \emph{non-conversational} text. Compared with bilateral conversations, non-conversational text are easier to obtain, more diverse and cover a much broader range of topics. We collect a large-scale non-conversational corpus from multi sources including forum comments, idioms and book snippets. We further present a training paradigm to effectively incorporate these text via iterative back translation. The resulting model is tested on two conversational datasets and is shown to produce significantly more diverse responses without sacrificing the relevance with context~\citep{su2018dialogue,su2020diversifying,mogadala2020integrating}. 

\section{Introduction}

Seq2seq models have achieved impressive success in a wide range of text generation tasks. In open-domain chitchat, however, people have found the model tends to strongly favor short, generic responses like ``I don't know" or ``OK"~\citep{vinyals2015neural,shen2017estimation,chang2021neural,chang2021selectgen}. The reason lies in the extreme one-to-many mapping relation between every context and its potential responses~\citep{zhao2017learning,su2018dialogue,zhao2019unsupervised,zhao2017gated,zhao2018comprehensive,del2021question}. Generic utterances, which can be in theory paired with most context, usually dominate the frequency distribution in the dialogue training corpus and thereby pushes the model to blindly produce these safe, dull responses ~\citep{su2019improving}

\begin{table}[!t]
    	 \small
    	\centering	
    		\scalebox{0.9}
    		{
    	\begin{tabular}{l|l}
    	  \multicolumn{2}{c}{\textbf{Conversational Text}}\\\hline 
    	\textbf{Context}& \begin{CJK*}{UTF8}{gbsn}暗恋的人却不喜欢我\end{CJK*} \\(Translation)& The one I have a crush on doesn't like me. \\ \hline
    	\multirow{2}{1.5cm}{\textbf{Response}} & \begin{CJK*}{UTF8}{gbsn} 摸摸头\end{CJK*} \\ & Head pat.\\ \hline

    	\multicolumn{2}{c}{\rule{0pt}{10pt}\textbf{Non-Conversational Text}}\\\hline
    	
    	\multirow{3}{1.5cm}{\textbf{Forum Comments}}& \begin{CJK*}{UTF8}{gbsn} 暗恋这碗酒，谁喝都会醉啊 \end{CJK*} \\& Crush is an alcoholic drink, whoever drinks\\& it will get intoxicated.\\ \hline
    	\multirow{2}{1.5cm}{\textbf{Idiom}} &  \begin{CJK*}{UTF8}{gbsn} 何必等待一个没有结果的等待\end{CJK*} \\& Why wait for a result without hope \\ \hline
    	\multirow{3}{1.5cm}{\textbf{Book Snippet}} &\begin{CJK*}{UTF8}{gbsn} 真诚的爱情之路永不会是平坦的\end{CJK*} \\& The course of true love never did run smooth \\ & (From \emph{A Midsummer Night's Dream})\\ \hline
    	\end{tabular}
}
    	\caption{\small A daily dialogue and non-conversational text from three sources. The contents of non-conversational text can be potentially utilized to enrich the response generation.}	\vspace{-5mm}
    	\label{tab:dialog}
    \end{table}

Current solutions can be roughly categorized into two classes: (1) Modify the seq2seq itself to bias toward diverse responses~\citep{li2015diversity,shen2019select}. However, the model is still trained on the \emph{limited dialogue corpus} which restricts its power at covering broad topics in open-domain chitchat. (2) Augment the training corpus with extra information like structured world knowledge, personality or emotions~\citep{li2016persona,dinan2018wizard,chang2021training}, which requires \emph{costly human annotation}.

In this work, we argue that training only based on conversational corpus can greatly constrain the usability of an open-domain chatbot system since many topics are not easily available in the dialogue format. With this in mind, we explore a cheap way to diversify dialogue generation by utilizing large amounts of \emph{non-conversational text}. Compared with bilateral conversations, non-conversational text covers a much broader range of topics, and can be easily obtained without further human annotation from multiple sources like forum comments, idioms and book snippets. More importantly, non-conversational text are usually \emph{more interesting and contentful} as they are written to convey some specific personal opinions or introduce a new topic, unlike in daily conversations where people often \emph{passively} reply to the last utterance. As can be seen in Table~\ref{tab:dialog}, the response from the daily conversation is a simple comfort of ``Head pat". Non-conversational text, on the contrary, exhibit diverse styles ranging from casual wording to poetic statements, which we believe can be potentially utilized to enrich the response generation.

To do so, we collect a large-scale corpus containing over 1M non-conversational utterances from multiple sources. To effectively integrate these utterances, we borrow the back translation idea from unsupervised neural machine translation~\citep{sennrich2016improving,lample2018phrase} and treat the collected utterances as unpaired responses. We first pre-train the forward and backward transduction model on the parallel conversational corpus. The forward and backward model are then iteratively tuned to find the optimal mapping relation between conversational context and non-conversational utterances~\citep{cotterell2018explaining}. By this means, the content of non-conversational utterances is gradually distilled into the dialogue generation model~\citep{kim2016sequence}, enlarging the space of generated responses to cover not only the original dialogue corpus, but also the wide topics reflected in the non-conversational utterances.

We test our model on two popular Chinese conversational datasets weibo~\citep{shang2015neural} and douban~\citep{wu2017sequential,xu2020data}. We compare our model against retrieval-based systems, style-transfer methods and several seq2seq variants which also target the diversity of dialogue generation. Automatic and human evaluation show that our model significantly improves the responses' diversity both semantically and syntactically without sacrificing the relevance with context, and is considered as most favorable judged by human evaluators~\footnote{Code and dataset available at \url{https://github.com/chin-gyou/Div-Non-Conv}}.
\section{Related Work}
The tendency to produce generic responses has been a long-standing problem in seq2seq-based open-domain dialogue generation~\citep{vinyals2015neural,li2015diversity}. Previous approaches to alleviate this issue can be grouped into two classes. 

The first class resorts to modifying the seq2seq architecture itself. For example, \citet{shen2018nexus,zhang2018generating,del2021question} changes the training objective to mutual information maximization and rely on continuous approximations or policy gradient to circumvent the non-differentiable issue for text. \citet{li2016deep,serban2017deep} treat open-domain chitchat as a reinforcement learning problem and manually define some rewards to encourage long-term conversations. There is also research that utilizes latent variable sampling~\citep{serban2016hierarchical,shen2018improving,shen2019improving}, adversarial learning~\citep{li2017adversarial,su2018dialogue}, replaces the beam search decoding with a more diverse sampling strategy~\citep{li2016simple,holtzman2020the} or applies reranking to filter generic responses~\citep{li2015diversity,wang2017steering}. All of the above are still trained on the original dialogue corpus and thereby cannot generate out-of-scope topics.

The second class seeks to bring in extra information into existing corpus like structured knowledge~\citep{zhao2018comprehensive,ghazvininejad2018knowledge,dinan2018wizard}, personal information~\citep{li2016persona,zhang2018personalizing} or emotions~\citep{shen2017conditional,zhou2018emotional}. However, corpus with such annotations can be extremely costly to obtain and is usually limited to a specific domain with small data size. Some recent research started to do dialogue style transfer based on personal speeches or TV scripts~\citep{niu2018polite,gao2019structuring,su2019personalized}. Our motivation differs from them in that we aim at enriching general dialogue generation with abundant non-conversational text instead of being constrained on one specific type of style.

Back translation is widely used in unsupervised machine translation~\citep{sennrich2016improving,lample2017unsupervised,artetxe2017unsupervised} and has been recently extended to similar areas like style transfer~\citep{subramanian2018multiple}, summarization~\citep{zhao2019unsupervised} and data-to-text~\citep{chang2020unsupervised}. To the best of our knowledge, it has never been applied to dialogue generation yet. Our work treats the context and non-conversational text as unpaired source-target data. The back-translation idea is naturally adopted to learn the mapping between them. The contents of non-conversational text can then be effectively utilized to enrich the dialogue generation.
\section{Dataset}
\begin{table}
    \centering
    \begin{adjustbox}{max width=\linewidth}
    \begin{tabular}{l c c }
    \toprule
    Resources & Size & Avg. length \\
    \midrule
    Comments & 781,847& 21.0 \\
    Idioms & 51,948 &  18.7  \\
    Book Snippets & 206,340 & 26.9  \\
    \bottomrule
    \end{tabular}
    \end{adjustbox}
 \caption{\label{tab:mono_data}\small Statistics of Non-Conversational Text.}
\end{table}
We would like to collect non-conversational utterances that stay close with daily-life topics and can be potentially used to augment the response space. The utterance should be neither too long nor too short, similar with our daily chitchats. Therefore, we collect data from the following three sources:
\begin{enumerate}
    \item Forum comments. We collect comments from zhihu~\footnote{\url{https://www.zhihu.com}}, a popular Chinese forums. Selected comments are restricted to have more than 10 likes and less than 30 words~\footnote{The posts are usually very long, describing a specific social phenomenon or news event, so building parallel conversational corpus from post-comment pairs is difficult. Nonetheless, these high-liked comments are normally high-quality themselves and can be used to augment the response space.}.
    \item Idioms. We crawl idioms, famous quotes, proverbs and locutions from several websites. These phrases are normally highly-refined and graceful, which we believe might provide a useful augmentation for responses.
    \item Book Snippets. We select top 1,000 favorite novels or prose from wechat read~\footnote{\url{https://weread.qq.com/}}. Snippets highlighted by readers, which are usually quintessential passages, and with the word length range 10-30 are kept.
\end{enumerate}
We further filter out sentences with offensive or discriminative languages by phrase matching against a large blocklist. The resulting corpus contains over 1M utterances. The statistics from each source are listed in Table~\ref{tab:mono_data}.
\section{Approach}
\label{sec: approach}
\begin{figure*}[!ht]
\centering
\centerline{\includegraphics[width=\columnwidth]{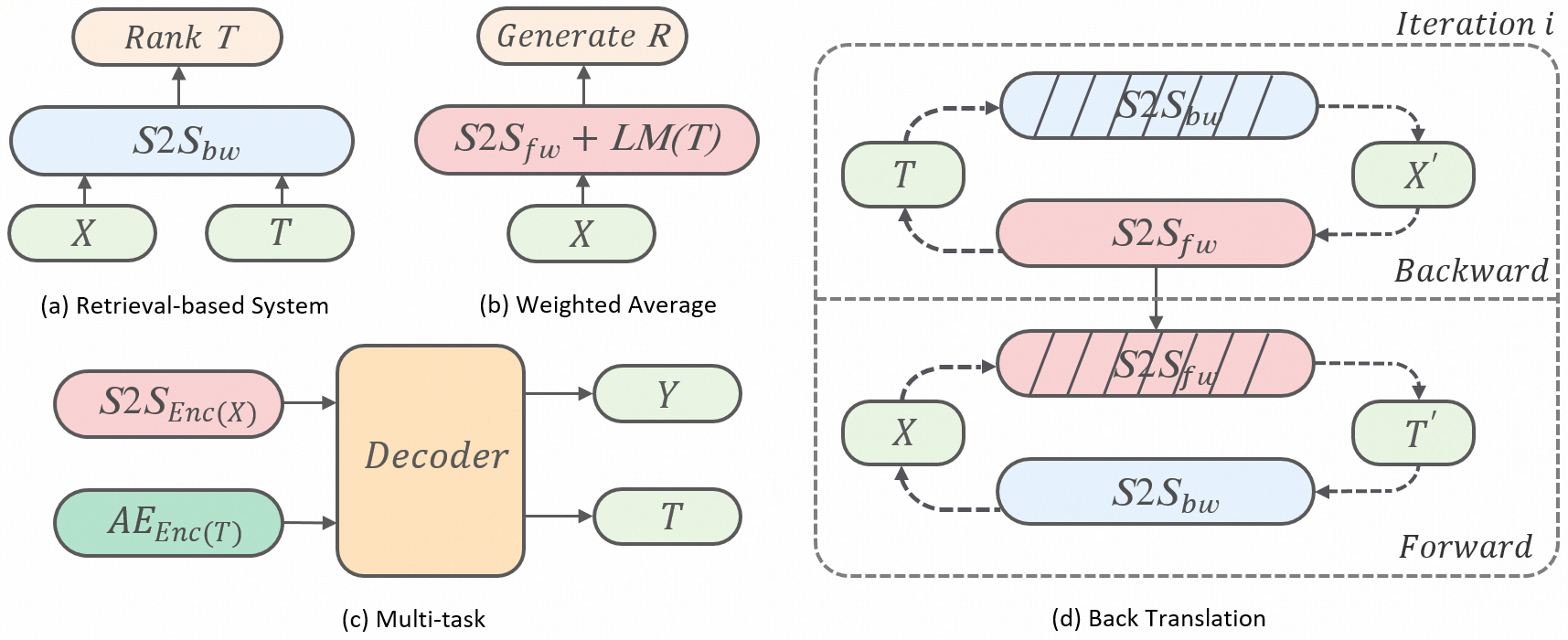}}
\caption{\small Comparison of four approaches leveraging the non-conversational text. $S2S_{fw}$, $S2S_{bw}$ and $LM$ indicate the forward, backward seq2seq and language model respectively. (d) visualizes the process of one iteration for the back translation approach. Striped component are not updated in each iteration.}
\label{fig: diversify_model}
\end{figure*}
Let $\mathcal{D}=\{(X_1, Y_1),(X_2, Y_2),\ldots,(X_N,Y_N)\}$ denote the parallel conversational corpus. $X_i$ is the context and $Y_i$ is the corresponding response. $\mathcal{D}_T=\{T_1, T_2,\ldots,T_M\}$ denotes our collected corpus where $T_i$ is a non-conversational utterance. As the standard seq2seq model trained only on $\mathcal{D}$ tends to generate over-generic responses, our purpose is to diversify the generated responses by leveraging the non-conversational corpus $\mathcal{D}_T$, which are semantically and syntactically much richer than responses contained in $\mathcal{D}$. In the following section, we first go through several baseline systems, then introduce our proposed method based on back translation.
\subsection{Retrieval-based System}
\label{sec: ret}
The first approach we consider is a retrieval-based system that considers all sentences contained in $\mathcal{D}_T$ as candidate responses. As the proportion of generic utterances in $\mathcal{D}_T$ is much lower than that in $\mathcal{D}$, the diversity will be largely improved. Standard retrieval algorithms based on context-matching~\citep{wu2017sequential,bartl2017retrieval} fail to apply here since non-conversational text does not come with its corresponding context. Therefore, we train a backward seq2seq model on the parallel conversational corpus $\mathcal{D}$ to maximize $p(X_i|Y_i)$. The score assigned by the backward model, which can be seen as an estimation of the point-wise mutual information, is used to rank the responses~\citep{li2015diversity}~\footnote{The backward seq2seq model measures the context relevance better than forward models since the latter highly biases generic utterances~\citep{li2015diversity,zhang2018generating}}.

The major limitation of the retrieval-based system is that it can only produce responses from a finite set of candidates. The model can work well only if an appropriate response already exists in the candidate bank. Nonetheless, due to the large size of the non-conversational corpus, this approach is a very strong baseline.
\subsection{Weighted Average}
\label{sec: weight}
The second approach is to take a weighted average score of a seq2seq model trained on $\mathcal{D}$ and a language model trained on $\mathbf{D}_T$ when decoding responses. The idea has been widely utilized on domain adaptation for text generation tasks~\citep{koehn2007experiments,wang2017steering,niu2018polite}. In our scenario, basically we hope the generated responses could share the diverse topics and styles of the non-conversational text, yet stay relevant with the dialogue context. The seq2seq model $S2S$ is trained on $\mathcal{D}$ as an indicator of how relevant each response is with the context. A language model $\mathcal{L}$ is trained on $\mathcal{D}_T$ to measure how the response matches the domain of $\mathcal{D}_T$. The decoding probability for generating word $w$ at time step $t$ is assigned by:
\begin{equation}
\label{eq: weight}
    p_t(w) = \alpha S2S_t(w) + (1-\alpha) L_t(w)
\end{equation}
where $\alpha$ is a hyperparameter to adjust the balance between the two. Setting $\alpha=1$ will make it degenerate into the standard seq2seq model while $\alpha=0$ will totally ignore the dialoge context.
\subsection{Multi-task}
\label{sec: multi}
The third approach is based on multi-task learning. A seq2seq model is trained on the parallel conversational corpus $\mathcal{D}$ while an autoencoder model is trained on the non-parallel monologue data $\mathcal{D}_T$. Both models share the decoder parameters to facilitate each other. The idea was first experimented on machine translation in order to leverage large amounts of target-side monolingual text~\citep{luong2015multi,sennrich2016improving}. \citet{luan2017multi} extended it to conversational models for speaker-role adaptation. The intuition is that by tying the decoder parameters, the seq2seq and autoencoder model can learn a shared latent space between the dialogue corpus and non-conversational text. When decoding, the model can generate responses with features from both sides.

\subsection{Back Translation}
\label{sec: bt}
Finally, we consider the back translation technique commonly used for unsupervised machine translation~\citep{artetxe2017unsupervised,lample2017unsupervised}. The basic idea is to first \emph{initialize} the model properly to provide a good starting point, then iteratively perform \emph{backward} and \emph{forward} translation to learn the correspondence between context and unpaired non-conversational utterances.
\paragraph{Initialization}
Unlike unsupervised machine translation, the source and target side in our case come from the same language, and we already have a parallel conversational corpus $\mathcal{D}$, so we can get rid of the careful embedding alignment and autoencoding steps as in \citet{lample2018phrase}. For the initialization, we simply train a forward and backward seq2seq model on $\mathcal{D}$. The loss function is:
\begin{equation}
\label{eq: init}
    \mathbb{E}_{X_i, Y_i \sim \mathcal{D}} -\log P_{f}(Y_i|X_i) - \log P_{b}(X_i|Y_i)
\end{equation}
where $P_f$ and $P_b$ are the decoding likelihood defined by the forward and backward seq2seq model respectively. We optimize Eq.~\ref{eq: init} until convergence. Afterwards, the forward and backward seq2seq can learn the backbone mapping relation between a context and its response in a conversational structure.
\paragraph{Backward}
After the initialization, we use the backward seq2seq to create pseudo parallel training examples from the non-conversational text $\mathcal{D}_T$. The forward seq2seq is then trained on the pseudo pairs. The objective is to minimize:
\begin{equation}
\label{eq: backward}
\begin{split}
    \mathbb{E}_{T_i \sim \mathcal{D}_T}& -\log P_{f}(T_i|b(T_i))\\
    b(T_i)&= \argmax_{u} P_b(u|T_i)
\end{split}
\end{equation}
where we approximate the $\argmax$ function by using a beam search decoder to decode from the backward model $P_b(u|T_i)$. Because of the non-differentiability of the $\argmax$ operator, the gradient is only passed through $P_f$ but not $P_b$~\footnote{As also noted in \citet{lample2018phrase}, backpropagating further through $P_b$ brings no improvement.}.

As $P_b$ is already well initialized by training on the parallel corpus $\mathcal{D}$, the back-translated pseudo pair $\{b(T_i), T_i\}$ can roughly follow the typical human conversational patterns. Training $P_f$ on top of them will encourage the forward decoder to generate utterances in the domain of $T_i$ while maintaining coherent as a conversation.
\paragraph{Forward}
The forward translation follows a similar step as back translation. The forward seq2seq $P_f$ translates context into a response, which in return form a pseudo pair to train the backward model $P_b$. The objective is to minimize:
\begin{equation}
\label{eq: diversify_forward}
\begin{split}
    \mathbb{E}_{X_i \sim \mathcal{D}}& -\log P_{b}(X_i|f(X_i))\\
    f(X_i)&= \argmax_{v} P_f(v|X_i)
\end{split}
\end{equation}
where the $\argmax$ function is again approximated with a beam search decoder and the gradient is only backpropagated through $P_b$. Though $X_i$ has its corresponding $Y_i$ in $\mathcal{D}$, we drop $Y_i$ and instead train on forward translated pseudo pairs $\{X_i,f(X_i)\}$. As $P_f$ is trained by leveraging data from $\mathcal{D}_T$, $f(X_i)$ can have superior diversity compared with $Y_i$.

The encoder parameters are shared between the forward and backward models while decoders are separate. The backward and forward translation are iteratively performed to close the gap between $P_f$ and $P_b$~\citep{hoang2018iterative,cotterell2018explaining}. The effects of non-conversational text are strengthened after each iteration. Eventually, the forward model will be able to produce diverse responses covering the wide topics in $\mathcal{D}_T$. Algorithm~\ref{algorithm} depicts the training process.

\begin{algorithm}[tb]
\caption{Model Training Process}
\label{algorithm}
\begin{algorithmic}
\STATE {\bfseries Inilialization:}
Train by minimizing Eq.~\ref{eq: init} until convergence;
\REPEAT
\STATE {\bfseries Backward:} Train by minimizing Eq.~\ref{eq: backward} until convergence\;
\STATE {\bfseries Forward:} Train by minimizing Eq.~\ref{eq: diversify_forward} until convergence\;
\UNTIL{Model converges}
\end{algorithmic}
\end{algorithm}

\section{Experiments}
\subsection{Datasets}
We conduct our experiments on two Chinese dialogue corpus Weibo~\citep{shang2015neural} and Douban~\citep{wu2017sequential}. Weibo~\footnote{\url{http://www.weibo.com/}} is a popular Twitter-like microblogging service in China, on which a user can post short messages, and other users make comment on a published post. The post-comment pairs are crawled as short-text conversations. Each utterance has 15.4 words on average and the data is split into train/valid/test subsets with 4M/40k/10k utterance pairs. Douban~\footnote{\url{https://www.douban.com/group}} is a Chinese social network service where people can chat about different topics online. The original data contains 1.1M multi-turn conversations. We split them into two-turn context-response pairs, resulting in 10M train, 500k valid and 100K test samples.
\subsection{General Setup}
For all models, we use a two-layer LSTM~\citep{hochreiter1997long} encoder/decoder structure with hidden size 500 and word embedding size 300. Models are trained with Adam optimizer~\citep{kingma2014adam} with an initial learning rate of 0.15. We set the batch size as 256 and use the gradients clipping of 5. We build out vocabulary with character-based segmentation for Chinese. For non-Chinese tokens, we simply split by space and keep all unique tokens that appear at least 5 times. Utterances are cut down to at most 50 tokens and fed to every batch. We implement our models based on the OpenNMT toolkit~\citep{klein2017opennmt} and other hyperparameters are set as the default values.
\subsection{Compared Models}
We compare our model with the standard seq2seq and four popular variants which were proposed to improve the diversity of generated utterances. All of them are trained only on the parallel conversational corpus:
\paragraph{Standard} The standard seq2seq with beam search decoding (size 5).
\paragraph{MMI} The maximum mutual information decoding which reranks the decoded responses with a backward seq2seq model~\citep{li2015diversity}. The hyperparameter $\lambda$ is set to 0.5 as suggested. 200 candidates
per context are sampled for re-ranking
\paragraph{Diverse Sampling} The diverse beam search strategy proposed in \citet{vijayakumar2016diverse} which explicitly controls for the exploration and
exploitation of the search space. We set the number of groups as 5, $\lambda=0.3$ and use the Hamming diversity as the penalty function as in the paper.
\paragraph{Nucleus Sampling} Proposed in \citet{holtzman2020the}, it allows for diverse sequence generations. Instead of decoding with a fixed beam size, it samples text from the dynamic nucleus. We use the default configuration and set $p=0.9$.
\paragraph{CVAE} The conditional variational autoencoder~\citep{serban2016hierarchical,zhao2017learning} which injects diversity by imposing stochastical latent variables. We use a latent variable with dimension 100 and utilize the KL-annealing strategy with step 350k and a word drop-out rate of 0.3 to alleviate the posterior collapse problem~\citep{bowman2016generating}.

Furthermore, we compare the 4 approaches mentioned in \cref{sec: approach} which incorporate the collected non-conversational text:
\paragraph{Retrieval-based}(\cref{sec: ret}) Due to the large size of the non-conversational corpus, exact ranking is extremely slow. Therefore, we first retrieve top 200 matched text with elastic search based on the similarity of Bert embeddings~\citep{devlin2019bert}. Specifically, we pass sentences through Bert and derive a fixed-sized vector by averaging the outputs from the second-to-last layer~\citep{may2019measuring}~\footnote{\url{https://github.com/hanxiao/bert-as-service}}. The 200 candidates are then ranked with the backward score~\footnote{This makes it similar to MMI reranking, whose 200 candidates are from seq2seq decodings instead of top-matched non-conversational utterances.}.
\paragraph{Weighted Average}(\cref{sec: weight}) We set $\lambda=0.5$ in eq.~\ref{eq: weight}, which considers context relevance and diversity with equal weights.
\paragraph{Multi-task}((\cref{sec: multi})) We concatenate each context-response pair with a non-conversational utterance and train with a mixed objective of seq2seq and autoencoding (by sharing the decoder).
\paragraph{Back Translation}(\cref{sec: bt}) We perform the iterative backward and forward translation 4 times for both datasets. We observe the forward cross entropy loss converges after 4 iterations.

\section{Results}
As for the experiment results, we report the automatic and human evaluation in \cref{sec: auto} and \cref{sec: human} respectively. Detailed analysis are shown in \cref{sec: analysis} to elaborate the differences among model performances and some case studies.

\subsection{Automatic Evaluation}
\label{sec: auto}
Evaluating dialogue generation is extremely difficult. Metrics which measure the word-level overlap like BLEU~\citep{papineni2002bleu} have been widely used for dialogue evaluation. However, these metrics do not fit into our setting well as we would like to diversify the response generation with an external corpus, the generations will inevitably differ greatly from the ground-truth references in the original conversational corpus. Though we report the BLEU score anyway and list all the results in Table~\ref{tab: auto}, it is worth mentioning that the BLEU score itself is by no means a reliable metric to measure the quality of dialogue generations.
\begin{table*}[t!]
  \small
  \begin{center}
    \resizebox{1\textwidth}{!}{
    \begin{tabular}{l|ccccc|ccccc}
    \hline 
     \textbf{Metrics} & 
     \multicolumn{5}{c|}{\textbf{Weibo}} & \multicolumn{5}{c}{\textbf{Douban}}  \\ 
     \textbf{Model}& \textbf{BLEU-2} &  \textbf{Dist-1} & \textbf{Dist-2} &   \textbf{Ent-4} & \textbf{Adver}&\textbf{BLEU-2} &  \textbf{Dist-1} & \textbf{Dist-2}  & \textbf{Ent-4}  & \textbf{Adver} \\
     \hline
     \textbf{\textsc{Standard}} & 0.0165  & 0.018 & 0.050  & 5.04 & 0.30 & 0.0285 & 0.071 & 0.206  & 7.55 & 0.19 \\
     \textbf{\textsc{MMI}} & 0.0161 & 0.025 & 0.069   & 5.98  & 0.42 & 0.0263 & 0.143 & 0.363  & 7.60  & \textbf{0.31}  \\
     \textbf{\textsc{Diverse}}  & 0.0175 & 0.019  & 0.054   & 6.20  & 0.38 & 0.0298 & 0.130 & 0.358  & 7.51   & 0.25  \\
     \textbf{\textsc{Nucleus}} & \textbf{0.0183} & \textbf{0.027}  & \textbf{0.074} & \textbf{7.41}  & \textbf{0.43} & \textbf{0.0312}  & 0.141 &  0.402  & \textbf{7.93} & 0.30  \\
     \textbf{\textsc{CVAE}} & 0.0171 & 0.023  & 0.061  & 6.63  & 0.36  & 0.0287 & \textbf{0.169} &  \textbf{0.496}  & 7.80  & 0.29 \\
    \hline
    \textbf{\textsc{Retrieval}}  & 0.0142 & \textbf{0.198} & \textbf{0.492}   & \textbf{12.5}  & 0.13 &\textbf{ 0.0276} & \textbf{0.203} & \textbf{0.510} & \textbf{13.3}  & \textbf{0.17} \\
     \textbf{\textsc{Weighted}} &\textbf{ 0.0152}  & 0.091  & 0.316   & 9.26  & 0.22 & 0.0188 & 0.172 & 0.407  & 8.73   & 0.14 \\
     \textbf{\textsc{Multi}}  & 0.0142 & 0.128  & 0.348  & 8.98  & \textbf{0.27}  & 0.0110 & 0.190 & 0.389  & 8.26& 0.16 \\    \hline
      \textbf{\textsc{BT (Iter=1)}}  & \textbf{0.0180} & 0.046  & 0.171  & 7.64  & 0.19  & \textbf{0.0274} &  0.106 & 0.313  & 8.16   & 0.15 \\
     \textbf{\textsc{BT (Iter=4)}} & 0.0176  & \textbf{0.175}   & \textbf{0.487}  & \textbf{11.2} & \textbf{0.35}  & 0.0269 &  \textbf{0.207} & \textbf{0.502}  & \textbf{11.0}   & \textbf{0.25} \\
     \hline
      \textbf{\textsc{Human}} & - & 0.171    & 0.452 & 9.23  & 0.88  & - & 0.209 & 0.514  &11.3   & 0.85 \\\hline
    \end{tabular}
    }
  \end{center}
   \caption{\small Automatic evaluation on Weibo and Douban datasets. Upper areas are models trained only on the conversational corpus. Middle areas are baseline models incorporating the non-conversational corpus. Bottom areas are our model with different number of iterations. Best results in every area are \textbf{bolded}.}
     \label{tab: auto}%
     \vspace{-0.3cm}
\end{table*}%
\paragraph{Diversity}
Diversity is a major concern for dialogue generation.  Same as in \citep{li2015diversity}, we measure the diversity by the ratio of distinct unigrams (\textbf{Dist-1}) and bigrams (\textbf{Dist-2}) in all generated responses. As the ratio itself ignores the frequency distribution of n-grams, we further calculate the entropy value for the empirical distribution of n-grams~\citep{zhang2018generating}. A larger entropy indicates more diverse distributions. We report the entropy of four-grams (\textbf{Ent-4}) in Table~\ref{tab: auto}. Among models trained only on the conversational corpus, the standard seq2seq performed worst as expected. All different variants improved the diversity more or less. Nucleus sampling and CVAE generated most diverse responses, especially Nucleus who wins on 6 out of the 8 metrics. By incorporating the non-conversational corpus, the diversity of generated responses improves dramatically. The retrieval-based system and our model perform best, in most cases even better than human references. This can happen as we enrich the response generation with external resources. The diversity would be more than the original conversational corpus. Weighted-average and multi-task models are relatively worse, though still greatly outperforming models trained only on the conversational corpus. We can also observe that our model improves over standard seq2seq only a bit after one iteration. As more iterations are added, the diversity improves gradually.
\paragraph{Relevance}
Measuring the context-response relevance automatically is tricky in our case. The typical way of using scores from forward or backward models as in \citet{li2016neural} is not suitable as our model borrowed information from extra resources. The generated responses are out-of-scope for the seq2seq model trained on only on the conversational corpus and thus would be assigned very low scores. Apart from the BLEU-2 score, we further evaluate the relevance by leveraging an adversarial discriminator~\citep{li2017adversarial}. As has been shown in previous research, discriminative models are generally less biased to high-frequent utterances and more robust against their generative counterparts~\citep{lu2017best,luo2018discriminability}. The discriminator is trained on the parallel conversational corpus distinguish correct responses from randomly sampled ones. We encode the context and response separately with two different LSTM neural networks and output a binary signal indicating relevant or not~\footnote{In our experiment, the discriminator performs reasonably well in the 4 scenarios outlined in \citet{li2017adversarial} and thus can be considered as a fair evaluation metric.}. The relevance score is defined as the success rate that the model fools the adversarial classifier into believing its generations (\textbf{Adver} in Table~\ref{tab: auto}). The retrieval-based model, who generates the most diverse generations, achieve the lowest score as for relevance with context. The restriction that it can only select from a set of fixed utterances do affect the relevance a lot~\footnote{The fact that we only rank on 200 most similar utterances might also affect. We tried increasing the size to 1,000 but observe no tangible improvement. The candidate size required for a decent relevance score can be unbearably large.}. Note that \emph{the discriminator is also trained on the same bilateral conversational corpus, putting our model into a naturally disadvantageous place due to the incorporation of out-of-scope non-conversational text.} Nonetheless, our model still achieves competitive relevance score even compared with models trained only on the conversational corpus. This suggests our model does learn the proper patterns in human conversations instead of randomly synthesizing diverse generations.
  \begin{table}[!hbtp] \addtolength{\tabcolsep}{-2pt}  
      \footnotesize
      \centering
      \begin{tabular}{l|ccc|ccc}
          \hline
             \textbf{Metrics} & 
     \multicolumn{3}{c|}{\textbf{Weibo}} & \multicolumn{3}{c}{\textbf{Douban}}  \\ 
          \textbf{Model} & \textbf{Rel} &\textbf{Inter}& \textbf{Flu} & \textbf{Rel} &\textbf{Inter}& \textbf{Flu} \\ \hline 
          STANDARD & 0.32 & 0.11 & 0.76 & 0.26 & 0.13 & 0.82 \\
          NUCLEUS & \textbf{0.46} & 0.19 & \textbf{0.78} & 0.38 & 0.21 & \textbf{0.83} \\ \hline 
          RETRIEVAL & 0.12 & 0.35 & - & 0.09 & 0.32 & - \\ 
          WEIGHTED & 0.19& 0.14& 0.52 & 0.15 & 0.17& 0.46 \\ 
          MULTI & 0.25& 0.21& 0.70 & 0.22 & 0.23& 0.66 \\ 
          BT (ITER=4) & 0.43 & \textbf{0.37} & 0.77 & \textbf{0.39} & \textbf{0.48} & 0.80 \\ \hline
      \end{tabular}
     \caption{\label{tab:human}\small Human Evaluation Results}
  \end{table}
\begin{table*}[t]
\centering
\begin{tabular}{l|l}
\hline
Context               & \makecell[l]{\begin{CJK*}{UTF8}{gbsn}一 直 单 身 怎 么 办\end{CJK*}\\(Being always single, what should I do?)}\\
\hline
Response                    & \makecell[l]{ {\begin{CJK*}{UTF8}{gbsn}勇 敢 一 点 多 去 加 好 友 啊\end{CJK*} }\\(Be brave and add more people to friends.) }\\ 
\hline
\multirow{5}{*}{Generation} & \makecell[l]{{[}Iteration 0{]}: \begin{CJK*}{UTF8}{gbsn}不知道该怎么办 \end{CJK*}\\ (I don't know what to do.) } \\ \cline{2-2} 
&\makecell[l]{{[}Iteration 1{]}: \begin{CJK*}{UTF8}{gbsn}单 身 不 可 怕 ， 单 身 不 可 怕 \end{CJK*} \\ (Being single is nothing, being single is nothing.)} \\ \cline{2-2} 
& \makecell[l]{{[}Iteration 4{]}:
\begin{CJK*}{UTF8}{gbsn}斯 人 若 彩 虹 ，遇 上 方 知 有\end{CJK*}\\ (Every once in a while you find someone who's iridescent,\\ and when you do, nothing will ever compare.)} \\ 
\hline
\end{tabular}

\caption{\small Example of response generation in different iterations.}
\label{table:iter}
\end{table*}
\subsection{Human Evaluation}
\label{sec: human}
Apart from automatic evaluations, we also employed crowdsourced judges to evaluate the quality of generations for 500 contexts of each dataset. We focus on evaluating the generated responses regarding the (1) relevance: if they coincide with the context (\textbf{Rel}), (2) interestingness: if they are interesting for people to continue the conversation (\textbf{Inter}) and (3) fluency: whether they are fluent by grammar (\textbf{Flu})~\footnote{We do not evaluate the retrieval-based model for the fluency score as the retrieved utterances are fluent by construct.}. Each sample gets one point if judged as yes and zero otherwise. Each pair is judged by three participants and the score supported by most people is adopted. The averaged scores are summarized in Table~\ref{tab:human}. We compare the standard seq2seq model, nucleus sampling which performs best among all seq2seq variants, and the four approaches leveraging the non-conversational text. All models perform decently well as for fluency except the weighted average one. The scores for diversity and relevance generally correlate well with the automatic evaluations. Overall the back-translation model are competitive with respect to fluency and relevance, while generating much more interesting responses to human evaluators. It also significantly outperforms the other three baseline approaches in its capability to properly make use of the non-conversational corpus.

\subsection{Analysis}
\label{sec: analysis}

\pgfplotsset{
axis background/.style={fill=gallery},
grid=both,
  xtick pos=left,
  ytick pos=left,
  tick style={
    major grid style={style=white,line width=1pt},
    minor grid style=bgc,
    draw=none
    },
  minor tick num=1,
  ymajorgrids,
	major grid style={draw=white},
	y axis line style={opacity=0},
	tickwidth=0pt,
}

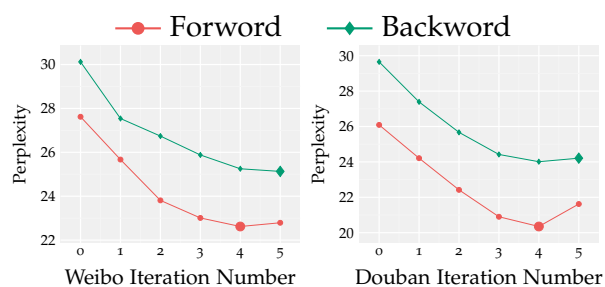
\begin{figure}[ht]
\centering
    \begin{tikzpicture}[scale=0.46]
	\begin{groupplot}[
	    group style={group size=2 by 1,
	        horizontal sep = 48pt}, 
	    xlabel=\Large  Weibo Iteration Number,
        ylabel=\large Perplexity,
        xticklabels={0,1,2,3,4,5},
        xtick={1,2,3,4,5,6},
        ymajorgrids,
        major grid style={draw=white},
        y axis line style={opacity=0},
        tickwidth=0pt,
        yticklabel style={
        /pgf/number format/fixed,
        /pgf/number format/precision=5
        },
        scaled y ticks=false,
        every axis title/.append style={at={(0.1,0.8)},font=\bfseries}
	    ]
		\nextgroupplot[
		legend style = {
		  font=\small,
          draw=none, 
          fill=none,
          column sep = 1pt, 
          /tikz/every even column/.append style={column sep=5mm},
          legend columns = -1, 
          legend to name = grouplegend},
		]

		\addplot[thick,color=flamingo,mark=*] coordinates {
          (1, 27.62)
          (2, 25.67 )
          (3, 23.81)
          (4, 23.01)
          (5, 22.62)
          (6, 22.79)
        }; \addlegendentry{\textcolor{black}{Forword}}
         \addplot[thick,color=free_speech_aquamarine,mark=diamond*] coordinates {
          (1, 30.12)
          (2, 27.54)
          (3, 26.74)
          (4, 25.88)
          (5, 25.25)
          (6, 25.13)
        }; \addlegendentry{\textcolor{black}{Backword}}

        \addplot[mark size=3.6pt, color=flamingo, mark=*,
        ] coordinates {
          (5, 22.62)
        }; 
         \addplot[mark size=4.2pt, color=free_speech_aquamarine, mark=diamond*] coordinates {
          (6, 25.13)
        };

        \nextgroupplot[
	    xlabel=\Large  Douban Iteration Number]
		\addplot[thick,color=flamingo,mark=*] coordinates {
          (1, 26.09)
          (2, 24.21)
          (3, 22.42)
          (4, 20.90)
          (5, 20.35 )
          (6, 21.62 )
        };
         \addplot[thick,color=free_speech_aquamarine,mark=diamond*] coordinates {
          (1, 29.65)
          (2, 27.39)
          (3, 25.67)
          (4, 24.42)
          (5, 24.01)
          (6, 24.21)

        }; 

        \addplot[mark size=3.6pt, color=flamingo, mark=*] coordinates {
          (5, 20.35 )
        }; 
         \addplot[mark size=4.2pt, color=free_speech_aquamarine, mark=diamond*] coordinates {
          (6, 24.21)
        };

	\end{groupplot}
\node at ($(group c1r1) + (110pt, 95pt)$) {\ref{grouplegend}};
\end{tikzpicture}

    \caption{Change of validation loss across iterations.}
    \label{fig: iter}
\end{figure}

\paragraph{Effect of Iterative Training}
To show the importance of the iterative training paradigm, we visualize the change of the validation loss in Figure~\ref{fig: iter}~\footnote{Iteration 0 means before the iteration starts but after the initialization stage, equal to a standard seq2seq.}. The forward validation loss is computed as the perplexity of the forward seq2seq on the pseudo context-response pairs obtained from the backward model, vice versa for backward loss. It approximately quantifies the KL divergence between them two~\citep{kim2016sequence,cotterell2018explaining}. As the iteration goes, the knowledge from the backward model is gradually distilled into the forward model. The divergence between them reaches the lowest point at iteration 4, where we stop our model. Table~\ref{table:iter} further displays examples for different iterations. Iteration 0 generates mostly generic responses. Iteration 1 starts to become more diverse but still struggle with fluency and relevance. In the final iteration, it can learn to incorporate novel topics from the non-conversational text yet maintaining the relevance with context.
\begin{table}[!hbtp] \addtolength{\tabcolsep}{-2pt}  
\centering
\begin{tabularx}{1\linewidth}{l|l}
\hline

\multirow{2}{*}{CXT}                  & \makecell[l]{\begin{CJK*}{UTF8}{gbsn}最近又长胖了\end{CJK*}}\\
&Fleshing out again recently.\\
\hline
\multirow{2}{*}{NS}                    & \makecell[l]{ {\begin{CJK*}{UTF8}{gbsn}我也是这样的\end{CJK*} }}\\
& Me too. \\ 
\hline
\multirow{2}{*}{BT}  &  \makecell[l]{\begin{CJK*}{UTF8}{gbsn}哈哈哈莫非已经\error{胖若两人}了\end{CJK*}}\\
& hahaha already \error{as fat as two people}? \\
\Xhline{1pt}

\multirow{2}{*}{CXT}                  & \makecell[l]{\begin{CJK*}{UTF8}{gbsn}爱一个人真的不能跟她表白吗？\end{CJK*}}\\
&Why loving someone but cannot confess?\\
\hline
\multirow{2}{*}{NS}                    & \makecell[l]{ {\begin{CJK*}{UTF8}{gbsn}不一定的\end{CJK*} }}\\
& Not necessarily. \\ 
\hline
\multirow{2}{*}{BT}  &  \makecell[l]{\begin{CJK*}{UTF8}{gbsn}爱一个人\error{不难，难的是}放下一个人。\end{CJK*}}\\
& \error{To} love \error{is easy, to} give up \error{is hard.} \\\hline
\end{tabularx}
\caption{\small Context (CXT), example generations from neucleus sampling (NS) and back-translation (BT). Novel words and syntax patterns are \error{highlighted}.}
\label{table: diverse-example}
\end{table}
\paragraph{Diversity of Generation}
We find the back translation model can generate \emph{both semantically and syntactically} novel responses. Some examples are shown in Table~\ref{table: diverse-example}. To find semantically novel responses, we segment them into phrases and find those containing novel phrases that do not exist on the conversational corpus. As in the first example of Table~\ref{table: diverse-example}, the word \begin{CJK*}{UTF8}{gbsn}胖若两人\end{CJK*} only exists in the non-conversational corpus. The model successfully learnt its semantic meaning and adopt it to generate novel responses. It is also common that the model learns frequent syntax structures from the non-conversational corpus. In the second example, it learnt the pattern of ``To ... is easy, to ... is hard", which appeared frequently in the non-conversational corpus, and utilized it to produce novel responses with the same structure. Note that both generations from the BT model \emph{never appear exactly in the non-conversational corpus}. It must generate them by correctly understanding the meaning of the phrase components instead of memorizing the utterances verbally.
\section{Conclusion}
We propose a novel way of diversifying dialogue generation by leveraging non-conversational text. To do so, we collect a large-scale corpus from forum comments, idioms and book snippets. By training the model through iterative back translation, it is able to significantly improve the diversity of generated responses both semantically and syntactically. We compare it with several strong baselines and find it achieved the best overall performance. The model can be potentially improved by filtering the corpus according to different domains, or augmenting with a retrieve-and-rewrite mechanism, which we leave for future work.
\cleardoublepage


\chapter{Conclusions and Future Prospects}
\label{chap: conclusions}
\lettrine[lines=3]{D}{eep} learning has reshaped the research of natural language processing with GPU parallel computing and large-scale databases. The task of text generation, which stands at the core place of natural language processing and machine intelligence, has been well studied and boosted with the population of recurrent neural networks and transformers. Recently, large-scale self-supervised pre-training further 
pushes the model limits of text generation and reaches or even surpasses human performance in many tasks~\citep{devlin2019bert}. Nonetheless, text generation is still far from being a solved problem. Specifically, deep neural networks tend to generate text with low diversity and lack in controllability and interpretability due to their blackbox nature. They also require extensive supervised training data in order to perform competitively. 

\section{Summary of Thesis}
In this thesis, we focus on addressing the above mentioned shortcomings of deep neural networks through an integration of latent-variable models. The text generation process is defined with a set of probablistic latent variables, through which we could diversify, interpret or control the text to be generated. It can also provide us with a principled way of utilizing non-paired text, enabling efficient unsupervised or semi-supervised learning. The integration of latent variables complicates the training as the exact log-likelihood is usually intractable. Nonetheless, we show the training can be done efficiently with well-designed approximation techniques like top-k sampling, dynamic programming or variational inference. In detail, we conclude this thesis with the following three points: 
\begin{itemize}
    \item latent-variable models can effectively diversify text generation by imposing an additional sampling process. We present algorithms to achieve this without sacrificing other attributes of generated text, while adding negligible training overhead (Chapter~\ref{chap: improve-vae},~\ref{chap: nexus}).
    \item By specifying the latent variable distribution to follow some human priors. It is able to automatically learn interpretable generation process. The text can be controlled via manipulating the latent variables (Chapter~\ref{chap: select},~\ref{chap: alignment},~\ref{chap: segment}).
    \item Non-paired text can be leveraged to boost the text generation model by treating the missing correspondence as latent variables. The optimization can be effectively done as in the EM algorithm(Chapter~\ref{chap: diversify}).
\end{itemize}

\section{Challenges}
However, latent-variable models still have many challenges when applied to text generation. 
A crucial unresolved challenge, and usually the core point of all latent variable applications, is the \textbf{training instability issue}. As we mentioned in Chapter~\ref{chap: optim}, the model can easily be trapped in a local optimum where latent variables are simply ignored, such that the whole model degrades into the standard seq2seq. Many approaches have been proposed to alleviate this problem, but a principled solution is not yet demonstrated. Most of them revolve it through some heuristic-based techniques, like to weight the learning objective, add handcrafted regularizations or careful initialization. The choice of them is highly data and task dependent, requiring a lot of human intuitions and designs. For example, in our work of controllable content selection (Chapter~\ref{chap: select}), we adopt the mutual information constraint to force meaningful info encoded into the latent variable. However, the set of the mutual information value is non-trivial. We set if by running human annotations to get a rough estimation, though an ideal model should be able to automate this process. In practice, even with careful design and initialization, the model might still behave unexpectedly while learning latent variables contrary to human intuitions. From one point, as the variables to be learnt are latent, errors made in the training process tend to be magnified, and can never be corrected back by help from explicit supervisions. From the other point, machines, in their strengths and weaknesses, are far different from human minds. Annotations from human supervisions, which serve as the ground truth for the human world, are not necessarily optimal for machines. Therefore, we can often see the cases where latent variables increase interpretability yet sacrifice the performance, as interpretability in the human sense might not be the optimal solution for machines. Before we design our own latent-variable models, we should think beforehand what we want to learn in the latent variables and what our final goal is. 
\begin{quote}
\emph{Do we expect the machines to perform similar to human minds to enable manual interpretation, or should we let machines learn their own way of thinking, to explore the most effective latent variables, even though uninterpretable to humans?}
\end{quote}

In the first case, proper supervision from humans is always helpful. We can use this signal to guide the latent variable learning in a semi-supervised setting. The second case is more challenging, as we are also uncertain about the format and meaning to be encoded into the latent variables. The current major solutions are intuition-based. We hope future works from the machine learning and natural language processing community come up with some principled, general-applicable metrics or algorithms to contribute to a more stable and robust latent-variable model.

Indeed, the recent wide applications of highly-complicated Transformer models further aggregate the challenge. The internal components of the neural network becomes more uninterpretable to humans. The representations at each position can be affected by its neighboring context, and each layer in the representation learn its different features. Building an interpretable latent-variable model on top of them becomes increasingly infeasible. 
\begin{quote}
    \emph{How do we effectively inorporate latent-variable models with Transformers? Or even, is latent-variable model and interpretability really needed? If Transformers can achieve state-of-the-art performance in their own magic way by simply using more resources and training data, is interpretability still an indispensable 
feature we want?}
\end{quote}
Exploring the answers to these questions would be an interesting future direction. The value of latent-variable models is general and will clearly not fade away due to the more complex Transformer architectures, but studying the combination of them is by all means important. After all, latent-variable models should be validated in the most powerful modern neural architectures.

The last point we would like to highlight is the re-utilization challenge. All the works mentioned in the thesis are task-dependent. To enable controllable text generation for different requirement, e.g., sentiment, content, style, etc, we need to define different probabilistic distributions and latent process. A model trained for one cannot be utilized for other applications at all. This is a large burden in real-life applications. User requirement changes along with time, and there are numerous latent factors and combinations of them that we might want to control. Listing the complete requirement from the beginning, and devising a comprehensive latent-variable model is unrealistic. Therefore, it brings us a new topic that is worth studying:
\begin{quote}
\emph{how do we effectively utilize trained models to adapt to newly introduced latent variables?}
\end{quote}
\cite{keskar2019ctrl} illustrate how to enable controllable generation from large-scaled pretrained language models in a lightweight way. However, their method only applies in a supervised way and all the variables are essentially not latent. They also only experimented with simple discrete controlling variables in a non-conditional setting, and did not shed light on how structured, finer-grained controlling can be fulfilled with conditional input. Re-utilizing pretrained model in practical applications has yet to be explored.
\cleardoublepage
%
\phantomsection
\addcontentsline{toc}{chapter}{\numberline{}\listfigurename}
\setcounter{lofdepth}{2}
\listoffigures
\clearoddpage 

\phantomsection
\addcontentsline{toc}{chapter}{\numberline{}\listtablename}
\setcounter{lotdepth}{2}
\listoftables
\clearoddpage

\addcontentsline{toc}{chapter}{\numberline{}Index}

\phantomsection
\addcontentsline{toc}{chapter}{\numberline{}\bibname} 
\bibliography{references}

\begin{thebibliography}{373}
{\markboth{\spacedlowsmallcaps{\bibname}}{\spacedlowsmallcaps{\bibname}}}
\expandafter\ifx\csname natexlab\endcsname\relax\def\natexlab#1{#1}\fi
\expandafter\ifx\csname url\endcsname\relax
  \def\url#1{{\tt #1}}\fi
\expandafter\ifx\csname urlprefix\endcsname\relax\def\urlprefix{URL }\fi

\bibitem[{Abadi \textit{et~al.}(2016)Abadi, Barham, Chen, Chen, Davis, Dean,
  Devin, Ghemawat, Irving, Isard, Kudlur, Levenberg, Monga, Moore, Murray,
  Steiner, Tucker, Vasudevan, Warden, Wicke, Yu, and
  Zheng}]{abadi2016tensorflow}
M.~Abadi, P.~Barham, J.~Chen, Z.~Chen, A.~Davis, J.~Dean, M.~Devin,
  S.~Ghemawat, G.~Irving, M.~Isard, M.~Kudlur, J.~Levenberg, R.~Monga,
  S.~Moore, D.~G. Murray, B.~Steiner, P.~Tucker, V.~Vasudevan, P.~Warden,
  M.~Wicke, Y.~Yu, and X.~Zheng (2016). TensorFlow: A System for Large-Scale
  Machine Learning, in {\em Proceedings of the 12th USENIX Conference on
  Operating Systems Design and Implementation\/} {\em 2016\/}.

\bibitem[{Adelani \textit{et~al.}(2021)Adelani, Zhang, Shen, Davody,
  Kleinbauer, and Klakow}]{adelani2021preventing}
D.~Adelani, M.~Zhang, X.~Shen, A.~Davody, T.~Kleinbauer, and D.~Klakow (2021).
  Preventing Author Profiling through Zero-Shot Multilingual Back-Translation,
  in {\em Proceedings of the 2021 Conference on Empirical Methods in Natural
  Language Processing\/} {\em 2021\/}.

\bibitem[{Aharoni and Goldberg(2017)}]{aharoni2017morphological}
R.~Aharoni and Y.~Goldberg (2017). Morphological Inflection Generation with
  Hard Monotonic Attention, in {\em Proceedings of the 55th Annual Meeting of
  the Association for Computational Linguistics (Volume 1: Long Papers)\/} {\em
  2017\/}.

\bibitem[{Alemi \textit{et~al.}(2018)Alemi, Poole, Fischer, Dillon, Saurous,
  and Murphy}]{pmlr-v80-alemi18a}
A.~Alemi, B.~Poole, I.~Fischer, J.~Dillon, R.~A. Saurous, and K.~Murphy (2018).
  Fixing a Broken {ELBO}, in {\em Proceedings of the 35th International
  Conference on Machine Learning\/} {\em 2018\/}.

\bibitem[{Allen and Hospedales(2019)}]{allen2019analogies}
C.~Allen and T.~Hospedales (2019). Analogies Explained: Towards Understanding
  Word Embeddings, {\em International Conference on Machine Learning (ICML)\/}.

\bibitem[{Angeli \textit{et~al.}(2010)Angeli, Liang, and
  Klein}]{angeli2010simple}
G.~Angeli, P.~Liang, and D.~Klein (2010). A simple domain-independent
  probabilistic approach to generation, in {\em Proceedings of the 2010
  Conference on Empirical Methods in Natural Language Processing\/} {\em
  2010\/}.

\bibitem[{Arjovsky and Bottou(2017)}]{arjovsky2017towards}
M.~Arjovsky and L.~Bottou (2017). Towards principled methods for training
  generative adversarial networks, {\em International Conference on Learning
  Representations(ICLR)\/}.

\bibitem[{Arjovsky \textit{et~al.}(2017)Arjovsky, Chintala, and
  Bottou}]{arjovsky2017wasserstein}
M.~Arjovsky, S.~Chintala, and L.~Bottou (2017). Wasserstein generative
  adversarial networks, in {\em International Conference on Machine Learning
  (ICLR)\/} {\em 2017\/}.

\bibitem[{Artetxe \textit{et~al.}(2018)Artetxe, Labaka, Agirre, and
  Cho}]{artetxe2017unsupervised}
M.~Artetxe, G.~Labaka, E.~Agirre, and K.~Cho (2018). Unsupervised neural
  machine translation, {\em International Conference on Learning
  Representations(ICLR)\/}.

\bibitem[{Backes \textit{et~al.}(2018)Backes, Berrang, Humbert, Shen, and
  Wolf}]{backes2018simulating}
M.~Backes, P.~Berrang, M.~Humbert, X.~Shen, and V.~Wolf (2018). Simulating the
  large-scale erosion of genomic privacy over time, {\em IEEE/ACM transactions
  on computational biology and bioinformatics\/}, vol.~15(5), pp.~1405--1412.

\bibitem[{Bahdanau \textit{et~al.}(2015)Bahdanau, Cho, and
  Bengio}]{bahdanau2015neural}
D.~Bahdanau, K.~Cho, and Y.~Bengio (2015). Neural machine translation by
  jointly learning to align and translate, {\em International Conference on
  Learning Representations(ICLR)\/}.

\bibitem[{Balakrishnan \textit{et~al.}(2019)Balakrishnan, Rao, Upasani, White,
  and Subba}]{balakrishnan2019constrained}
A.~Balakrishnan, J.~Rao, K.~Upasani, M.~White, and R.~Subba (2019). Constrained
  Decoding for Neural NLG from Compositional Representations in Task-Oriented
  Dialogue, in {\em Proceedings of the 57th Annual Meeting of the Association
  for Computational Linguistics\/} {\em 2019\/}.

\bibitem[{Banerjee and Lavie(2005)}]{banerjee2005meteor}
S.~Banerjee and A.~Lavie (2005). METEOR: An automatic metric for MT evaluation
  with improved correlation with human judgments, in {\em Proceedings of the
  acl workshop on intrinsic and extrinsic evaluation measures for machine
  translation and/or summarization\/} {\em 2005\/}.

\bibitem[{Bannard and Callison-Burch(2005)}]{bannard2005paraphrasing}
C.~Bannard and C.~Callison-Burch (2005). Paraphrasing with bilingual parallel
  corpora, in {\em Proceedings of the 43rd Annual Meeting on Association for
  Computational Linguistics\/} {\em 2005\/}.

\bibitem[{Barany \textit{et~al.}(2007)Barany, Vu
  \textit{et~al.}}]{barany2007central}
I.~Barany, V.~Vu, \textit{et~al.} (2007). Central limit theorems for Gaussian
  polytopes, {\em The Annals of Probability\/}, vol.~35(4), pp.~1593--1621.

\bibitem[{Bartl and Spanakis(2017)}]{bartl2017retrieval}
A.~Bartl and G.~Spanakis (2017). A retrieval-based dialogue system utilizing
  utterance and context embeddings, in {\em 2017 16th IEEE International
  Conference on Machine Learning and Applications (ICMLA)\/} {\em 2017\/}.

\bibitem[{Barzilay and Lapata(2005)}]{barzilay2005collective}
R.~Barzilay and M.~Lapata (2005). Collective content selection for
  concept-to-text generation, in {\em Proceedings of the conference on Human
  Language Technology and Empirical Methods in Natural Language Processing\/}
  {\em 2005\/}.

\bibitem[{Beaudry and Renner(2012)}]{beaudry2012intuitive}
N.~J. Beaudry and R.~Renner (2012). An intuitive proof of the data processing
  inequality, {\em Quantum Information \& Computation\/}, vol.~12(5-6),
  pp.~432--441.

\bibitem[{Belghazi \textit{et~al.}(2018)Belghazi, Baratin, Rajeshwar, Ozair,
  Bengio, Hjelm, and Courville}]{belghazi2018mutual}
M.~I. Belghazi, A.~Baratin, S.~Rajeshwar, S.~Ozair, Y.~Bengio, D.~Hjelm, and
  A.~Courville (2018). Mutual Information Neural Estimation, in {\em
  Proceedings of the 35th International Conference on Machine Learning\/} {\em
  2018\/}.

\bibitem[{Belz(2008)}]{belz2008automatic}
A.~Belz (2008). Automatic generation of weather forecast texts using
  comprehensive probabilistic generation-space models, {\em Natural Language
  Engineering\/}, vol.~14(4), pp.~431--455.

\bibitem[{Belz and Reiter(2006)}]{belz2006comparing}
A.~Belz and E.~Reiter (2006). Comparing automatic and human evaluation of NLG
  systems, in {\em 11th Conference of the European Chapter of the Association
  for Computational Linguistics\/} {\em 2006\/}.

\bibitem[{Bengio \textit{et~al.}(2015)Bengio, Vinyals, Jaitly, and
  Shazeer}]{bengio2015scheduled}
S.~Bengio, O.~Vinyals, N.~Jaitly, and N.~Shazeer (2015). Scheduled sampling for
  sequence prediction with recurrent neural networks, in {\em Advances in
  Neural Information Processing Systems\/} {\em 2015\/}.

\bibitem[{Bengio \textit{et~al.}(2003)Bengio, Ducharme, Vincent, and
  Jauvin}]{bengio2003neural}
Y.~Bengio, R.~Ducharme, P.~Vincent, and C.~Jauvin (2003). A neural
  probabilistic language model, {\em Journal of machine learning research\/},
  vol.~3(Feb), pp.~1137--1155.

\bibitem[{Berthelot \textit{et~al.}(2017)Berthelot, Schumm, and
  Metz}]{berthelot2017began}
D.~Berthelot, T.~Schumm, and L.~Metz (2017). Began: Boundary equilibrium
  generative adversarial networks, {\em arXiv preprint arXiv:1703.10717\/}.

\bibitem[{Blei \textit{et~al.}(2003)Blei, Ng, and Jordan}]{blei2003latent}
D.~M. Blei, A.~Y. Ng, and M.~I. Jordan (2003). Latent dirichlet allocation,
  {\em Journal of machine Learning research\/}, vol.~3(Jan), pp.~993--1022.

\bibitem[{Bordes \textit{et~al.}(2013)Bordes, Usunier, Garcia-Duran, Weston,
  and Yakhnenko}]{bordes2013translating}
A.~Bordes, N.~Usunier, A.~Garcia-Duran, J.~Weston, and O.~Yakhnenko (2013).
  Translating embeddings for modeling multi-relational data, in {\em Advances
  in neural information processing systems\/} {\em 2013\/}.

\bibitem[{Borgwardt \textit{et~al.}(2021)Borgwardt, Chang, Chapman, Demberg,
  Kovtunova, and Yeh}]{borgwardt2021logic}
S.~Borgwardt, E.~Chang, K.~Chapman, V.~Demberg, A.~Kovtunova, and H.-S. Yeh
  (2021). Logic-Guided Neural Utterance Generation from Drone Sensory Data, in
  {\em 34th International Workshop on Description Logics\/} {\em 2021\/}.

\bibitem[{Bowman \textit{et~al.}(2016)Bowman, Vilnis, Vinyals, Dai, Jozefowicz,
  and Bengio}]{bowman2016generating}
S.~R. Bowman, L.~Vilnis, O.~Vinyals, A.~Dai, R.~Jozefowicz, and S.~Bengio
  (2016). Generating Sentences from a Continuous Space, in {\em Proceedings of
  The 20th SIGNLL Conference on Computational Natural Language Learning\/} {\em
  2016\/}.

\bibitem[{Britz \textit{et~al.}(2017)Britz, Guan, and
  Luong}]{britz2017efficient}
D.~Britz, M.~Guan, and M.-T. Luong (2017). Efficient Attention using a
  Fixed-Size Memory Representation, in {\em Proceedings of the 2017 Conference
  on Empirical Methods in Natural Language Processing\/} {\em 2017\/}.

\bibitem[{Brown \textit{et~al.}(1993)Brown, Pietra, Pietra, and
  Mercer}]{brown1993mathematics}
P.~F. Brown, V.~J.~D. Pietra, S.~A.~D. Pietra, and R.~L. Mercer (1993). The
  mathematics of statistical machine translation: Parameter estimation, {\em
  Computational linguistics\/}, vol.~19(2), pp.~263--311.

\bibitem[{Burda \textit{et~al.}(2015)Burda, Grosse, and
  Salakhutdinov}]{burda2015importance}
Y.~Burda, R.~B. Grosse, and R.~Salakhutdinov (2015). Importance Weighted
  Autoencoders, {\em CoRR\/}, vol.~abs/1509.00519.

\bibitem[{Cao \textit{et~al.}(2018)Cao, Wei, Li, and Li}]{cao2018faithful}
Z.~Cao, F.~Wei, W.~Li, and S.~Li (2018). Faithful to the original: Fact aware
  neural abstractive summarization, in {\em Thirty-Second AAAI Conference on
  Artificial Intelligence\/} {\em 2018\/}.

\bibitem[{Chang(2018)}]{chang2018generative}
E.~Chang (2018). {\em A Generative-Discriminative Framework for Title
  Generation in the E-commerce Domain\/}.

\bibitem[{Chang \textit{et~al.}(2020{\natexlab{a}})Chang, Adelani, Shen, and
  Demberg}]{chang2020unsupervised}
E.~Chang, D.~Adelani, X.~Shen, and V.~Demberg (2020{\natexlab{a}}).
  Unsupervised Pidgin Text Generation By Pivoting English Data and
  Self-Training, in {\em In Proceedings of Workshop at ICLR\/} {\em 2020\/}.

\bibitem[{Chang \textit{et~al.}(2020{\natexlab{b}})Chang, Caplinger, Marin,
  Shen, and Demberg}]{chang2020dart}
E.~Chang, J.~Caplinger, A.~Marin, X.~Shen, and V.~Demberg (2020{\natexlab{b}}).
  DART: A Lightweight Quality-Suggestive Data-to-Text Annotation Tool, {\em
  arXiv preprint arXiv:2010.04141\/}.

\bibitem[{Chang \textit{et~al.}(2021{\natexlab{a}})Chang, Demberg, and
  Marin}]{chang2021jointly}
E.~Chang, V.~Demberg, and A.~Marin (2021{\natexlab{a}}). Jointly Improving
  Language Understanding and Generation with Quality-Weighted Weak Supervision
  of Automatic Labeling, {\em Proceedings of EACL 2021\/}.

\bibitem[{Chang \textit{et~al.}(2021{\natexlab{b}})Chang, Shen, Marin, and
  Demberg}]{chang2021selectgen}
E.~Chang, X.~Shen, A.~Marin, and V.~Demberg (2021{\natexlab{b}}). The SelectGen
  Challenge: Finding the Best Training Samples for Few-Shot Neural Text
  Generation, in {\em Proceedings of the 14th INLG\/} {\em 2021\/}.

\bibitem[{Chang \textit{et~al.}(2021{\natexlab{c}})Chang, Shen, Yeh, and
  Demberg}]{chang2021training}
E.~Chang, X.~Shen, H.-S. Yeh, and V.~Demberg (2021{\natexlab{c}}). On Training
  Instance Selection for Few-Shot Neural Text Generation, in {\em Proceedings
  of ACL 2021\/} {\em 2021\/}.

\bibitem[{Chang \textit{et~al.}(2021{\natexlab{d}})Chang, Shen, Zhu, Demberg,
  and Su}]{chang2021neural}
E.~Chang, X.~Shen, D.~Zhu, V.~Demberg, and H.~Su (2021{\natexlab{d}}). Neural
  Data-to-text Generation with LM-based Text Augmentation, {\em Proceedings of
  EACL 2021\/}.

\bibitem[{Chang \textit{et~al.}(2021{\natexlab{e}})Chang, Shiue, Yeh, and
  Demberg}]{chang2021time}
E.~Chang, Y.-T. Shiue, H.-S. Yeh, and V.~Demberg (2021{\natexlab{e}}).
  Time-Aware Ancient Chinese Text Translation and Inference, in {\em LChange @
  ACL 2021\/} {\em 2021\/}.

\bibitem[{Chang \textit{et~al.}(2021{\natexlab{f}})Chang, Yeh, and
  Demberg}]{chang2021does}
E.~Chang, H.-S. Yeh, and V.~Demberg (2021{\natexlab{f}}). Does the Order of
  Training Samples Matter? Improving Neural Data-to-Text Generation with
  Curriculum Learning, {\em Proceedings of EACL 2021\/}.

\bibitem[{Chen \textit{et~al.}(2018)Chen, Lampouras, and
  Vlachos}]{chen2018sheffield}
M.~Chen, G.~Lampouras, and A.~Vlachos (2018). Sheffield at E2E: structured
  prediction approaches to end-to-end language generation, {\em arxiv\/}.

\bibitem[{Chen and Goodman(1996)}]{chen1996empirical}
S.~F. Chen and J.~Goodman (1996). An empirical study of smoothing techniques
  for language modeling, in {\em Proceedings of the 34th annual meeting on
  Association for Computational Linguistics\/} {\em 1996\/}.

\bibitem[{Chen \textit{et~al.}(2016)Chen, Duan, Houthooft, Schulman, Sutskever,
  and Abbeel}]{chen2016infogan}
X.~Chen, Y.~Duan, R.~Houthooft, J.~Schulman, I.~Sutskever, and P.~Abbeel
  (2016). Infogan: Interpretable representation learning by information
  maximizing generative adversarial nets, in {\em Advances in Neural
  Information Processing Systems\/} {\em 2016\/}.

\bibitem[{Chen \textit{et~al.}(2017)Chen, Kingma, Salimans, Duan, Dhariwal,
  Schulman, Sutskever, and Abbeel}]{chen2016variational}
X.~Chen, D.~P. Kingma, T.~Salimans, Y.~Duan, P.~Dhariwal, J.~Schulman,
  I.~Sutskever, and P.~Abbeel (2017). Variational Lossy Autoencoder, {\em
  International Conference on Learning Representations(ICLR)\/}.

\bibitem[{Chen and Bansal(2018)}]{chen2018fast}
Y.-C. Chen and M.~Bansal (2018). Fast Abstractive Summarization with
  Reinforce-Selected Sentence Rewriting, in {\em Proceedings of the 56th Annual
  Meeting of the Association for Computational Linguistics (Volume 1: Long
  Papers)\/} {\em 2018\/}.

\bibitem[{Child \textit{et~al.}(2019)Child, Gray, Radford, and
  Sutskever}]{child2019generating}
R.~Child, S.~Gray, A.~Radford, and I.~Sutskever (2019). Generating long
  sequences with sparse transformers, {\em arXiv preprint arXiv:1904.10509\/}.

\bibitem[{Chiu and Raffel(2018)}]{chiu2017monotonic}
C.-C. Chiu and C.~Raffel (2018). Monotonic chunkwise attention, {\em
  International Conference on Learning Representations(ICLR)\/}.

\bibitem[{Cho \textit{et~al.}(2014)Cho, van Merrienboer, Gulcehre, Bahdanau,
  Bougares, Schwenk, and Bengio}]{cho2014learning}
K.~Cho, B.~van Merrienboer, C.~Gulcehre, D.~Bahdanau, F.~Bougares, H.~Schwenk,
  and Y.~Bengio (2014). Learning Phrase Representations using RNN
  Encoder--Decoder for Statistical Machine Translation, in {\em Proceedings of
  the 2014 Conference on Empirical Methods in Natural Language Processing
  (EMNLP)\/} {\em 2014\/}.

\bibitem[{Clarke and Lapata(2010)}]{clarke2010discourse}
J.~Clarke and M.~Lapata (2010). Discourse constraints for document compression,
  {\em Computational Linguistics\/}, vol.~36(3), pp.~411--441.

\bibitem[{Cotterell and Kreutzer(2018)}]{cotterell2018explaining}
R.~Cotterell and J.~Kreutzer (2018). Explaining and Generalizing
  Back-Translation through Wake-Sleep, {\em arXiv preprint arXiv:1806.04402\/}.

\bibitem[{Creswell and Bharath(2016)}]{creswell2016task}
A.~Creswell and A.~A. Bharath (2016). Task specific adversarial cost function,
  {\em arXiv preprint arXiv:1609.08661\/}.

\bibitem[{Dale \textit{et~al.}(2012)Dale, Anisimoff, and
  Narroway}]{dale2012hoo}
R.~Dale, I.~Anisimoff, and G.~Narroway (2012). HOO 2012: A report on the
  preposition and determiner error correction shared task, in {\em Proceedings
  of the Seventh Workshop on Building Educational Applications Using NLP\/}
  {\em 2012\/}.

\bibitem[{Daum{\'e}~III and Marcu(2005)}]{daume2005induction}
H.~Daum{\'e}~III and D.~Marcu (2005). Induction of word and phrase alignments
  for automatic document summarization, {\em Computational Linguistics\/},
  vol.~31(4), pp.~505--530.

\bibitem[{de~Souza \textit{et~al.}(2018)de~Souza, Kozielski, Mathur, Chang,
  Guerini, Negri, Turchi, and Matusov}]{de2018generating}
J.~G. de~Souza, M.~Kozielski, P.~Mathur, E.~Chang, M.~Guerini, M.~Negri,
  M.~Turchi, and E.~Matusov (2018). Generating e-commerce product titles and
  predicting their quality, in {\em Proceedings of INLG\/} {\em 2018\/}.

\bibitem[{Del~Tredici \textit{et~al.}(2021)Del~Tredici, Barlacchi, Shen, Cheng,
  and de~Gispert}]{del2021question}
M.~Del~Tredici, G.~Barlacchi, X.~Shen, W.~Cheng, and A.~de~Gispert (2021).
  Question Rewriting for Open-Domain Conversational QA: Best Practices and
  Limitations, in {\em Proceedings of the 30th ACM International Conference on
  Information \& Knowledge Management\/} {\em 2021\/}.

\bibitem[{Dempster \textit{et~al.}(1977)Dempster, Laird, and
  Rubin}]{dempster1977maximum}
A.~P. Dempster, N.~M. Laird, and D.~B. Rubin (1977). Maximum likelihood from
  incomplete data via the EM algorithm, {\em Journal of the Royal Statistical
  Society: Series B (Methodological)\/}, vol.~39(1), pp.~1--22.

\bibitem[{Deng \textit{et~al.}(2020)Deng, Bakhtin, Ott, Szlam, and
  Ranzato}]{deng2020residual}
Y.~Deng, A.~Bakhtin, M.~Ott, A.~Szlam, and M.~Ranzato (2020). Residual
  Energy-Based Models for Text Generation, in {\em International Conference on
  Learning Representations\/} {\em 2020\/}.

\bibitem[{Deng \textit{et~al.}(2017)Deng, Kanervisto, Ling, and
  Rush}]{deng2017image}
Y.~Deng, A.~Kanervisto, J.~Ling, and A.~M. Rush (2017). Image-to-markup
  generation with coarse-to-fine attention, in {\em Proceedings of the 34th
  International Conference on Machine Learning-Volume 70\/} {\em 2017\/}.

\bibitem[{Deng \textit{et~al.}(2018)Deng, Kim, Chiu, Guo, and
  Rush}]{deng2018latent}
Y.~Deng, Y.~Kim, J.~Chiu, D.~Guo, and A.~Rush (2018). Latent alignment and
  variational attention, in {\em Advances in Neural Information Processing
  Systems\/} {\em 2018\/}.

\bibitem[{Devlin \textit{et~al.}(2019{\natexlab{a}})Devlin, Chang, Lee, and
  Toutanova}]{devlin2018bert}
J.~Devlin, M.-W. Chang, K.~Lee, and K.~Toutanova (2019{\natexlab{a}}). Bert:
  Pre-training of deep bidirectional transformers for language understanding,
  {\em The North American Chapter of the Association for Computational
  Linguistics\/}.

\bibitem[{Devlin \textit{et~al.}(2019{\natexlab{b}})Devlin, Chang, Lee, and
  Toutanova}]{devlin2019bert}
J.~Devlin, M.-W. Chang, K.~Lee, and K.~Toutanova (2019{\natexlab{b}}). BERT:
  Pre-training of Deep Bidirectional Transformers for Language Understanding,
  in {\em Proceedings of the 2019 Conference of the North American Chapter of
  the Association for Computational Linguistics: Human Language Technologies,
  Volume 1 (Long and Short Papers)\/} {\em 2019\/}.

\bibitem[{Dinan \textit{et~al.}(2019)Dinan, Roller, Shuster, Fan, Auli, and
  Weston}]{dinan2018wizard}
E.~Dinan, S.~Roller, K.~Shuster, A.~Fan, M.~Auli, and J.~Weston (2019). Wizard
  of wikipedia: Knowledge-powered conversational agents, {\em International
  Conference on Learning Representations(ICLR)\/}.

\bibitem[{Dinh \textit{et~al.}(2014)Dinh, Krueger, and Bengio}]{dinh2014nice}
L.~Dinh, D.~Krueger, and Y.~Bengio (2014). NICE: Non-linear independent
  components estimation, {\em International Conference on Learning
  Representation, Big Learn workshop\/}.

\bibitem[{Donahue \textit{et~al.}(2017)Donahue, Kr{\"a}henb{\"u}hl, and
  Darrell}]{donahue2016adversarial}
J.~Donahue, P.~Kr{\"a}henb{\"u}hl, and T.~Darrell (2017). Adversarial feature
  learning, {\em International Conference on Learning Representations(ICLR)\/}.

\bibitem[{Donsker and Varadhan(1983)}]{donsker1983asymptotic}
M.~D. Donsker and S.~S. Varadhan (1983). Asymptotic evaluation of certain
  Markov process expectations for large time. IV, {\em Communications on Pure
  and Applied Mathematics\/}, vol.~36(2), pp.~183--212.

\bibitem[{Dosovitskiy and Brox(2016)}]{dosovitskiy2016generating}
A.~Dosovitskiy and T.~Brox (2016). Generating images with perceptual similarity
  metrics based on deep networks, in {\em Advances in Neural Information
  Processing Systems\/} {\em 2016\/}.

\bibitem[{Dumoulin \textit{et~al.}(2017)Dumoulin, Belghazi, Poole, Lamb,
  Arjovsky, Mastropietro, and Courville}]{dumoulin2016adversarially}
V.~Dumoulin, I.~Belghazi, B.~Poole, A.~Lamb, M.~Arjovsky, O.~Mastropietro, and
  A.~Courville (2017). Adversarially learned inference, {\em International
  Conference on Learning Representations(ICLR)\/}.

\bibitem[{Du{\v{s}}ek \textit{et~al.}(2018)Du{\v{s}}ek, Novikova, and
  Rieser}]{duvsek2018findings}
O.~Du{\v{s}}ek, J.~Novikova, and V.~Rieser (2018). Findings of the E2E NLG
  Challenge, in {\em Proceedings of the 11th International Conference on
  Natural Language Generation\/} {\em 2018\/}.

\bibitem[{Du{\v{s}}ek \textit{et~al.}(2020)Du{\v{s}}ek, Novikova, and
  Rieser}]{duvsek2020evaluating}
O.~Du{\v{s}}ek, J.~Novikova, and V.~Rieser (2020). Evaluating the
  state-of-the-art of end-to-end natural language generation: The E2E NLG
  Challenge, {\em Computer Speech \& Language\/}, vol.~59, pp.~123--156.

\bibitem[{Eisner(2002)}]{eisner2002parameter}
J.~Eisner (2002). Parameter estimation for probabilistic finite-state
  transducers, in {\em Proceedings of the 40th Annual Meeting of the
  Association for Computational Linguistics\/} {\em 2002\/}.

\bibitem[{Eisner(2016)}]{eisner2016inside}
J.~Eisner (2016). Inside-outside and forward-backward algorithms are just
  backprop (tutorial paper), in {\em Proceedings of the Workshop on Structured
  Prediction for NLP\/} {\em 2016\/}.

\bibitem[{Elder \textit{et~al.}(2019)Elder, Foster, Barry, and
  OConnor}]{elder2019designing}
H.~Elder, J.~Foster, J.~Barry, and A.~OConnor (2019). Designing a Symbolic
  Intermediate Representation for Neural Surface Realization, in {\em
  Proceedings of the Workshop on Methods for Optimizing and Evaluating Neural
  Language Generation\/} {\em 2019\/}.

\bibitem[{Fan \textit{et~al.}(2018{\natexlab{a}})Fan, Grangier, and
  Auli}]{fan2017controllable}
A.~Fan, D.~Grangier, and M.~Auli (2018{\natexlab{a}}). Controllable Abstractive
  Summarization, in {\em Proceedings of the 2nd Workshop on Neural Machine
  Translation and Generation\/} {\em 2018\/}.

\bibitem[{Fan \textit{et~al.}(2018{\natexlab{b}})Fan, Grangier, and
  Auli}]{fan2018controllable}
A.~Fan, D.~Grangier, and M.~Auli (2018{\natexlab{b}}). Controllable Abstractive
  Summarization, in {\em Proceedings of the 2nd Workshop on Neural Machine
  Translation and Generation\/} {\em 2018\/}.

\bibitem[{Fedus \textit{et~al.}(2018)Fedus, Goodfellow, and
  Dai}]{fedus2018maskgan}
W.~Fedus, I.~Goodfellow, and A.~M. Dai (2018). MaskGAN: Better Text Generation
  via Filling in the \_\_\_\_\_\_\_, in {\em International Conference on
  Learning Representations\/} {\em 2018\/}.

\bibitem[{Ferreira \textit{et~al.}(2019)Ferreira, van~der Lee, van Miltenburg,
  and Krahmer}]{ferreira2019neural}
T.~C. Ferreira, C.~van~der Lee, E.~van Miltenburg, and E.~Krahmer (2019).
  Neural data-to-text generation: A comparison between pipeline and end-to-end
  architectures, {\em Conference on Empirical Methods in Natural Language
  Processing\/}.

\bibitem[{Fleiss(1971)}]{fleiss1971measuring}
J.~L. Fleiss (1971). Measuring nominal scale agreement among many raters., {\em
  Psychological bulletin\/}, vol.~76(5), p.~378.

\bibitem[{Frey(1998)}]{frey1998graphical}
B.~J. Frey (1998). {\em Graphical models for machine learning and digital
  communication\/}, MIT press.

\bibitem[{Frey \textit{et~al.}(1996)Frey, Hinton, and Dayan}]{frey1996does}
B.~J. Frey, G.~E. Hinton, and a.~P. Dayan (1996). Does the wake-sleep algorithm
  produce good density estimators?, in {\em Advances in neural information
  processing systems\/} {\em 1996\/}.

\bibitem[{Ganchev \textit{et~al.}(2010)Ganchev, Gillenwater, Taskar
  \textit{et~al.}}]{ganchev2010posterior}
K.~Ganchev, J.~Gillenwater, B.~Taskar, \textit{et~al.} (2010). Posterior
  regularization for structured latent variable models, {\em Journal of Machine
  Learning Research\/}, vol.~11(Jul), pp.~2001--2049.

\bibitem[{Gao \textit{et~al.}(2019)Gao, Zhang, Lee, Galley, Brockett, Gao, and
  Dolan}]{gao2019structuring}
X.~Gao, Y.~Zhang, S.~Lee, M.~Galley, C.~Brockett, J.~Gao, and B.~Dolan (2019).
  Structuring Latent Spaces for Stylized Response Generation, in {\em
  Proceedings of the 2019 Conference on Empirical Methods in Natural Language
  Processing and the 9th International Joint Conference on Natural Language
  Processing (EMNLP-IJCNLP)\/} {\em 2019\/}.

\bibitem[{Gardent \textit{et~al.}(2017)Gardent, Shimorina, Narayan, and
  Perez-Beltrachini}]{gardent2017webnlg}
C.~Gardent, A.~Shimorina, S.~Narayan, and L.~Perez-Beltrachini (2017). The
  webnlg challenge: Generating text from rdf data, in {\em Proceedings of the
  10th International Conference on Natural Language Generation\/} {\em 2017\/}.

\bibitem[{Ge \textit{et~al.}(2021)Ge, Huang, Shen, Li, and Hu}]{ge2021learning}
J.~Ge, Y.~Huang, X.~Shen, C.~Li, and W.~Hu (2021). Learning Fine-Grained
  Fact-Article Correspondence in Legal Cases, {\em IEEE/ACM Transactions on
  Audio, Speech, and Language Processing\/}, vol.~29, pp.~3694--3706.

\bibitem[{Gehring \textit{et~al.}(2017)Gehring, Auli, Grangier, Yarats, and
  Dauphin}]{gehring2017convolutional}
J.~Gehring, M.~Auli, D.~Grangier, D.~Yarats, and Y.~N. Dauphin (2017).
  Convolutional Sequence to Sequence Learning, in {\em International Conference
  on Machine Learning\/} {\em 2017\/}.

\bibitem[{Gehrmann \textit{et~al.}(2018)Gehrmann, Deng, and
  Rush}]{gehrmann2018bottom}
S.~Gehrmann, Y.~Deng, and A.~Rush (2018). Bottom-Up Abstractive Summarization,
  in {\em Proceedings of the 2018 Conference on Empirical Methods in Natural
  Language Processing\/} {\em 2018\/}.

\bibitem[{Germain \textit{et~al.}(2015)Germain, Gregor, Murray, and
  Larochelle}]{germain2015made}
M.~Germain, K.~Gregor, I.~Murray, and H.~Larochelle (2015). MADE: masked
  autoencoder for distribution estimation, in {\em Proceedings of the 32nd
  International Conference on Machine Learning (ICML-15)\/} {\em 2015\/}.

\bibitem[{Ghader and Monz(2017)}]{ghader2017does}
H.~Ghader and C.~Monz (2017). What does Attention in Neural Machine Translation
  Pay Attention to?, in {\em Proceedings of the Eighth International Joint
  Conference on Natural Language Processing (Volume 1: Long Papers)\/} {\em
  2017\/}.

\bibitem[{Ghazvininejad \textit{et~al.}(2018)Ghazvininejad, Brockett, Chang,
  Dolan, Gao, Yih, and Galley}]{ghazvininejad2018knowledge}
M.~Ghazvininejad, C.~Brockett, M.-W. Chang, B.~Dolan, J.~Gao, W.-t. Yih, and
  M.~Galley (2018). A knowledge-grounded neural conversation model, in {\em
  Thirty-Second AAAI Conference on Artificial Intelligence\/} {\em 2018\/}.

\bibitem[{Glynn(1990)}]{glynn1990likelilood}
P.~W. Glynn (1990). Likelihood Ratio Gradient Estimation for Stochastic
  Systems, {\em Commun. ACM\/}, vol.~33(10), pp.~75--84.

\bibitem[{Godfrey and Holliman()}]{godfreyswitchboard}
J.~Godfrey and E.~Holliman (). SWITCHBOARD-1 Release 2, 1997, {\em Linguistic
  Data Consortium, Philadelphia\/}.

\bibitem[{Goldberg \textit{et~al.}(1994)Goldberg, Driedger, and
  Kittredge}]{goldberg1994using}
E.~Goldberg, N.~Driedger, and R.~I. Kittredge (1994). Using natural-language
  processing to produce weather forecasts, {\em IEEE Expert\/}, vol.~9(2),
  pp.~45--53.

\bibitem[{Goodfellow(2016)}]{goodfellow2016nips}
I.~Goodfellow (2016). NIPS 2016 tutorial: Generative adversarial networks, {\em
  Advances in neural information processing systems tutorial\/}.

\bibitem[{Goodfellow \textit{et~al.}(2014)Goodfellow, Pouget-Abadie, Mirza, Xu,
  Warde-Farley, Ozair, Courville, and Bengio}]{goodfellow2014generative}
I.~Goodfellow, J.~Pouget-Abadie, M.~Mirza, B.~Xu, D.~Warde-Farley, S.~Ozair,
  A.~Courville, and Y.~Bengio (2014). Generative adversarial nets, in {\em
  Advances in neural information processing systems\/} {\em 2014\/}.

\bibitem[{Grathwohl \textit{et~al.}(2018)Grathwohl, Choi, Wu, Roeder, and
  Duvenaud}]{grathwohl2018backpropagation}
W.~Grathwohl, D.~Choi, Y.~Wu, G.~Roeder, and D.~Duvenaud (2018).
  Backpropagation through the Void: Optimizing control variates for black-box
  gradient estimation, in {\em International Conference on Learning
  Representations\/} {\em 2018\/}.

\bibitem[{Graves(2012)}]{graves2012sequence}
A.~Graves (2012). Sequence Transduction with Recurrent Neural Networks, {\em
  CoRR\/}, vol.~abs/1211.3711.

\bibitem[{Gu \textit{et~al.}(2018{\natexlab{a}})Gu, Bradbury, Xiong, Li, and
  Socher}]{gu2018nonautoregressive}
J.~Gu, J.~Bradbury, C.~Xiong, V.~O. Li, and R.~Socher (2018{\natexlab{a}}).
  Non-Autoregressive Neural Machine Translation, in {\em International
  Conference on Learning Representations\/} {\em 2018\/}.

\bibitem[{Gu \textit{et~al.}(2018{\natexlab{b}})Gu, Im, and Li}]{gu2017neural}
J.~Gu, D.~J. Im, and V.~O. Li (2018{\natexlab{b}}). Neural machine translation
  with gumbel-greedy decoding, {\em Association for the Advancement of
  Artificial Intelligence\/}, pp. 5125--5132.

\bibitem[{Gu \textit{et~al.}(2016)Gu, Lu, Li, and Li}]{gu2016incorporating}
J.~Gu, Z.~Lu, H.~Li, and V.~O. Li (2016). Incorporating Copying Mechanism in
  Sequence-to-Sequence Learning, in {\em Proceedings of the 54th Annual Meeting
  of the Association for Computational Linguistics (Volume 1: Long Papers)\/}
  {\em 2016\/}.

\bibitem[{Gulcehre \textit{et~al.}(2016)Gulcehre, Ahn, Nallapati, Zhou, and
  Bengio}]{gulcehre2016pointing}
C.~Gulcehre, S.~Ahn, R.~Nallapati, B.~Zhou, and Y.~Bengio (2016). Pointing the
  Unknown Words, in {\em Proceedings of the 54th Annual Meeting of the
  Association for Computational Linguistics (Volume 1: Long Papers)\/} {\em
  2016\/}.

\bibitem[{Gulrajani \textit{et~al.}(2017)Gulrajani, Ahmed, Arjovsky, Dumoulin,
  and Courville}]{gulrajani2017improved}
I.~Gulrajani, F.~Ahmed, M.~Arjovsky, V.~Dumoulin, and A.~Courville (2017).
  Improved training of wasserstein gans, {\em Advances in Neural Information
  Processing Systems\/}.

\bibitem[{Guo \textit{et~al.}(2017)Guo, Lu, Cai, Zhang, Yu, and
  Wang}]{guo2017long}
J.~Guo, S.~Lu, H.~Cai, W.~Zhang, Y.~Yu, and J.~Wang (2017). Long Text
  Generation via Adversarial Training with Leaked Information, {\em Association
  for the Advancement of Artificial Intelligence\/}.

\bibitem[{Hakami \textit{et~al.}(2018)Hakami, Hayashi, and
  Bollegala}]{hakami2018does}
H.~Hakami, K.~Hayashi, and D.~Bollegala (2018). Why does PairDiff work?-A
  Mathematical Analysis of Bilinear Relational Compositional Operators for
  Analogy Detection, in {\em Proceedings of the 27th International Conference
  on Computational Linguistics\/} {\em 2018\/}.

\bibitem[{He \textit{et~al.}(2016{\natexlab{a}})He, Xia, Qin, Wang, Yu, Liu,
  and Ma}]{he2016dual}
D.~He, Y.~Xia, T.~Qin, L.~Wang, N.~Yu, T.~Liu, and W.-Y. Ma
  (2016{\natexlab{a}}). Dual learning for machine translation, in {\em Advances
  in Neural Information Processing Systems\/} {\em 2016\/}.

\bibitem[{He \textit{et~al.}(2020)He, Wang, Neubig, and
  Berg-Kirkpatrick}]{he2020a}
J.~He, X.~Wang, G.~Neubig, and T.~Berg-Kirkpatrick (2020). A Probabilistic
  Formulation of Unsupervised Text Style Transfer, in {\em International
  Conference on Learning Representations\/} {\em 2020\/}.

\bibitem[{He \textit{et~al.}(2016{\natexlab{b}})He, Zhang, Ren, and
  Sun}]{he2016deep}
K.~He, X.~Zhang, S.~Ren, and J.~Sun (2016{\natexlab{b}}). Deep residual
  learning for image recognition, in {\em Proceedings of the IEEE conference on
  computer vision and pattern recognition\/} {\em 2016\/}.

\bibitem[{He \textit{et~al.}(2018)He, Haffari, and Norouzi}]{he2018sequence}
X.~He, G.~Haffari, and M.~Norouzi (2018). Sequence to Sequence Mixture Model
  for Diverse Machine Translation, in {\em Proceedings of the 22nd Conference
  on Computational Natural Language Learning\/} {\em 2018\/}.

\bibitem[{Hermann \textit{et~al.}(2015)Hermann, Kocisky, Grefenstette,
  Espeholt, Kay, Suleyman, and Blunsom}]{hermann2015teaching}
K.~M. Hermann, T.~Kocisky, E.~Grefenstette, L.~Espeholt, W.~Kay, M.~Suleyman,
  and P.~Blunsom (2015). Teaching machines to read and comprehend, in {\em
  Advances in Neural Information Processing Systems\/} {\em 2015\/}.

\bibitem[{Higgins \textit{et~al.}(2017)Higgins, Matthey, Pal, Burgess, Glorot,
  Botvinick, Mohamed, and Lerchner}]{higgins2017beta}
I.~Higgins, L.~Matthey, A.~Pal, C.~Burgess, X.~Glorot, M.~Botvinick,
  S.~Mohamed, and A.~Lerchner (2017). beta-VAE: Learning basic visual concepts
  with a constrained variational framework, in {\em In Proceedings of the
  International Conference on Learning Representations (ICLR)\/} {\em 2017\/}.

\bibitem[{Hinton \textit{et~al.}(1995)Hinton, Dayan, Frey, and
  Neal}]{hinton1995wake}
G.~E. Hinton, P.~Dayan, B.~J. Frey, and R.~M. Neal (1995). The" wake-sleep"
  algorithm for unsupervised neural networks, {\em Science\/}, vol.~268(5214),
  p.~1158.

\bibitem[{Hjelm \textit{et~al.}(2017)Hjelm, Jacob, Che, Cho, and
  Bengio}]{hjelm2017boundary}
R.~D. Hjelm, A.~P. Jacob, T.~Che, K.~Cho, and Y.~Bengio (2017).
  Boundary-Seeking Generative Adversarial Networks, {\em arXiv preprint
  arXiv:1702.08431\/}.

\bibitem[{Hoang \textit{et~al.}(2018)Hoang, Koehn, Haffari, and
  Cohn}]{hoang2018iterative}
V.~C.~D. Hoang, P.~Koehn, G.~Haffari, and T.~Cohn (2018). Iterative
  Back-Translation for Neural Machine Translation, in {\em Proceedings of the
  2nd Workshop on Neural Machine Translation and Generation\/} {\em 2018\/}.

\bibitem[{Hochreiter and Schmidhuber(1997)}]{hochreiter1997long}
S.~Hochreiter and J.~Schmidhuber (1997). Long Short-Term Memory, {\em Neural
  Computation\/}, vol.~9(8), pp.~1735--1780.

\bibitem[{Holtzman \textit{et~al.}(2020)Holtzman, Buys, Du, Forbes, and
  Choi}]{holtzman2020the}
A.~Holtzman, J.~Buys, L.~Du, M.~Forbes, and Y.~Choi (2020). The Curious Case of
  Neural Text Degeneration, in {\em International Conference on Learning
  Representations\/} {\em 2020\/}.

\bibitem[{Hong \textit{et~al.}(2019)Hong, Chang, and
  Demberg}]{hong2019improving}
X.~Hong, E.~Chang, and V.~Demberg (2019). Improving language generation from
  feature-rich tree-structured data with relational graph convolutional
  encoders, in {\em Proceedings of the 2nd Workshop on Multilingual Surface
  Realisation (MSR 2019)\/} {\em 2019\/}.

\bibitem[{Hsu \textit{et~al.}(2018)Hsu, Lin, Lee, Min, Tang, and
  Sun}]{hsu2018unified}
W.-T. Hsu, C.-K. Lin, M.-Y. Lee, K.~Min, J.~Tang, and M.~Sun (2018). A Unified
  Model for Extractive and Abstractive Summarization using Inconsistency Loss,
  in {\em Proceedings of the 56th Annual Meeting of the Association for
  Computational Linguistics (Volume 1: Long Papers)\/} {\em 2018\/}.

\bibitem[{Hu \textit{et~al.}(2017)Hu, Yang, Liang, Salakhutdinov, and
  Xing}]{hu2017controllable}
Z.~Hu, Z.~Yang, X.~Liang, R.~Salakhutdinov, and E.~P. Xing (2017). Controllable
  Text Generation, {\em International Conference on Machine Learning\/}.

\bibitem[{Hu \textit{et~al.}(2018)Hu, Yang, Salakhutdinov, and
  Xing}]{hu2017unifying}
Z.~Hu, Z.~Yang, R.~Salakhutdinov, and E.~P. Xing (2018). On Unifying Deep
  Generative Models, {\em International Conference on Learning
  Representation\/}.

\bibitem[{Huang \textit{et~al.}(2018)Huang, Wang, Huang, Zhou, and
  Deng}]{huang2017towards}
P.-S. Huang, C.~Wang, S.~Huang, D.~Zhou, and L.~Deng (2018). Towards neural
  phrase-based machine translation, {\em International Conference on Learning
  Representations(ICLR)\/}.

\bibitem[{Huang \textit{et~al.}(2021)Huang, Shen, Li, Ge, and
  Luo}]{huang2021dependency}
Y.~Huang, X.~Shen, C.~Li, J.~Ge, and B.~Luo (2021). Dependency Learning for
  Legal Judgment Prediction with a Unified Text-to-Text Transformer, {\em arXiv
  preprint arXiv:2112.06370\/}.

\bibitem[{Husz{\'a}r(2015)}]{huszar2015not}
F.~Husz{\'a}r (2015). How (not) to Train your Generative Model: Scheduled
  Sampling, Likelihood, Adversary?, {\em arXiv preprint arXiv:1511.05101\/}.

\bibitem[{Husz{\'a}r(2017)}]{huszar2017variational}
F.~Husz{\'a}r (2017). Variational Inference using Implicit Distributions, {\em
  arXiv preprint arXiv:1702.08235\/}.

\bibitem[{Inan \textit{et~al.}(2017)Inan, Khosravi, and Socher}]{inan2016tying}
H.~Inan, K.~Khosravi, and R.~Socher (2017). Tying word vectors and word
  classifiers: A loss framework for language modeling, {\em International
  Conference on Learning Representations(ICLR)\/}.

\bibitem[{Jacobs \textit{et~al.}(1991)Jacobs, Jordan, Nowlan, and
  Hinton}]{jacobs1991adaptive}
R.~A. Jacobs, M.~I. Jordan, S.~J. Nowlan, and G.~E. Hinton (1991). Adaptive
  mixtures of local experts, {\em Neural computation\/}, vol.~3(1), pp.~79--87.

\bibitem[{Jain and Wallace(2019)}]{jain2019attention}
S.~Jain and B.~C. Wallace (2019). Attention is not Explanation, {\em The North
  American Chapter of the Association for Computational Linguistics\/}.

\bibitem[{Jang \textit{et~al.}(2017)Jang, Gu, and Poole}]{jang2016categorical}
E.~Jang, S.~Gu, and B.~Poole (2017). Categorical reparameterization with
  gumbel-softmax, {\em International Conference on Learning
  Representations(ICLR)\/}.

\bibitem[{Jordan \textit{et~al.}(1999)Jordan, Ghahramani, Jaakkola, and
  Saul}]{jordan1999introduction}
M.~I. Jordan, Z.~Ghahramani, T.~S. Jaakkola, and L.~K. Saul (1999). An
  introduction to variational methods for graphical models, {\em Machine
  learning\/}, vol.~37(2), pp.~183--233.

\bibitem[{Juraska \textit{et~al.}(2018)Juraska, Karagiannis, Bowden, and
  Walker}]{juraska2018deep}
J.~Juraska, P.~Karagiannis, K.~Bowden, and M.~Walker (2018). A Deep Ensemble
  Model with Slot Alignment for Sequence-to-Sequence Natural Language
  Generation, in {\em Proceedings of the 2018 Conference of the North American
  Chapter of the Association for Computational Linguistics: Human Language
  Technologies, Volume 1 (Long Papers)\/} {\em 2018\/}.

\bibitem[{Kaiser \textit{et~al.}(2018)Kaiser, Bengio, Roy, Vaswani, Parmar,
  Uszkoreit, and Shazeer}]{kaiser2018fast}
L.~Kaiser, S.~Bengio, A.~Roy, A.~Vaswani, N.~Parmar, J.~Uszkoreit, and
  N.~Shazeer (2018). Fast Decoding in Sequence Models Using Discrete Latent
  Variables, in {\em Proceedings of the 35th International Conference on
  Machine Learning\/} {\em 2018\/}.

\bibitem[{Kawakami \textit{et~al.}(2019)Kawakami, Dyer, and
  Blunsom}]{kawakami2018unsupervised}
K.~Kawakami, C.~Dyer, and P.~Blunsom (2019). Unsupervised Word Discovery with
  Segmental Neural Language Models, {\em Annual Meeting of the Association for
  Computational Linguistics\/}.

\bibitem[{Ke \textit{et~al.}(2018{\natexlab{a}})Ke, GOYAL, Bilaniuk, Binas,
  Mozer, Pal, and Bengio}]{ke2018sparse}
N.~R. Ke, A.~G. A.~P. GOYAL, O.~Bilaniuk, J.~Binas, M.~C. Mozer, C.~Pal, and
  Y.~Bengio (2018{\natexlab{a}}). Sparse attentive backtracking: Temporal
  credit assignment through reminding, in {\em Advances in Neural Information
  Processing Systems\/} {\em 2018\/}.

\bibitem[{Ke \textit{et~al.}(2018{\natexlab{b}})Ke, {\.Z}o{\l}na, Sordoni, Lin,
  Trischler, Bengio, Pineau, Charlin, and Pal}]{pmlr-v80-ke18a}
N.~R. Ke, K.~{\.Z}o{\l}na, A.~Sordoni, Z.~Lin, A.~Trischler, Y.~Bengio,
  J.~Pineau, L.~Charlin, and C.~Pal (2018{\natexlab{b}}). Focused Hierarchical
  {RNN}s for Conditional Sequence Processing, in {\em Proceedings of the 35th
  International Conference on Machine Learning\/} {\em 2018\/}.

\bibitem[{Keskar \textit{et~al.}(2019)Keskar, McCann, Varshney, Xiong, and
  Socher}]{keskar2019ctrl}
N.~S. Keskar, B.~McCann, L.~R. Varshney, C.~Xiong, and R.~Socher (2019). Ctrl:
  A conditional transformer language model for controllable generation, {\em
  arXiv preprint arXiv:1909.05858\/}.

\bibitem[{Kim and Mooney(2010)}]{kim2010generative}
J.~Kim and R.~J. Mooney (2010). Generative alignment and semantic parsing for
  learning from ambiguous supervision, in {\em Proceedings of the 23rd
  International Conference on Computational Linguistics: Posters\/} {\em
  2010\/}.

\bibitem[{Kim \textit{et~al.}(2017)Kim, Denton, Hoang, and
  Rush}]{kim2017structured}
Y.~Kim, C.~Denton, L.~Hoang, and A.~M. Rush (2017). Structured attention
  networks, {\em International Conference on Learning Representations(ICLR)\/}.

\bibitem[{Kim and Rush(2016)}]{kim2016sequence}
Y.~Kim and A.~M. Rush (2016). Sequence-Level Knowledge Distillation, in {\em
  Proceedings of the 2016 Conference on Empirical Methods in Natural Language
  Processing\/} {\em 2016\/}.

\bibitem[{Kim \textit{et~al.}(2018{\natexlab{a}})Kim, Wiseman, and
  Rush}]{kim2018tutorial}
Y.~Kim, S.~Wiseman, and A.~M. Rush (2018{\natexlab{a}}). A tutorial on deep
  latent variable models of natural language, {\em arXiv preprint
  arXiv:1812.06834\/}.

\bibitem[{Kim \textit{et~al.}(2018{\natexlab{b}})Kim, Zhang, Rush, LeCun
  \textit{et~al.}}]{kim2017adversarially}
Y.~Kim, K.~Zhang, A.~M. Rush, Y.~LeCun, \textit{et~al.} (2018{\natexlab{b}}).
  Adversarially Regularized Autoencoders, {\em International Conference on
  Machine Learning (ICML)\/}.

\bibitem[{Kingma and Ba(2015)}]{kingma2014adam}
D.~Kingma and J.~Ba (2015). Adam: A method for stochastic optimization, {\em
  International Conference on Learning Representations(ICLR)\/}.

\bibitem[{Kingma \textit{et~al.}(2016{\natexlab{a}})Kingma, Salimans,
  Jozefowicz, Chen, Sutskever, and Welling}]{kingma2016improved}
D.~P. Kingma, T.~Salimans, R.~Jozefowicz, X.~Chen, I.~Sutskever, and M.~Welling
  (2016{\natexlab{a}}). Improved variational inference with inverse
  autoregressive flow, in {\em Advances in Neural Information Processing
  Systems\/} {\em 2016\/}.

\bibitem[{Kingma \textit{et~al.}(2016{\natexlab{b}})Kingma, Salimans, and
  Welling}]{kingma2016improving}
D.~P. Kingma, T.~Salimans, and M.~Welling (2016{\natexlab{b}}). Improving
  variational inference with inverse autoregressive flow, {\em Advances in
  Neural Information Processing Systems\/}.

\bibitem[{Kingma and Welling(2014)}]{kingma2014auto}
D.~P. Kingma and M.~Welling (2014). Auto-encoding variational bayes, {\em
  International Conference on Learning Representations(ICLR)\/}.

\bibitem[{Kipf and Welling(2017)}]{kipf2016semi}
T.~N. Kipf and M.~Welling (2017). Semi-supervised classification with graph
  convolutional networks, {\em International Conference on Learning
  Representations(ICLR)\/}.

\bibitem[{Kiyono \textit{et~al.}(2018)Kiyono, Takase, Suzuki, Okazaki, Inui,
  and Nagata}]{kiyono2018unsupervised}
S.~Kiyono, S.~Takase, J.~Suzuki, N.~Okazaki, K.~Inui, and M.~Nagata (2018).
  Unsupervised Token-wise Alignment to Improve Interpretation of
  Encoder-Decoder Models, in {\em Proceedings of the 2018 EMNLP Workshop
  BlackboxNLP: Analyzing and Interpreting Neural Networks for NLP\/} {\em
  2018\/}.

\bibitem[{Klein \textit{et~al.}(2017)Klein, Kim, Deng, Senellart, and
  Rush}]{klein2017opennmt}
G.~Klein, Y.~Kim, Y.~Deng, J.~Senellart, and A.~Rush (2017). OpenNMT:
  Open-Source Toolkit for Neural Machine Translation, in {\em Proceedings of
  ACL 2017, System Demonstrations\/} {\em 2017\/}.

\bibitem[{Koehn(2004)}]{koehn2004statistical}
P.~Koehn (2004). Statistical significance tests for machine translation
  evaluation, in {\em Proceedings of the 2004 conference on empirical methods
  in natural language processing\/} {\em 2004\/}.

\bibitem[{Koehn and Knowles(2017)}]{koehn2017six}
P.~Koehn and R.~Knowles (2017). Six Challenges for Neural Machine Translation,
  in {\em Proceedings of the First Workshop on Neural Machine Translation\/}
  {\em 2017\/}.

\bibitem[{Koehn and Schroeder(2007)}]{koehn2007experiments}
P.~Koehn and J.~Schroeder (2007). Experiments in domain adaptation for
  statistical machine translation, in {\em Proceedings of the second workshop
  on statistical machine translation\/} {\em 2007\/}.

\bibitem[{Koncel-Kedziorski \textit{et~al.}(2014)Koncel-Kedziorski, Hajishirzi,
  and Farhadi}]{koncel2014multi}
R.~Koncel-Kedziorski, H.~Hajishirzi, and A.~Farhadi (2014). Multi-resolution
  language grounding with weak supervision, in {\em Proceedings of the 2014
  Conference on Empirical Methods in Natural Language Processing (EMNLP)\/}
  {\em 2014\/}.

\bibitem[{Kondadadi \textit{et~al.}(2013)Kondadadi, Howald, and
  Schilder}]{kondadadi2013statistical}
R.~Kondadadi, B.~Howald, and F.~Schilder (2013). A statistical nlg framework
  for aggregated planning and realization, in {\em Proceedings of the 51st
  Annual Meeting of the Association for Computational Linguistics (Volume 1:
  Long Papers)\/} {\em 2013\/}.

\bibitem[{Konstas and Lapata(2013)}]{konstas2013global}
I.~Konstas and M.~Lapata (2013). A global model for concept-to-text generation,
  {\em Journal of Artificial Intelligence Research\/}, vol.~48, pp.~305--346.

\bibitem[{Kry{\'s}ci{\'n}ski \textit{et~al.}(2018)Kry{\'s}ci{\'n}ski, Paulus,
  Xiong, and Socher}]{kryscinski2018improving}
W.~Kry{\'s}ci{\'n}ski, R.~Paulus, C.~Xiong, and R.~Socher (2018). Improving
  Abstraction in Text Summarization, in {\em Proceedings of the 2018 Conference
  on Empirical Methods in Natural Language Processing\/} {\em 2018\/}.

\bibitem[{Kuang \textit{et~al.}(2018)Kuang, Li, Branco, Luo, and
  Xiong}]{kuang2018attention}
S.~Kuang, J.~Li, A.~Branco, W.~Luo, and D.~Xiong (2018). Attention Focusing for
  Neural Machine Translation by Bridging Source and Target Embeddings, in {\em
  Proceedings of the 56th Annual Meeting of the Association for Computational
  Linguistics (Volume 1: Long Papers)\/} {\em 2018\/}.

\bibitem[{Kukich(1983)}]{kukich1983design}
K.~Kukich (1983). Design of a knowledge-based report generator, in {\em
  Proceedings of the 21st annual meeting on Association for Computational
  Linguistics\/} {\em 1983\/}.

\bibitem[{Lamb \textit{et~al.}(2016{\natexlab{a}})Lamb, Dumoulin, and
  Courville}]{lamb2016discriminative}
A.~Lamb, V.~Dumoulin, and A.~Courville (2016{\natexlab{a}}). Discriminative
  regularization for generative models, {\em arXiv preprint
  arXiv:1602.03220\/}.

\bibitem[{Lamb \textit{et~al.}(2016{\natexlab{b}})Lamb, GOYAL, Zhang, Zhang,
  Courville, and Bengio}]{lamb2016professor}
A.~M. Lamb, A.~G. A.~P. GOYAL, Y.~Zhang, S.~Zhang, A.~C. Courville, and
  Y.~Bengio (2016{\natexlab{b}}). Professor forcing: A new algorithm for
  training recurrent networks, in {\em Advances In Neural Information
  Processing Systems\/} {\em 2016\/}.

\bibitem[{Lample \textit{et~al.}(2018{\natexlab{a}})Lample, Conneau, Denoyer,
  and Ranzato}]{lample2017unsupervised}
G.~Lample, A.~Conneau, L.~Denoyer, and M.~Ranzato (2018{\natexlab{a}}).
  Unsupervised machine translation using monolingual corpora only, {\em
  International Conference on Learning Representations(ICLR)\/}.

\bibitem[{Lample \textit{et~al.}(2018{\natexlab{b}})Lample, Ott, Conneau,
  Denoyer \textit{et~al.}}]{lample2018phrase}
G.~Lample, M.~Ott, A.~Conneau, L.~Denoyer, \textit{et~al.}
  (2018{\natexlab{b}}). Phrase-Based \& Neural Unsupervised Machine
  Translation, in {\em Proceedings of the 2018 Conference on Empirical Methods
  in Natural Language Processing\/} {\em 2018\/}.

\bibitem[{Larsen \textit{et~al.}(2016)Larsen, S{\o}nderby, Larochelle, and
  Winther}]{larsen2015autoencoding}
A.~B.~L. Larsen, S.~K. S{\o}nderby, H.~Larochelle, and O.~Winther (2016).
  Autoencoding beyond pixels using a learned similarity metric, {\em
  International Conference on Machine Learning (ICML)\/}.

\bibitem[{Lawson \textit{et~al.}(2018)Lawson, Chiu, Tucker, Raffel, Swersky,
  and Jaitly}]{lawson2018learning}
D.~Lawson, C.-C. Chiu, G.~Tucker, C.~Raffel, K.~Swersky, and N.~Jaitly (2018).
  Learning hard alignments with variational inference, in {\em 2018 IEEE
  International Conference on Acoustics, Speech and Signal Processing
  (ICASSP)\/} {\em 2018\/}.

\bibitem[{Lebret \textit{et~al.}(2016)Lebret, Grangier, and
  Auli}]{lebret2016neural}
R.~Lebret, D.~Grangier, and M.~Auli (2016). Neural Text Generation from
  Structured Data with Application to the Biography Domain, in {\em Proceedings
  of the 2016 Conference on Empirical Methods in Natural Language Processing\/}
  {\em 2016\/}.

\bibitem[{Lee \textit{et~al.}(2018)Lee, Mansimov, and
  Cho}]{lee2018deterministic}
J.~Lee, E.~Mansimov, and K.~Cho (2018). Deterministic Non-Autoregressive Neural
  Sequence Modeling by Iterative Refinement, in {\em Proceedings of the 2018
  Conference on Empirical Methods in Natural Language Processing\/} {\em
  2018\/}.

\bibitem[{Lee \textit{et~al.}(2016)Lee, Prakash, Cogswell, Ranjan, Crandall,
  and Batra}]{lee2016stochastic}
S.~Lee, S.~P.~S. Prakash, M.~Cogswell, V.~Ranjan, D.~Crandall, and D.~Batra
  (2016). Stochastic multiple choice learning for training diverse deep
  ensembles, in {\em Advances in Neural Information Processing Systems\/} {\em
  2016\/}.

\bibitem[{Lei \textit{et~al.}(2016)Lei, Barzilay, and
  Jaakkola}]{lei2016rationalizing}
T.~Lei, R.~Barzilay, and T.~Jaakkola (2016). Rationalizing Neural Predictions,
  in {\em Proceedings of the 2016 Conference on Empirical Methods in Natural
  Language Processing\/} {\em 2016\/}.

\bibitem[{Li \textit{et~al.}(2016{\natexlab{a}})Li, Galley, Brockett, Gao, and
  Dolan}]{li2015diversity}
J.~Li, M.~Galley, C.~Brockett, J.~Gao, and B.~Dolan (2016{\natexlab{a}}). A
  Diversity-Promoting Objective Function for Neural Conversation Models, in
  {\em Proceedings of the 2016 Conference of the North American Chapter of the
  Association for Computational Linguistics: Human Language Technologies\/}
  {\em 2016\/}.

\bibitem[{Li \textit{et~al.}(2016{\natexlab{b}})Li, Galley, Brockett,
  Spithourakis, Gao, and Dolan}]{li2016persona}
J.~Li, M.~Galley, C.~Brockett, G.~P. Spithourakis, J.~Gao, and B.~Dolan
  (2016{\natexlab{b}}). A persona-based neural conversation model, {\em arXiv
  preprint arXiv:1603.06155\/}.

\bibitem[{Li and Jurafsky(2017)}]{li2016neural}
J.~Li and D.~Jurafsky (2017). Neural net models for open-domain discourse
  coherence, {\em Conference on Empirical Methods in Natural Language
  Processing\/}.

\bibitem[{Li \textit{et~al.}(2016{\natexlab{c}})Li, Monroe, and
  Jurafsky}]{li2016simple}
J.~Li, W.~Monroe, and D.~Jurafsky (2016{\natexlab{c}}). A Simple, Fast Diverse
  Decoding Algorithm for Neural Generation, {\em CoRR\/}, vol.~abs/1611.08562.

\bibitem[{Li \textit{et~al.}(2016{\natexlab{d}})Li, Monroe, Ritter, Jurafsky,
  Galley, and Gao}]{li2016deep}
J.~Li, W.~Monroe, A.~Ritter, D.~Jurafsky, M.~Galley, and J.~Gao
  (2016{\natexlab{d}}). Deep Reinforcement Learning for Dialogue Generation, in
  {\em Proceedings of the 2016 Conference on Empirical Methods in Natural
  Language Processing\/} {\em 2016\/}.

\bibitem[{Li \textit{et~al.}(2017{\natexlab{a}})Li, Monroe, Shi, Jean, Ritter,
  and Jurafsky}]{li2017adversarial}
J.~Li, W.~Monroe, T.~Shi, S.~Jean, A.~Ritter, and D.~Jurafsky
  (2017{\natexlab{a}}). Adversarial Learning for Neural Dialogue Generation, in
  {\em Proceedings of the 2017 Conference on Empirical Methods in Natural
  Language Processing\/} {\em 2017\/}.

\bibitem[{Li \textit{et~al.}(2019)Li, Zhao, Wen, and Song}]{li2019generating}
J.~Li, W.~X. Zhao, J.-R. Wen, and Y.~Song (2019). Generating Long and
  Informative Reviews with Aspect-Aware Coarse-to-Fine Decoding, in {\em
  Proceedings of the 57th Annual Meeting of the Association for Computational
  Linguistics\/} {\em 2019\/}.

\bibitem[{Li \textit{et~al.}(2017{\natexlab{b}})Li, Lam, Bing, and
  Wang}]{li2017deep}
P.~Li, W.~Lam, L.~Bing, and Z.~Wang (2017{\natexlab{b}}). Deep Recurrent
  Generative Decoder for Abstractive Text Summarization, in {\em Proceedings of
  the 2017 Conference on Empirical Methods in Natural Language Processing\/}
  {\em 2017\/}.

\bibitem[{Li \textit{et~al.}(2018)Li, Xiao, Lyu, and Wang}]{li2018improving}
W.~Li, X.~Xiao, Y.~Lyu, and Y.~Wang (2018). Improving Neural Abstractive
  Document Summarization with Explicit Information Selection Modeling, {\em
  Conference on Empirical Methods in Natural Language Processing\/}.

\bibitem[{Li \textit{et~al.}(2017{\natexlab{c}})Li, Su, Shen, Li, Cao, and
  Niu}]{li2017dailydialog}
Y.~Li, H.~Su, X.~Shen, W.~Li, Z.~Cao, and S.~Niu (2017{\natexlab{c}}).
  DailyDialog: A Manually Labelled Multi-turn Dialogue Dataset, in {\em
  Proceedings of the Eighth International Joint Conference on Natural Language
  Processing (Volume 1: Long Papers)\/} {\em 2017\/}.

\bibitem[{Liang \textit{et~al.}(2009)Liang, Jordan, and
  Klein}]{liang2009learning}
P.~Liang, M.~I. Jordan, and D.~Klein (2009). Learning semantic correspondences
  with less supervision, in {\em Proceedings of the Joint Conference of the
  47th Annual Meeting of the ACL and the 4th International Joint Conference on
  Natural Language Processing of the AFNLP: Volume 1-Volume 1\/} {\em 2009\/}.

\bibitem[{Lin(2004)}]{lin2004rouge}
C.-Y. Lin (2004). Rouge: A package for automatic evaluation of summaries, {\em
  Text Summarization Branches Out\/}, pp. 74--81.

\bibitem[{Lin and Och(2004)}]{lin2004orange}
C.-Y. Lin and F.~J. Och (2004). Orange: a method for evaluating automatic
  evaluation metrics for machine translation, in {\em Proceedings of the 20th
  international conference on Computational Linguistics\/} {\em 2004\/}.

\bibitem[{Lin \textit{et~al.}(2018)Lin, SUN, Ma, and Su}]{lin2018global}
J.~Lin, X.~SUN, S.~Ma, and Q.~Su (2018). Global Encoding for Abstractive
  Summarization, in {\em Proceedings of the 56th Annual Meeting of the
  Association for Computational Linguistics (Volume 2: Short Papers)\/} {\em
  2018\/}.

\bibitem[{Ling and Rush(2017)}]{ling2017coarse}
J.~Ling and A.~Rush (2017). Coarse-to-Fine Attention Models for Document
  Summarization, in {\em Proceedings of the Workshop on New Frontiers in
  Summarization\/} {\em 2017\/}.

\bibitem[{Liu \textit{et~al.}(2016)Liu, Lowe, Serban, Noseworthy, Charlin, and
  Pineau}]{liu2016not}
C.-W. Liu, R.~Lowe, I.~Serban, M.~Noseworthy, L.~Charlin, and J.~Pineau (2016).
  How NOT To Evaluate Your Dialogue System: An Empirical Study of Unsupervised
  Evaluation Metrics for Dialogue Response Generation, in {\em Proceedings of
  the 2016 Conference on Empirical Methods in Natural Language Processing\/}
  {\em 2016\/}.

\bibitem[{Liu \textit{et~al.}(2018)Liu, Wang, Sha, Chang, and
  Sui}]{liu2018table}
T.~Liu, K.~Wang, L.~Sha, B.~Chang, and Z.~Sui (2018). Table-to-text generation
  by structure-aware seq2seq learning, in {\em Thirty-Second AAAI Conference on
  Artificial Intelligence\/} {\em 2018\/}.

\bibitem[{Lu \textit{et~al.}(2017{\natexlab{a}})Lu, Kannan, Yang, Parikh, and
  Batra}]{lu2017best}
J.~Lu, A.~Kannan, J.~Yang, D.~Parikh, and D.~Batra (2017{\natexlab{a}}). Best
  of both worlds: Transferring knowledge from discriminative learning to a
  generative visual dialog model, in {\em Advances in Neural Information
  Processing Systems\/} {\em 2017\/}.

\bibitem[{Lu \textit{et~al.}(2017{\natexlab{b}})Lu, Keung, Zhang, Sun, and
  Bhardwaj}]{DBLP:journals/corr/LuKZSB17}
Y.~Lu, P.~Keung, S.~Zhang, J.~Sun, and V.~Bhardwaj (2017{\natexlab{b}}). A
  practical approach to dialogue response generation in closed domains, {\em
  CoRR\/}, vol.~abs/1703.09439.

\bibitem[{Luan \textit{et~al.}(2017)Luan, Brockett, Dolan, Gao, and
  Galley}]{luan2017multi}
Y.~Luan, C.~Brockett, B.~Dolan, J.~Gao, and M.~Galley (2017). Multi-Task
  Learning for Speaker-Role Adaptation in Neural Conversation Models, in {\em
  Proceedings of the Eighth International Joint Conference on Natural Language
  Processing (Volume 1: Long Papers)\/} {\em 2017\/}.

\bibitem[{Luo \textit{et~al.}(2018)Luo, Price, Cohen, and
  Shakhnarovich}]{luo2018discriminability}
R.~Luo, B.~Price, S.~Cohen, and G.~Shakhnarovich (2018). Discriminability
  objective for training descriptive captions, in {\em Proceedings of the IEEE
  Conference on Computer Vision and Pattern Recognition\/} {\em 2018\/}.

\bibitem[{Luong \textit{et~al.}(2016)Luong, Le, Sutskever, Vinyals, and
  Kaiser}]{luong2015multi}
M.-T. Luong, Q.~V. Le, I.~Sutskever, O.~Vinyals, and L.~Kaiser (2016).
  Multi-task sequence to sequence learning, {\em International Conference on
  Learning Representations(ICLR)\/}.

\bibitem[{Luong \textit{et~al.}(2015)Luong, Pham, and
  Manning}]{luong2015effective}
T.~Luong, H.~Pham, and C.~D. Manning (2015). Effective Approaches to
  Attention-based Neural Machine Translation, in {\em Proceedings of the 2015
  Conference on Empirical Methods in Natural Language Processing\/} {\em
  2015\/}.

\bibitem[{Ma \textit{et~al.}(2017)Ma, Gao, Hu, Yu, Deng, and
  Hovy}]{ma2017dropout}
X.~Ma, Y.~Gao, Z.~Hu, Y.~Yu, Y.~Deng, and E.~Hovy (2017). Dropout with
  expectation-linear regularization, in {\em International Conference on
  Learning Representations\/} {\em 2017\/}.

\bibitem[{Maddison \textit{et~al.}(2017)Maddison, Mnih, and
  Teh}]{maddison2016concrete}
C.~J. Maddison, A.~Mnih, and Y.~W. Teh (2017). The concrete distribution: A
  continuous relaxation of discrete random variables, {\em International
  Conference on Learning Representations(ICLR)\/}.

\bibitem[{Makhzani \textit{et~al.}(2016)Makhzani, Shlens, Jaitly, Goodfellow,
  and Frey}]{makhzani2015adversarial}
A.~Makhzani, J.~Shlens, N.~Jaitly, I.~Goodfellow, and B.~Frey (2016).
  Adversarial autoencoders, {\em International Conference on Learning
  Representations(ICLR)\/}.

\bibitem[{Mao \textit{et~al.}(2015)Mao, Xu, Yang, Wang, Huang, and
  Yuille}]{mao2014deep}
J.~Mao, W.~Xu, Y.~Yang, J.~Wang, Z.~Huang, and A.~Yuille (2015). Deep
  captioning with multimodal recurrent neural networks (m-rnn), {\em
  International Conference on Learning Representations(ICLR)\/}.

\bibitem[{Mao \textit{et~al.}(2016)Mao, Li, Xie, Lau, Wang, and
  Smolley}]{mao2016least}
X.~Mao, Q.~Li, H.~Xie, R.~Y. Lau, Z.~Wang, and S.~P. Smolley (2016). Least
  squares generative adversarial networks, {\em arXiv preprint
  ArXiv:1611.04076\/}.

\bibitem[{Martins and Astudillo(2016)}]{martins2016softmax}
A.~Martins and R.~Astudillo (2016). From softmax to sparsemax: A sparse model
  of attention and multi-label classification, in {\em International Conference
  on Machine Learning\/} {\em 2016\/}.

\bibitem[{May \textit{et~al.}(2019)May, Wang, Bordia, Bowman, and
  Rudinger}]{may2019measuring}
C.~May, A.~Wang, S.~Bordia, S.~Bowman, and R.~Rudinger (2019). On Measuring
  Social Biases in Sentence Encoders, in {\em Proceedings of the 2019
  Conference of the North American Chapter of the Association for Computational
  Linguistics: Human Language Technologies, Volume 1 (Long and Short Papers)\/}
  {\em 2019\/}.

\bibitem[{McCann \textit{et~al.}(2017)McCann, Bradbury, Xiong, and
  Socher}]{mccann2017learned}
B.~McCann, J.~Bradbury, C.~Xiong, and R.~Socher (2017). Learned in translation:
  Contextualized word vectors, in {\em Advances in Neural Information
  Processing Systems\/} {\em 2017\/}.

\bibitem[{McCann \textit{et~al.}(2019)McCann, Keskar, Xiong, and
  Socher}]{mccann2019the}
B.~McCann, N.~S. Keskar, C.~Xiong, and R.~Socher (2019). {\em The Natural
  Language Decathlon: Multitask Learning as Question Answering\/}.

\bibitem[{McKeown(1992)}]{mckeown1992text}
K.~McKeown (1992). {\em Text generation\/}, Cambridge University Press.

\bibitem[{Mei \textit{et~al.}(2016)Mei, UChicago, Bansal, and
  Walter}]{mei2016talk}
H.~Mei, T.~UChicago, M.~Bansal, and M.~R. Walter (2016). What to talk about and
  how? Selective Generation using LSTMs with Coarse-to-Fine Alignment, in {\em
  Proceedings of NAACL-HLT\/} {\em 2016\/}.

\bibitem[{Merity \textit{et~al.}(2017)Merity, Xiong, Bradbury, and
  Socher}]{merity2016pointer}
S.~Merity, C.~Xiong, J.~Bradbury, and R.~Socher (2017). Pointer sentinel
  mixture models, {\em International Conference on Learning
  Representations(ICLR)\/}.

\bibitem[{Mescheder \textit{et~al.}(2017)Mescheder, Nowozin, and
  Geiger}]{mescheder2017adversarial}
L.~Mescheder, S.~Nowozin, and A.~Geiger (2017). Adversarial variational bayes:
  Unifying variational autoencoders and generative adversarial networks, {\em
  International Conference on Machine Learning (ICML)\/}.

\bibitem[{Metz \textit{et~al.}(2017)Metz, Poole, Pfau, and
  Sohl-Dickstein}]{metz2016unrolled}
L.~Metz, B.~Poole, D.~Pfau, and J.~Sohl-Dickstein (2017). Unrolled generative
  adversarial networks, {\em International Conference on Learning
  Representations(ICLR)\/}.

\bibitem[{Miao \textit{et~al.}(2016)Miao, Yu, and Blunsom}]{miao2016neural}
Y.~Miao, L.~Yu, and P.~Blunsom (2016). Neural variational inference for text
  processing, in {\em International Conference on Machine Learning\/} {\em
  2016\/}.

\bibitem[{Mikolov \textit{et~al.}(2013{\natexlab{a}})Mikolov, Chen, Corrado,
  and Dean}]{mikolov2013efficient}
T.~Mikolov, K.~Chen, G.~Corrado, and J.~Dean (2013{\natexlab{a}}). Efficient
  estimation of word representations in vector space, {\em International
  Conference on Learning Representation workshop\/}.

\bibitem[{Mikolov \textit{et~al.}(2013{\natexlab{b}})Mikolov, Sutskever, Chen,
  Corrado, and Dean}]{mikolov2013distributed}
T.~Mikolov, I.~Sutskever, K.~Chen, G.~S. Corrado, and J.~Dean
  (2013{\natexlab{b}}). Distributed representations of words and phrases and
  their compositionality, in {\em Advances in neural information processing
  systems\/} {\em 2013\/}.

\bibitem[{Mnih and Gregor(2014)}]{mnih2014neural}
A.~Mnih and K.~Gregor (2014). Neural Variational Inference and Learning in
  Belief Networks, in {\em International Conference on Machine Learning\/} {\em
  2014\/}.

\bibitem[{Mnih and Rezende(2016)}]{mnih2016variational}
A.~Mnih and D.~J. Rezende (2016). Variational Inference for Monte Carlo
  Objectives, in {\em Proceedings of the 33rd International Conference on
  International Conference on Machine Learning - Volume 48\/} {\em 2016\/}.

\bibitem[{Mogadala \textit{et~al.}(2020)Mogadala, Shen, and
  Klakow}]{mogadala2020integrating}
A.~Mogadala, X.~Shen, and D.~Klakow (2020). Integrating Image Captioning with
  Rule-based Entity Masking, {\em arXiv preprint arXiv:2007.11690\/}.

\bibitem[{Mohamed and Lakshminarayanan(2017)}]{mohamed2016learning}
S.~Mohamed and B.~Lakshminarayanan (2017). Learning in implicit generative
  models, {\em International Conference on Learning Representation workshop\/}.

\bibitem[{Moon(1996)}]{moon1996expectation}
T.~K. Moon (1996). The expectation-maximization algorithm, {\em IEEE Signal
  processing magazine\/}, vol.~13(6), pp.~47--60.

\bibitem[{Moryossef \textit{et~al.}(2019)Moryossef, Goldberg, and
  Dagan}]{moryossef2019step}
A.~Moryossef, Y.~Goldberg, and I.~Dagan (2019). Step-by-Step: Separating
  Planning from Realization in Neural Data-to-Text Generation, in {\em
  Proceedings of the 2019 Conference of the North American Chapter of the
  Association for Computational Linguistics: Human Language Technologies,
  Volume 1 (Long and Short Papers)\/} {\em 2019\/}.

\bibitem[{Nallapati \textit{et~al.}(2017)Nallapati, Zhai, and
  Zhou}]{nallapati2017summarunner}
R.~Nallapati, F.~Zhai, and B.~Zhou (2017). SummaRuNNer: A Recurrent Neural
  Network Based Sequence Model for Extractive Summarization of Documents., in
  {\em Association for the Advancement of Artificial Intelligence\/} {\em
  2017\/}.

\bibitem[{Napoles \textit{et~al.}(2012)Napoles, Gormley, and
  Van~Durme}]{napoles2012annotated}
C.~Napoles, M.~Gormley, and B.~Van~Durme (2012). Annotated gigaword, in {\em
  Proceedings of the Joint Workshop on Automatic Knowledge Base Construction
  and Web-scale Knowledge Extraction\/} {\em 2012\/}.

\bibitem[{Narayan \textit{et~al.}(2018)Narayan, Cohen, and
  Lapata}]{narayan2018don}
S.~Narayan, S.~B. Cohen, and M.~Lapata (2018). Don't Give Me the Details, Just
  the Summary! Topic-Aware Convolutional Neural Networks for Extreme
  Summarization, in {\em Proceedings of the 2018 Conference on Empirical
  Methods in Natural Language Processing\/} {\em 2018\/}.

\bibitem[{Neal and Hinton(1998)}]{neal1998view}
R.~M. Neal and G.~E. Hinton (1998). A view of the EM algorithm that justifies
  incremental, sparse, and other variants, in {\em Learning in graphical
  models\/} {\em 1998\/}, pp. 355--368, Springer.

\bibitem[{Nema \textit{et~al.}(2017)Nema, Khapra, Laha, and
  Ravindran}]{nema2017diversity}
P.~Nema, M.~M. Khapra, A.~Laha, and B.~Ravindran (2017). Diversity driven
  attention model for query-based abstractive summarization, in {\em
  Proceedings of the 55th Annual Meeting of the Association for Computational
  Linguistics (Volume 1: Long Papers)\/} {\em 2017\/}.

\bibitem[{Nguyen and Tran(2018)}]{nguyen2018structurebased}
D.~T. Nguyen and T.~Tran (2018). Structurebased Generation System for E2E NLG
  Challenge, {\em arxiv\/}.

\bibitem[{Nguyen and Chiang(2018)}]{nguyen2018improving}
T.~Nguyen and D.~Chiang (2018). Improving Lexical Choice in Neural Machine
  Translation, in {\em Proceedings of the 2018 Conference of the North American
  Chapter of the Association for Computational Linguistics: Human Language
  Technologies, Volume 1 (Long Papers)\/} {\em 2018\/}.

\bibitem[{Niu and Bansal(2018)}]{niu2018polite}
T.~Niu and M.~Bansal (2018). Polite dialogue generation without parallel data,
  {\em Transactions of the Association for Computational Linguistics\/},
  vol.~6, pp.~373--389.

\bibitem[{Novikova \textit{et~al.}(2017{\natexlab{a}})Novikova, Du{\v{s}}ek,
  Curry, and Rieser}]{novikova2017we}
J.~Novikova, O.~Du{\v{s}}ek, A.~C. Curry, and V.~Rieser (2017{\natexlab{a}}).
  Why We Need New Evaluation Metrics for NLG, in {\em Proceedings of the 2017
  Conference on Empirical Methods in Natural Language Processing\/} {\em
  2017\/}.

\bibitem[{Novikova \textit{et~al.}(2017{\natexlab{b}})Novikova, Du{\v{s}}ek,
  and Rieser}]{novikova2017e2e}
J.~Novikova, O.~Du{\v{s}}ek, and V.~Rieser (2017{\natexlab{b}}). The E2E
  Dataset: New Challenges For End-to-End Generation, in {\em Proceedings of the
  18th Annual SIGdial Meeting on Discourse and Dialogue\/} {\em 2017\/}.

\bibitem[{Nowozin \textit{et~al.}(2016)Nowozin, Cseke, and
  Tomioka}]{nowozin2016f}
S.~Nowozin, B.~Cseke, and R.~Tomioka (2016). f-GAN: Training generative neural
  samplers using variational divergence minimization, in {\em Advances in
  Neural Information Processing Systems\/} {\em 2016\/}.

\bibitem[{Och and Ney(2000)}]{och2000comparison}
F.~J. Och and H.~Ney (2000). A comparison of alignment models for statistical
  machine translation, in {\em COLING 2000 Volume 2: The 18th International
  Conference on Computational Linguistics\/} {\em 2000\/}.

\bibitem[{Och \textit{et~al.}(1999)Och, Tillmann, and Ney}]{och1999improved}
F.~J. Och, C.~Tillmann, and H.~Ney (1999). Improved alignment models for
  statistical machine translation, in {\em 1999 Joint SIGDAT Conference on
  Empirical Methods in Natural Language Processing and Very Large Corpora\/}
  {\em 1999\/}.

\bibitem[{Oord \textit{et~al.}(2016)Oord, Kalchbrenner, and
  Kavukcuoglu}]{oord2016pixel}
A.~v.~d. Oord, N.~Kalchbrenner, and K.~Kavukcuoglu (2016). Pixel recurrent
  neural networks, {\em arXiv preprint arXiv:1601.06759\/}.

\bibitem[{Over \textit{et~al.}(2007)Over, Dang, and Harman}]{over2007duc}
P.~Over, H.~Dang, and D.~Harman (2007). DUC in context, {\em Information
  Processing \& Management\/}, vol.~43(6), pp.~1506--1520.

\bibitem[{Oya \textit{et~al.}(2014)Oya, Mehdad, Carenini, and
  Ng}]{oya2014template}
T.~Oya, Y.~Mehdad, G.~Carenini, and R.~Ng (2014). A template-based abstractive
  meeting summarization: Leveraging summary and source text relationships, in
  {\em Proceedings of the 8th International Natural Language Generation
  Conference (INLG)\/} {\em 2014\/}.

\bibitem[{Paisley \textit{et~al.}(2012)Paisley, Blei, and
  Jordan}]{paisley2012variational}
J.~Paisley, D.~Blei, and M.~Jordan (2012). Variational Bayesian inference with
  stochastic search, {\em International Conference on Machine Learning
  (ICML)\/}.

\bibitem[{Papineni \textit{et~al.}(2002)Papineni, Roukos, Ward, and
  Zhu}]{papineni2002bleu}
K.~Papineni, S.~Roukos, T.~Ward, and W.-J. Zhu (2002). BLEU: a method for
  automatic evaluation of machine translation, in {\em Proceedings of the 40th
  annual meeting on association for computational linguistics\/} {\em 2002\/}.

\bibitem[{Parshakova \textit{et~al.}(2019)Parshakova, Andreoli, and
  Dymetman}]{parshakova2019distributional}
T.~Parshakova, J.-M. Andreoli, and M.~Dymetman (2019). Distributional
  Reinforcement Learning for Energy-Based Sequential Models, {\em arXiv
  preprint arXiv:1912.08517\/}.

\bibitem[{Paszke \textit{et~al.}(2017)Paszke, Gross, Chintala, Chanan, Yang,
  DeVito, Lin, Desmaison, Antiga, and Lerer}]{paszke2017automatic}
A.~Paszke, S.~Gross, S.~Chintala, G.~Chanan, E.~Yang, Z.~DeVito, Z.~Lin,
  A.~Desmaison, L.~Antiga, and A.~Lerer (2017). Automatic differentiation in
  PyTorch, in {\em Advances in Neural Information Processing Systems
  workshop\/} {\em 2017\/}.

\bibitem[{Paszke \textit{et~al.}(2019)Paszke, Gross, Massa, Lerer, Bradbury,
  Chanan, Killeen, Lin, Gimelshein, Antiga \textit{et~al.}}]{pytorch}
A.~Paszke, S.~Gross, F.~Massa, A.~Lerer, J.~Bradbury, G.~Chanan, T.~Killeen,
  Z.~Lin, N.~Gimelshein, L.~Antiga, \textit{et~al.} (2019). PyTorch: An
  imperative style, high-performance deep learning library, in {\em Advances in
  Neural Information Processing Systems\/} {\em 2019\/}.

\bibitem[{Paulus \textit{et~al.}(2018)Paulus, Xiong, and
  Socher}]{paulus2017deep}
R.~Paulus, C.~Xiong, and R.~Socher (2018). A deep reinforced model for
  abstractive summarization, {\em International Conference on Learning
  Representations(ICLR)\/}.

\bibitem[{Pennington \textit{et~al.}(2014)Pennington, Socher, and
  Manning}]{pennington2014glove}
J.~Pennington, R.~Socher, and C.~D. Manning (2014). GloVe: Global Vectors for
  Word Representation, in {\em Empirical Methods in Natural Language Processing
  (EMNLP)\/} {\em 2014\/}.

\bibitem[{Peters \textit{et~al.}(2018)Peters, Neumann, Iyyer, Gardner, Clark,
  Lee, and Zettlemoyer}]{peters2018deep}
M.~Peters, M.~Neumann, M.~Iyyer, M.~Gardner, C.~Clark, K.~Lee, and
  L.~Zettlemoyer (2018). Deep Contextualized Word Representations, in {\em
  Proceedings of the 2018 Conference of the North American Chapter of the
  Association for Computational Linguistics: Human Language Technologies,
  Volume 1 (Long Papers)\/} {\em 2018\/}.

\bibitem[{Press \textit{et~al.}(2017)Press, Bar, Bogin, Berant, and
  Wolf}]{press2017language}
O.~Press, A.~Bar, B.~Bogin, J.~Berant, and L.~Wolf (2017). Language Generation
  with Recurrent Generative Adversarial Networks without Pre-training, {\em
  arXiv preprint arXiv:1706.01399\/}.

\bibitem[{Press and Wolf(2017)}]{press2017using}
O.~Press and L.~Wolf (2017). Using the Output Embedding to Improve Language
  Models, in {\em Proceedings of the 15th Conference of the European Chapter of
  the Association for Computational Linguistics: Volume 2, Short Papers\/} {\em
  2017\/}.

\bibitem[{Pu \textit{et~al.}(2017)Pu, Chen, Dai, Wang, Li, and
  Carin}]{pu2017symmetric}
Y.~Pu, L.~Chen, S.~Dai, W.~Wang, C.~Li, and L.~Carin (2017). Symmetric
  Variational Autoencoder and Connections to Adversarial Learning, {\em
  Advances in Neural Information Processing Systems\/}.

\bibitem[{Puduppully \textit{et~al.}(2019)Puduppully, Dong, and
  Lapata}]{puduppully2019data}
R.~Puduppully, L.~Dong, and M.~Lapata (2019). Data-to-text generation with
  content selection and planning, in {\em Proceedings of the AAAI Conference on
  Artificial Intelligence\/} {\em 2019\/}.

\bibitem[{Puzikov and Gurevych(2018)}]{puzikov2018e2e}
Y.~Puzikov and I.~Gurevych (2018). E2e nlg challenge: Neural models vs.
  templates, in {\em Proceedings of the 11th International Conference on
  Natural Language Generation\/} {\em 2018\/}.

\bibitem[{Qin \textit{et~al.}(2018)Qin, Yao, Wang, Wang, and
  Lin}]{qin2018learning}
G.~Qin, J.-G. Yao, X.~Wang, J.~Wang, and C.-Y. Lin (2018). Learning Latent
  Semantic Annotations for Grounding Natural Language to Structured Data, in
  {\em Proceedings of the 2018 Conference on Empirical Methods in Natural
  Language Processing\/} {\em 2018\/}.

\bibitem[{Qiu \textit{et~al.}(2020)Qiu, Xu, Zhang, Wang, Shen, De~Melo, Long,
  and Li}]{qiu2020easyaug}
S.~Qiu, B.~Xu, J.~Zhang, Y.~Wang, X.~Shen, G.~De~Melo, C.~Long, and X.~Li
  (2020). Easyaug: An automatic textual data augmentation platform for
  classification tasks, in {\em Companion Proceedings of the Web Conference
  2020\/} {\em 2020\/}.

\bibitem[{Quandt(1972)}]{quandt1972new}
R.~E. Quandt (1972). A new approach to estimating switching regressions, {\em
  Journal of the American statistical association\/}, vol.~67(338),
  pp.~306--310.

\bibitem[{Rabiner(1989)}]{rabiner1989tutorial}
L.~R. Rabiner (1989). A tutorial on hidden Markov models and selected
  applications in speech recognition, {\em Proceedings of the IEEE\/},
  vol.~77(2), pp.~257--286.

\bibitem[{Radford \textit{et~al.}(2019)Radford, Wu, Child, Luan, Amodei, and
  Sutskever}]{radford2019language}
A.~Radford, J.~Wu, R.~Child, D.~Luan, D.~Amodei, and I.~Sutskever (2019).
  Language models are unsupervised multitask learners, {\em arXiv\/}.

\bibitem[{Raffel \textit{et~al.}(2019)Raffel, Shazeer, Roberts, Lee, Narang,
  Matena, Zhou, Li, and Liu}]{raffel2019exploring}
C.~Raffel, N.~Shazeer, A.~Roberts, K.~Lee, S.~Narang, M.~Matena, Y.~Zhou,
  W.~Li, and P.~J. Liu (2019). Exploring the limits of transfer learning with a
  unified text-to-text transformer, {\em arXiv preprint arXiv:1910.10683\/}.

\bibitem[{Raiko \textit{et~al.}(2015)Raiko, Berglund, Alain, and
  Dinh}]{raiko2014techniques}
T.~Raiko, M.~Berglund, G.~Alain, and L.~Dinh (2015). Techniques for learning
  binary stochastic feedforward neural networks, {\em International Conference
  on Learning Representations(ICLR)\/}.

\bibitem[{Reed \textit{et~al.}(2018)Reed, Oraby, and Walker}]{reed2018can}
L.~Reed, S.~Oraby, and M.~Walker (2018). Can Neural Generators for Dialogue
  Learn Sentence Planning and Discourse Structuring?, in {\em Proceedings of
  the 11th International Conference on Natural Language Generation\/} {\em
  2018\/}.

\bibitem[{Reiter and Belz(2009)}]{reiter2009investigation}
E.~Reiter and A.~Belz (2009). An investigation into the validity of some
  metrics for automatically evaluating natural language generation systems,
  {\em Computational Linguistics\/}, vol.~35(4), pp.~529--558.

\bibitem[{Reiter and Dale(1997)}]{reiter1997building}
E.~Reiter and R.~Dale (1997). Building applied natural language generation
  systems, {\em Natural Language Engineering\/}, vol.~3(1), pp.~57--87.

\bibitem[{Reiter and Dale(2000)}]{reiter2000building}
E.~Reiter and R.~Dale (2000). {\em Building natural language generation
  systems\/}, Cambridge university press.

\bibitem[{Reiter \textit{et~al.}(1995)Reiter, Mellish, and
  Levine}]{reiter1995automatic}
E.~Reiter, C.~Mellish, and J.~Levine (1995). Automatic generation of technical
  documentation, {\em Applied Artificial Intelligence an International
  Journal\/}, vol.~9(3), pp.~259--287.

\bibitem[{Rezende and Mohamed(2015)}]{rezende2015variational}
D.~J. Rezende and S.~Mohamed (2015). Variational inference with normalizing
  flows, {\em arXiv preprint arXiv:1505.05770\/}.

\bibitem[{Rezende \textit{et~al.}(2014)Rezende, Mohamed, and
  Wierstra}]{rezende2014stochastic}
D.~J. Rezende, S.~Mohamed, and D.~Wierstra (2014). Stochastic backpropagation
  and approximate inference in deep generative models, {\em arXiv preprint
  arXiv:1401.4082\/}.

\bibitem[{Riezler and Maxwell(2005)}]{riezler2005some}
S.~Riezler and J.~T. Maxwell (2005). On some pitfalls in automatic evaluation
  and significance testing for MT, in {\em Proceedings of the ACL workshop on
  intrinsic and extrinsic evaluation measures for machine translation and/or
  summarization\/} {\em 2005\/}.

\bibitem[{Ritter \textit{et~al.}(2011)Ritter, Cherry, and
  Dolan}]{ritter2011data}
A.~Ritter, C.~Cherry, and W.~B. Dolan (2011). Data-driven response generation
  in social media, in {\em Proceedings of the conference on empirical methods
  in natural language processing\/} {\em 2011\/}.

\bibitem[{Rosca \textit{et~al.}(2017)Rosca, Lakshminarayanan, Warde-Farley, and
  Mohamed}]{rosca2017variational}
M.~Rosca, B.~Lakshminarayanan, D.~Warde-Farley, and S.~Mohamed (2017).
  Variational Approaches for Auto-Encoding Generative Adversarial Networks,
  {\em arXiv preprint arXiv:1706.04987\/}.

\bibitem[{Roth \textit{et~al.}(2017)Roth, Lucchi, Nowozin, and
  Hofmann}]{roth2017stabilizing}
K.~Roth, A.~Lucchi, S.~Nowozin, and T.~Hofmann (2017). Stabilizing Training of
  Generative Adversarial Networks through Regularization, {\em arXiv preprint
  arXiv:1705.09367\/}.

\bibitem[{Rus and Lintean(2012)}]{rus2012comparison}
V.~Rus and M.~Lintean (2012). A comparison of greedy and optimal assessment of
  natural language student input using word-to-word similarity metrics, in {\em
  Proceedings of the Seventh Workshop on Building Educational Applications
  Using NLP\/} {\em 2012\/}.

\bibitem[{Rush \textit{et~al.}(2015)Rush, Chopra, and Weston}]{rush2015neural}
A.~M. Rush, S.~Chopra, and J.~Weston (2015). A Neural Attention Model for
  Abstractive Sentence Summarization, in {\em Proceedings of the 2015
  Conference on Empirical Methods in Natural Language Processing\/} {\em
  2015\/}.

\bibitem[{Salakhutdinov \textit{et~al.}(2003)Salakhutdinov, Roweis, and
  Ghahramani}]{salakhutdinov2003optimization}
R.~Salakhutdinov, S.~T. Roweis, and Z.~Ghahramani (2003). Optimization with EM
  and expectation-conjugate-gradient, in {\em Proceedings of the 20th
  International Conference on Machine Learning (ICML-03)\/} {\em 2003\/}.

\bibitem[{Salimans \textit{et~al.}(2016)Salimans, Goodfellow, Zaremba, Cheung,
  Radford, and Chen}]{salimans2016improved}
T.~Salimans, I.~Goodfellow, W.~Zaremba, V.~Cheung, A.~Radford, and X.~Chen
  (2016). Improved techniques for training gans, in {\em Advances in Neural
  Information Processing Systems\/} {\em 2016\/}.

\bibitem[{Salimans \textit{et~al.}(2015)Salimans, Kingma, and
  Welling}]{salimans2015markov}
T.~Salimans, D.~Kingma, and M.~Welling (2015). Markov chain monte carlo and
  variational inference: Bridging the gap, in {\em Proceedings of the 32nd
  International Conference on Machine Learning (ICML-15)\/} {\em 2015\/}.

\bibitem[{Schluter(2017)}]{schluter2017limits}
N.~Schluter (2017). The limits of automatic summarisation according to rouge,
  in {\em Proceedings of the 15th Conference of the European Chapter of the
  Association for Computational Linguistics: Volume 2, Short Papers\/} {\em
  2017\/}.

\bibitem[{Schmidt \textit{et~al.}(2019)Schmidt, Mandt, and
  Hofmann}]{schmidt2019autoregressive}
F.~Schmidt, S.~Mandt, and T.~Hofmann (2019). Autoregressive Text Generation
  Beyond Feedback Loops, {\em arXiv preprint arXiv:1908.11658\/}.

\bibitem[{See \textit{et~al.}(2017)See, Liu, and Manning}]{see2017get}
A.~See, P.~J. Liu, and C.~D. Manning (2017). Get To The Point: Summarization
  with Pointer-Generator Networks, in {\em Proceedings of the 55th Annual
  Meeting of the Association for Computational Linguistics (Volume 1: Long
  Papers)\/} {\em 2017\/}.

\bibitem[{Semeniuta \textit{et~al.}(2017)Semeniuta, Severyn, and
  Barth}]{semeniuta2017hybrid}
S.~Semeniuta, A.~Severyn, and E.~Barth (2017). A Hybrid Convolutional
  Variational Autoencoder for Text Generation, {\em arXiv preprint
  arXiv:1702.02390\/}.

\bibitem[{Sennrich \textit{et~al.}(2016{\natexlab{a}})Sennrich, Haddow, and
  Birch}]{sennrich2016improving}
R.~Sennrich, B.~Haddow, and A.~Birch (2016{\natexlab{a}}). Improving Neural
  Machine Translation Models with Monolingual Data, in {\em Proceedings of the
  54th Annual Meeting of the Association for Computational Linguistics (Volume
  1: Long Papers)\/} {\em 2016\/}.

\bibitem[{Sennrich \textit{et~al.}(2016{\natexlab{b}})Sennrich, Haddow, and
  Birch}]{sennrich2016neural}
R.~Sennrich, B.~Haddow, and A.~Birch (2016{\natexlab{b}}). Neural Machine
  Translation of Rare Words with Subword Units, in {\em Proceedings of the 54th
  Annual Meeting of the Association for Computational Linguistics (Volume 1:
  Long Papers)\/} {\em 2016\/}.

\bibitem[{Serban \textit{et~al.}(2017{\natexlab{a}})Serban, II, Pineau, and
  Courville}]{serbanpiecewise}
I.~V. Serban, A.~G.~O. II, J.~Pineau, and A.~Courville (2017{\natexlab{a}}).
  Piecewise Latent Variables for Neural Variational Text Processing, {\em
  Conference on Empirical Methods in Natural Language Processing\/}.

\bibitem[{Serban \textit{et~al.}(2017{\natexlab{b}})Serban, Sankar, Germain,
  Zhang, Lin, Subramanian, Kim, Pieper, Chandar, Ke
  \textit{et~al.}}]{serban2017deep}
I.~V. Serban, C.~Sankar, M.~Germain, S.~Zhang, Z.~Lin, S.~Subramanian, T.~Kim,
  M.~Pieper, S.~Chandar, N.~R. Ke, \textit{et~al.} (2017{\natexlab{b}}). A deep
  reinforcement learning chatbot, {\em arXiv preprint arXiv:1709.02349\/}.

\bibitem[{Serban \textit{et~al.}(2016)Serban, Sordoni, Bengio, Courville, and
  Pineau}]{serban2015building}
I.~V. Serban, A.~Sordoni, Y.~Bengio, A.~Courville, and J.~Pineau (2016).
  Building end-to-end dialogue systems using generative hierarchical neural
  network models, in {\em Proceedings of the Thirtieth AAAI Conference on
  Artificial Intelligence\/} {\em 2016\/}.

\bibitem[{Serban \textit{et~al.}(2017{\natexlab{c}})Serban, Sordoni, Lowe,
  Charlin, Pineau, Courville, and Bengio}]{serban2016hierarchical}
I.~V. Serban, A.~Sordoni, R.~Lowe, L.~Charlin, J.~Pineau, A.~Courville, and
  Y.~Bengio (2017{\natexlab{c}}). A Hierarchical Latent Variable
  Encoder-Decoder Model for Generating Dialogues, in {\em Thirty-First AAAI
  Conference on Artificial Intelligence\/} {\em 2017\/}.

\bibitem[{Serban \textit{et~al.}(2017{\natexlab{d}})Serban, Sordoni, Lowe,
  Charlin, Pineau, Courville, and Bengio}]{serban2017hierarchical}
I.~V. Serban, A.~Sordoni, R.~Lowe, L.~Charlin, J.~Pineau, A.~C. Courville, and
  Y.~Bengio (2017{\natexlab{d}}). A Hierarchical Latent Variable
  Encoder-Decoder Model for Generating Dialogues., in {\em Association for the
  Advancement of Artificial Intelligence\/} {\em 2017\/}.

\bibitem[{Shang \textit{et~al.}(2015)Shang, Lu, and Li}]{shang2015neural}
L.~Shang, Z.~Lu, and H.~Li (2015). Neural responding machine for short-text
  conversation, {\em arXiv preprint arXiv:1503.02364\/}.

\bibitem[{Shankar \textit{et~al.}(2018)Shankar, Garg, and
  Sarawagi}]{shankar2018surprisingly}
S.~Shankar, S.~Garg, and S.~Sarawagi (2018). Surprisingly easy hard-attention
  for sequence to sequence learning, in {\em Proceedings of the 2018 Conference
  on Empirical Methods in Natural Language Processing\/} {\em 2018\/}.

\bibitem[{Shankar and Sarawagi(2019)}]{shankar2018posterior}
S.~Shankar and S.~Sarawagi (2019). Posterior Attention Models for Sequence to
  Sequence Learning, in {\em International Conference on Learning
  Representations\/} {\em 2019\/}.

\bibitem[{Shen \textit{et~al.}(2017{\natexlab{a}})Shen, Zhang, Henao, Su, and
  Carin}]{shen2017deconvolutional}
D.~Shen, Y.~Zhang, R.~Henao, Q.~Su, and L.~Carin (2017{\natexlab{a}}).
  Deconvolutional Latent-Variable Model for Text Sequence Matching, {\em arXiv
  preprint arXiv:1709.07109\/}.

\bibitem[{Shen \textit{et~al.}(2017{\natexlab{b}})Shen, Lei, Barzilay, and
  Jaakkola}]{shen2017style}
T.~Shen, T.~Lei, R.~Barzilay, and T.~Jaakkola (2017{\natexlab{b}}). Style
  transfer from non-parallel text by cross-alignment, in {\em Advances in
  neural information processing systems\/} {\em 2017\/}.

\bibitem[{Shen \textit{et~al.}(2019{\natexlab{a}})Shen, Ott, Auli, and
  Ranzato}]{shen2019mixture}
T.~Shen, M.~Ott, M.~Auli, and M.~Ranzato (2019{\natexlab{a}}). Mixture Models
  for Diverse Machine Translation: Tricks of the Trade, in {\em International
  Conference on Machine Learning\/} {\em 2019\/}.

\bibitem[{Shen \textit{et~al.}(2018{\natexlab{a}})Shen, Zhou, Long, Jiang,
  Wang, and Zhang}]{shen2018reinforced}
T.~Shen, T.~Zhou, G.~Long, J.~Jiang, S.~Wang, and C.~Zhang
  (2018{\natexlab{a}}). Reinforced Self-Attention Network: a Hybrid of Hard and
  Soft Attention for Sequence Modeling, in {\em Proceedings of the
  Twenty-Seventh International Joint Conference on Artificial Intelligence,
  {IJCAI-18}\/} {\em 2018\/}.

\bibitem[{Shen(2021)}]{shen2021deep}
X.~Shen (2021). Deep latent-variable models for neural text generation.

\bibitem[{Shen \textit{et~al.}(2020)Shen, Chang, Su, Niu, and
  Klakow}]{shen2020neural}
X.~Shen, E.~Chang, H.~Su, C.~Niu, and D.~Klakow (2020). Neural Data-to-Text
  Generation via Jointly Learning the Segmentation and Correspondence, in {\em
  Proceedings of the 58th Annual Meeting of the Association for Computational
  Linguistics\/} {\em 2020\/}.

\bibitem[{Shen \textit{et~al.}(2017{\natexlab{c}})Shen, Oualil, Greenberg,
  Singh, and Klakow}]{shen2017estimation}
X.~Shen, Y.~Oualil, C.~Greenberg, M.~Singh, and D.~Klakow (2017{\natexlab{c}}).
  Estimation of Gap Between Current Language Models and Human Performance, {\em
  Proc. Interspeech 2017\/}, pp. 553--557.

\bibitem[{Shen and Su(2018)}]{shen2018towards}
X.~Shen and H.~Su (2018). Towards better variational encoder-decoders in
  seq2seq tasks, in {\em Thirty-Second AAAI Conference on Artificial
  Intelligence\/} {\em 2018\/}.

\bibitem[{Shen \textit{et~al.}(2018{\natexlab{b}})Shen, Su, Li, and
  Klakow}]{shen2018nexus}
X.~Shen, H.~Su, W.~Li, and D.~Klakow (2018{\natexlab{b}}). Nexus network:
  Connecting the preceding and the following in dialogue generation, in {\em
  Proceedings of the 2018 Conference on Empirical Methods in Natural Language
  Processing\/} {\em 2018\/}.

\bibitem[{Shen \textit{et~al.}(2017{\natexlab{d}})Shen, Su, Li, Li, Niu, Zhao,
  Aizawa, and Long}]{shen2017conditional}
X.~Shen, H.~Su, Y.~Li, W.~Li, S.~Niu, Y.~Zhao, A.~Aizawa, and G.~Long
  (2017{\natexlab{d}}). A Conditional Variational Framework for Dialog
  Generation, in {\em Proceedings of the 55th Annual Meeting of the Association
  for Computational Linguistics (Volume 2: Short Papers)\/} {\em 2017\/}.

\bibitem[{Shen \textit{et~al.}(2018{\natexlab{c}})Shen, Su, Niu, and
  Demberg}]{shen2018improving}
X.~Shen, H.~Su, S.~Niu, and V.~Demberg (2018{\natexlab{c}}). Improving
  Variational Encoder-Decoders in Dialogue Generation, {\em Association for the
  Advancement of Artificial Intelligence\/}, pp. 5456--5463.

\bibitem[{Shen \textit{et~al.}(2017{\natexlab{e}})Shen, Su, Niu, and
  Klakow}]{shen2017wake}
X.~Shen, H.~Su, S.~Niu, and D.~Klakow (2017{\natexlab{e}}). Wake-Sleep
  Variational Autoencoders for Language Modeling, in {\em International
  Conference on Neural Information Processing\/} {\em 2017\/}.

\bibitem[{Shen \textit{et~al.}(2019{\natexlab{b}})Shen, Suzuki, Inui, Su,
  Klakow, and Sekine}]{shen2019select}
X.~Shen, J.~Suzuki, K.~Inui, H.~Su, D.~Klakow, and S.~Sekine
  (2019{\natexlab{b}}). Select and Attend: Towards Controllable Content
  Selection in Text Generation, {\em arXiv preprint arXiv:1909.04453\/}.

\bibitem[{Shen \textit{et~al.}(2019{\natexlab{c}})Shen, Zhao, Su, and
  Klakow}]{shen2019improving}
X.~Shen, Y.~Zhao, H.~Su, and D.~Klakow (2019{\natexlab{c}}). Improving Latent
  Alignment in Text Summarization by Generalizing the Pointer Generator, in
  {\em Proceedings of the 2019 Conference on Empirical Methods in Natural
  Language Processing and the 9th International Joint Conference on Natural
  Language Processing (EMNLP-IJCNLP)\/} {\em 2019\/}.

\bibitem[{Shetty \textit{et~al.}(2017)Shetty, Rohrbach, Hendricks, Fritz, and
  Schiele}]{shetty2017speaking}
R.~Shetty, M.~Rohrbach, L.~A. Hendricks, M.~Fritz, and B.~Schiele (2017).
  Speaking the same language: Matching machine to human captions by adversarial
  training, in {\em Proceedings of the IEEE International Conference on
  Computer Vision (ICCV)\/} {\em 2017\/}.

\bibitem[{Sohn \textit{et~al.}(2015)Sohn, Lee, and Yan}]{sohn2015learning}
K.~Sohn, H.~Lee, and X.~Yan (2015). Learning structured output representation
  using deep conditional generative models, in {\em Advances in Neural
  Information Processing Systems\/} {\em 2015\/}.

\bibitem[{S{\o}nderby \textit{et~al.}(2016)S{\o}nderby, Caballero, Theis, Shi,
  and Husz{\'a}r}]{sonderby2016amortised}
C.~K. S{\o}nderby, J.~Caballero, L.~Theis, W.~Shi, and F.~Husz{\'a}r (2016).
  Amortised map inference for image super-resolution, {\em arXiv preprint
  arXiv:1610.04490\/}.

\bibitem[{Sordoni \textit{et~al.}(2015)Sordoni, Galley, Auli, Brockett, Ji,
  Mitchell, Nie, Gao, and Dolan}]{sordoni2015neural}
A.~Sordoni, M.~Galley, M.~Auli, C.~Brockett, Y.~Ji, M.~Mitchell, J.-Y. Nie,
  J.~Gao, and B.~Dolan (2015). A Neural Network Approach to Context-Sensitive
  Generation of Conversational Responses, in {\em Proceedings of the 2015
  Conference of the North American Chapter of the Association for Computational
  Linguistics: Human Language Technologies\/} {\em 2015\/}.

\bibitem[{Sriram \textit{et~al.}(2018)Sriram, Jun, Satheesh, and
  Coates}]{sriram2018cold}
A.~Sriram, H.~Jun, S.~Satheesh, and A.~Coates (2018). Cold Fusion: Training
  Seq2Seq Models Together with Language Models, {\em Proc. Interspeech 2018\/},
  pp. 387--391.

\bibitem[{Srivastava and Sutton(2017)}]{srivastava2017autoencoding}
A.~Srivastava and C.~Sutton (2017). Autoencoding Variational Inference For
  Topic Models, {\em International Conference on Learning
  Representations(ICLR)\/}.

\bibitem[{Srivastava \textit{et~al.}(2017)Srivastava, Valkov, Russell, Gutmann,
  and Sutton}]{srivastava2017veegan}
A.~Srivastava, L.~Valkov, C.~Russell, M.~Gutmann, and C.~Sutton (2017). VEEGAN:
  Reducing Mode Collapse in GANs using Implicit Variational Learning, {\em
  Advances in Neural Information Processing Systems\/}.

\bibitem[{Su \textit{et~al.}(2019{\natexlab{a}})Su, Hsu, Tuan, and
  Lee}]{su2019personalized}
F.-G. Su, A.~R. Hsu, Y.-L. Tuan, and H.-Y. Lee (2019{\natexlab{a}}).
  Personalized Dialogue Response Generation Learned from Monologues, {\em Proc.
  Interspeech 2019\/}, pp. 4160--4164.

\bibitem[{Su \textit{et~al.}(2018)Su, Shen, Hu, Li, and Chen}]{su2018dialogue}
H.~Su, X.~Shen, P.~Hu, W.~Li, and Y.~Chen (2018). Dialogue generation with GAN,
  in {\em Thirty-Second AAAI Conference on Artificial Intelligence\/} {\em
  2018\/}.

\bibitem[{Su \textit{et~al.}(2020{\natexlab{a}})Su, Shen, Xiao, Zhang, Chang,
  Zhang, Niu, and Zhou}]{su2020moviechats}
H.~Su, X.~Shen, Z.~Xiao, Z.~Zhang, E.~Chang, C.~Zhang, C.~Niu, and J.~Zhou
  (2020{\natexlab{a}}). MovieChats: Chat like Humans in a Closed Domain, in
  {\em Proceedings of EMNLP 2020\/} {\em 2020\/}.

\bibitem[{Su \textit{et~al.}(2019{\natexlab{b}})Su, Shen, Zhang, Sun, Hu, Niu,
  and Zhou}]{su2019improving}
H.~Su, X.~Shen, R.~Zhang, F.~Sun, P.~Hu, C.~Niu, and J.~Zhou
  (2019{\natexlab{b}}). Improving Multi-turn Dialogue Modelling with Utterance
  ReWriter, in {\em Proceedings of the 57th Annual Meeting of the Association
  for Computational Linguistics\/} {\em 2019\/}.

\bibitem[{Su \textit{et~al.}(2020{\natexlab{b}})Su, Shen, Zhao, Xiao, Hu,
  Zhong, Niu, and Zhou}]{su2020diversifying}
H.~Su, X.~Shen, S.~Zhao, Z.~Xiao, P.~Hu, R.~Zhong, C.~Niu, and J.~Zhou
  (2020{\natexlab{b}}). Diversifying Dialogue Generation with
  Non-Conversational Text, in {\em Proceedings of the 58th Annual Meeting of
  the Association for Computational Linguistics\/} {\em 2020\/}.

\bibitem[{Subramanian \textit{et~al.}(2019)Subramanian, Lample, Smith, Denoyer,
  Ranzato, and Boureau}]{subramanian2018multiple}
S.~Subramanian, G.~Lample, E.~M. Smith, L.~Denoyer, M.~Ranzato, and Y.-L.
  Boureau (2019). Multiple-Attribute Text Style Transfer, {\em International
  Conference on Learning Representations(ICLR)\/}.

\bibitem[{Subramanian \textit{et~al.}(2017)Subramanian, Rajeswar, Dutil, Pal,
  and Courville}]{subramanian2017adversarial}
S.~Subramanian, S.~Rajeswar, F.~Dutil, C.~Pal, and A.~Courville (2017).
  Adversarial Generation of Natural Language, in {\em Proceedings of the 2nd
  Workshop on Representation Learning for NLP\/} {\em 2017\/}.

\bibitem[{Sutskever \textit{et~al.}(2014)Sutskever, Vinyals, and
  Le}]{sutskever2014sequence}
I.~Sutskever, O.~Vinyals, and Q.~V. Le (2014). Sequence to sequence learning
  with neural networks, in {\em Advances in neural information processing
  systems\/} {\em 2014\/}.

\bibitem[{Sutton \textit{et~al.}(2012)Sutton, McCallum
  \textit{et~al.}}]{sutton2012introduction}
C.~Sutton, A.~McCallum, \textit{et~al.} (2012). An introduction to conditional
  random fields, {\em Foundations and Trends{\textregistered} in Machine
  Learning\/}, vol.~4(4), pp.~267--373.

\bibitem[{Tang \textit{et~al.}(2021)Tang, Li, Ge, Shen, Zhu, and
  Luo}]{tang2021ast}
Z.~Tang, C.~Li, J.~Ge, X.~Shen, Z.~Zhu, and B.~Luo (2021). AST-Transformer:
  Encoding Abstract Syntax Trees Efficiently for Code Summarization, in {\em
  2021 36th IEEE/ACM International Conference on Automated Software Engineering
  (ASE)\/} {\em 2021\/}.

\bibitem[{Thompson(1977)}]{thompson1977strategy}
H.~Thompson (1977). Strategy and tactics: A model for language production, in
  {\em Papers from the... Regional Meeting. Chicago Ling. Soc. Chicago, Ill\/}
  {\em 1977\/}.

\bibitem[{Tomczak and Welling(2018)}]{tomczak2018vae}
J.~Tomczak and M.~Welling (2018). VAE with a VampPrior, in {\em International
  Conference on Artificial Intelligence and Statistics\/} {\em 2018\/}.

\bibitem[{Trouillon \textit{et~al.}(2016)Trouillon, Welbl, Riedel, Gaussier,
  and Bouchard}]{trouillon2016complex}
T.~Trouillon, J.~Welbl, S.~Riedel, {\'E}.~Gaussier, and G.~Bouchard (2016).
  Complex embeddings for simple link prediction, in {\em International
  Conference on Machine Learning\/} {\em 2016\/}.

\bibitem[{Tucker \textit{et~al.}(2019)Tucker, Lawson, Gu, and
  Maddison}]{tucker2019doubly}
G.~Tucker, D.~Lawson, S.~Gu, and C.~J. Maddison (2019). Doubly Reparameterized
  Gradient Estimators for Monte Carlo Objectives, in {\em International
  Conference on Learning Representations\/} {\em 2019\/}.

\bibitem[{Tucker \textit{et~al.}(2017)Tucker, Mnih, Maddison, Lawson, and
  Sohl-Dickstein}]{tucker2017rebar}
G.~Tucker, A.~Mnih, C.~J. Maddison, J.~Lawson, and J.~Sohl-Dickstein (2017).
  Rebar: Low-variance, unbiased gradient estimates for discrete latent variable
  models, in {\em Advances in Neural Information Processing Systems\/} {\em
  2017\/}.

\bibitem[{Ulyanov \textit{et~al.}(2017)Ulyanov, Vedaldi, and
  Lempitsky}]{ulyanov2017adversarial}
D.~Ulyanov, A.~Vedaldi, and V.~Lempitsky (2017). Adversarial Generator-Encoder
  Networks, {\em arXiv preprint arXiv:1704.02304\/}.

\bibitem[{van~den Oord \textit{et~al.}(2017)van~den Oord, Vinyals
  \textit{et~al.}}]{van2017neural}
A.~van~den Oord, O.~Vinyals, \textit{et~al.} (2017). Neural discrete
  representation learning, in {\em Advances in Neural Information Processing
  Systems\/} {\em 2017\/}.

\bibitem[{Vaswani \textit{et~al.}(2017)Vaswani, Shazeer, Parmar, Uszkoreit,
  Jones, Gomez, Kaiser, and Polosukhin}]{vaswani2017attention}
A.~Vaswani, N.~Shazeer, N.~Parmar, J.~Uszkoreit, L.~Jones, A.~N. Gomez,
  {\L}.~Kaiser, and I.~Polosukhin (2017). Attention is all you need, in {\em
  Advances in Neural Information Processing Systems\/} {\em 2017\/}.

\bibitem[{Vedantam \textit{et~al.}(2015)Vedantam, Lawrence~Zitnick, and
  Parikh}]{vedantam2015cider}
R.~Vedantam, C.~Lawrence~Zitnick, and D.~Parikh (2015). Cider: Consensus-based
  image description evaluation, in {\em Proceedings of the IEEE conference on
  computer vision and pattern recognition\/} {\em 2015\/}.

\bibitem[{Vijayakumar \textit{et~al.}(2018)Vijayakumar, Cogswell, Selvaraju,
  Sun, Lee, Crandall, and Batra}]{vijayakumar2016diverse}
A.~K. Vijayakumar, M.~Cogswell, R.~R. Selvaraju, Q.~Sun, S.~Lee, D.~J.
  Crandall, and D.~Batra (2018). Diverse Beam Search for Improved Description
  of Complex Scenes., {\em Association for the Advancement of Artificial
  Intelligence\/}, pp. 7371--7379.

\bibitem[{Villani(2008)}]{villani2008optimal}
C.~Villani (2008). {\em Optimal transport: old and new\/}, vol. 338, Springer
  Science \& Business Media.

\bibitem[{Vinyals \textit{et~al.}(2015)Vinyals, Fortunato, and
  Jaitly}]{vinyals2015pointer}
O.~Vinyals, M.~Fortunato, and N.~Jaitly (2015). Pointer networks, in {\em
  Advances in Neural Information Processing Systems\/} {\em 2015\/}.

\bibitem[{Vinyals and Le(2015)}]{vinyals2015neural}
O.~Vinyals and Q.~V. Le (2015). A Neural Conversational Model, {\em CoRR\/},
  vol.~abs/1506.05869.

\bibitem[{Vylomova \textit{et~al.}(2016)Vylomova, Rimell, Cohn, and
  Baldwin}]{vylomova2016take}
E.~Vylomova, L.~Rimell, T.~Cohn, and T.~Baldwin (2016). Take and Took, Gaggle
  and Goose, Book and Read: Evaluating the Utility of Vector Differences for
  Lexical Relation Learning, in {\em Proceedings of the 54th Annual Meeting of
  the Association for Computational Linguistics (Volume 1: Long Papers)\/} {\em
  2016\/}.

\bibitem[{Wang \textit{et~al.}(2017{\natexlab{a}})Wang, Wang, Huang, Mohamed,
  Zhou, and Deng}]{wang2017sequence}
C.~Wang, Y.~Wang, P.-S. Huang, A.~Mohamed, D.~Zhou, and L.~Deng
  (2017{\natexlab{a}}). Sequence modeling via segmentations, in {\em
  Proceedings of the 34th International Conference on Machine Learning-Volume
  70\/} {\em 2017\/}.

\bibitem[{Wang \textit{et~al.}(2017{\natexlab{b}})Wang, Jojic, Brockett, and
  Nyberg}]{wang2017steering}
D.~Wang, N.~Jojic, C.~Brockett, and E.~Nyberg (2017{\natexlab{b}}). Steering
  Output Style and Topic in Neural Response Generation, in {\em Proceedings of
  the 2017 Conference on Empirical Methods in Natural Language Processing\/}
  {\em 2017\/}.

\bibitem[{Wang \textit{et~al.}(2017{\natexlab{c}})Wang, Schwing, and
  Lazebnik}]{wang2017diverse}
L.~Wang, A.~Schwing, and S.~Lazebnik (2017{\natexlab{c}}). Diverse and accurate
  image description using a variational auto-encoder with an additive gaussian
  encoding space, in {\em Advances in Neural Information Processing Systems\/}
  {\em 2017\/}.

\bibitem[{Wang \textit{et~al.}(2020)Wang, Shen, de~Melo, and
  Weikum}]{wang2020cross}
L.~Wang, X.~Shen, G.~de~Melo, and G.~Weikum (2020). Cross-Domain Learning for
  Classifying Propaganda in Online Contents, {\em arXiv preprint
  arXiv:2011.06844\/}.

\bibitem[{Wang \textit{et~al.}(2019)Wang, Hu, Yang, Shi, Xu, and
  Xing}]{wang2019toward}
W.~Wang, Z.~Hu, Z.~Yang, H.~Shi, F.~Xu, and E.~Xing (2019). Toward Unsupervised
  Text Content Manipulation, {\em arXiv preprint arXiv:1901.09501\/}.

\bibitem[{Wang \textit{et~al.}(2018{\natexlab{a}})Wang, Zhu, Alkhouli, Gan, and
  Ney}]{wang2018neural}
W.~Wang, D.~Zhu, T.~Alkhouli, Z.~Gan, and H.~Ney (2018{\natexlab{a}}). Neural
  hidden markov model for machine translation, in {\em Proceedings of the 56th
  Annual Meeting of the Association for Computational Linguistics (Volume 2:
  Short Papers)\/} {\em 2018\/}.

\bibitem[{Wang \textit{et~al.}(2018{\natexlab{b}})Wang, Pham, Yin, and
  Neubig}]{wang2018tree}
X.~Wang, H.~Pham, P.~Yin, and G.~Neubig (2018{\natexlab{b}}). A Tree-based
  Decoder for Neural Machine Translation, in {\em Proceedings of the 2018
  Conference on Empirical Methods in Natural Language Processing\/} {\em
  2018\/}.

\bibitem[{Weaver and Tao(2001)}]{weaver2001optimal}
L.~Weaver and N.~Tao (2001). The Optimal Reward Baseline for Gradient-based
  Reinforcement Learning, in {\em Proceedings of the Seventeenth Conference on
  Uncertainty in Artificial Intelligence\/} {\em 2001\/}.

\bibitem[{Wen \textit{et~al.}(2015)Wen, Gasic, Mrk{\v{s}}i{\'c}, Su, Vandyke,
  and Young}]{wen2015semantically}
T.-H. Wen, M.~Gasic, N.~Mrk{\v{s}}i{\'c}, P.-H. Su, D.~Vandyke, and S.~Young
  (2015). Semantically Conditioned LSTM-based Natural Language Generation for
  Spoken Dialogue Systems, in {\em Proceedings of the 2015 Conference on
  Empirical Methods in Natural Language Processing\/} {\em 2015\/}.

\bibitem[{Wen \textit{et~al.}(2017)Wen, Miao, Blunsom, and
  Young}]{wen2017latent}
T.-H. Wen, Y.~Miao, P.~Blunsom, and S.~Young (2017). Latent Intention Dialogue
  Models, in {\em Proceedings of the 34th International Conference on Machine
  Learning\/} {\em 2017\/}.

\bibitem[{Wiehr \textit{et~al.}(2020)Wiehr, Hirsch, Daiber, Kruger, Kovtunova,
  Borgwardt, Chang, Demberg, Steinmetz, and Jorg}]{wiehr2020safe}
F.~Wiehr, A.~Hirsch, F.~Daiber, A.~Kruger, A.~Kovtunova, S.~Borgwardt,
  E.~Chang, V.~Demberg, M.~Steinmetz, and H.~Jorg (2020). Safe Handover in
  Mixed-Initiative Control for Cyber-Physical Systems.

\bibitem[{Wiehr \textit{et~al.}(2021)Wiehr, Hirsch, Schmitz, Knieriemen,
  Kr{\"u}ger, Kovtunova, Borgwardt, Chang, Demberg, Steinmetz
  \textit{et~al.}}]{wiehr2021have}
F.~Wiehr, A.~Hirsch, L.~Schmitz, N.~Knieriemen, A.~Kr{\"u}ger, A.~Kovtunova,
  S.~Borgwardt, E.~Chang, V.~Demberg, M.~Steinmetz, \textit{et~al.} (2021). Why
  Do I Have to Take Over Control? Evaluating Safe Handovers with Advance Notice
  and Explanations in HAD, in {\em Proceedings of the 2021 International
  Conference on Multimodal Interaction\/} {\em 2021\/}.

\bibitem[{Williams(1992)}]{williams1992simple}
R.~J. Williams (1992). Simple statistical gradient-following algorithms for
  connectionist reinforcement learning, {\em Machine learning\/}, vol.~8(3-4),
  pp.~229--256.

\bibitem[{Wiseman \textit{et~al.}(2017)Wiseman, Shieber, and
  Rush}]{wiseman2017challenges}
S.~Wiseman, S.~Shieber, and A.~Rush (2017). Challenges in Data-to-Document
  Generation, in {\em Proceedings of the 2017 Conference on Empirical Methods
  in Natural Language Processing\/} {\em 2017\/}.

\bibitem[{Wiseman \textit{et~al.}(2018)Wiseman, Shieber, and
  Rush}]{wiseman2018learning}
S.~Wiseman, S.~M. Shieber, and A.~M. Rush (2018). Learning Neural Templates for
  Text Generation, {\em Conference on Empirical Methods in Natural Language
  Processing\/}.

\bibitem[{Wu \textit{et~al.}(2019)Wu, Socher, and Xiong}]{wu2018globaltolocal}
C.-S. Wu, R.~Socher, and C.~Xiong (2019). Global-to-local Memory Pointer
  Networks for Task-Oriented Dialogue, in {\em International Conference on
  Learning Representations\/} {\em 2019\/}.

\bibitem[{Wu \textit{et~al.}(2018)Wu, Shapiro, and Cotterell}]{wu2018hard}
S.~Wu, P.~Shapiro, and R.~Cotterell (2018). Hard Non-Monotonic Attention for
  Character-Level Transduction, in {\em Proceedings of the 2018 Conference on
  Empirical Methods in Natural Language Processing\/} {\em 2018\/}.

\bibitem[{Wu \textit{et~al.}(2016)Wu, Schuster, Chen, Le, Norouzi, Macherey,
  Krikun, Cao, Gao, Macherey \textit{et~al.}}]{wu2016google}
Y.~Wu, M.~Schuster, Z.~Chen, Q.~V. Le, M.~Norouzi, W.~Macherey, M.~Krikun,
  Y.~Cao, Q.~Gao, K.~Macherey, \textit{et~al.} (2016). Google's neural machine
  translation system: Bridging the gap between human and machine translation,
  {\em arXiv preprint arXiv:1609.08144\/}.

\bibitem[{Wu \textit{et~al.}(2017)Wu, Wu, Xing, Zhou, and
  Li}]{wu2017sequential}
Y.~Wu, W.~Wu, C.~Xing, M.~Zhou, and Z.~Li (2017). Sequential Matching Network:
  A New Architecture for Multi-turn Response Selection in Retrieval-Based
  Chatbots, in {\em Proceedings of the 55th Annual Meeting of the Association
  for Computational Linguistics (Volume 1: Long Papers)\/} {\em 2017\/}.

\bibitem[{Xie(2017)}]{xie2017neural}
Z.~Xie (2017). Neural text generation: A practical guide, {\em arXiv preprint
  arXiv:1711.09534\/}.

\bibitem[{Xie \textit{et~al.}(2017)Xie, Wang, Li, L{\'e}vy, Nie, Jurafsky, and
  Ng}]{xie2017data}
Z.~Xie, S.~I. Wang, J.~Li, D.~L{\'e}vy, A.~Nie, D.~Jurafsky, and A.~Y. Ng
  (2017). Data Noising as Smoothing in Neural Network Language Models, {\em
  International Conference on Learning Representations(ICLR)\/}.

\bibitem[{Xu \textit{et~al.}(2020)Xu, Qiu, Zhang, Wang, Shen, and
  de~Melo}]{xu2020data}
B.~Xu, S.~Qiu, J.~Zhang, Y.~Wang, X.~Shen, and G.~de~Melo (2020). Data
  Augmentation for Multiclass Utterance Classification--A Systematic Study, in
  {\em Proceedings of the 28th International Conference on Computational
  Linguistics\/} {\em 2020\/}.

\bibitem[{Xu \textit{et~al.}(2015)Xu, Ba, Kiros, Cho, Courville, Salakhudinov,
  Zemel, and Bengio}]{xu2015show}
K.~Xu, J.~Ba, R.~Kiros, K.~Cho, A.~Courville, R.~Salakhudinov, R.~Zemel, and
  Y.~Bengio (2015). Show, attend and tell: Neural image caption generation with
  visual attention, in {\em International conference on machine learning\/}
  {\em 2015\/}.

\bibitem[{Yan \textit{et~al.}(2016)Yan, Yang, Sohn, and
  Lee}]{yan2016attribute2image}
X.~Yan, J.~Yang, K.~Sohn, and H.~Lee (2016). Attribute2image: Conditional image
  generation from visual attributes, in {\em European Conference on Computer
  Vision\/} {\em 2016\/}.

\bibitem[{Yang \textit{et~al.}(2017{\natexlab{a}})Yang, Blunsom, Dyer, and
  Ling}]{yang2017reference}
Z.~Yang, P.~Blunsom, C.~Dyer, and W.~Ling (2017{\natexlab{a}}). Reference-Aware
  Language Models, in {\em Proceedings of the 2017 Conference on Empirical
  Methods in Natural Language Processing\/} {\em 2017\/}.

\bibitem[{Yang \textit{et~al.}(2018)Yang, Dai, Salakhutdinov, and
  Cohen}]{yang2018breaking}
Z.~Yang, Z.~Dai, R.~Salakhutdinov, and W.~W. Cohen (2018). Breaking the Softmax
  Bottleneck: A High-Rank {RNN} Language Model, in {\em International
  Conference on Learning Representations\/} {\em 2018\/}.

\bibitem[{Yang \textit{et~al.}(2017{\natexlab{b}})Yang, Hu, Salakhutdinov, and
  Berg-Kirkpatrick}]{yang2017improved}
Z.~Yang, Z.~Hu, R.~Salakhutdinov, and T.~Berg-Kirkpatrick (2017{\natexlab{b}}).
  Improved Variational Autoencoders for Text Modeling using Dilated
  Convolutions, {\em International Conference on Machine Learning (ICML)\/}.

\bibitem[{Yang \textit{et~al.}(2019)Yang, wu, Yang, Xu, and
  li}]{yang-etal-2019-low}
Z.~Yang, w.~wu, J.~Yang, C.~Xu, and z.~li (2019). Low-Resource Response
  Generation with Template Prior, in {\em Proceedings of the 2019 Conference on
  Empirical Methods in Natural Language Processing and the 9th International
  Joint Conference on Natural Language Processing (EMNLP-IJCNLP)\/} {\em
  2019\/}.

\bibitem[{Yao \textit{et~al.}(2019)Yao, Peng, Ralph, Knight, Zhao, and
  Yan}]{yao2018plan}
L.~Yao, N.~Peng, W.~Ralph, K.~Knight, D.~Zhao, and R.~Yan (2019).
  Plan-And-Write: Towards Better Automatic Storytelling, {\em Association for
  the Advancement of Artificial Intelligence\/}.

\bibitem[{Yu \textit{et~al.}(2016)Yu, Buys, and Blunsom}]{yu2016online}
L.~Yu, J.~Buys, and P.~Blunsom (2016). Online Segment to Segment Neural
  Transduction, in {\em Proceedings of the 2016 Conference on Empirical Methods
  in Natural Language Processing\/} {\em 2016\/}.

\bibitem[{Yu \textit{et~al.}(2017)Yu, Zhang, Wang, and Yu}]{yu2017seqgan}
L.~Yu, W.~Zhang, J.~Wang, and Y.~Yu (2017). SeqGAN: Sequence Generative
  Adversarial Nets with Policy Gradient., in {\em Association for the
  Advancement of Artificial Intelligence\/} {\em 2017\/}.

\bibitem[{Yu \textit{et~al.}(2018)Yu, Zhang, Huang, and Zhu}]{yu2018operation}
N.~Yu, J.~Zhang, M.~Huang, and X.~Zhu (2018). An Operation Network for
  Abstractive Sentence Compression, in {\em Proceedings of the 27th
  International Conference on Computational Linguistics\/} {\em 2018\/}.

\bibitem[{Zhang \textit{et~al.}(2016{\natexlab{a}})Zhang, Xiong, Su, Duan, and
  Zhang}]{zhang2016variational}
B.~Zhang, D.~Xiong, J.~Su, H.~Duan, and M.~Zhang (2016{\natexlab{a}}).
  Variational neural machine translation, {\em arXiv preprint
  arXiv:1605.07869\/}.

\bibitem[{Zhang \textit{et~al.}(2018{\natexlab{a}})Zhang, Yao, and
  Yan}]{zhang2018abstractiveness}
F.~Zhang, J.-g. Yao, and R.~Yan (2018{\natexlab{a}}). On the Abstractiveness of
  Neural Document Summarization, in {\em Proceedings of the 2018 Conference on
  Empirical Methods in Natural Language Processing\/} {\em 2018\/}.

\bibitem[{Zhang \textit{et~al.}(2021)Zhang, Gu, Shen, and
  Su}]{zhang2021knowledge}
R.~Zhang, Y.~Gu, X.~Shen, and H.~Su (2021). Knowledge-enhanced Session-based
  Recommendation with Temporal Transformer, {\em arXiv preprint
  arXiv:2112.08745\/}.

\bibitem[{Zhang \textit{et~al.}(2018{\natexlab{b}})Zhang, Dinan, Urbanek,
  Szlam, Kiela, and Weston}]{zhang2018personalizing}
S.~Zhang, E.~Dinan, J.~Urbanek, A.~Szlam, D.~Kiela, and J.~Weston
  (2018{\natexlab{b}}). Personalizing Dialogue Agents: I have a dog, do you
  have pets too?, in {\em Proceedings of the 56th Annual Meeting of the
  Association for Computational Linguistics (Volume 1: Long Papers)\/} {\em
  2018\/}.

\bibitem[{Zhang \textit{et~al.}(2018{\natexlab{c}})Zhang, Galley, Gao, Gan, Li,
  Brockett, and Dolan}]{zhang2018generating}
Y.~Zhang, M.~Galley, J.~Gao, Z.~Gan, X.~Li, C.~Brockett, and B.~Dolan
  (2018{\natexlab{c}}). Generating informative and diverse conversational
  responses via adversarial information maximization, in {\em Advances in
  Neural Information Processing Systems\/} {\em 2018\/}.

\bibitem[{Zhang \textit{et~al.}(2016{\natexlab{b}})Zhang, Gan, and
  Carin}]{zhang2016generating}
Y.~Zhang, Z.~Gan, and L.~Carin (2016{\natexlab{b}}). Generating text via
  adversarial training, in {\em Advances in Neural Information Processing
  Systems workshop on Adversarial Training\/} {\em 2016\/}.

\bibitem[{Zhang \textit{et~al.}(2017)Zhang, Gan, Fan, Chen, Henao, Shen, and
  Carin}]{zhang2017adversarial}
Y.~Zhang, Z.~Gan, K.~Fan, Z.~Chen, R.~Henao, D.~Shen, and L.~Carin (2017).
  Adversarial Feature Matching for Text Generation, {\em International
  Conference on Machine Learning (ICML)\/}.

\bibitem[{Zhao \textit{et~al.}(2017{\natexlab{a}})Zhao, Mathieu, and
  LeCun}]{zhao2016energy}
J.~Zhao, M.~Mathieu, and Y.~LeCun (2017{\natexlab{a}}). Energy-based generative
  adversarial network, {\em International Conference on Learning
  Representations(ICLR)\/}.

\bibitem[{Zhao \textit{et~al.}(2017{\natexlab{b}})Zhao, Song, and
  Ermon}]{zhao2017towards}
S.~Zhao, J.~Song, and S.~Ermon (2017{\natexlab{b}}). Towards Deeper
  Understanding of Variational Autoencoding Models, {\em arXiv preprint
  arXiv:1702.08658\/}.

\bibitem[{Zhao \textit{et~al.}(2018{\natexlab{a}})Zhao, Song, and
  Ermon}]{zhao2018information}
S.~Zhao, J.~Song, and S.~Ermon (2018{\natexlab{a}}). The Information
  Autoencoding Family: A Lagrangian Perspective on Latent Variable Generative
  Models, {\em UAI\/}.

\bibitem[{Zhao \textit{et~al.}(2017{\natexlab{c}})Zhao, Zhao, and
  Eskenazi}]{zhao2017learning}
T.~Zhao, R.~Zhao, and M.~Eskenazi (2017{\natexlab{c}}). Learning
  Discourse-level Diversity for Neural Dialog Models using Conditional
  Variational Autoencoders, in {\em Proceedings of the 55th Annual Meeting of
  the Association for Computational Linguistics (Volume 1: Long Papers)\/} {\em
  2017\/}.

\bibitem[{Zhao \textit{et~al.}(2017{\natexlab{d}})Zhao, Senuma, Shen, and
  Aizawa}]{zhao2017gated}
Y.~Zhao, H.~Senuma, X.~Shen, and A.~Aizawa (2017{\natexlab{d}}). Gated neural
  network for sentence compression using linguistic knowledge, in {\em
  International Conference on Applications of Natural Language to Information
  Systems\/} {\em 2017\/}.

\bibitem[{Zhao \textit{et~al.}(2019)Zhao, Shen, Bi, and
  Aizawa}]{zhao2019unsupervised}
Y.~Zhao, X.~Shen, W.~Bi, and A.~Aizawa (2019). Unsupervised Rewriter for
  Multi-Sentence Compression, in {\em Proceedings of the 57th Annual Meeting of
  the Association for Computational Linguistics\/} {\em 2019\/}.

\bibitem[{Zhao \textit{et~al.}(2018{\natexlab{b}})Zhao, Shen, Senuma, and
  Aizawa}]{zhao2018comprehensive}
Y.~Zhao, X.~Shen, H.~Senuma, and A.~Aizawa (2018{\natexlab{b}}). A
  comprehensive study: Sentence compression with linguistic knowledge-enhanced
  gated neural network, {\em Data \& Knowledge Engineering\/}, vol.~117,
  pp.~307--318.

\bibitem[{Zhou and Neubig(2017)}]{zhou2017multi}
C.~Zhou and G.~Neubig (2017). Multi-space Variational Encoder-Decoders for
  Semi-supervised Labeled Sequence Transduction, {\em Annual Meeting of the
  Association for Computational Linguistics\/}.

\bibitem[{Zhou \textit{et~al.}(2018)Zhou, Huang, Zhang, Zhu, and
  Liu}]{zhou2018emotional}
H.~Zhou, M.~Huang, T.~Zhang, X.~Zhu, and B.~Liu (2018). Emotional chatting
  machine: Emotional conversation generation with internal and external memory,
  in {\em Thirty-Second AAAI Conference on Artificial Intelligence\/} {\em
  2018\/}.

\bibitem[{Zhou \textit{et~al.}(2017)Zhou, Yang, Wei, and
  Zhou}]{zhou2017selective}
Q.~Zhou, N.~Yang, F.~Wei, and M.~Zhou (2017). Selective Encoding for
  Abstractive Sentence Summarization, in {\em Proceedings of the 55th Annual
  Meeting of the Association for Computational Linguistics (Volume 1: Long
  Papers)\/} {\em 2017\/}.

\bibitem[{Zhou \textit{et~al.}(2020)Zhou, Ge, Xu, Wei, and
  Zhou}]{Zhou2020Self-Adversarial}
W.~Zhou, T.~Ge, K.~Xu, F.~Wei, and M.~Zhou (2020). Self-Adversarial Learning
  with Comparative Discrimination for Text Generation, in {\em International
  Conference on Learning Representations\/} {\em 2020\/}.

\bibitem[{Zhu \textit{et~al.}(2018)Zhu, Lu, Zheng, Guo, Zhang, Wang, and
  Yu}]{zhu2018texygen}
Y.~Zhu, S.~Lu, L.~Zheng, J.~Guo, W.~Zhang, J.~Wang, and Y.~Yu (2018). Texygen:
  a benchmarking platform for text generation models, in {\em The 41st
  International ACM SIGIR Conference on Research \& Development in Information
  Retrieval\/} {\em 2018\/}.

\bibitem[{Zhuang and Chang(2017)}]{zhuang2017neobility}
W.~Zhuang and E.~Chang (2017). Neobility at SemEval-2017 Task 1: An
  attention-based sentence similarity model., in {\em In Proceedings of
  SemEval-2017 at ACL 2017.\/} {\em 2017\/}.

\end{thebibliography}
\cleardoublepage


\end{document}